%% file: Main.tex
\documentclass{article}

\usepackage[a4paper, margin=1in]{geometry}
\usepackage[english]{babel}
\usepackage{hyperref}       
\usepackage{url}     
\usepackage[T1]{fontenc}
\usepackage[utf8]{inputenc}
\usepackage{adjustbox}
\usepackage{graphicx}
\usepackage{subfigure}
\usepackage{authblk}
\usepackage{amsfonts}
\usepackage{amsmath}
\usepackage{color}
\usepackage{bm}
\usepackage[sort,numbers]{natbib}
\usepackage{xcolor}
\usepackage[ruled,linesnumbered]{algorithm2e}
\usepackage{multirow}

\usepackage{booktabs}
\usepackage{makecell}
\usepackage{hhline}
\usepackage[strings]{underscore}
\usepackage{threeparttable}

\newcommand{\mc}{\mathcal}

\providecommand{\keywords}[1]
{
  \small	
  \textbf{\textit{Keywords---}} #1
}

\title{Geometric Rectification of Creased Document Images based on Isometric Mapping}

\author{Dong Luo \authorcr \href{mailto:borges.d.luo@gmail.com}{borges.d.luo@gmail.com}}
\affil{Harbin Institute of Technology, Weihai, 264209, China}
\author{Pengbo Bo \thanks{Corresponding author} 
\authorcr 
\href{mailto:pbbo@hit.edu.cn}{pbbo@hit.edu.cn}}
\affil{Harbin Institute of Technology, Weihai, 264209, China}
\date{}

\begin{document}
\maketitle
\begin{abstract}
Geometric rectification of images of distorted documents finds wide applications in document digitization and Optical Character Recognition (OCR). Although smoothly curved deformations have been widely investigated by many works, the most challenging distortions, e.g. complex creases and large foldings, have not been studied in particular. The performance of existing approaches, when applied to largely creased or folded documents, is far from satisfying, leaving substantial room for improvement. To tackle this task, knowledge about document rectification should be incorporated into the computation, among which the developability of 3D document models and particular textural features in the images, such as straight lines, are the most essential ones. For this purpose, we propose a general framework of document image rectification in which a computational isometric mapping model is utilized for expressing a 3D document model and its flattening in the plane. Based on this framework, both model developability and textural features are considered in the computation.  The experiments and comparisons to the state-of-the-art approaches demonstrated the effectiveness and outstanding performance of the proposed method. Our method is also flexible in that the rectification results can be enhanced  by any other methods that extract high-quality feature lines in the images.
\end{abstract}


\input{Introduction}
\input{RelatedWork}
\input{Constraints}

\input{Algorithm}

\input{Experiments}
\input{Conclusion}

\paragraph{Acknowledgments} We thank the authors of the works in comparison for sharing their data and models online. We thank the authors of  DocScanner for their help with generating the results for comparison.

\bibliography{bibtex}
\end{document}

%% file: Introduction.tex
\section{Introduction}
Rectification of distorted document images aims to compute an image of the planar state of a document from one or several photos of the document with deformations. It finds wide applications in various fields, such as OCR and document digitization. The factors that impact rectification quality include image background, geometric deformation, and illumination condition. In this paper, we focus on geometric rectification of distorted document images and propose a novel method which is capable of dealing with the challenging case of creased documents.

The difficulty level of geometric rectification depends on the type and complexity of the distortion. Traditional methods often assume a smoothly curved surface which can be expressed by cylinders. Non-smooth surfaces with folds and creases are recently investigated using deep learning-based or multi-view methods. However, the existing approaches do not work well for documents with complex creases exhibiting a fractal shape or a large number of folds of varying sizes. To tackle this challenging task, we need to incorporate some knowledge about document rectification in the method.  Firstly, the deformed document is known to be a developable surface. Therefore, enforcing developability in the document model will benefit image rectification through improving the accuracy of document development. Moreover, some particular textural features in the document can serve as constraints for rectification, e.g. straightening feature lines in the rectified image which are known to be straight in the planar state of the document. Although these two kinds of constraints have been investigated for rectifying smoothly curved documents, incorporating them into a method for creased documents is non-trivial due to the lack of explicit expressions for creased document models.

We propose a method based on  the 3D  reconstruction model $\mc{M}$ of the document and its 2D unfolding $\mc{M}^{\prime}$ in which both the developability property and textural features can be used as constraints. This is made possible by incorporating an computational isometric mapping model which enables  simultaneous computation of  $\mc{M}$ and $\mc{M}^{\prime}$ and this distinguishes our method from existing ones  which separate 3D model reconstruction and its unwrapping. The benefit of our method is thus a direct control on the quality of the rectified image using the textural features in the input image. The contributions of our work are summarized as follows.

\begin{itemize}
\item Isometric mapping between freeform meshes is investigated in document image rectification to give a constraint which leads to a method capable of dealing with documents with complex creases. 
\item Our method directly computes the final rectified document without an isolated unfolding process, thus avoiding additional errors produced by unfolding.
\item  The textural features in the image, such as straight lines and text lines, are naturally integrated into our method as constraints to serve as direct control over the rectified document.  
\end{itemize}

%% file: RelatedWork.tex
\section{Related works}

We classify previous works on document rectification into three categories as follows.

\subsection{Methods based on image deformation} Document rectification can be treated as an image deformation problem and the deformation is mainly driven by straightening the text lines or boundaries of the document.  Therefore, this method is widely considered for documents with text. Various curves are used to represent text lines, including cubic splines~\cite{lavialle2001active, ezaki2005dewarping} and thin-plate splines~\cite{zhang2008arbitrary, liu2015restoring}. Efforts have been made to improve the accuracy of text line extraction. ~\citeauthor{ulges2005document}~\cite{ulges2005document} estimates quadrilateral cell for each letter based on local baseline detection. ~\citeauthor{shijianlu2006document}~\cite{shijianlu2006document} exploits the vertical stroke boundary (VSB)~\cite{lu2005perspective}  to estimate vertical text direction and then divides curved text lines into quadrilateral grids on the image. ~\citeauthor{schneider2007robust}~\cite{schneider2007robust} uses text baselines to generate vector fields for wrapped surface reconstruction. ~\citeauthor{stamatopoulos2011goaloriented}~\cite{stamatopoulos2011goaloriented} finds the boundaries of the text region to do a coarse rectification. In image deformation methods, the document is assumed to be smoothly curved. Some works apply the shape-from-shading (SFS) method for document rectification by unfolding the reconstructed surface model~\cite{wada1997shape, chewlimtan2006restoring, courteille2004shape, courteille2007shape, zhang2004estimation, zhang2009unified} and are limited to the cases with simple geometric distortions.

\subsection{Methods based on 3D model representation}

Since the document surface is a developable surface, various kinds of developable surface models, including continuous ones or discrete ones, have been used for the document model.  

\paragraph{General cylindrical model.}  The cylindrical surface model (CSM) is proposed for the curled pages in opened books, assuming the parallelism of the generatrix of the book surface with the image plane~\cite{cao2003cylindrical}~\cite{koo2013segmentation}~\cite{koo2009composition}. This parallelism  assumption is relaxed in the method in ~\cite{fu2007modelbased}. The general cylindrical surface (GCS) model was proposed to deal with waving shapes which can be constructed from text lines~\cite{meng2012metric, kim2015document}, horizontal edges of rectangular regions in the image~\cite{kil2017robust}, vector fields indicating curved baselines~\cite{meng2018exploiting}, or document boundaries~\cite{he2013book}. General cylindrical models are suitable for curved pages of books without folds but can not be used for curved documents where the generatrices are far from being parallel.

\paragraph{Ruled surface.} The ruled surfaces are considered to deal with surface models in which the generatrices intersect with each other, which is thus able to handle more complex curls and even simple folds. In ~\cite{brown2006geometric, tsoi2007multiview}, the ruled surface is constructed from boundary curves of the document, and the folds are modeled by ruling lines, resulting in a piecewise cylindrical model. In~\cite{meng2020baselines}, assuming a developable ruled surface, the baselines in the image are extracted and the slope fields from text lines are computed to generate the mesh model. Dual to the ruling representation, a sequence of strips is used to generate a piece of discrete developable surface~\cite{jianliang2005flattening, liang2008geometric}. In ~\cite{tian2011rectification}, the document surface is represented by a 3D quad mesh composed of parallelogram cells which is a discrete developable model. This method assumes that the text in the document has the same font type and size and is not flexible enough to represent a general developable surface.

\paragraph{General surfaces.} More general surfaces are needed for documents with complicated deformations such as large foldings and creasings. In~\cite{perriollat2013computational}, the model of multiple ruled surfaces connected by planar transition regions is generated from the 3D data points using the structure-from-motion (SfM) method. ~\citeauthor{you2018multiview}~\cite{you2018multiview} uses a general mesh to represent the folded documents. It reconstructs the 3d mesh surface from the data points given by the SfM method with specific treatment to sharp ridges. However, surface developability is not enforced in the reconstruction. The reconstructed mesh is then unwrapped to the plane to obtain the rectified document.  In contrast to the reconstruction-unfolding strategy employed in existing methods, our method simultaneously solves for a developable model and its flattening state, thus avoiding additional errors introduced by unwrapping. Furthermore,  the quality of the rectified document is directly controlled by the textural features in the input image with our method.

\subsection{Methods based on deep learning} 
Deep learning methods have been applied to document rectification and exhibit superiorities over traditional methods. Most deep-learning strategies learn the features implicitly which do not have explicit meanings related to document-specific features such as textlines~\cite{li2019document, liu2020geometric, ma2018docunet, xie2021dewarping, xie2022document}. The other methods try to incorporate the structural features in the images or the 3D geometry models of the documents in the network, as explained as follows.

\paragraph{Structural information.} Documents contain a range of structural details that are helpful to unwrapping, including boundaries, text lines, creases, etc.  ~\citeauthor{das2017common}~\cite{das2017common}  divides the document into several patches along the detected folds. The patches are subsequently unwarped independently using the Coon’s Patch.  CREASE~\cite{markovitz2020can} takes into account the slope angles of the texts in the reconstruction.  Both DocTr~\cite{feng2021doctr} and DocScanner~\cite{feng2021docscanner} implement semantic segmentation modules to extract boundaries for prepossessing, resulting in improved performance in the cases of complex backgrounds. Moreover, DocScanner identifies the pixel-wise positions of each distorted straight line in the document and straightens them. However, the extracted lines are inaccurate in the regions around complex cases since the feature lines such as text lines are not directly relied on.

\paragraph{3D information.} Some deep-learning methods take advantage of the 3D document models, by restricting the 3D models in a physically limited domain~\cite{das2019dewarpnet, markovitz2020can, das2021endtoend}. DewarpNet~\cite{das2019dewarpnet} replaces the 2D forward mapping in DocUNet~\cite{ma2018docunet} by the 3D shape and texture networks. CREASE \cite{markovitz2020can} introduces additional content information for estimating 3D shapes based on  DewarpNet~\cite{das2019dewarpnet}. ~\citeauthor{das2021endtoend}~\cite{das2021endtoend} estimates the 3D shape and piece-wise unwrapping to propose a fully-differentiable stitching module with global constraints for 3D patches. However, the property of being developable has not been incorporated in these methods for building the mapping from 3D shapes to 2D images. 

%% file: Constraints.tex
 \section{Document image rectification using isometric mapping}
\label{sec:constraints}

Our approach to document image rectification is based on the 3D reconstruction of the document and its unfolding into the plane. Unlike existing methods that perform 3D reconstruction and model unfolding in two separate steps, our method computes both models simultaneously while preserving an isometric mapping between them. The isometric mapping also establishes a connection between the input image and the rectified image, making it possible to define constraints for the flattened document using the textural features in the input image. In this section, we first introduce a general framework of document image rectification involving the isometric mapping model. Then a variety of constraints are proposed, which are used to formulate an optimization problem for document image rectification. The detailed algorithm will be given in Sec.~\ref{sec:alg}.

\subsection{Framework of document image rectification}
\label{sec:framework}

\begin{figure}[!t]     
\centering
\includegraphics[width=1\textwidth]{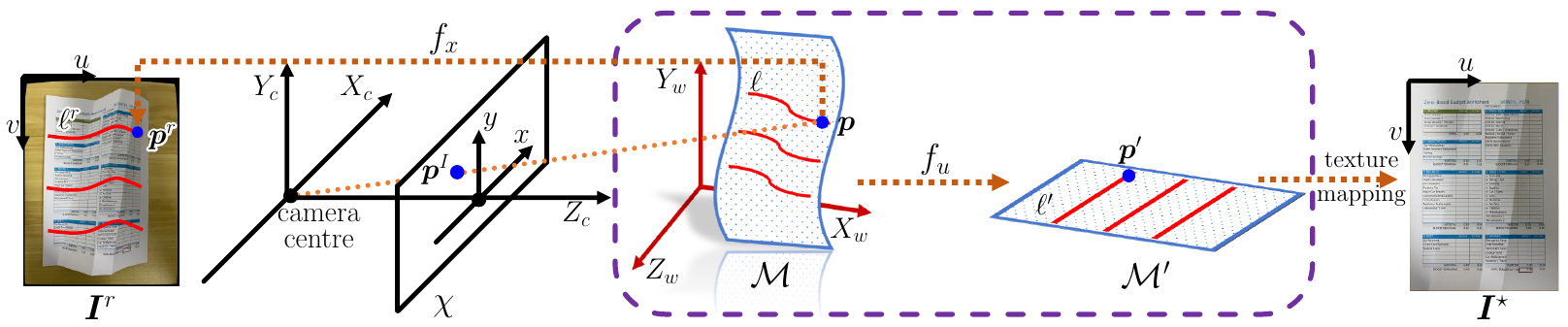}
\caption{Framework of document image rectification.} \label{f:general_model}
\end{figure}

 Fig.~\ref{f:general_model} illustrates the framework of document image rectification. In addition to the input image $\bm{I}^{r}$ and the output image $\bm{I}^*$, three objects are included, which are the image plane $\chi$, the 3D model $\mc{M}$, and the unfold model $\mc{M}^{\prime}$. In our discussions, a point in $\bm{I}^{r}$ is expressed by image coordinates in pixels, also called pixel coordinates. The models  $\mc{M}$ and $\mc{M}^{\prime}$ are represented in the camera coordinate system. 
 
The left part of the framework in Fig.~\ref{f:general_model} containing $\bm{I}^r$,$\chi$, and $\mc{M}$ expresses the known pinhole camera model.  A point in $\mc{M}$ is related to a point in $\bm{I}^{r}$ by the mapping  $f_x\colon \mc{M} \mapsto \bm{I}^{r}$ which is defined by the intrinsic matrix,  $\bm{K}=\tiny\left[\begin{array}{lll}
f * k_{u} & & c_{u} \\
& f * k_{v} &  c_{v} \\
& & 1
\end{array}\right]$, with $(c_{u},c_{v})$ being the principal point, $k_{u}$ and $k_{v}$ being the scale factors, and $f$ being the focal length. Inversely, for a point $\bm{p}^{r}$ in $\bm{I}^{r}$ with the pixel coordinate $(u,v)$, the corresponding point $\bm{p}^{I}$ in $\chi$ is easily known  and the depth of $\bm{p}^{r}$ can be obtained.  This is easily seen  from the fact that $\bm{p}^{I}$ is the projection of a point $\bm{p}$ in $\mc{M}$ that is the intersection point between  $\mc{M}$ and the viewing ray passing through the camera center and $\bm{p}^{I}$.

The right part of the framework contains $\mc{M}^{\prime}$  which expresses the flattened state of the document from which the rectified image can be generated.  $\mc{M}^{\prime}$  is related to the pinhole camera model by the mapping $f_u\colon   \mc{M} \mapsto \mc{M}^{\prime}$ which defines the unfolding of $\mc{M}$ to $\mc{M}^{\prime}$ and is known to be an isometric mapping. Furthermore,  a point $\bm{p}^{\prime}$ in the flat model is related to a point   $\bm{p}^{r}$  in the input image $\bm{I}^r$ via $\bm{p}^{r}= f_{x} \circ f_u^{-1}(\bm{p}^{\prime})$.  The rectified image $\bm{I}^\star$ can be obtained by performing texture mapping in $\mc{M}^\prime$ with $\bm{I}^r$ being the texture image.

Based on the presented framework in Fig.~\ref{f:general_model}, the rectified image $\bm{I}^{*}$ depends on the models $\mc{M}$ and $\mc{M}^{\prime}$.  The vertices of both $\mc{M}$ and $\mc{M}^{\prime}$ are unknown variables, which are computed simultaneously in our method while preserving an isometric mapping between them.  The benefits of this strategy are two-fold. Firstly, the error caused by unfolding can be directly controlled and minimized. Secondly, the textural features in  $\bm{I}^r$ can be used to directly control the accuracy of the rectified image $\bm{I}^\star$, for example by  enforcing straightness of straight lines in the flattened document.

To compute $\mc{M}$ and $\mc{M}^{\prime}$, we adopt the computational isometric model in \cite{jiang2020quadmesh} to represent  $\mc{M}$ and it's isometric unfolding $\mc{M}^\prime$ by combinatorially equivalent quad meshes.  In this way, $\mc{M}$ is a discrete developable surface and this representation is known to be more flexible than other ones, such as planar quad meshes and orthogonal geodesic nets~\cite{jiang2020quadmesh}.  To compute $\mc{M}$ and $\mc{M}^\prime$ in the framework of document image rectification, we formulate an optimization problem involving some appropriate constraints given in Sec.~\ref{sec:constraintSec}.

\subsection{Constraints for document rectification}
\label{sec:constraintSec}
Two types of features of document image rectification can serve as constraints. The first one concerns the developability property of the deformed geometric model. The second one uses textural features in the document whose shapes are known a priori, e.g. straight lines.  The framework in Fig.~\ref{f:general_model} establishes connections between the input image, the rectified image, and the 3D geometric model of the document,  making the incorporation of both types of constraints possible. 

\subsubsection{Developability constraints}
\label{sec:isomapping}

\begin{figure}[!t]     
\centering
\includegraphics[width=.6\textwidth]{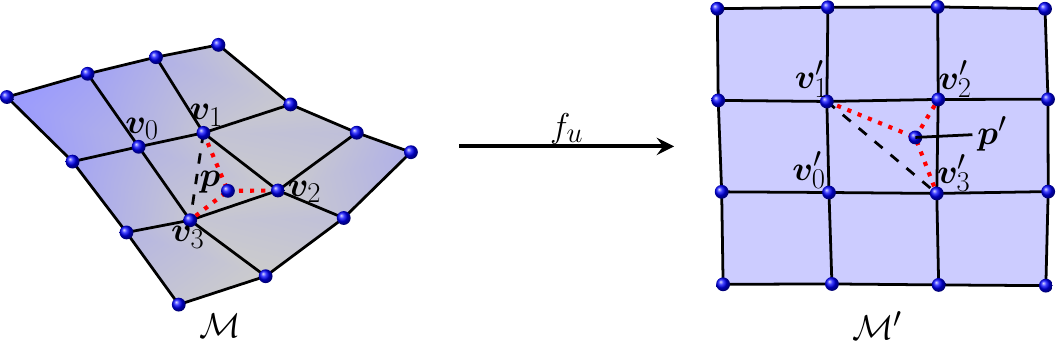}
\caption{Isometric mapping between two quad meshes. $\bm{v}_0\bm{v}_1\bm{v}_2\bm{v}_3$ and $\bm{v}^{\prime}_0\bm{v}^{\prime}_1\bm{v}^{\prime}_2\bm{v}^{\prime}_3$ are the corresponding two faces which satisfy the isometric mapping. $\bm{p}^\prime$ is the corresponding point of $\bm{p}$ in $\mc{M}^\prime$ and both can be expressed in the same barycentric coordinates in the triangle $\bm{v}_1\bm{v}_2\bm{v}_3$ and $\bm{v}_1^\prime\bm{v}_2^\prime\bm{v}_3^\prime$, respectively.} \label{f:iso_meshes}
\end{figure}

\paragraph{Isometric mapping constraint}
The isometric mapping constraint proposed by \cite{jiang2020quadmesh} is defined based on two combinatorially equivalent quad meshes, i.e., in our setting, the deformed document mesh $\mc{M}$ and the flattened mesh $\mc{M}^{\prime}$.  Denote the 3D mesh by $\mc{M}=(\mc{V}, \mc{E}, \mc{F})$ where $\mc{V}$, $\mc{E}$ and $\mc{F}$ refer to the sets of vertices, edges, and faces, respectively. Similarly, the planar mesh is denoted by $\mc{M}^\prime=(\mc{V}^{\prime}, \mc{E}^{\prime}, \mc{F}^{\prime})$. A face and vertex of $\mc{M}$ are denoted by $f$ resp. $\bm{v}$, and their planar correspondences are $f^{\prime}$ resp. $\bm{v}^{\prime}$.  Let the vertex positions of a quad face $f \in \mc{F}$ be $\bm{v}_0\bm{v}_1\bm{v}_2\bm{v}_3$ and the corresponding quad face $f^{\prime} \in \mc{F}^\prime$ be $\bm{v}^{\prime}_0\bm{v}^{\prime}_1\bm{v}^{\prime}_2\bm{v}^{\prime}_3$, cf. Fig.~\ref{f:iso_meshes}. The mapping between $\mc{M}$ and $\mc{M}^\prime$ is isometric if  and only if the following equations hold for each pair of corresponding faces $f$ and $f^{\prime}$:

\begin{equation}
\begin{aligned}
&c_{\text {iso}, 1}(f)=\left\|\bm{v}_{0}-\bm{v}_{2}\right\|^{2}-\left\|\bm{v}^{\prime}_{0}-\bm{v}^{\prime}_{2}\right\|^{2}=0, \\
&c_{\text {iso}, 2}(f)=\left\|\bm{v}_{1}-\bm{v}_{3}\right\|^{2}-\left\|\bm{v}^{\prime}_{1}-\bm{v}^{\prime}_{3}\right\|^{2}=0, \\
&c_{\text {iso}, 3}(f)=\left\langle \bm{v}_{0}-\bm{v}_{2}, \bm{v}_{1}-\bm{v}_{3}\right\rangle-\left\langle \bm{v}^{\prime}_{0}-\bm{v}^{\prime}_{2}, \bm{v}^{\prime}_{1}-\bm{v}^{\prime}_{3}\right\rangle=0.
\label{eq:iso_local}
\end{aligned}
\end{equation}

The isometric mapping constraint is expressed by minimizing an energy function $E_{\text{iso}}$, which is defined by

\begin{equation}
E_{\text {iso}}(\mc{V},\mc{V}^{\prime})=\sum_{f \in \mc{F}} \sum_{i=1}^{3} c_{\text {iso}, i}(f)^{2}.
\label{c:iso}
\end{equation}

By minimizing Eq.~(\ref{c:iso}), the preservation of the developability of the deformed document defines a restricted solution space, naturally serving as a constraint  to improve the accuracy of document image rectification.

\paragraph{Regularization}
As in existing mesh optimization work, a regularization term is  employed to prevent zigzagging polylines or self-intersections of the mesh. This term is defined based on the fairness of streamlines in the mesh. For a polyline $\bm{v}_1\bm{v}_2\ldots\bm{v}_L$ in $\mc{M}$, the fairness at $v_i$ is expressed by $2\bm{v}_i=\bm{v}_{i-1}+\bm{v}_{i+1}$. Employing this rule, we define the regularization terms $E_{\text {fair},\mc{M}}$ and $E_{\text {fair},\mc{M}^\prime}$  for $\mc{M}$ and $\mc{M}^\prime$, respectively: 

\begin{equation}
\begin{aligned}
E_{\text{fair},\mc{M}}(\mc{V})=\sum\nolimits_{}\left\|\bm{v}_{i}-2 \bm{v}_{j}+\bm{v}_{k}\right\|^{2},~\bm{v}_{i} \bm{v}_{j} \bm{v}_{k} \text{~successive vertices in~} \mc{M},
\label{c:f3}
\end{aligned}
\end{equation}
\begin{equation}
\begin{aligned}
E_{\text{fair},\mc{M}^\prime}(\mc{V}^{\prime})=\sum\nolimits_{}\left\|\bm{v}^{\prime}_{i}-2 \bm{v}^{\prime}_{j}+\bm{v}^{\prime}_{k}\right\|^{2},~\bm{v}^{\prime}_{i} \bm{v}^{\prime}_{j} \bm{v}^{\prime}_{k} \text{~successive vertices in~} \mc{M}^{\prime}.
\label{c:f2}
\end{aligned}
\end{equation}

\paragraph{Shape fidelity}
To provide shape guidance for $\mc{M}$, we employ the structure-from-motion (SfM) method to obtain a point cloud $\hat{\mc{P}}$, which is represented in the camera coordinate system corresponding to a reference image $\bm{I}^{r}$  chosen from the set of multi-view images. The camera parameters and images with optical distortion correction are also given by the SfM method.  

The original point cloud obtained by SfM, $\hat{\mc{P}}$,  may contain noise or outliers and is processed using a noise filtering operation detailed in Sec~\ref{sec:outlier} to give the valid data points $\mc{P}$.   $\mc{P}$ are the target points that $\mc{M}$ should approximate. This model approximation problem is resolved by minimizing the distance between $\mc{M}$ and $\mc{P}$, which leads to a highly nonlinear function.  Therefore, it is commonly solved iteratively and a quadratic distance error function is minimized in each iteration. For each  point $\bm{x}\in\mc{P}$, let the footpoint of it in $\mc{M}$ be $\bm{x}^{\pi}$, which lies in a triangle part $\bm{v}_a\bm{v}_b\bm{v}_c$ of a quad face of $\mc{M}$, and $\bm{N}$ be the unit normal vector of the triangle. The distance error function $E_{\text{dist},\mc{M}}$ is defined by

\begin{equation}
		E_{\text {dist}, \mc{M}}(\mc{V})= \frac{1}{H_1}\sum_{\bm{x} \in \mc{P}}{\left( \beta_1 \Vert \bm{x} - \bm{x}^{\pi} \Vert ^2 + \beta_2 <\bm{x} -\bm{x}^{\pi}, \bm{N}>^2 \right) },
	\label{c:dis}
\end{equation}
where $\bm{x}^{\pi}=\begin{bmatrix}{\bm{v}_a}&{\bm{v}_b}&{\bm{v}_c}\end{bmatrix} \bm{\omega}$,  with $\bm{\omega}$ being the barycentric coordinates and $H_1$ being the number of data points in $\mc{P}$. The first term in Eq.~(\ref{c:dis}) is known to be the point-point distance and the second term is the point-tangent distance error. A proper combination of them has been used to define a distance error term that works fast and stable \cite{bo2012Revisit}. We set $\beta_1=1$ and $\beta_2=0.1$ in our implementation.

\paragraph{Remarks:} Distance functions based on $L_1$ norm  are used in~\cite{tsoi2007multiview} for processing noisy data points where it works better than the $L_2$ norm. In our method, due to additional constraints concerning document developability and textural features,  $L_2$ norm works well when combined with the progressive noise processing strategy detailed in Sec.~\ref{sec:alg}.

\subsubsection{Feature line constraints}
\label{sec:feature_line_con}
A document frequently contains straight lines, namely \emph{feature lines}, such as the document boundaries, edges of rectangle regions or tables,  and text lines. Enforcing straightness of feature lines in the rectified image can improve document rectification accuracy. In this work, we only consider the most frequent feature lines, i.e., the horizontal and vertical lines. Other geometric features, e.g., sloping lines and circles, can be handled analogously as long as they have analytical equations and can be detected reliably in the image.

Referring to Fig.~\ref{f:general_model}, a feature line ${\ell}^{r}$  in $\bm{I}^r$  is composed of a set of pixels $\bm{p}^{r} \in \ell^{r}$ in the pixel coordinates. Suppose corresponding feature lines in $\mc{M}$ and $\mc{M}^\prime$, denoted by $\ell$ and ${\ell}^{\prime}$ respectively,  are composed of data points $\bm{p}$ resp. $\bm{p}^{\prime}$.  To define the straight line constraint relating to ${\ell}^{r}$, we need to find the points $\bm{p}^{\prime} \in \ell^{\prime}$ in $\mc{M}^{\prime}$ relating to this feature line and make them close to a straight line.  We observe that $\bm{p}^r$ and $\bm{p}^{\prime}$ are related  through $\bm{p}$  by two mappings, i.e., $\bm{p}^r= f_x(\bm{p})$ and $\bm{p}^{\prime} = f_u(\bm{p})$. Therefore, the feature line constraints can be formulated by three classes of rules, two of which correspond to the two mappings, i.e., $f_x$ and $f_u$, with the third one straightening feature lines in $\mc{M}^{\prime}$. Since $f_u$ is the isometric mapping which has already been explained in Sec.~\ref{sec:isomapping}, we define the other constraints here, i.e., the \emph{projection constraint} relating to $f_x$ and the \emph{straightness constraint} for feature straightening.  In this treatment of feature line constraints, $\mc{M}$ and $\mc{M}^{\prime}$ serve as auxiliary variables in optimization to provide a fair solution space.

\paragraph{Projection constraint}
Refer to Fig.~\ref{f:general_model}.  Let  ${\ell}^{I}$ denote a feature line in $\chi$ composed of feature points $\bm{p}^I$. Suppose  the straight line $R$ passing through the camera centre $C$ and $\bm{p}^{I}$ is defined by $\bm{a}^{T}\bm{x} = 0$. Since $\bm{p}^{I}$ is the perspective projection of $\bm{p}$ onto the image plane, $\bm{p}$ must lie in $R$ which is expressed by Eq.~(\ref{e:projconst_p})

\begin{equation}
\label{e:projconst_p}
c_{\text {proj}}(\bm{p})= \bm{a}^{T}\bm{p} =0.
\end{equation}

 Suppose $\mc{L}^{r}$ is a set of feature lines  in $\bm{I}^r$. Let $\mc{L}$  and $\mc{L}^{\prime}$ be the corresponding sets in $\mc{M}$ and $\mc{M}^\prime$, respectively.  The \emph{projection constraint} with respect to all feature lines is defined by minimizing the following function 
 
\begin{equation}
\label{eqn:rayLine}
E_{\text {ray}, \mc{M}}(\mc{V}) = \frac{1}{H_2} \sum_{\ell \in \mc{L}} \sum_{\bm{p} \in \ell}{\Vert  c_{\text{proj}}(\bm{p}) \Vert ^2},
\end{equation}
where $H_2$ is the total number of points in $\mc{L}$, i.e., the number of data points in all feature lines.

\paragraph{Straightness constraint}
The points  $\bm{p}^{\prime} \in \ell^{\prime}$ of the feature line ${\ell}^{\prime}$ in $\mc{M}^\prime$ are required to lie in a straight line,  cf. Fig.~\ref{f:general_model}. Suppose the equation of the straight line is $\bm{b}^{T}\bm{x}=0$. The \emph{straightness constraint}  for the feature line $\ell^{\prime}$ is defined by 

\begin{equation}
c_{\text{straight}}(\ell^{\prime}) = \sum_{\bm{p}^{\prime} \in \ell^{\prime}}{\Vert \bm{b}^{T} \bm{p}^{\prime} \Vert ^2}=0.
\end{equation}

The straightness constraint involving all feature lines is defined by minimizing the following function
\begin{equation}
\label{eqn:straightLine}
E_{\text {line}, \mc{M}^\prime}(\mc{V}^{\prime},\mc{A}_{line}) = \frac{1}{H_2}\sum_{{\ell}^{\prime} \in \mc{L}^{\prime}}{c_{\text{straight}}(\ell^{\prime}) }.
\end{equation}

The fitting lines to the feature lines  are also variables in optimization and  $\mc{A}_{\text{line}}$ denotes the parameters in the equations of the fitting lines.

We emphasize that, in our setup, the feature line constraints are defined by minimizing $E_{iso}$, $E_{line,\mc{M}^{\prime}}$ and $E_{ray,\mc{M}}$ simultaneously, with $E_{iso}$ establishing a correspondence between a point in $\mc{M}$ and a point in $\mc{M}^{\prime}$. Suppose $\bm{p}$ is the point of a feature line in $\mc{M}$ which lies in a triangular region $t$ formed by three vertices $\bm{v}_1\bm{v}_2\bm{v}_3$ of a quad face of $\mc{M}$. $\bm{p}$ is expressed by the barycentric coordinates $\bm{\omega}$ in $t$, i.e., $\bm{p}=\begin{bmatrix}{\bm{v}_1}&{\bm{v}_2}&{\bm{v}_3}\end{bmatrix} \bm{\omega}$. Analogously, we have $t^{\prime}$, $\bm{p}^{\prime}$ and ${\bm{\omega}}^{\prime}$ defined  in the planar mesh $\mc{M}^{\prime}$.  Due to the isometric mapping  between $\mc{M}$ and $\mc{M}^{\prime}$, we have $\bm{\omega} = \bm{\omega}^{\prime}$ and  $\bm{p}^{\prime} = \begin{bmatrix}{\bm{v}^{\prime}_1}&{\bm{v}^{\prime}_2}&{\bm{v}^{\prime}_3}\end{bmatrix} {\bm{\omega}}^{\prime}$, cf. Fig.~\ref{f:iso_meshes}.

\subsection{Document rectification via optimization}
\label{sec:alg_overview}
The quad meshes $\mc{M}$ and ${\mc{M}}^{\prime}$ meeting the constraints in Sec.~\ref{sec:constraintSec} give naturally a rectified image of the document. Computing $\mc{M}$ and ${\mc{M}}^{\prime}$ is realized by solving an optimization problem. In each iteration of the optimization, both $\mc{M}$ and ${\mc{M}}^{\prime}$  are updated  simultaneously by  minimizing an objective function  $F$ as the combination of the constraints, which is defined by 

\begin{equation}
\begin{aligned}
F(\mc{X}) = &\lambda_1 E_{\text {iso}}+ \lambda_2 E_{\text {dist}, \mc{M}}+ \lambda_3 E_{\text{fair}, \mc{M}}+\\
& \lambda_4 E_{\text {fair}, {\mc{M}}^{\prime}}+ \lambda_5 E_{\text{line}, {\mc{M}}^{\prime}}+  \lambda_6 E_{\text {ray}, \mc{M}}.
\label{eq:objective_all}
\end{aligned}
\end{equation}

The vector of knowns is $\mc{X}=(\mc{V}, \mc{V}^{\prime}, \mc{A}_{\text{line}})$ where $\mc{V}$ and $\mc{V}^{\prime}$ are vertex positions of the meshes $\mc{M}$ and ${\mc{M}}^{\prime}$, respectively, and $\mc{A}_{\text{line}}$ are parameters of the equations of straight lines fitting to feature lines in ${\mc{M}}^{\prime}$. $\lambda_i, i=1, \ldots, 6$ are weights to function terms.

%% file: Algorithm.tex
\section{Algorithm}
\label{sec:alg}

As we have discussed, our method updates $\mc{M}$ and $\mc{M}^{\prime}$ iteratively. To provide a complete algorithm, accurate extraction of feature lines in the image is the most crucial issue which is also quite challenging due to the non-smoothness of feature lines in creased document images. Moreover, identifying noise points can also benefit image rectification by enhancing shape fidelity. To solve these problems,  we propose a progressive method in which feature line detection and noisy points processing are integrated into the iterative process of our algorithm, by referring to the active models $\mc{M}$ and ${\mc{M}}^{\prime}$ which get their quality improved throughout the iteration.

\subsection{Algorithm outline}
\begin{algorithm}[!htbp] \SetKwData{Left}{left}\SetKwData{This}{this}\SetKwData{Up}{up} \SetKwFunction{Union}{Union}\SetKwFunction{FindCompress}{FindCompress} \SetKwInOut{Input}{input}\SetKwInOut{Output}{output}
	\Input{Multiple images}
	\Output{A rectified image}
	 \BlankLine 
	  Employ SfM to obtain $\hat{\mc{P}}$\;
	  Choose reference ${\bm{I}}^r$ and extract boundary lines $\mc{L}_B^{r}$\;
	  Extract feature segments $\mc{L}^r_T$, $\mc{L}^r_E$ in ${\bm{I}}^r$\; 
	  Initialize $\mc{M}$ and ${\mc{M}}^{\prime}$\;
	  \While{quality of the result is not satisfied}
	  {
	   \Repeat{$|\frac{F^{n}-F^{n-1}}{F^{n-1}}|<\varepsilon$ or iteration number exceeds $Q$}
	    {
	      Perform noise processing to obtain $\mc{P}$, and update point projections and associated normal vectors\;
	      Compute feature segments in $\mc{M}$ and $\mc{M}^{\prime}$\; 
	      Generate feature lines ${\mc{L}}^{\prime} =\mc{L}_{F}^{\prime} \cup  \mc{L}_{B}^{\prime}$\; 
	      Computing fitting lines to feature lines in $\mc{L}^{\prime}$ and perform feature line projection\; 		 
	      Solve $F \rightarrow$~min and update $\mc{M}$ and ${\mc{M}}^{\prime}$\;
        }
		Subdivide $\mc{M}$ and ${\mc{M}}^{\prime}$\; 	
	    Update weights $\lambda_5$ and $\lambda_6$ in Eq.~(\ref{eq:objective_all}).
     }
	 Generate a rectified image ${\bm{I}}^{\star}$\;
	\textbf{return} ${\bm{I}}^\star.$ 
\caption{Document image rectification}
\label{a:algo} 
\end{algorithm}
The pseudocode of the whole algorithm is given in  Algorithm~\ref{a:algo} in which the minimization of $F$  and the update of $\mc{M}$ and ${\mc{M}}^{\prime}$ make up a nested loop.
The main task in the inner loop is solving the minimization problem defined in Eq.~(\ref{eq:objective_all}) and updating $\mc{M}$, ${\mc{M}}^{\prime}$ and the straight  fitting  lines in ${\mc{M}}^{\prime}$. In each inner-loop, the  footpoint $\bm{x}^{\pi}$ for each valid point and the corresponding normal vector $\bm{N}$ in Eq.~(\ref{c:dis}) are re-computed. The operation of feature line projection is performed to step over local minimums. The inner loop stops when a convergence condition is met or the maximum number of iterations is reached (the parameters in the algorithm are set to $\varepsilon=0.01$ and $Q=100$). Then both $\mc{M}$ and ${\mc{M}}^{\prime}$ are subdivided using the Catmull-Clark subdivision method to provide more degrees of freedom when necessary, and another round of optimization is repeated until the rectified image is satisfactory. Three rounds of subdivision are sufficient for all examples in the experiments. The rectified image is obtained based on the unfolded mesh ${\mc{M}}^{\prime}$ by performing texture mapping with $\bm{I}^{r}$ being the texture image. The texture coordinates of the vertices of $\mc{M}^{\prime}$ are naturally obtained using the relations between $\bm{I}^{r}$, $\mc{M}$ and $\mc{M}^{\prime}$.

\subsection{Algorithm details}
\label{sec:alg_details}

\subsubsection{Reference image and document boundary}
\label{sec:refImage}
An image ${\bm{I}}^r$  among the multi-view images is selected as the reference image for mesh initialization and texture mapping.  We also use ${\bm{I}}^r$  as the input to deep learning-based methods in the experiments.  Any image can be used as a reference, but missing parts of the document caused by self-occlusion are not desired. Moreover, although our method can work without any document boundary information, a complete boundary  in the reference image is preferred. In the experiments, we choose the reference image manually.

The boundary lines of the document region can be extracted from background segmentation, and are required to become straight lines in the rectified image, serving as feature line constraints.
Moreover, data points and feature lines outside the document region are regarded as noise, and removing them can avoid their negative impact on rectification. For most examples in our experiments, the background in the images is of moderate complexity, and the traditional threshold-based segmentation method can be used~\cite{he2013book}. However, a thorough investigation of background segmentation is beyond the scope of the present work.

\subsubsection{Initialization}
\label{sec:initialM}

\begin{figure}[!t]
	\centering
	\subfigure[]{
	\includegraphics[width=.14\textwidth]{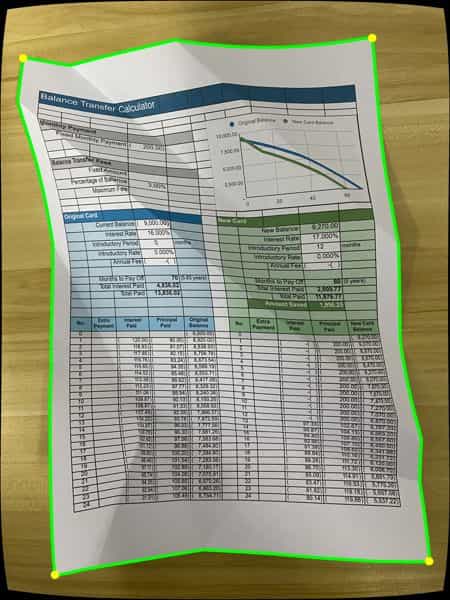}
		\label{f:Initial_0}
	}\hspace{.01\textwidth}%
		\subfigure[]{
	\includegraphics[width=.14\textwidth]{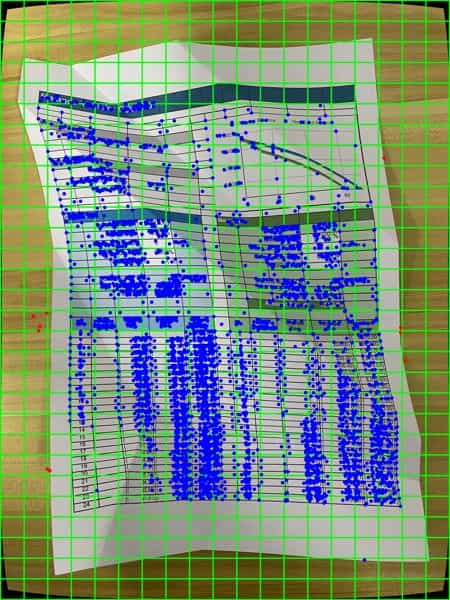}
		\label{f:Initial_1}
	}\hspace{.01\textwidth}%
    	\subfigure[]{
	\includegraphics[width=.14\textwidth]{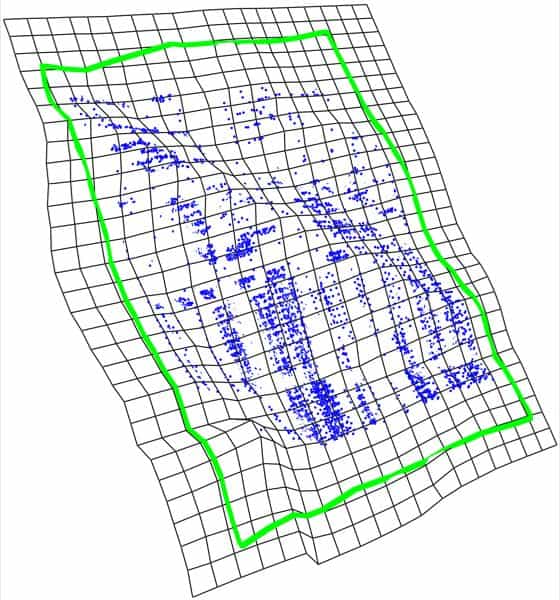}
		\label{f:Initial_2}
	}
    	\subfigure[]{
    		\includegraphics[width=.14\textwidth]{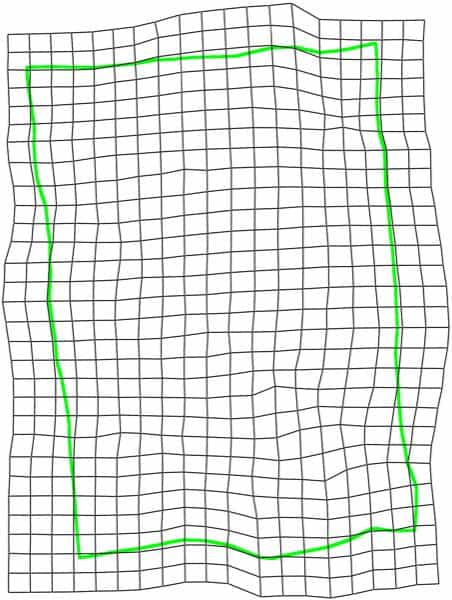}
		\label{f:Initial_3}    
		}\hspace{.01\textwidth}%
		    	\subfigure[]{
    		\includegraphics[width=.14\textwidth]{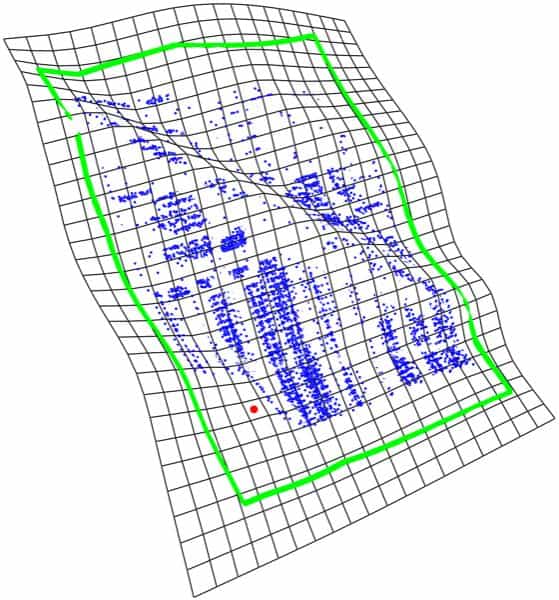}
		\label{f:Initial_4}    
		}\hspace{.01\textwidth}%
		    	\subfigure[]{
    		\includegraphics[width=.14\textwidth]{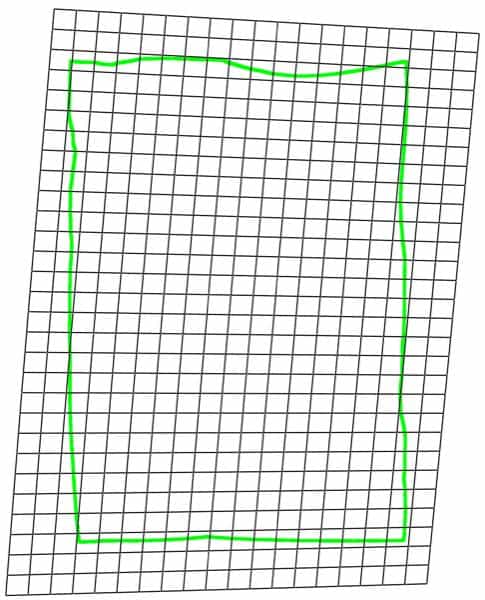}
		\label{f:Initial_5}    
		}
  \vspace{-.015\textwidth}%
	\caption{Quad mesh initialization. (a) The reference image ${\bm{I}}^r$ and extracted boundary curves. (b) A quad mesh $\mc{M}^{r}$ by uniform sampling in ${\bm{I}}^r$. The data points outside the document regions are rendered in red. (c) Initial 3D quad mesh $\mc{M}$ with data points. (d) Planar quad mesh ${{\mc{M}}^{\prime}}$ unfolded from the 3D mesh in (c). (e) Initial 3D quad mesh $\mc{M}$ after refinement. (f) Initial planar mesh ${\mc{M}}^{\prime}$ after refinement.}
	\label{f:quadmesh_initialization}
\end{figure}

Our iterative algorithm  needs initialization of $\mc{M}$ and ${\mc{M}}^{\prime}$. See Fig.~\ref{f:quadmesh_initialization} for the process of quad mesh initialization which consists of the following steps. 

Step 1: Image mesh generation. We start with generating a 2D quad mesh $\mc{M}^{r}$ in the reference image ${\bm{I}}^r$ with uniformly distributed edges. See Fig.~\ref{f:Initial_1} for an example of  $\mc{M}^{r}$. 
Let the positions of mesh vertices be denoted by $\bm{v}_i^r$,  $i=1, \ldots ,N_1 \times N_2$,  with $N_1$, $N_2$ being the dimension of the mesh. When the document boundary is known,  an alternative manner for creating $\mc{M}^{r}$ is generating a Coons surface bounded by the document boundary and sampling iso-parametric lines for mesh edges~\cite{brown2006geometric}. 

Step 2: Initialization of $\mc{M}$. $\mc{M}$ has the same combinatorial structure as $\mc{M}^{r}$. The position of its vertices $\bm{v}_i=(x_i, y_i, z_i)^{\mathsf{T}}$ are computed as follows. Let the point in ${\bm{I}}^r$ corresponding to $\bm{v}_i$ be denoted as $\bm{v}_i^r=(u_i, v_i)^{\mathsf{T}}$ in pixel coordinates. We first find the $k$ closest  ($k=3$ in our implementation) points to $\bm{v}_i^r$ in $\bm{I}^r$, denoted by $\{\bm{x}^r_s,s=1,...,k\}$,   among the points detected by SfM whose depths are already known. Then, $z_i$ is computed by $z_i =\frac{1}{\sum_{s=1}^{k}{\phi(r_s)}}\sum_{s=1}^{k}{{z}_s \phi(r_s)}$ where $r_s = \Vert \bm{x}_s^r - \bm{v}_i^r \Vert$, and $\phi(r)=\frac{1}{1+r^2}$ is a radial basis function.  Then $x_i$ and $y_i$ follow directly from the pinhole camera model, which are $x_i=\frac{(u_i-c_{u})\cdot z_i}{f\cdot k_{u}}$ and
$y_i=\frac{(v_i-c_{v})\cdot z_i}{f \cdot k_{v}}$, respectively. 

Step 3: Initialization of ${\mc{M}}^{\prime}$.  $\mc{M}$ is unfolded into the plane to obtain ${{\mc{M}}^{\prime}}$, see Fig.~\ref{f:Initial_3}.  Our optimization is not sensitive to the unfolding method, and we use the boundary-first-flattening (BFF) method~\cite{sawhney2018boundary} in our implementation.

Step 4: Refinement of $\mc{M}$ and ${\mc{M}}^{\prime}$. $\mc{M}$ and ${\mc{M}}^{\prime}$ are refined by performing a surface fitting operation with respect to the target points.  This procedure iteratively minimizes the function Eq.~(\ref{eq:objective_all}) with $\lambda_5=0$ and $\lambda_6=0$, updating the footpoints of target points and normal vectors. See Fig.~\ref{f:Initial_4} and Fig.~\ref{f:Initial_5}  for the refined $\mc{M}$ and ${\mc{M}}^{\prime}$, respectively. ${\mc{M}}^{\prime}$ is put upright by adjusting the Oriented Bounding Box (OBB) of the document boundaries. If the boundary is incomplete, we can adjust the OBB of the entire planar mesh $\mc{M}^\prime$ instead.

\subsubsection{Noise processing}
\label{sec:outlier}
\begin{figure}[!t]
\centering
    \subfigure[]{\centering
    \begin{minipage}[b]{.19\textwidth}\centering
    \includegraphics[angle=25,width=1\textwidth]{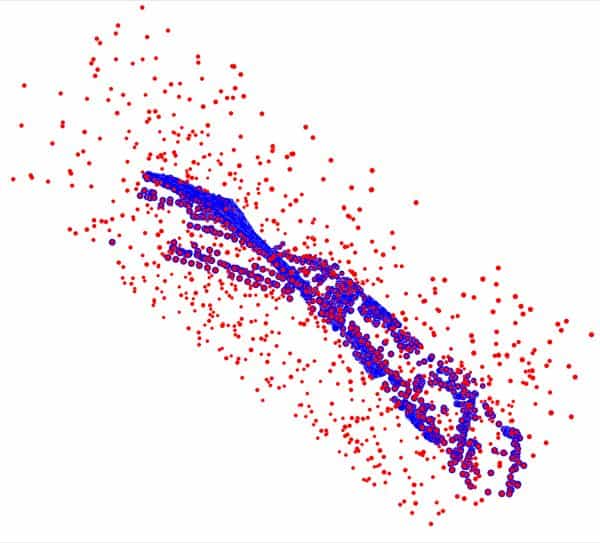}
         \vspace{-0.3\textwidth}\\
    \includegraphics[angle=15,width=1\textwidth]{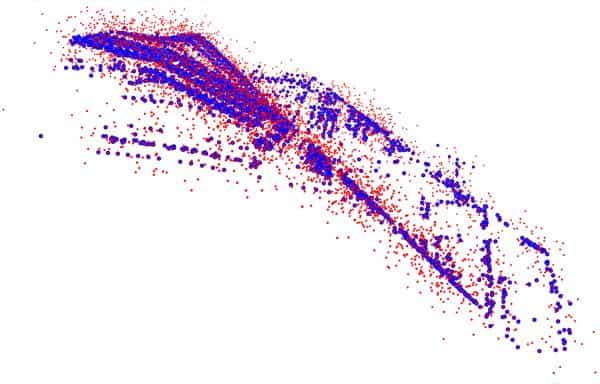}
        \vspace{-0.4\textwidth}\\
        \label{f:noise0} 
    \end{minipage}}\hspace{.005\textwidth}%
    \subfigure[]{\centering
   \begin{minipage}[b]{.19\textwidth}\centering
        \includegraphics[width=1\textwidth]{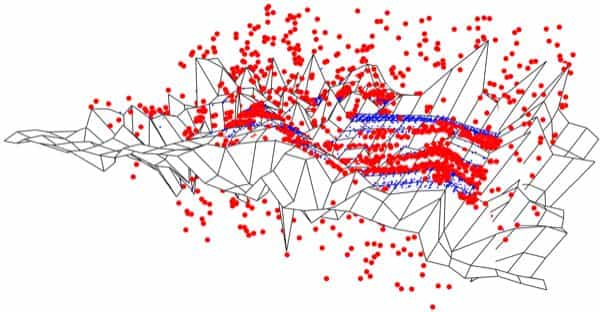} \vspace{.15\textwidth}\\
        \includegraphics[width=1\textwidth]{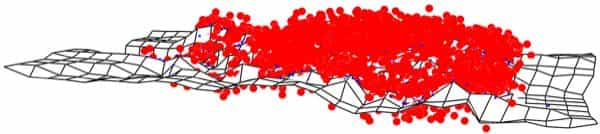} 
        \vspace{-.1\textwidth}\\
        \label{f:noise1}
    \end{minipage}}\hspace{.005\textwidth}%
    \subfigure[]{\centering
    \begin{minipage}[b]{.19\textwidth}\centering
         \includegraphics[width=1\textwidth]{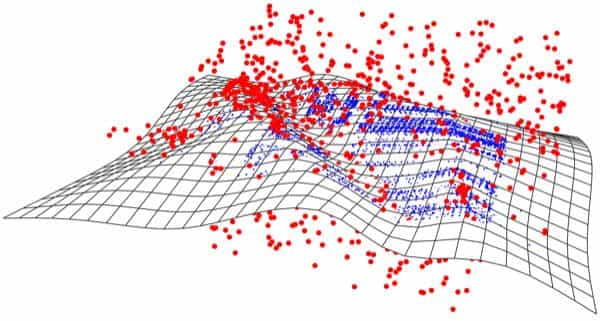} \vspace{.15\textwidth}\\
        \includegraphics[width=1\textwidth]{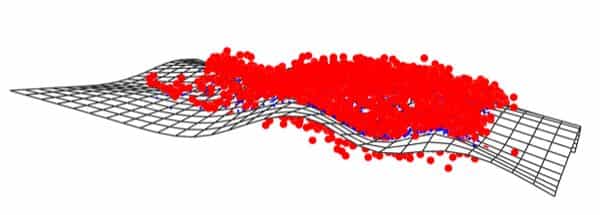} 
        \vspace{-.2\textwidth}\\  
        \label{f:noise2}
    \end{minipage}}\hspace{.005\textwidth}%
    \subfigure[]{\centering
    \begin{minipage}[b]{.19\textwidth}\centering
         \includegraphics[width=1\textwidth]{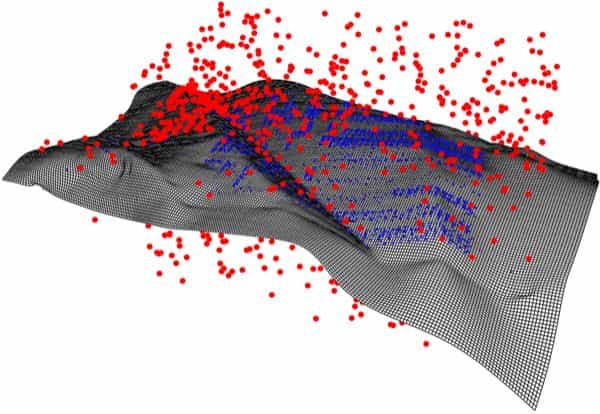} \vspace{.05\textwidth}\\
        \includegraphics[width=1\textwidth]{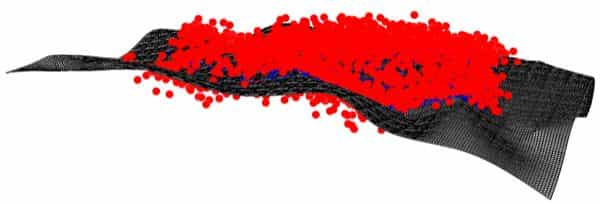} 
        \vspace{-.2\textwidth}\\
          \label{f:noise3}
    \end{minipage}}\hspace{.02\textwidth}%
    \subfigure[]{\centering
    \begin{minipage}[b]{.19\textwidth}\centering
    \includegraphics[angle=90,width=.8\textwidth]{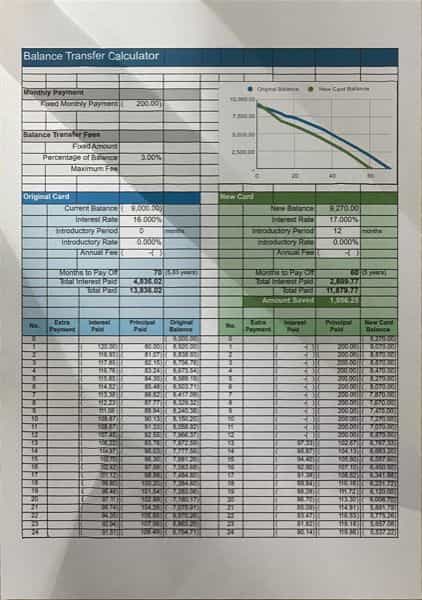} \vspace{.05\textwidth}\\
    \includegraphics[angle=90,width=.8\textwidth]{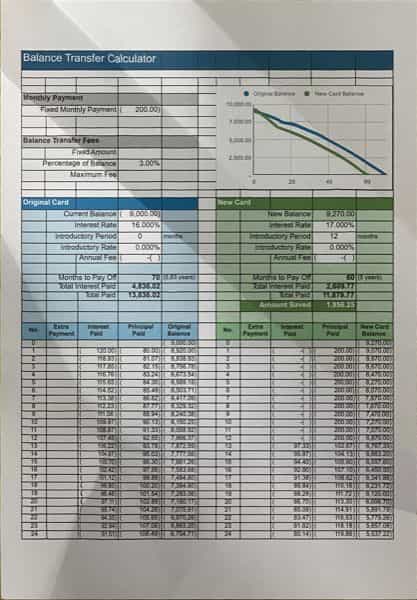} 
    \vspace{-.05\textwidth}\\
        \label{f:noise4}
    \end{minipage}}
\begin{minipage}{1\textwidth}
\vspace{-0.02\textwidth} 
\caption{Progressive noise point filtering. The first row shows the experiment with $\hat{\mc{P}}$ and one thousand outliers. The second row shows an experiment with Gaussian noise. In columns (b), (c), and (d), the mesh is rendered with the data points where the valid points are rendered in blue and the identified noise points in red. (a) Original points (in blue) and added noise points (in red). (b) Initial mesh $\mc{M}$ before refinement. (c) Initial mesh $\mc{M}$ after refinement. (d) The obtained mesh $\mc{M}$ by our rectification algorithm. (e) The rectified image rotated by 90 degrees.
}  
\label{f:noise}
\end{minipage}
\end{figure}

The 3D points obtained by the SfM method may contain noise or outliers that should be ignored in model approximation. Our approach uses the active mesh surface  $\mc{M}$  as a reference, and identifies a point as noisy when its distance to $\mc{M}$ is larger than a predefined tolerance $\varphi$ (typically $\varphi = 0.06$ is used for a normalized model).  This simple strategy is combined with our iterative algorithm, which progressively refines the surface model $\mc{M}$. As a result, the quality of noise point identification also gets improved gradually. The data set after noise removal is denoted by $\mc{P}$, which is updated in each round since $\mc{M}$ is updated.

To demonstrate the performance of our noise processing method, we perform two experiments using the same data as in Fig.~\ref{f:quadmesh_initialization}.  In the first experiment, we add one thousand random points as outliers inside the axis-aligned bounding box (AABB) of $\hat{\mc{P}}$. In the second experiment, we produce Gaussian noise around each point in $\hat{\mc{P}}$. The mediate and final results are shown in Fig.~\ref{f:noise}. Compared to the result of data 2 in Fig.~\ref{f:results1} which uses the original data without artificial noise,  the difference in the rectified images is hardly noticeable.

\subsubsection{Feature line extraction} 
\label{sec:featureExtraction}

\begin{figure}[!b]
	\centering
	\subfigure[]{
\includegraphics[width=0.2\textwidth]{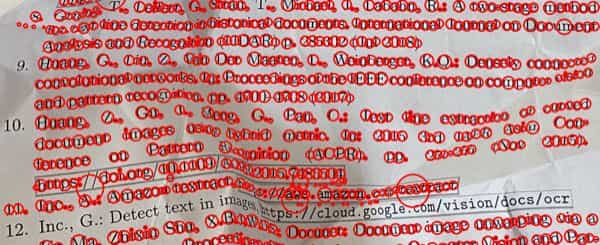}
		\label{f:textline_0}
	}
	\subfigure[]{
   \includegraphics[width=0.2\textwidth]{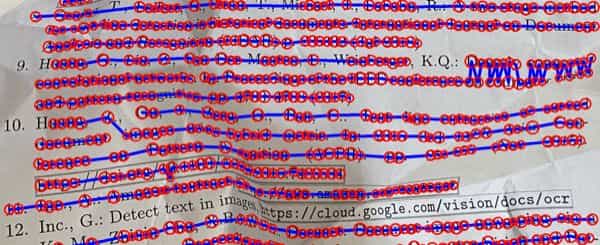}
		\label{f:textline_1}
    	}
    \subfigure[]{
   \includegraphics[width=0.16\textwidth]{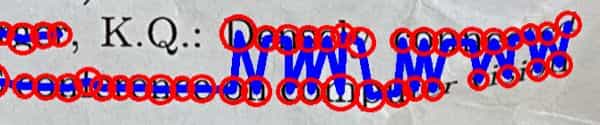}
		\label{f:textline_2}
    	}
    \subfigure[]{
    \includegraphics[width=0.16\textwidth]{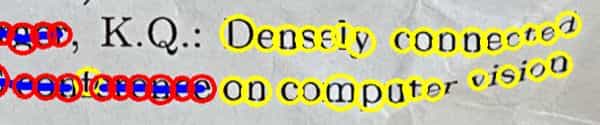}
		\label{f:textline_3}
    	}
    \subfigure[]{
    \includegraphics[width=0.16\textwidth]{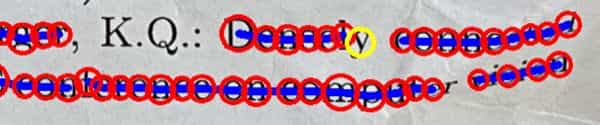}
		\label{f:textline_4}
    	}
\vspace{-.02\textwidth}
	\caption{Feature segment extraction for text lines. (a) Detection results by the method in~\cite{koo2010state}. (b) Text lines by connecting the remained circle centers of characters. (c) A part of the original text lines that containing the wrong connections. (d) Disconnection of text lines after removing wrong edges. Isolated circle centers are shown in yellow. (e) Merging isolated circle centers into text feature segments.}
	\label{f:textline}
\end{figure}
\begin{figure}[!t]
\centering
    \subfigure[]{\centering
    \begin{minipage}[b]{.14\textwidth}\centering
        \includegraphics[width=1\textwidth, height=1.333333\textwidth]{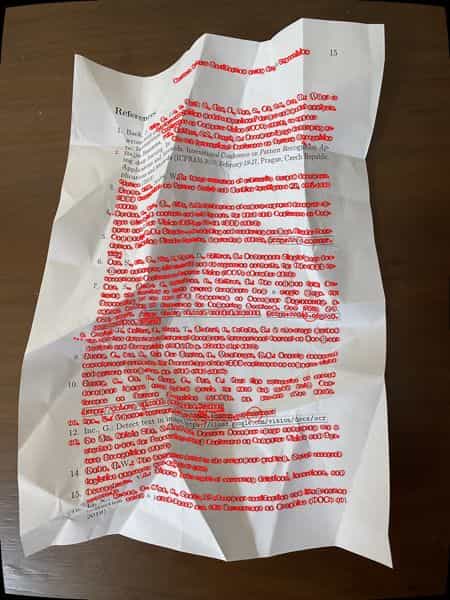}
         \vspace{-.15\textwidth}\\
        \includegraphics[width=1\textwidth, height=1.333333\textwidth]{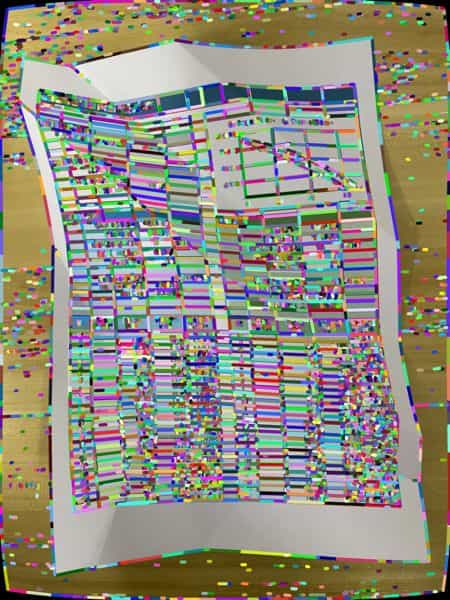}
        \vspace{-.20\textwidth}\\
        \label{f:cluster_0} 
    \end{minipage}}\hspace{.05\textwidth}%
    \subfigure[]{\centering
   \begin{minipage}[b]{.14\textwidth}\centering
        \includegraphics[width=1\textwidth, height=1.333333\textwidth]{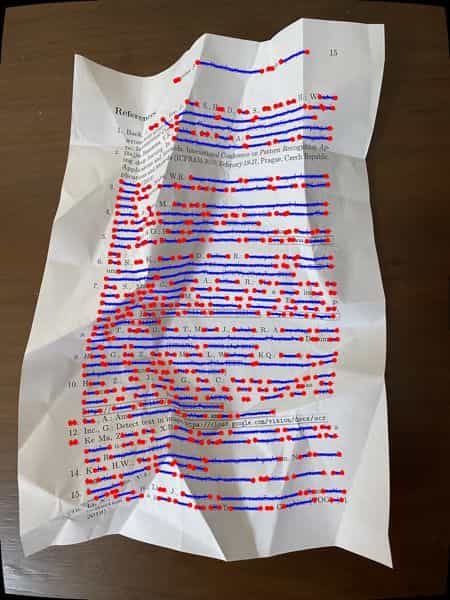} \vspace{-.15\textwidth}\\
        \includegraphics[width=1\textwidth, height=1.333333\textwidth]{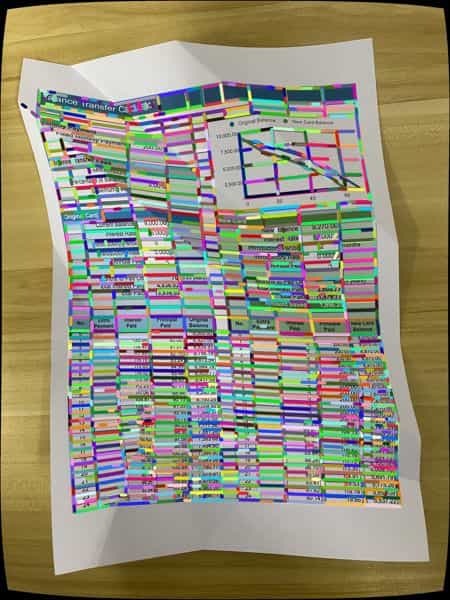} 
        \vspace{-.20\textwidth}\\
        \label{f:cluster_1}
    \end{minipage}}\hspace{.05\textwidth}%
    \subfigure[]{\centering
    \begin{minipage}[b]{.14\textwidth}\centering
        \includegraphics[width=1\textwidth, height=1.333333\textwidth]{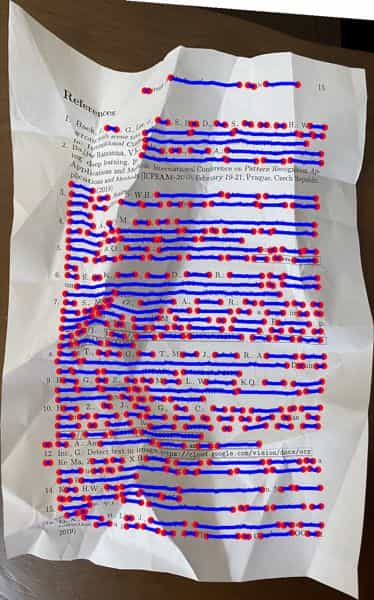} \vspace{-.15\textwidth}\\
        \includegraphics[width=1\textwidth, height=1.333333\textwidth]{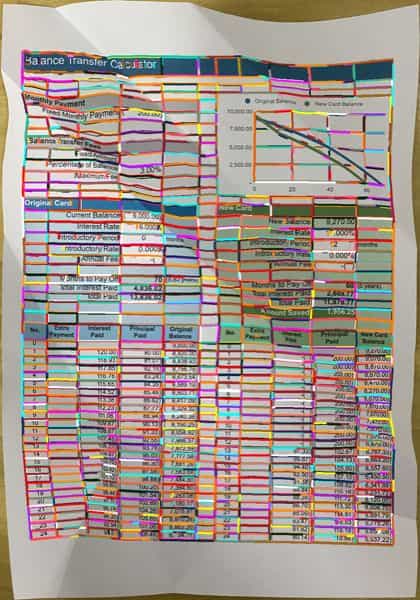}  
        \vspace{-.20\textwidth}\\
        \label{f:cluster_2}
    \end{minipage}}\hspace{.05\textwidth}%
    \subfigure[]{\centering
    \begin{minipage}[b]{.14\textwidth}\centering
        \includegraphics[width=1\textwidth, height=1.333333\textwidth]{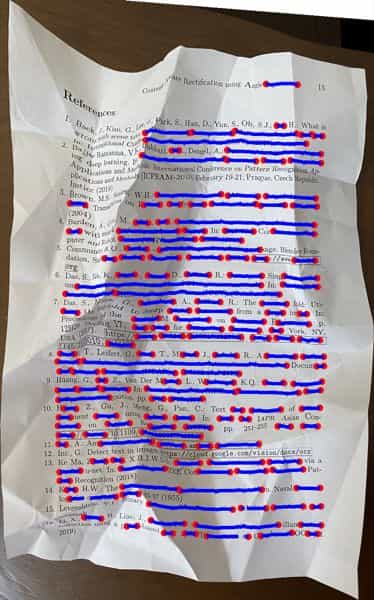} \vspace{-.15\textwidth}\\
        \includegraphics[width=1\textwidth, height=1.333333\textwidth]{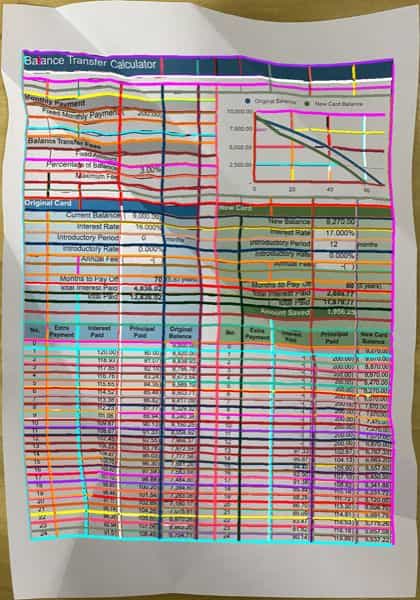} 
        \vspace{-.20\textwidth}\\
        \label{f:cluster_3}
    \end{minipage}}\hspace{.05\textwidth}%
    \subfigure[]{\centering
    \begin{minipage}[b]{.14\textwidth}\centering
        \includegraphics[width=1\textwidth, height=1.333333\textwidth]{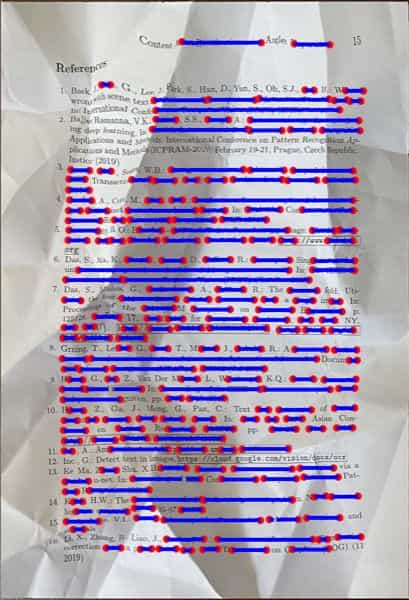} \vspace{-.15\textwidth}\\
        \includegraphics[width=1\textwidth, height=1.333333\textwidth]{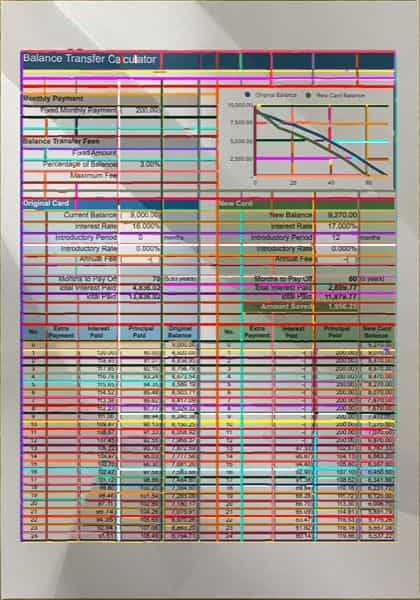} 
        \vspace{-.20\textwidth}\\
        \label{f:cluster_4}
    \end{minipage}}
    \vspace{-.02\textwidth}\\
\caption{Feature line generation. The top row shows an example of text line extraction. From left to right: detection results by the method in~\cite{koo2010state}, text segments $\mc{L}^{r}_T$,  text segments in ${\mc{M}}^{\prime}$, text lines after the first iteration of segment merging, text lines in the final result.  The bottom row shows an example of edge line detection. From left to right: edge segments detected by the LSD method in ~\citep{gromponevongioi2012lsd}, edge segments $\mc{L}^{r}_E$,  edge segments in ${\mc{M}}^{\prime}$, edge lines after the first iteration of segment merging, edge lines in the final result.}
	\label{f:cluster}
\end{figure}

Existing feature line extraction methods that work well for flat or smoothly curved documents often fail for documents with considerable creasing or folding deformations. We observe that it is relatively easier to detect short line segments, namely \emph{feature segments}, by applying the existing line detection method to ${\bm{I}}^r$ locally. Therefore, based on the fact the feature segments coming from the same feature line approximate one identical straight line in the unfolded mesh, we propose to form feature lines by merging feature segments in ${\mc{M}}^{\prime}$. Since  ${\mc{M}}^{\prime}$  is updated throughout the iteration of our optimization, the accuracy of feature line generation is also enhanced progressively. We consider two kinds of feature lines in this work, i.e.,  the \emph{text lines}, which are lines following texts, and \emph{edge lines} such as lines in tables or boundaries of box regions. As already explained, the boundary lines of the document region, denoted by $\mc{L}^{r}_{B}$, are extracted using traditional threshold-based segmentation methods.

\paragraph{Feature segment detection} For detecting feature segments for text lines, denoted by $\mc{L}^r_T$, we apply the approach in~\cite{koo2010state} and remove obviously misidentified characters to give lines connecting the character centers. A disconnection operation is applied to remove the faulty connections caused by the non-smooth deformation of the document, resulting in a set of locally reliable feature segments. A faulty connection edge forms a large angle with adjacent edges or has a large length, regarding some predefined tolerances. Isolated character centers are merged into nearby segments or combined to form new segments when they are close enough and their connection does not produces faulty connections. See Fig.~\ref{f:textline} and Fig. \ref{f:cluster_1} for examples of  generating text feature segments.

For detecting feature segments for edge lines, denoted by $\mc{L}^r_E$, the Line Segment Detector (LSD) method~\citep{gromponevongioi2012lsd} is applied to ${\bm{I}}^r$, which gives a set of short segments which are further cleaned by removing tiny ones and those lying outside of the document region.  See Fig.~\ref{f:cluster_0} and Fig.~\ref{f:cluster_1} for an example.

\paragraph{Feature line generation}  Each feature segment in $\bm{I}^r$ has a corresponding segment in the flat mesh $\mc{M}^{\prime}$. Feature line generation is done by merging the cluster of feature segments in $\mc{M}^{\prime}$ which are close to the same straight line. Since both $\mc{M}$ and $\mc{M}^{\prime}$ are updated iteratively in our method, we need to compute the feature lines in $\mc{M}^{\prime}$ in each iteration as follows. For a point $\bm{p}^{r}$ in $\bm{I}^{r}$, the intersection of $\mc{M}$ and the viewing ray corresponding to $\bm{p}^{r}$ is computed, which is denoted by $\bm{p}$. Then the corresponding point $\bm{p}^{\prime}$ in $\mc{M}^{\prime}$ is obtained by preserving the barycentric coordinates of $\bm{p}$ in the corresponding mesh face.

Feature line generation works as follows. Let $\mc{L}^{\prime}_{F}$ denote the feature lines in $\mc{M}^{\prime}$, which is initially the set of all feature segments in  ${\mc{M}}^{\prime}$.  The merging operation iterates over each pair of feature lines  $({\ell}^{\prime}_1,{\ell}^{\prime}_2)$ in $\mc{L}^{\prime}_{F}$ and merge them if their two closest endpoints to each other lie in a  tolerance box and the merging result preserve straightness. In our implementation, the KD-Tree structure is used for the closest point computation. The straightness of a line is defined by $h/w$, with $h$ and $w$ being the height and width of the OBB of the line, respectively. In the experiments, the straightness tolerance is set to $0.1$.  See Fig.~\ref{f:cluster} for some feature lines formed by segment merging where the lines not closely horizontal or vertical are also removed. The merging operation is applied to the updated set  $\mc{L}^{\prime}_{F}$ repeatedly until no merging happens. The final $\mc{L}^{\prime}_{F}$ is the set of all text feature lines and edge feature lines. 

$\mc{L}^{\prime}=\mc{L}^\prime_B \cup \mc{L}^\prime_F$ are all feature lines  in ${\mc{M}}^{\prime}$ for constraints in optimization, where  $\mc{L}^{\prime}_B$ denotes corresponding boundary lines of the document region in $\mc{M}^{\prime}$.  See  Fig.~\ref{f:cluster_4} for the final feature lines of an example.

\paragraph{Remarks:} There are several tolerance values in our feature line detection method, and it is hard to find the optimal values for them to obtain all correct feature lines without introducing any wrong feature lines. Therefore, instead of trying to find all the correct lines, we choose the values such that the wrong feature lines are excluded to avoid their negative impact on the results. Our image rectification method works well  with only a few feature lines. Feature line detection in a creased document image is very challenging, and more advanced methods can be directly used in our framework.

\subsubsection{Feature line projection}
\label{sec:featurelineUpdate}

\begin{figure}[!t]
\centering
\includegraphics[width=.7\textwidth]{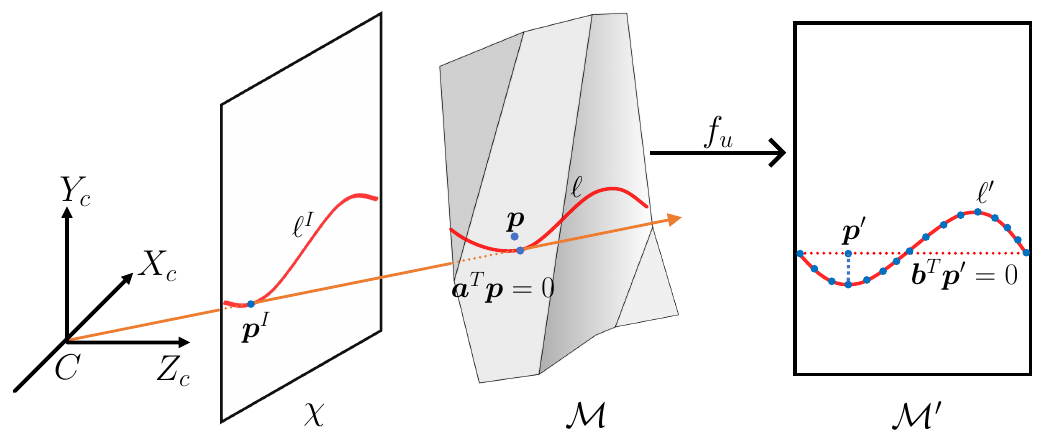}
\vspace{-.02\textwidth}
\caption{Feature line projection. The feature points in both $\mc{M}^{\prime}$ and $\mc{M}$ are modified. }\label{f:rayline}
\end{figure}
\begin{figure}[!b]
	\centering
	\subfigure[]{
		\centering
\includegraphics[width=.15\textwidth,height=.2115\textwidth]{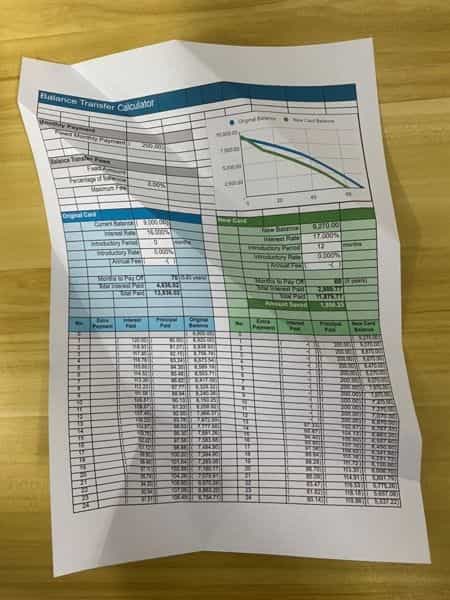}
		\label{f:update_o}
	}\hspace{.04\textwidth}%
	\subfigure[]{
		\centering
			\includegraphics[width=.15\textwidth,height=.2115\textwidth]{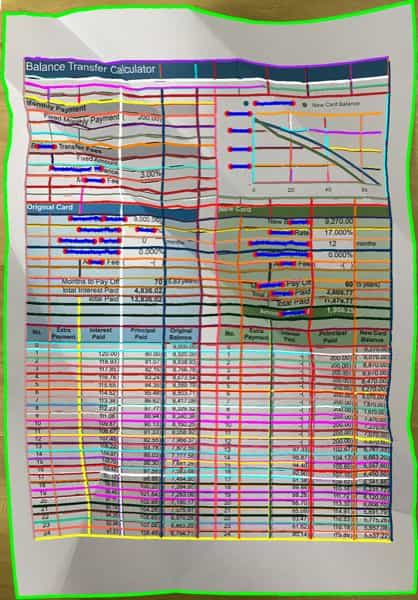}
		\label{f:update_0}
	}\hspace{.04\textwidth}%
    	\subfigure[]{
    		\centering
   		 	\includegraphics[width=.15\textwidth,height=.2115\textwidth]{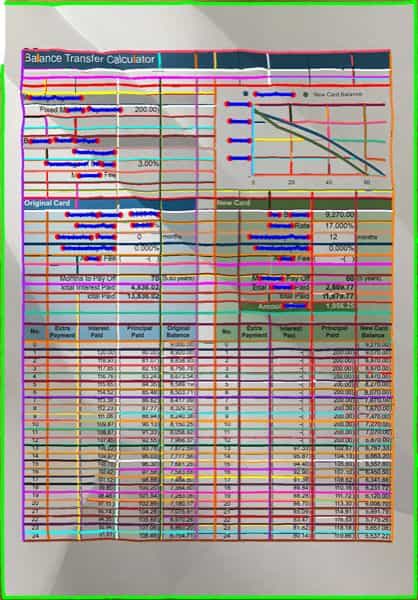}
		\label{f:update_1}
    	}\hspace{.04\textwidth}%
    \subfigure[]{
   		 	\includegraphics[width=0.15\textwidth,height=.2115\textwidth]{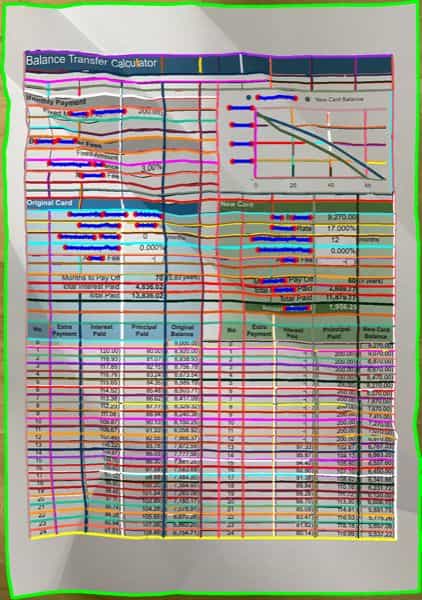}
		\label{f:update_2}
    	}\hspace{.04\textwidth}%
    \subfigure[]{  		\includegraphics[width=0.15\textwidth,height=.2115\textwidth]{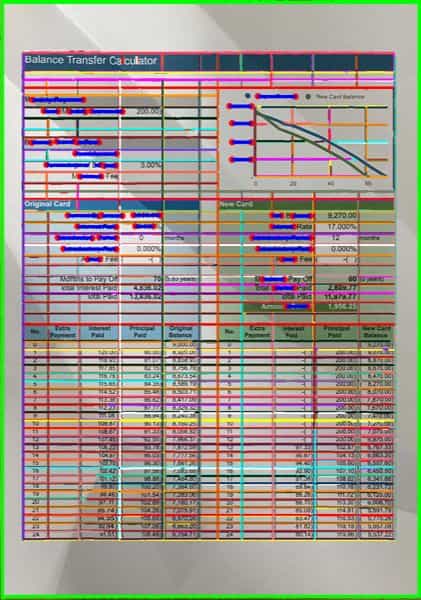}
		\label{f:update_3}
    	}
    	\vspace{-0.02\textwidth} 
	\caption{Experiment of feature line projection. (a) Original image. (b) Rectified image by our method  without feature line projection, no subdivision. (c) Results without the feature line projection strategy, three  rounds of subdivisions are applied. (d) Results with the feature line projection strategy, no subdivision. (e) Results with the feature line projection strategy and three rounds of subdivision. }
	\label{f:update}
\end{figure}

The straightness constraint defined in Eq.~(\ref{eqn:straightLine}) relating to each feature line involves a varying straight line fitting to each feature line ${\ell}^{\prime}$ in ${\mc{M}}^{\prime}$.  This straight line is initialized  with the least squared method.

Recall that the points in $\mc{M}$ are computed by ray-model intersection. Therefore, the constraint in Eq.~(\ref{eqn:rayLine}) is strictly satisfied and the optimization decreases the value of the function in Eq.~(\ref{eqn:straightLine}). However, we notice that this treatment often leads to unsatisfactory local minimums, see Fig.~\ref{f:update} for an example.  We believe this is caused by the  fact that a developable surface only has a unique isometric planar state, and the planar mesh $\mc{M}^{\prime}$ is not flexible enough due to its low dimension. 

Our method to deal with this issue is based on the observation that the 3D mesh $\mc{M}$ is much more flexible than $\mc{M}^{\prime}$ since it exists in a higher dimensional space and there are infinite 3D surfaces isometric to a planar mesh. We propose an operation called \emph{feature line projection} to step out of a local minimum by updating the feature points. Refer to Fig.~\ref{f:rayline} for an illustration. This operation moves the feature points $\bm{p}^{\prime}$ in ${\mc{M}}^{\prime}$ to corresponding projection points at the fitting straight line $\bm{b}^{T}\bm{x}=0$. Under isometric mapping, the face and the corresponding barycentric coordinates associated with the updated points are preserved. Therefore, the feature points $\bm{p}$ in $\mc{M}$ are updated accordingly for which the projection constraint $\bm{a}^{T}\bm{p}=0$ applies. The straightness constraint of Eq.~(\ref{eqn:straightLine}) is strictly satisfied by employing the feature line projection. The 3D mesh $\mc{M}$ is optimized due to the non-zero gradients of the objective function in Eq.~(\ref{eqn:rayLine}). Experiments show that this can result in a better result than without feature line projection.  Refer to Fig.~\ref{f:update} for an experiment showing the performance of the feature line projection strategy.

\subsubsection{Image generation of rectified document}
\label{sec:imageGeneration}
The last step of our method is to generate an image of the rectified document ${\bm{I}}^{\star}$  based on the planar mesh model ${\mc{M}}^{\prime}$. This is achieved by employing an operation of texture mapping to ${\mc{M}}^{\prime}$ with the reference image $\bm{I}^r$ being the texture image. The texture coordinates of each vertex $\bm{v}^{\prime} \in {\mc{V}}^{\prime}$ are directly obtained from the planar mesh ${\mc{M}}^{\prime}$. The color of $\bm{v}^{\prime}$ is the linear interpolation of the colors of the neighboring pixels of $f_{x} \circ f_u^{-1}(\bm{v}^{\prime})$ in $\bm{I}^r$.

The rectangular region enclosed by the straight lines fitting to the boundaries $\mc{L}^\prime_B$ is rendered as the resulting image to exclude useless background information. If the boundary information is incomplete and there is no region enclosed by the boundary lines,  an AABB of all feature lines in $\mc{M}^\prime$ is used as the output region.

\begin{figure}[!t]
\centering
\includegraphics[width=.8\textwidth]{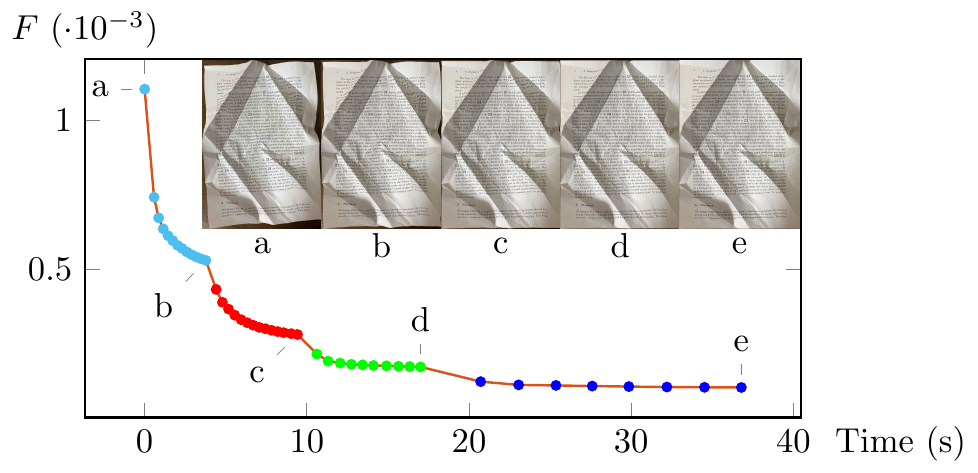}
 \vspace{-.02\textwidth}
\caption{Convergence behavior of Algorithm~\ref{a:algo}, which shows the variation of the value of the objective function $F$. The horizontal line indicates the computation time in seconds. Each dot corresponds to one iteration of the inner loop and the color of the dot is used to distinguish different rounds of optimization. The input image and rectified images associated with each round of optimization are also rendered.}
\label{f:subdivison}
\end{figure}

To demonstrate the convergence behavior of the algorithm, we show in Fig.~\ref{f:subdivison} by one example the variation of the value of the objective function $F$ through iterations where three rounds of the subdivision are conducted. We can see that the objective function converges in a small number of iterations (around 10) in each round of optimization. The subdivision operation, followed by another round of optimization, yields a further decrease in the objective function and improved quality of the rectified image.

%% file: Experiments.tex
\section{Experiments}

\subsection{Implementation details}
We have implemented the proposed image rectification algorithm in C++, on an Intel Core i9-9900KF 3.6 GHz Windows PC with 32GB RAM. The SfM implementation in the OpenMVG library ~\cite{moulon2017openmvg} is used to generate sparse point clouds from the multi-view images. The LBFGS method is applied to solve the minimization problem of Eq.~(\ref{eq:objective_all}). No parallel processing or other acceleration techniques are applied. 

Extensive experiments indicate the robustness of our method to the choice of $\lambda_1$,$\lambda_2$, $\lambda_4$, $\lambda_5$ and $\lambda_6$. We set $\lambda_1 = 1$, $\lambda_2= 1$, $\lambda_4=0.1$ in experiments and no perceptible difference is observed in the results if these values range in $[0.1, 1]$. The weights corresponding to feature line constraints are initially set $\lambda_5=1$ and $\lambda_6=1$ and multiplied by 4 after each subdivision step. The choice of $\lambda_3$ leads to noticeable variation in the results in the way that a larger value of it results in a fairer shape with increased approximation error to the target data points. In practice, we use a tiny value $\lambda_4 =1\cdot 10^{-4}$ to avoid over-smooth of the creases in the 3D mesh $\mc{M}$.  The dimensions  of $\mc{M}$ and $M^{\prime}$  are initially set to $20 \times 30$. For each example, we perform four rounds of optimization, interleaved with three subdivisions, which provide sufficient degrees of freedom. The parameters in feature line generation are selected empirically. More elegant methods for extracting feature lines in the image of creased documents can be directly used for replacing current feature line detection methods used in our framework.

\subsection{Comparisons with existing methods}
Traditional methods for document image rectification often assumed smoothly curved shapes and are essentially not suitable for creased or folded documents. For deformed documents with creases and folds, the multi-view methods and deep learning methods perform much better than traditional methods, which have been demonstrated in existing works. Therefore, we compare our method with the state-of-the-art multi-view and deep learning methods in the experiments. 

\subsubsection{Comparison with the multi-view method} 
\begin{figure}[!t]
\centering
    \subfigure[]{\centering
    \begin{minipage}[b]{.09\textwidth}\centering
        \includegraphics[width=1\textwidth, height=1.414\textwidth]{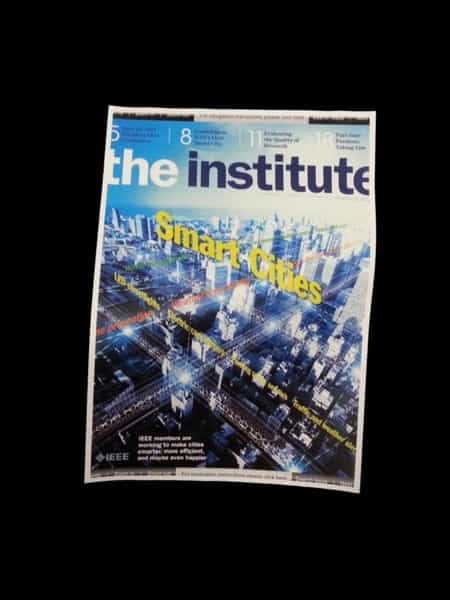}\\
        \vspace{.02\textwidth}%
        \includegraphics[width=1\textwidth, height=1.414\textwidth]{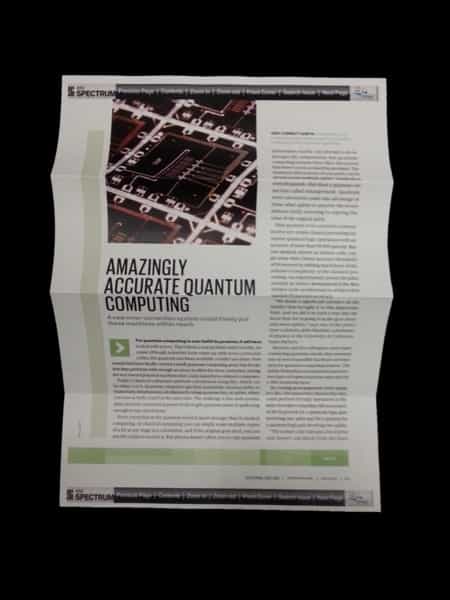}\\
        \vspace{.02\textwidth}%
        \includegraphics[width=1\textwidth, height=1.414\textwidth]{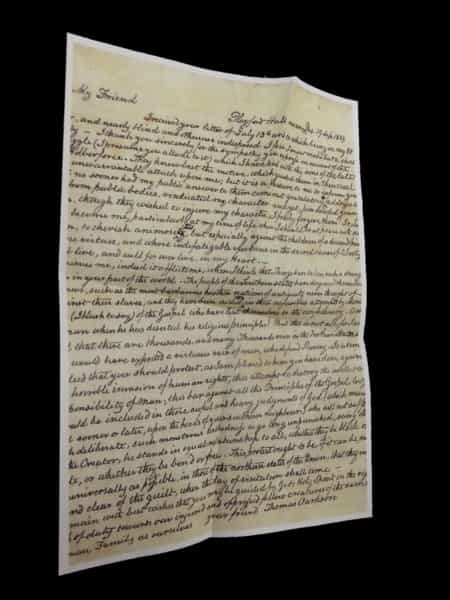}
    \end{minipage}}\hspace{.01\textwidth}%
   \subfigure[]{\centering
    \begin{minipage}[b]{.09\textwidth}\centering
        \includegraphics[width=1\textwidth, height=1.414\textwidth]{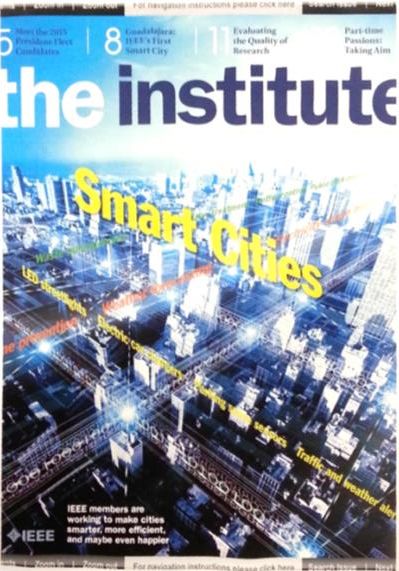}\\
        \vspace{.02\textwidth}%
        \includegraphics[width=1\textwidth, height=1.414\textwidth]{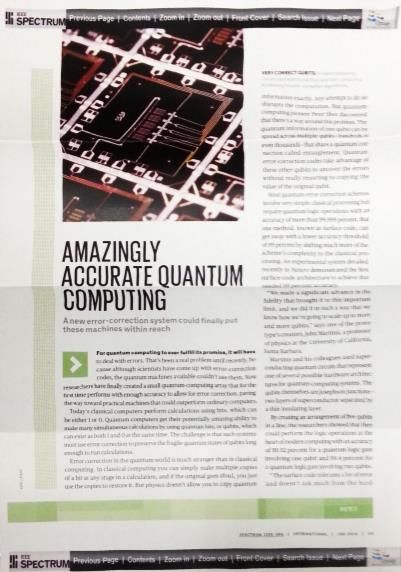}\\
        \vspace{.02\textwidth}%
        \includegraphics[width=1\textwidth, height=1.414\textwidth]{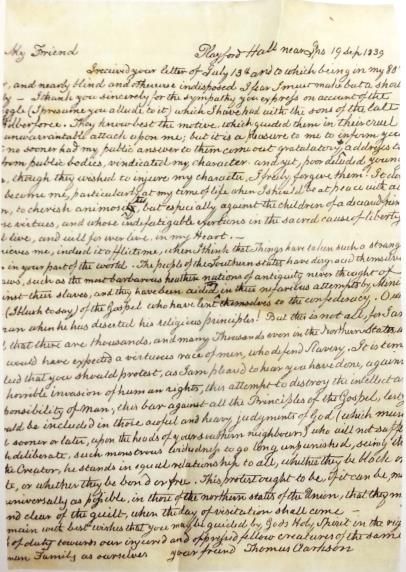}
    \end{minipage}}\hspace{.01\textwidth}%
    \subfigure[]{\centering
    \begin{minipage}[b]{.09\textwidth}\centering
        \includegraphics[width=1\textwidth, height=1.414\textwidth]{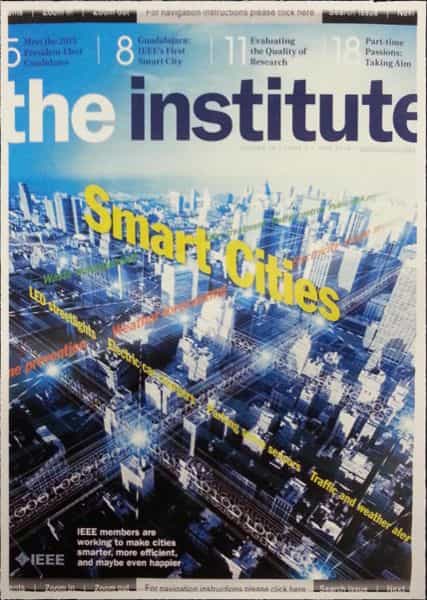}\\
        \vspace{.02\textwidth}%
        \includegraphics[width=1\textwidth, height=1.414\textwidth]{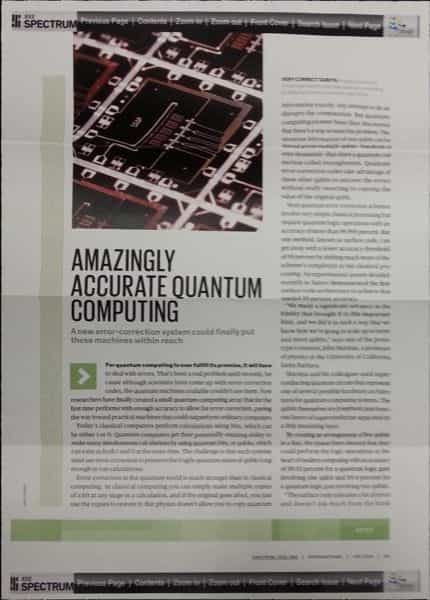}\\
        \vspace{.02\textwidth}%
        \includegraphics[width=1\textwidth, height=1.414\textwidth]{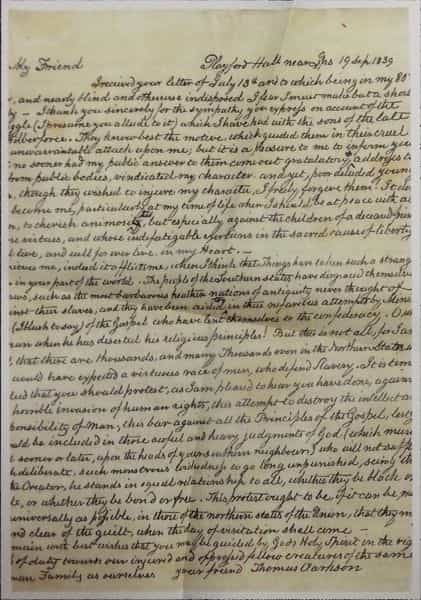}
    \end{minipage}}\hspace{.01\textwidth}%
    \subfigure[]{\centering
    \begin{minipage}[b]{.09\textwidth}\centering
        \includegraphics[width=1\textwidth, height=1.414\textwidth]{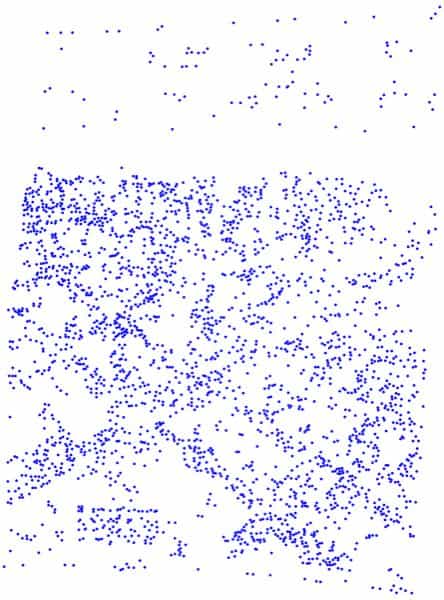}\\
        \vspace{.02\textwidth}%
        \includegraphics[width=1\textwidth, height=1.414\textwidth]{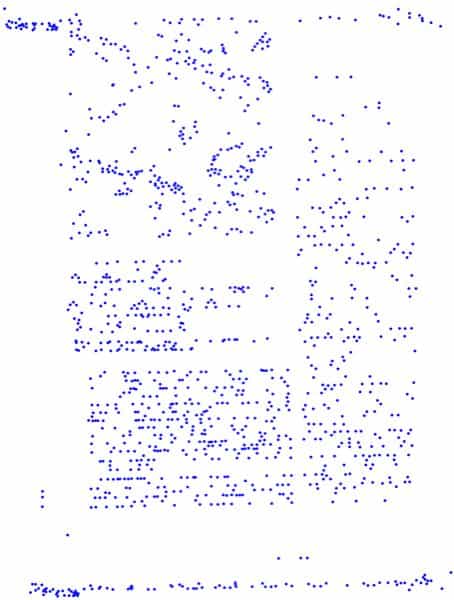}\\
        \vspace{.02\textwidth}%
        \includegraphics[width=1\textwidth, height=1.414\textwidth]{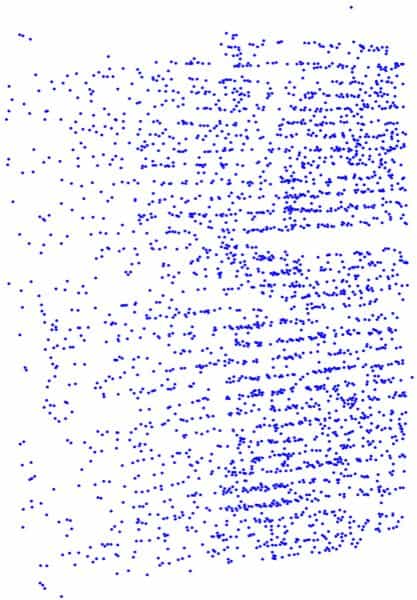}
    \end{minipage}}\hspace{.01\textwidth}%
    \subfigure[]{\centering
    \begin{minipage}[b]{.09\textwidth}\centering
        \includegraphics[width=1\textwidth, height=1.414\textwidth]{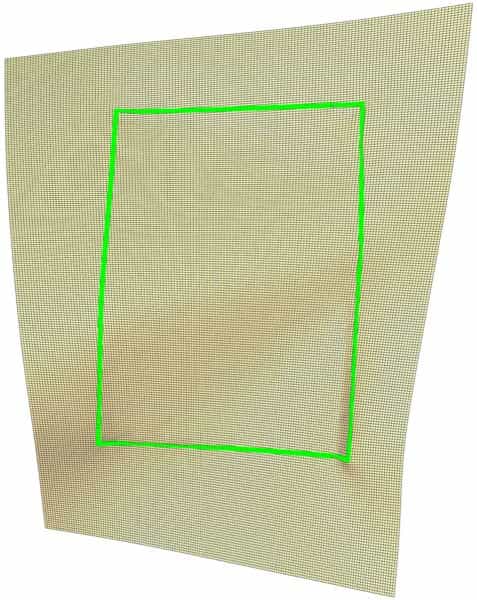}\\
        \vspace{.02\textwidth}%
        \includegraphics[width=1\textwidth, height=1.414\textwidth]{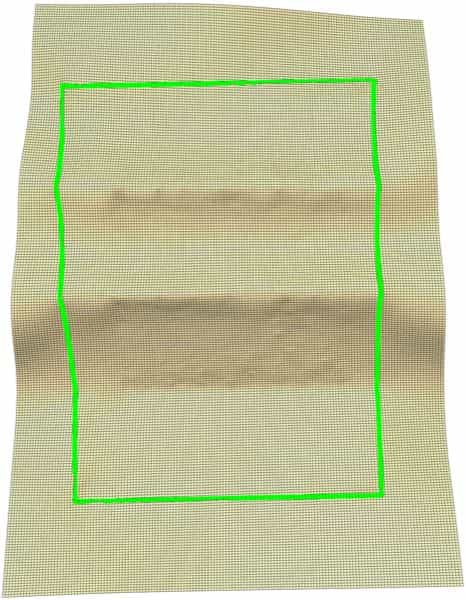}\\
        \vspace{.02\textwidth}%
        \includegraphics[width=1\textwidth, height=1.414\textwidth]{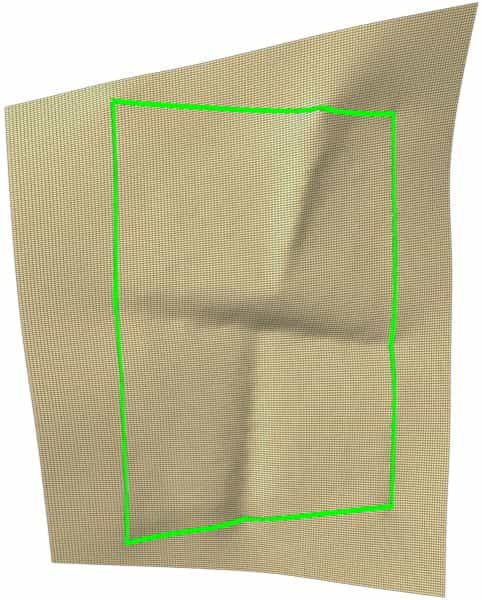}
    \end{minipage}}\hspace{.01\textwidth}%
    \subfigure[]{\centering
    \begin{minipage}[b]{.09\textwidth}\centering
        \includegraphics[width=1\textwidth, height=1.414\textwidth]{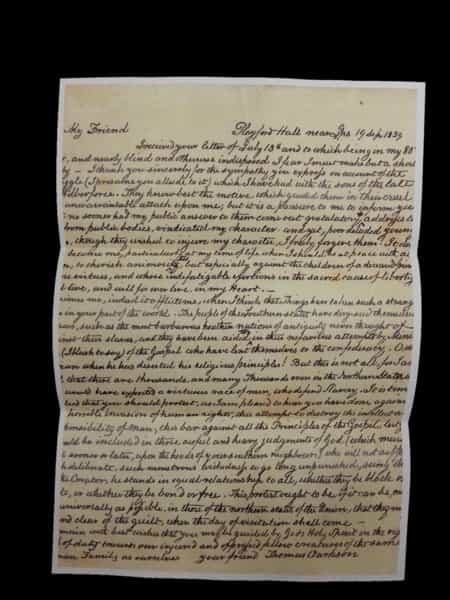}\\
        \vspace{.02\textwidth}%
        \includegraphics[width=1\textwidth, height=1.414\textwidth]{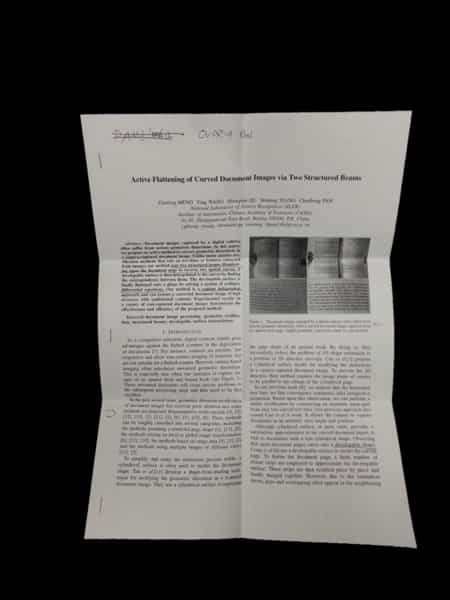}\\
        \vspace{.02\textwidth}%
        \includegraphics[width=1\textwidth, height=1.414\textwidth]{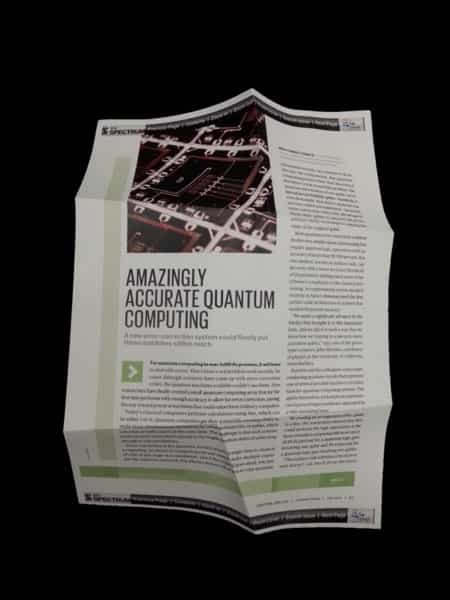}
    \end{minipage}}\hspace{.01\textwidth}%
   \subfigure[]{\centering
    \begin{minipage}[b]{.09\textwidth}\centering
        \includegraphics[width=1\textwidth, height=1.414\textwidth]{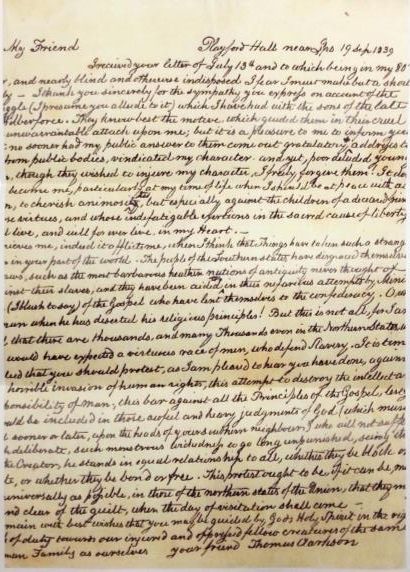}\\
        \vspace{.02\textwidth}%
        \includegraphics[width=1\textwidth, height=1.414\textwidth]{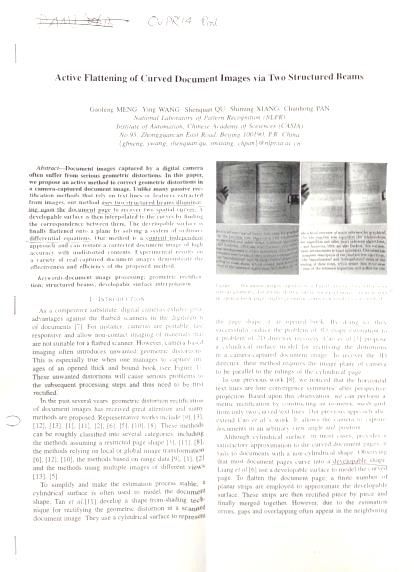}\\
        \vspace{.02\textwidth}%
        \includegraphics[width=1\textwidth, height=1.414\textwidth]{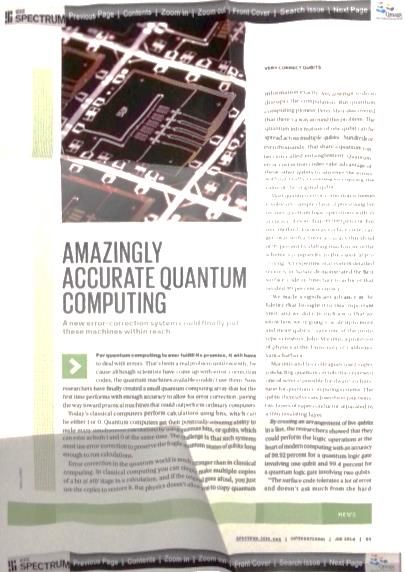}
    \end{minipage}}\hspace{.01\textwidth}%
    \subfigure[]{\centering
    \begin{minipage}[b]{.09\textwidth}\centering
        \includegraphics[width=1\textwidth, height=1.414\textwidth]{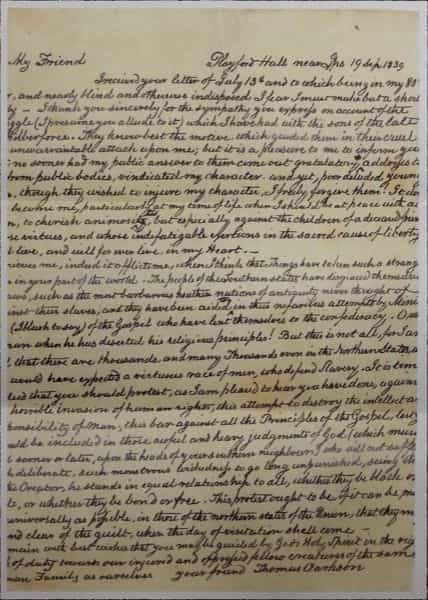}\\
        \vspace{.02\textwidth}%
        \includegraphics[width=1\textwidth, height=1.414\textwidth]{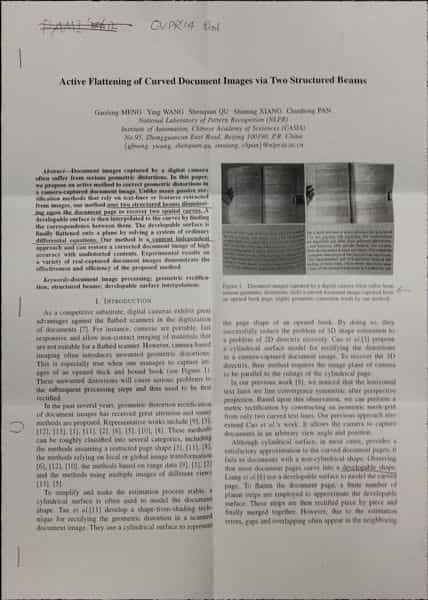}\\
        \vspace{.02\textwidth}%
        \includegraphics[width=1\textwidth, height=1.414\textwidth]{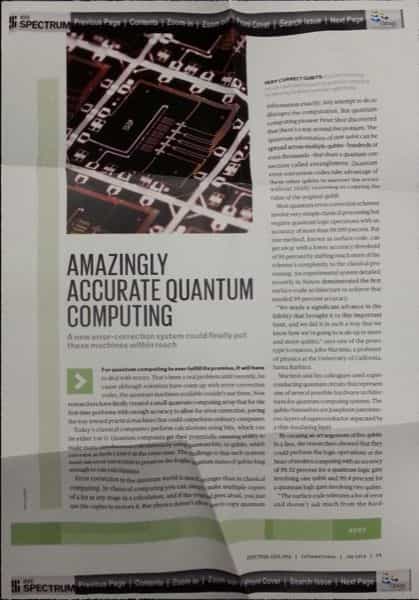}
    \end{minipage}}\hspace{.01\textwidth}%
     \subfigure[]{\centering
    \begin{minipage}[b]{.09\textwidth}\centering
        \includegraphics[width=1\textwidth, height=1.414\textwidth]{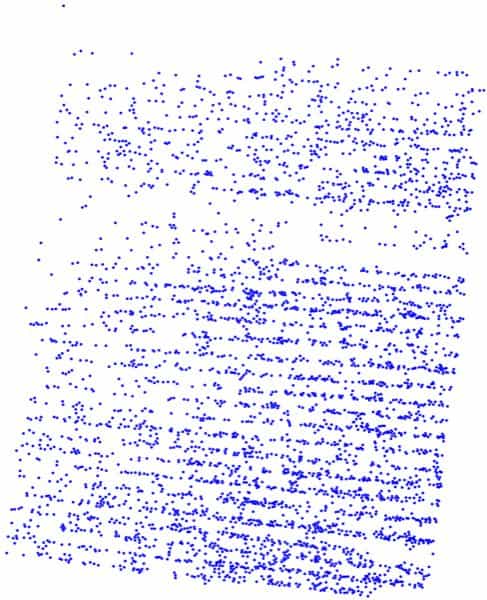}\\
        \vspace{.02\textwidth}%
        \includegraphics[width=1\textwidth, height=1.414\textwidth]{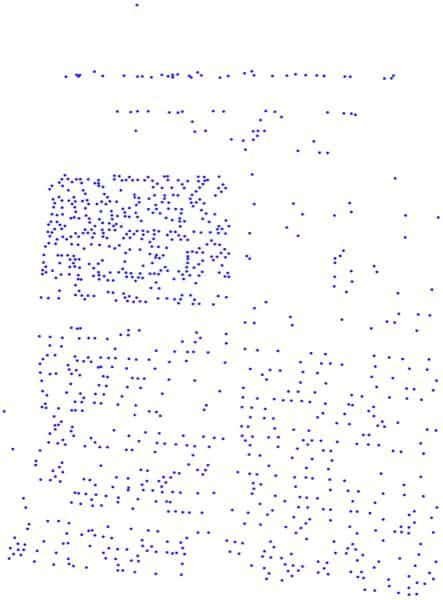}\\
        \vspace{.02\textwidth}%
        \includegraphics[width=1\textwidth, height=1.414\textwidth]{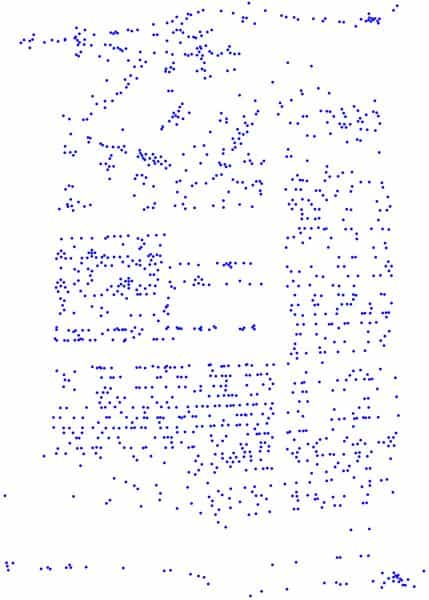}
    \end{minipage}}\hspace{.01\textwidth}%
    \subfigure[]{\centering
    \begin{minipage}[b]{.09\textwidth}\centering
        \includegraphics[width=1\textwidth, height=1.414\textwidth]{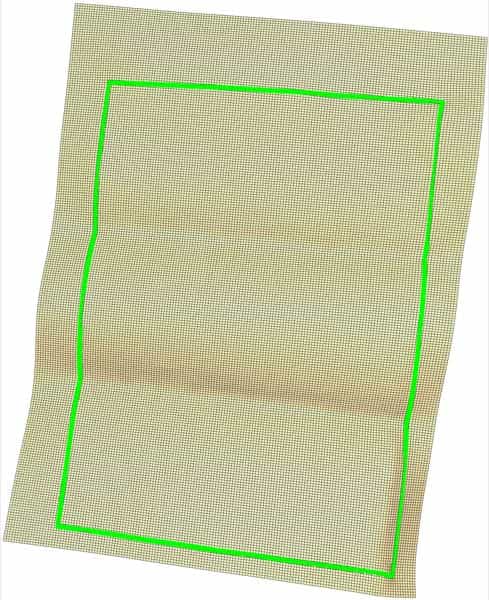}\\
        \vspace{.02\textwidth}%
        \includegraphics[width=1\textwidth, height=1.414\textwidth]{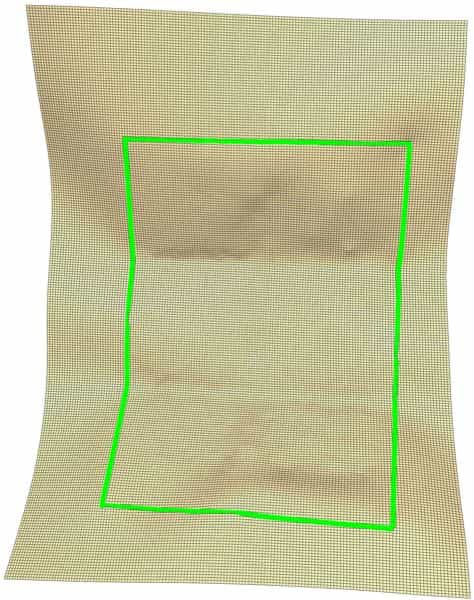}\\
        \vspace{.02\textwidth}%
        \includegraphics[width=1\textwidth, height=1.414\textwidth]{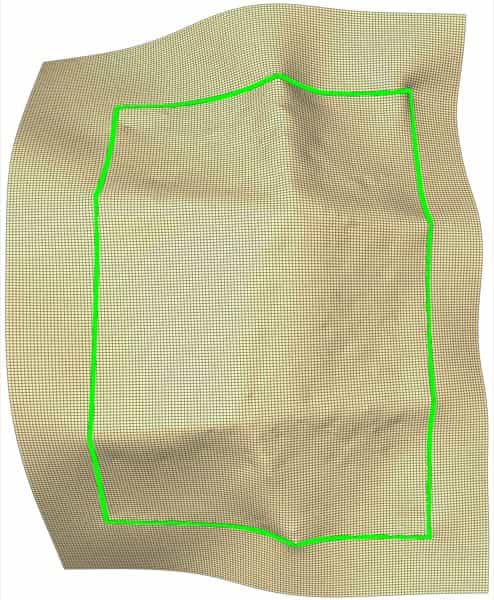}
    \end{minipage}}
\vspace{-.015\textwidth}
\caption{Comparison with the multi-view method. Original images are shown in (a) and (f). Results 
 of \citeauthor{you2018multiview} are shown in (b) and (g). 
 Our results are shown in (c) and (h). Our final $\mc{P}$ are shown in (d) and (i). Our final $\mc{M}$ with boundary are shown in (e) and (j).}\label{f:you}
\end{figure}
\citeauthor{you2018multiview} proposed a method which reconstructs the 3D document model from point clouds generated from multi-view images~\cite{you2018multiview}. Its main contribution is a ridge-aware strategy for reconstructing sharp folding edges. The boundary lines are straightened in the flattened model to benefit the unfolding of the document model. 

We take several most challenging cases from~\cite{you2018multiview} and give the results produced by our method for comparison.  There are 7 to 10 images provided for each example in \cite{you2018multiview} and we only use 3 images for each example in our method, which increases the difficulty of reconstruction. In the experiments, only the boundary lines are used as constraints in our method, which are also used in~\cite{you2018multiview}. From the results in Fig.~\ref{f:you}, we observe an improvement in rectification quality with our method than the method in~\cite{you2018multiview}. The improvements are even more noticeable for the examples with more complicated folds. The 3D reconstructed meshes obtained by our method are also shown, which correctly recover the overall shape of the document as well as the position of folds. The recovered folding edges are not as sharp as the original ones, since no sharpening operation similar to the one used in~\cite{you2018multiview} is applied to them. These experiments show that the overall reconstruction quality and developability of the document play essential roles in document rectification.

\subsubsection{Comparison with  deep learning methods}

We compare our method with the state-of-the-art deep learning methods including  DewarpNet~ \cite{das2019dewarpnet}, FCN-based~ \cite{xie2021dewarping},  Points-based~\cite{xie2022document}, DocTr~ \cite{feng2021doctr}, and DocScanner~ \cite{feng2021docscanner}. We use the trained models provided by the authors in the experiments.

\paragraph{Datasets generation} Since folded or creased documents with large deformation are not particularly studied in existing works and such kinds of datasets are unavailable, we create some datasets by manually folding or creasing the documents commonly found in our daily life.  We take photos using an iPhone 11, and 3–6 images are used for each example. The selected datasets cover a large variety of contents and we classify them into three categories. 
\begin{figure}[!t]
\centering
    \subfigure[]{\centering 
    \begin{minipage}[b]{.09\textwidth}\centering
    \includegraphics[width=1\textwidth, height=1.414\textwidth]{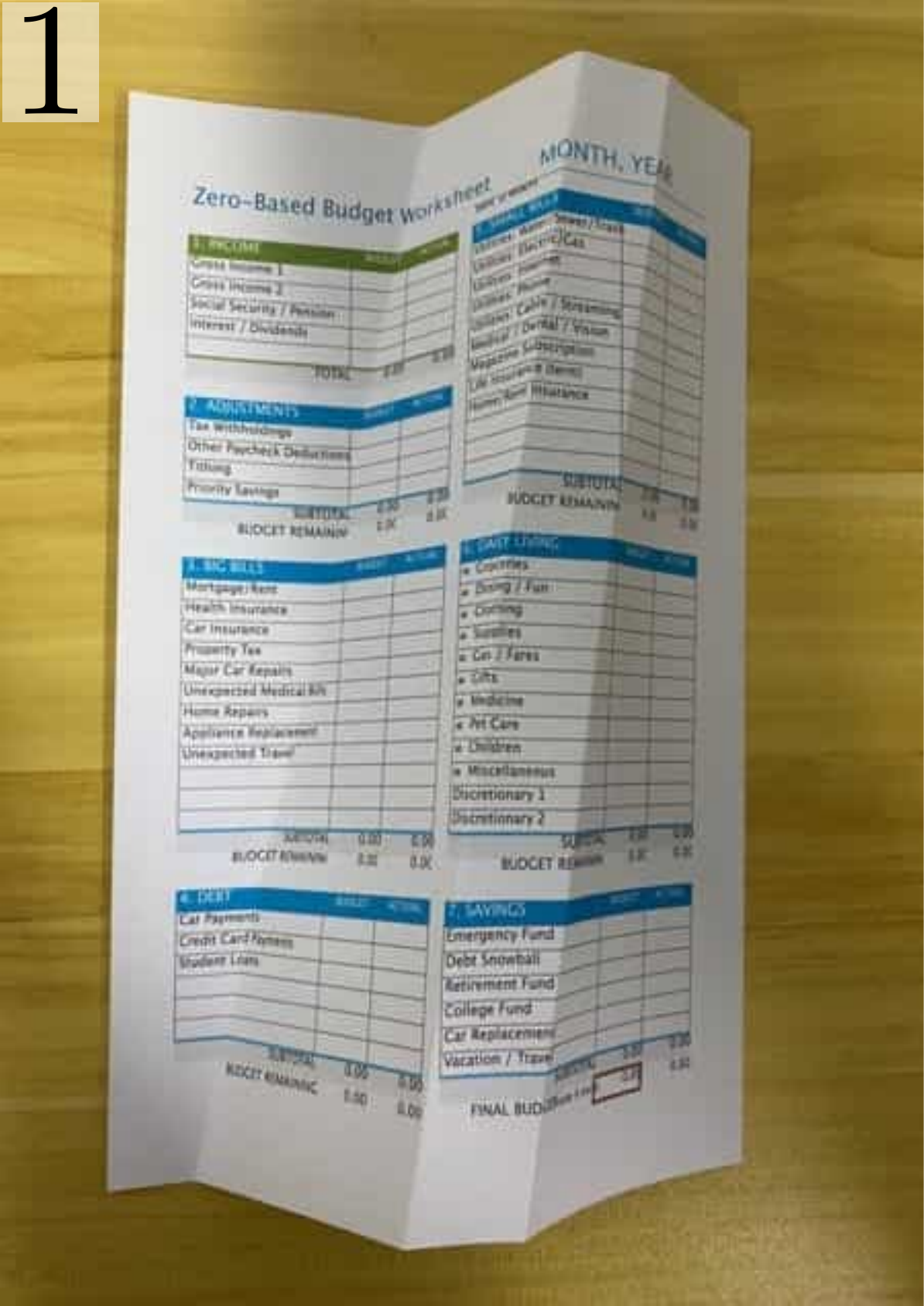}\\
    \vspace{.02\textwidth}%
    \includegraphics[width=1\textwidth, height=1.414\textwidth]{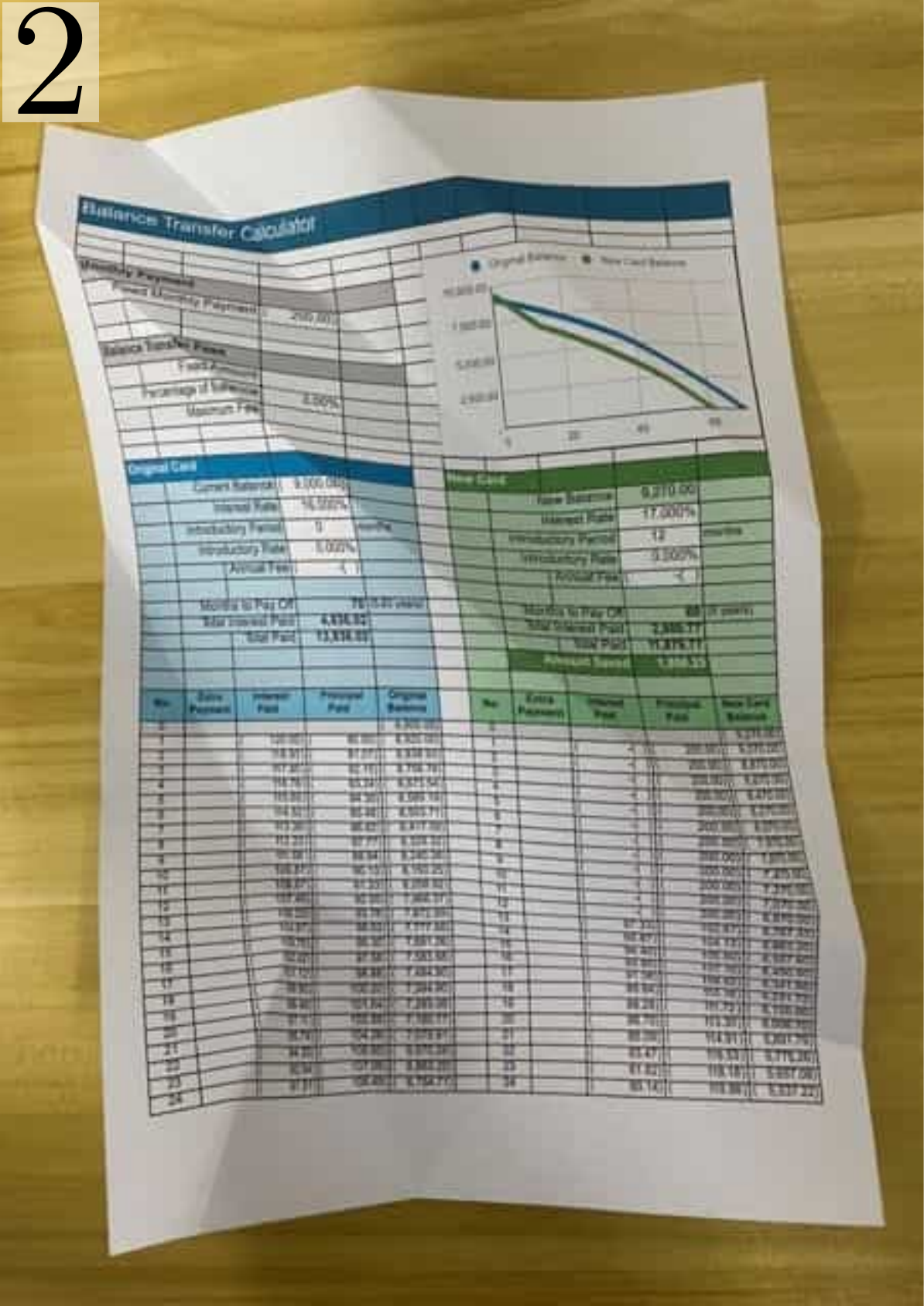}\\
    \vspace{.02\textwidth}%
    \includegraphics[width=1\textwidth, height=1.414\textwidth]{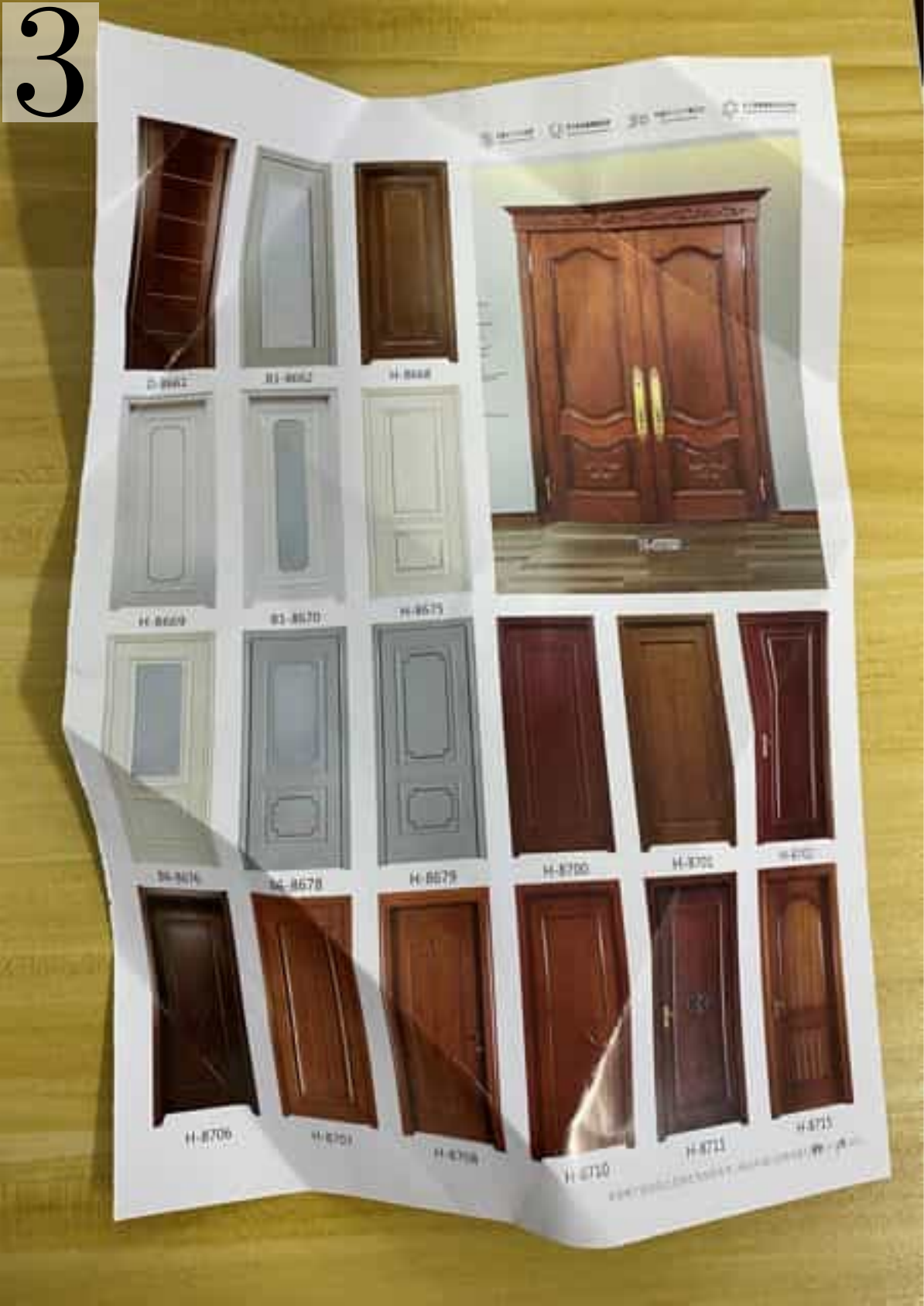}\\
    \vspace{.02\textwidth}%
    \includegraphics[width=1\textwidth, height=1.414\textwidth]{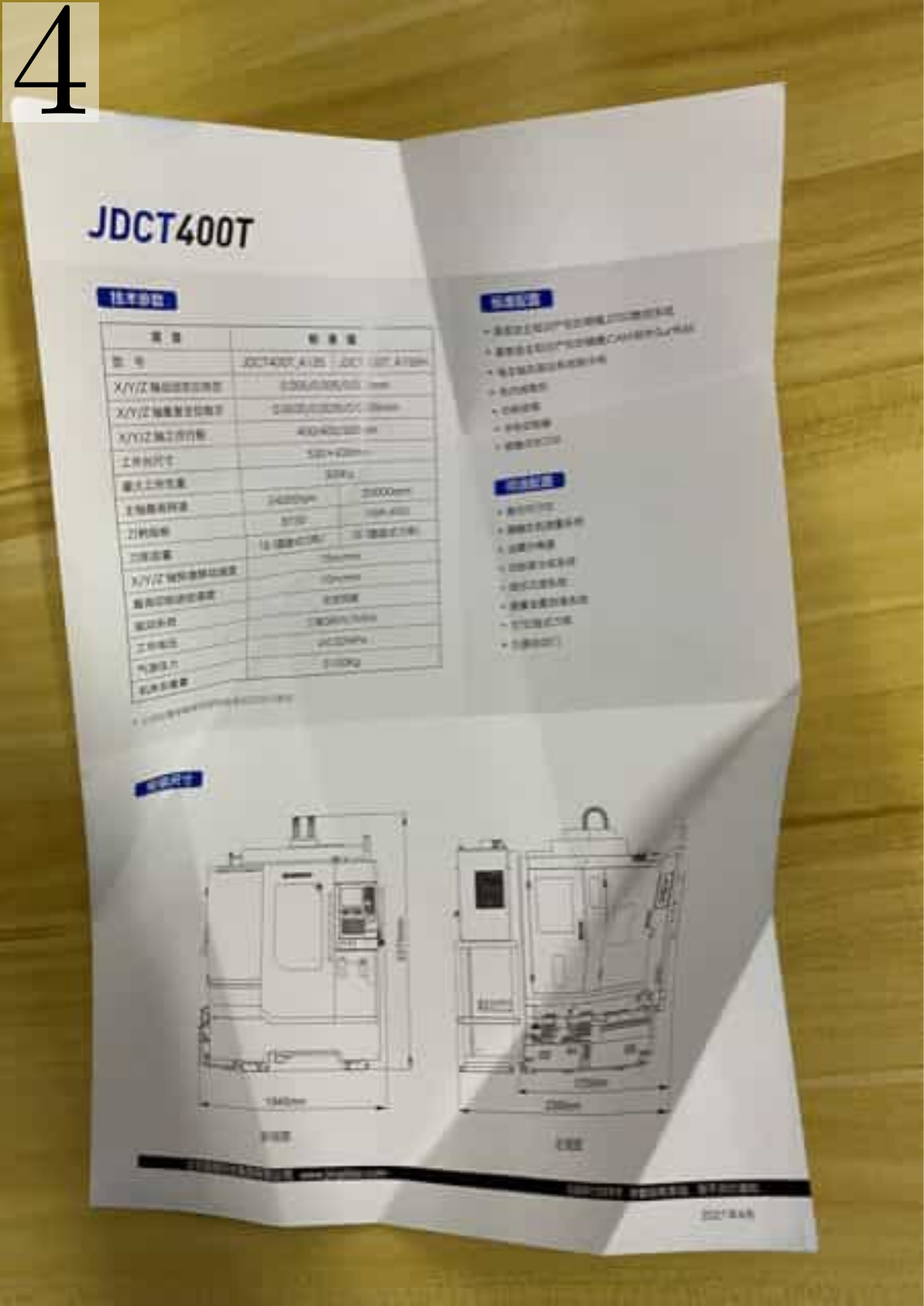}\\
    \vspace{.02\textwidth}%
    \includegraphics[width=1\textwidth, height=1.414\textwidth]{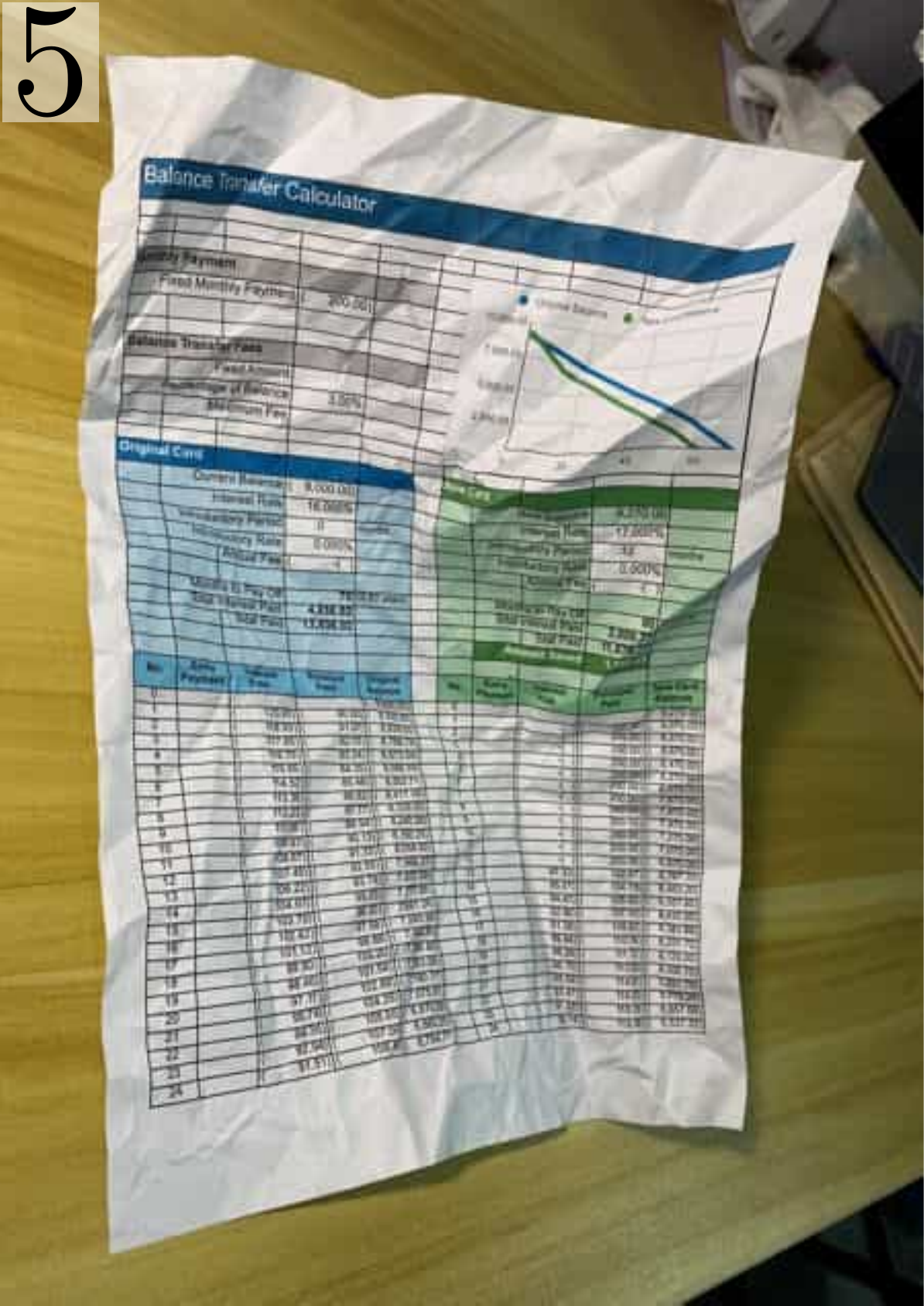}\\
    \vspace{.02\textwidth}%
    \includegraphics[width=1\textwidth, height=1.414\textwidth]{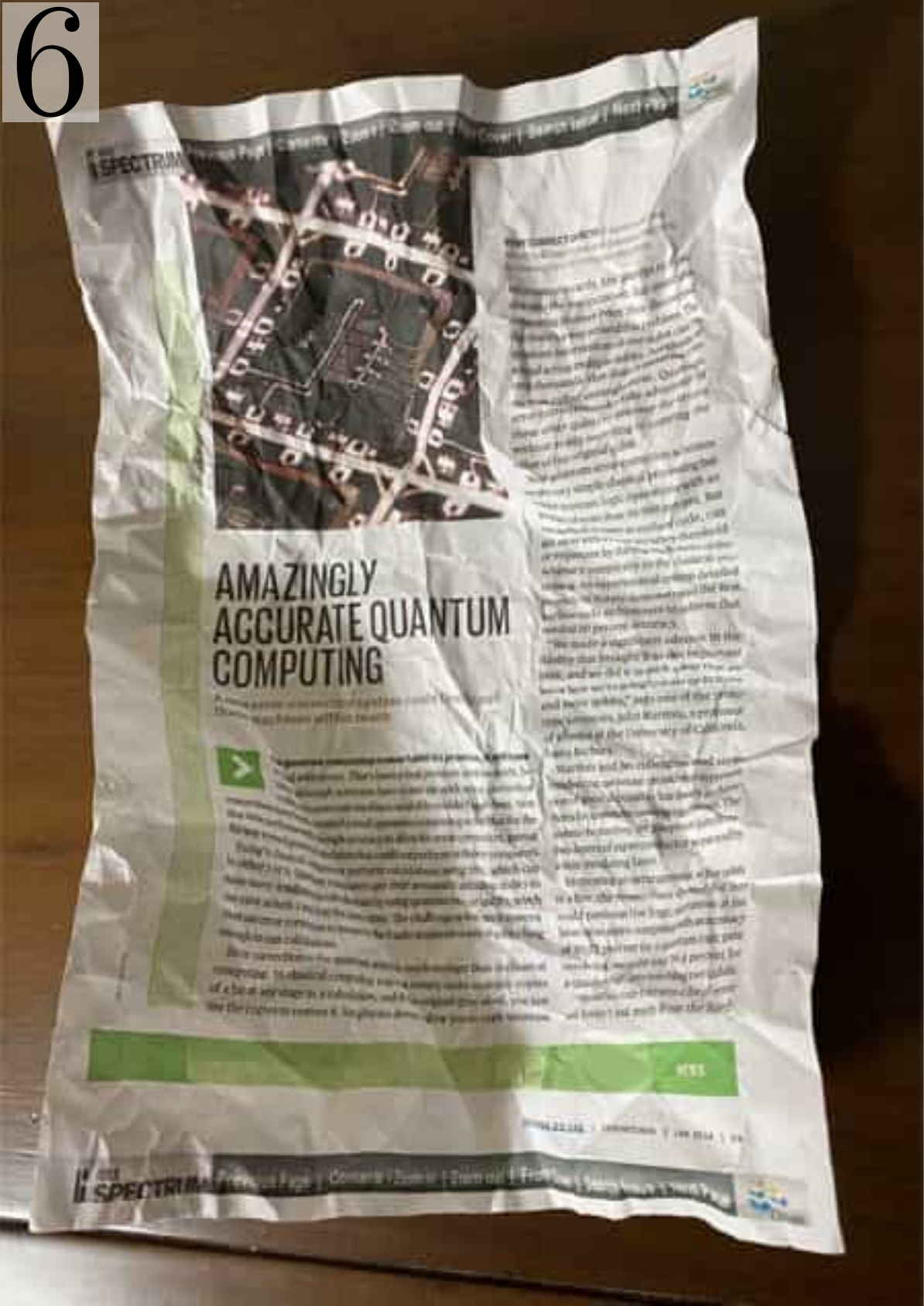}
    \end{minipage}}\hspace{.01\textwidth}%
    \subfigure[]{\centering
    \begin{minipage}[b]{.09\textwidth}\centering
        \includegraphics[width=1\textwidth, height=1.414\textwidth]{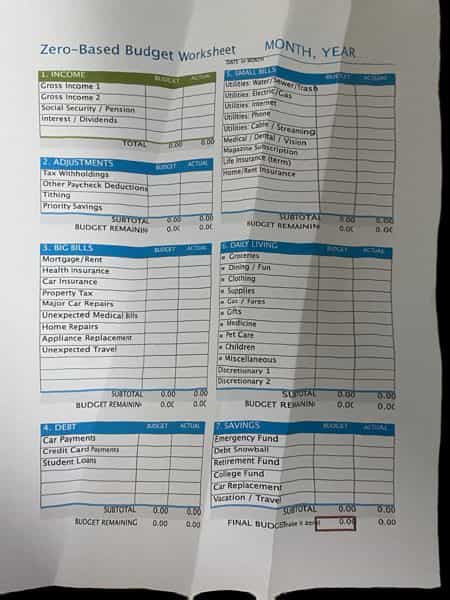}\\
        \vspace{.02\textwidth}%
        \includegraphics[width=1\textwidth, height=1.414\textwidth]{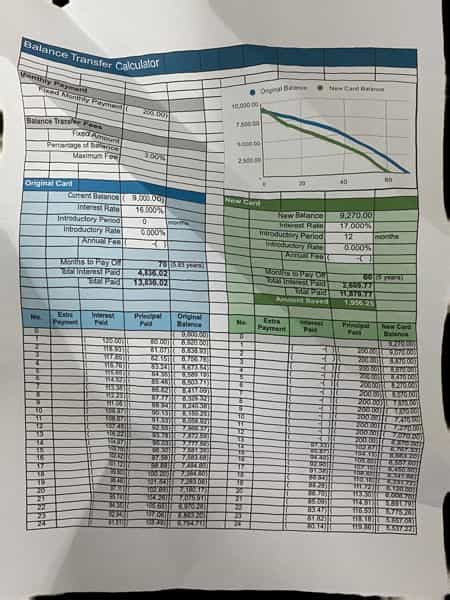}\\
        \vspace{.02\textwidth}%
        \includegraphics[width=1\textwidth, height=1.414\textwidth]{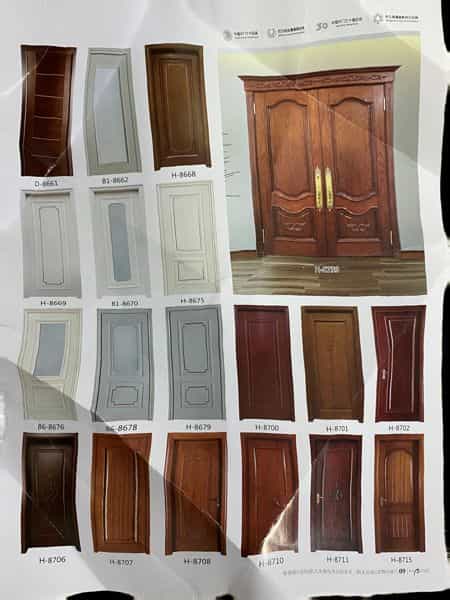}\\
        \vspace{.02\textwidth}%
        \includegraphics[width=1\textwidth, height=1.414\textwidth]{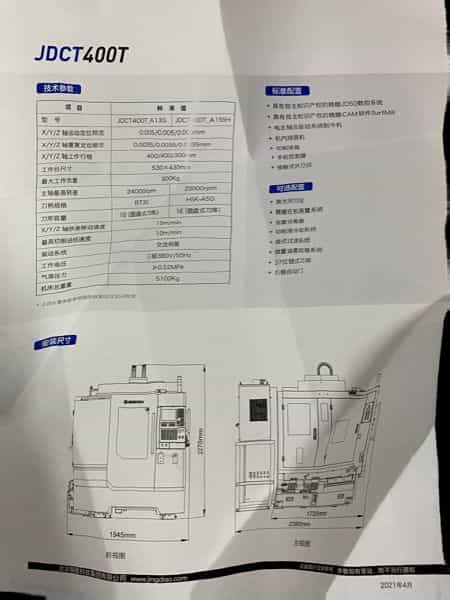}\\
        \vspace{.02\textwidth}%
        \includegraphics[width=1\textwidth, height=1.414\textwidth]{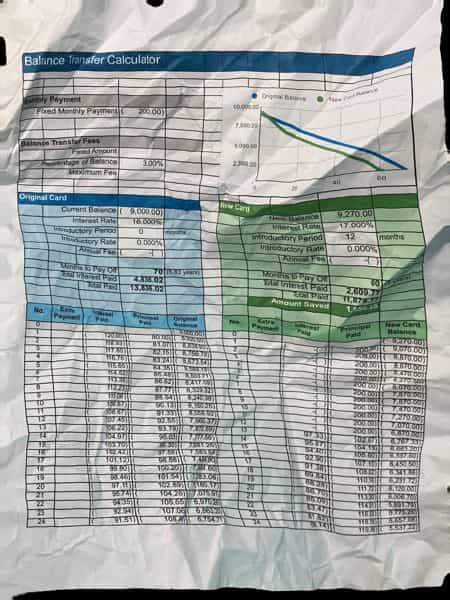}\\
        \vspace{.02\textwidth}%
        \includegraphics[width=1\textwidth, height=1.414\textwidth]{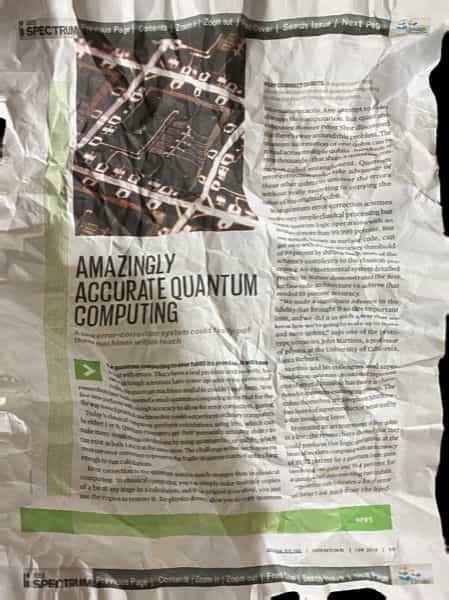}
    \end{minipage}}\hspace{.01\textwidth}%
    \subfigure[]{\centering
    \begin{minipage}[b]{.09\textwidth}\centering
        \includegraphics[width=1\textwidth, height=1.414\textwidth]{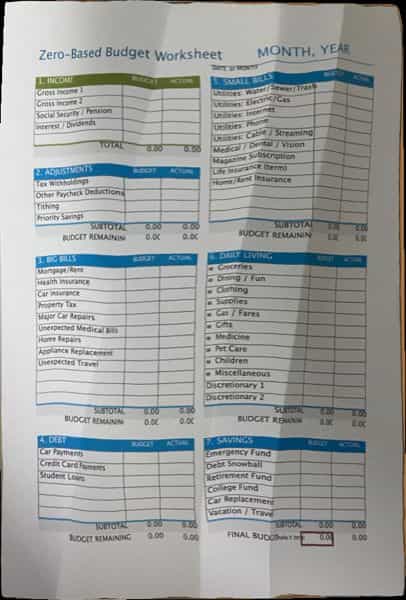}\\
        \vspace{.02\textwidth}%
        \includegraphics[width=1\textwidth, height=1.414\textwidth]{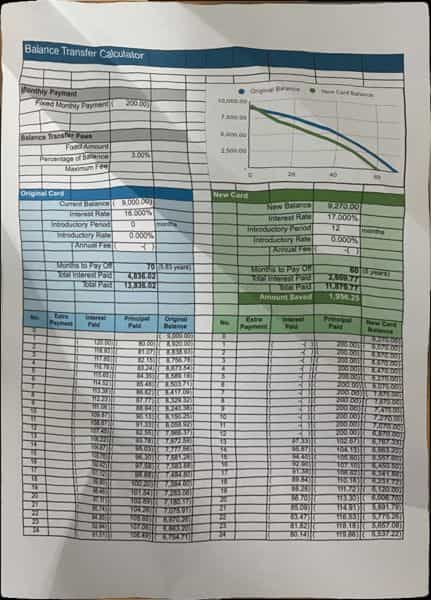}\\
        \vspace{.02\textwidth}%
        \includegraphics[width=1\textwidth, height=1.414\textwidth]{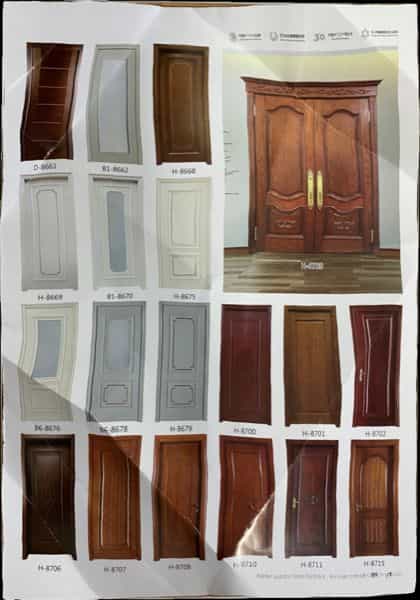}\\
        \vspace{.02\textwidth}%
        \includegraphics[width=1\textwidth, height=1.414\textwidth]{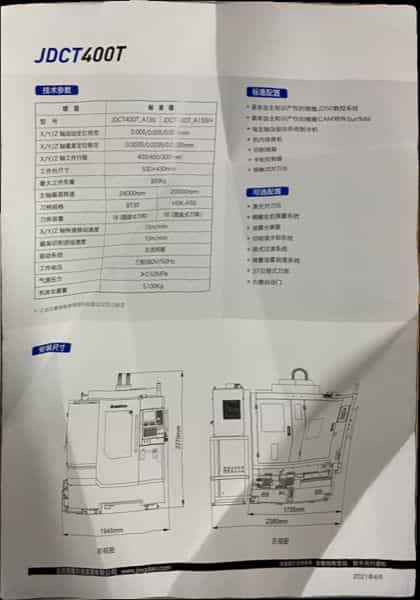}\\
        \vspace{.02\textwidth}%
        \includegraphics[width=1\textwidth, height=1.414\textwidth]{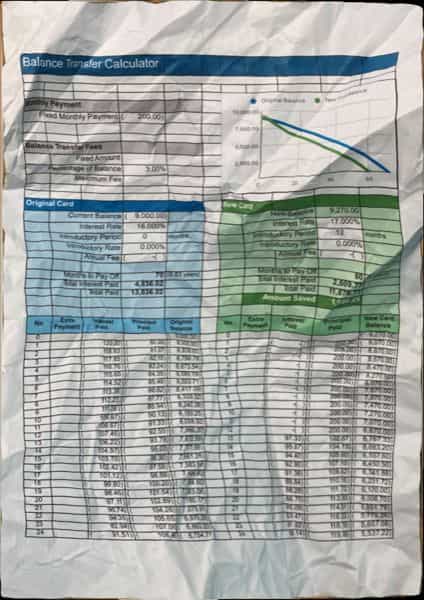}\\
        \vspace{.02\textwidth}%
        \includegraphics[width=1\textwidth, height=1.414\textwidth]{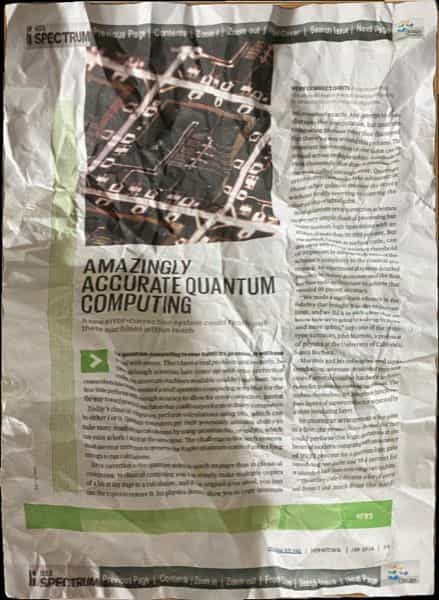}
    \end{minipage}}\hspace{.01\textwidth}%
    \subfigure[]{\centering
    \begin{minipage}[b]{.09\textwidth}\centering
        \includegraphics[width=1\textwidth, height=1.414\textwidth]{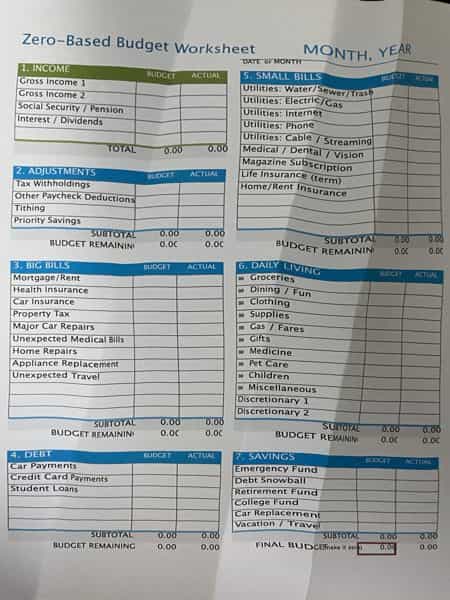}\\
        \vspace{.02\textwidth}%
        \includegraphics[width=1\textwidth, height=1.414\textwidth]{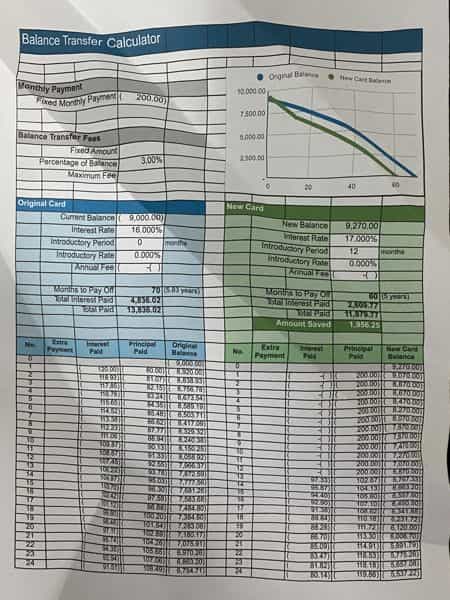}\\
        \vspace{.02\textwidth}%
        \includegraphics[width=1\textwidth, height=1.414\textwidth]{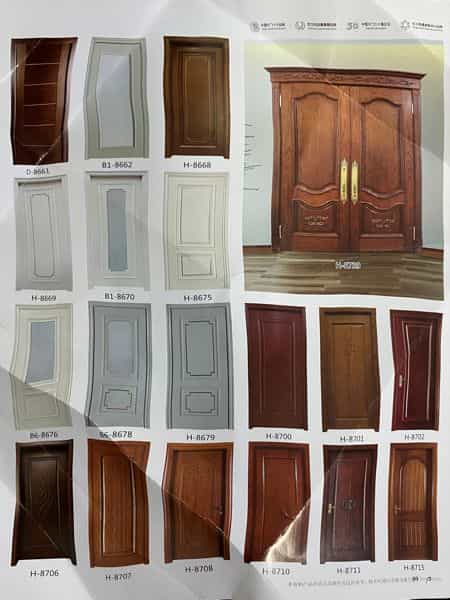}\\
        \vspace{.02\textwidth}%
        \includegraphics[width=1\textwidth, height=1.414\textwidth]{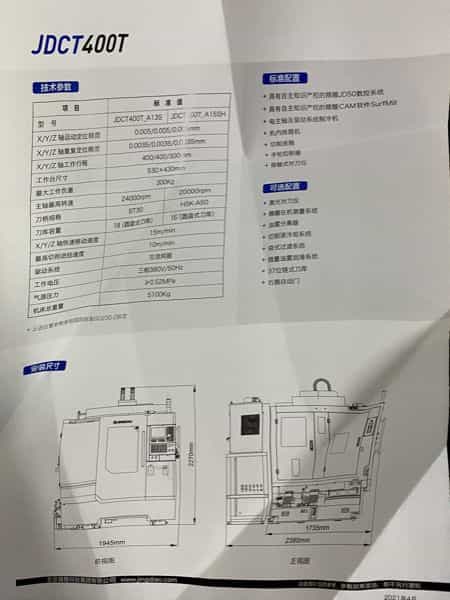}\\
        \vspace{.02\textwidth}%
        \includegraphics[width=1\textwidth, height=1.414\textwidth]{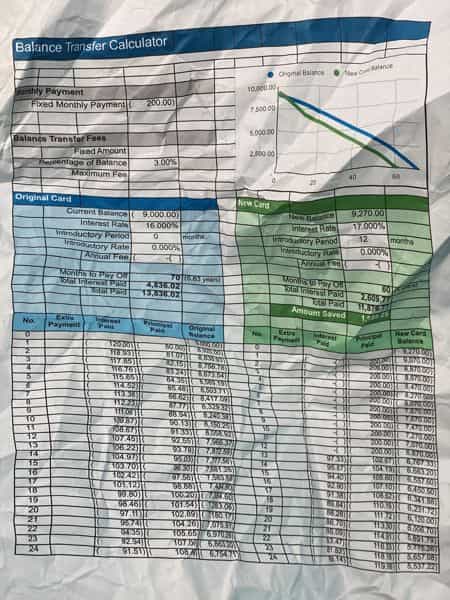}\\
        \vspace{.02\textwidth}%
        \includegraphics[width=1\textwidth, height=1.414\textwidth]{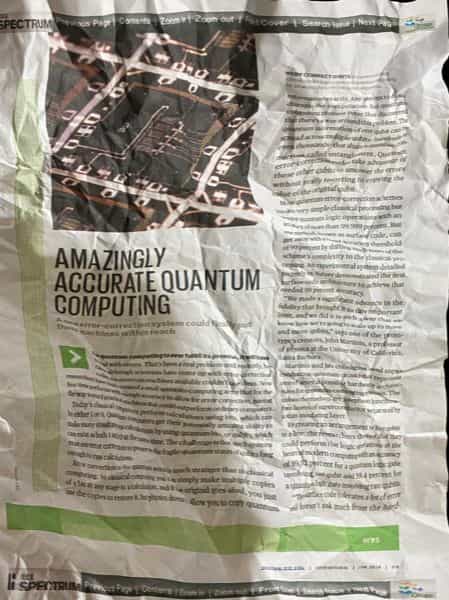}
    \end{minipage}}\hspace{.01\textwidth}%
    \subfigure[]{\centering
    \begin{minipage}[b]{.09\textwidth}\centering
        \includegraphics[width=1\textwidth, height=1.414\textwidth]{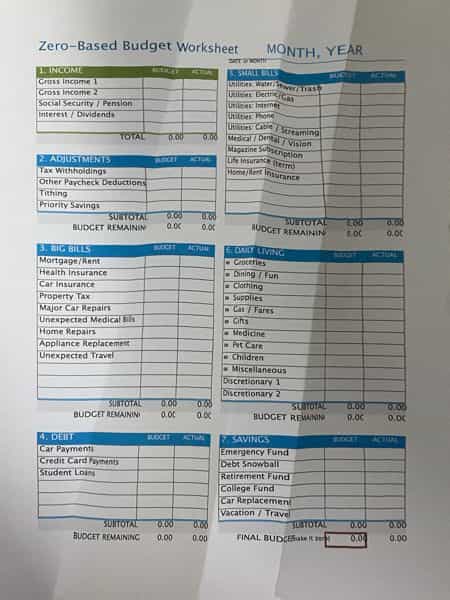}\\
        \vspace{.02\textwidth}%
        \includegraphics[width=1\textwidth, height=1.414\textwidth]{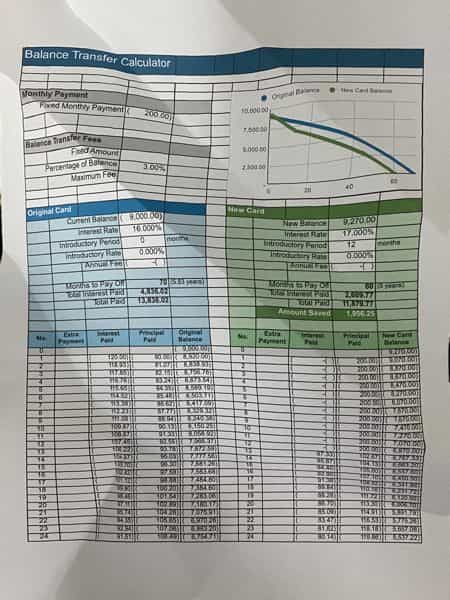}\\
        \vspace{.02\textwidth}%
        \includegraphics[width=1\textwidth, height=1.414\textwidth]{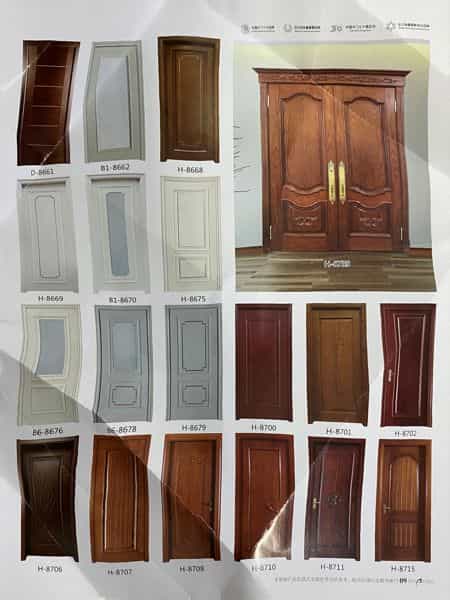}\\
        \vspace{.02\textwidth}%
        \includegraphics[width=1\textwidth, height=1.414\textwidth]{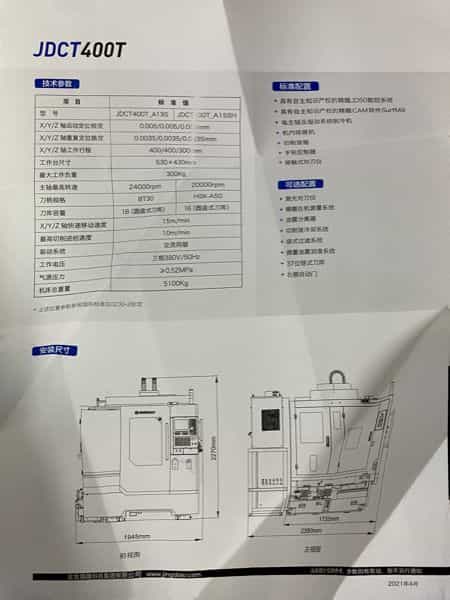}\\
        \vspace{.02\textwidth}%
        \includegraphics[width=1\textwidth, height=1.414\textwidth]{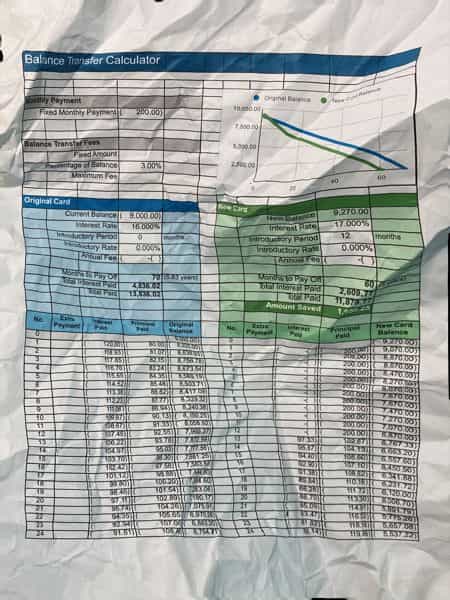}\\
        \vspace{.02\textwidth}%
        \includegraphics[width=1\textwidth, height=1.414\textwidth]{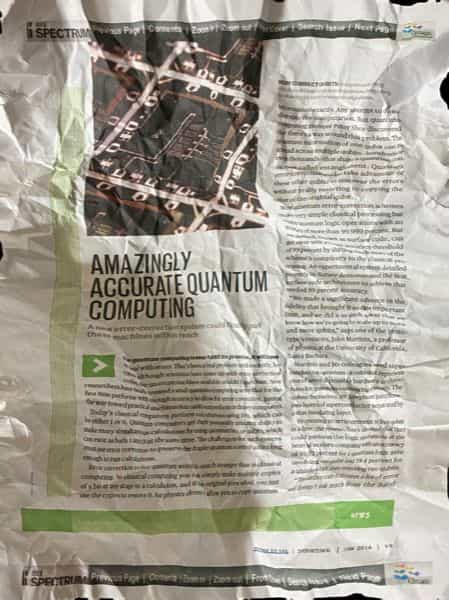}
    \end{minipage}}\hspace{.01\textwidth}%
    \subfigure[]{\centering
    \begin{minipage}[b]{.09\textwidth}\centering
        \includegraphics[width=1\textwidth, height=1.414\textwidth]{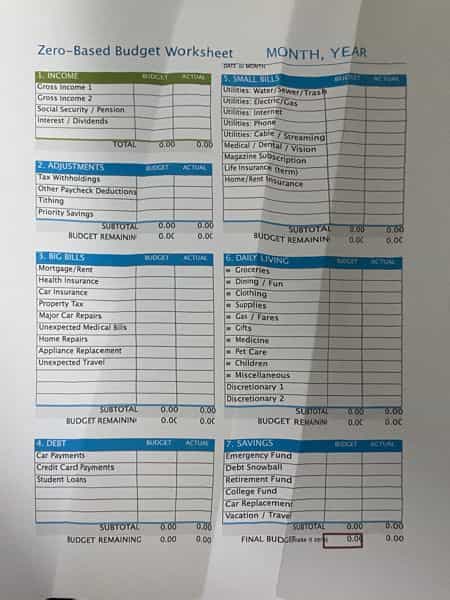}\\
        \vspace{.02\textwidth}%
        \includegraphics[width=1\textwidth, height=1.414\textwidth]{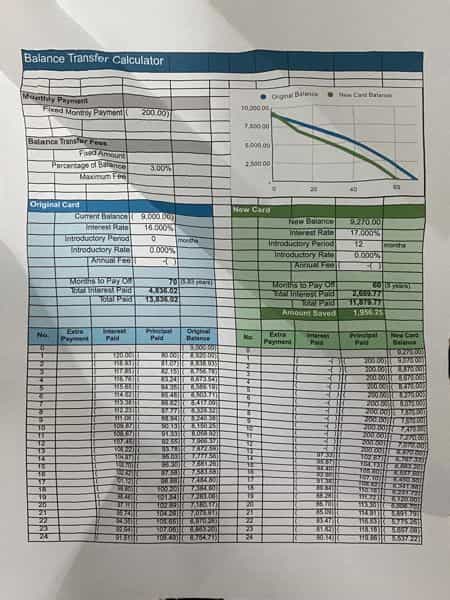}\\
        \vspace{.02\textwidth}%
        \includegraphics[width=1\textwidth, height=1.414\textwidth]{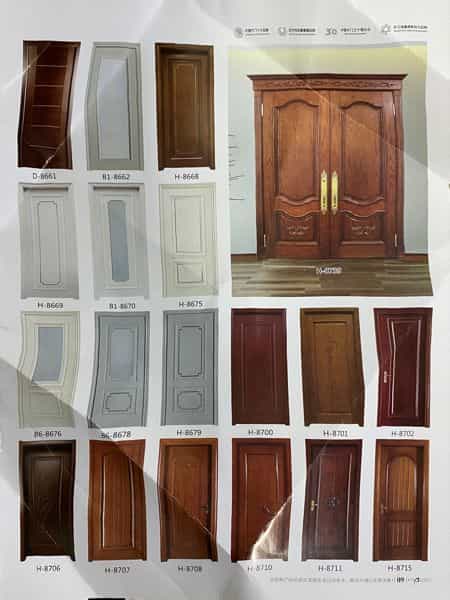}\\
        \vspace{.02\textwidth}%
        \includegraphics[width=1\textwidth, height=1.414\textwidth]{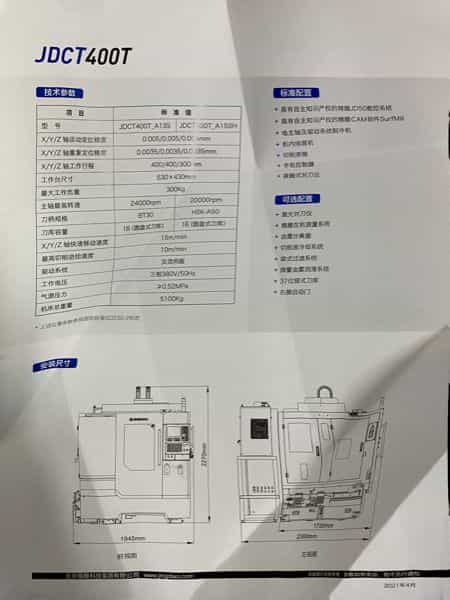}\\
        \vspace{.02\textwidth}%
        \includegraphics[width=1\textwidth, height=1.414\textwidth]{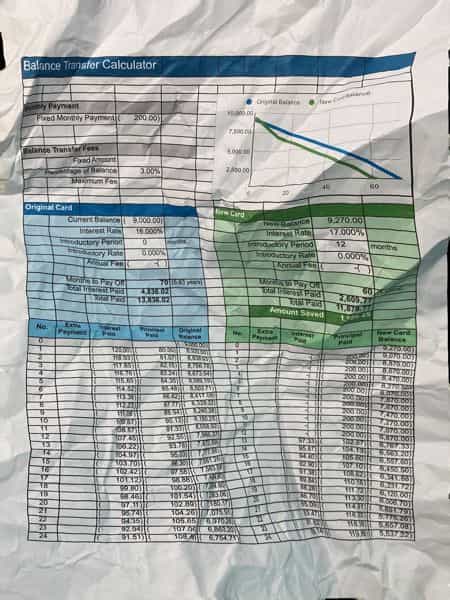}\\
        \vspace{.02\textwidth}%
        \includegraphics[width=1\textwidth, height=1.414\textwidth]{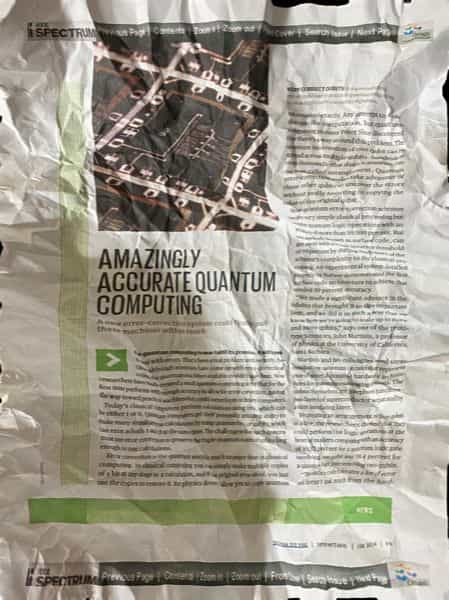}
    \end{minipage}}\hspace{.01\textwidth}%
    \subfigure[]{\centering
    \begin{minipage}[b]{.09\textwidth}\centering
        \includegraphics[width=1\textwidth, height=1.414\textwidth]{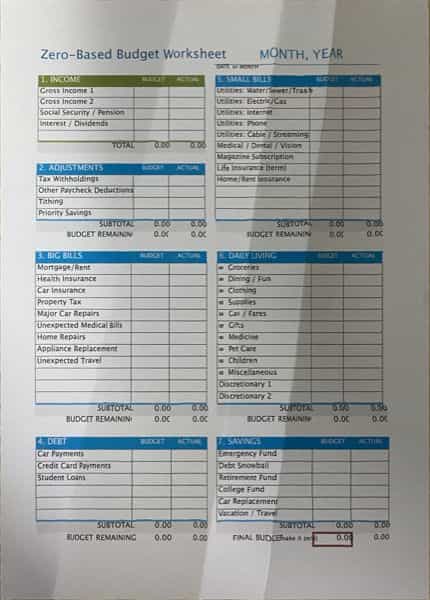}\\
        \vspace{.02\textwidth}%
        \includegraphics[width=1\textwidth, height=1.414\textwidth]{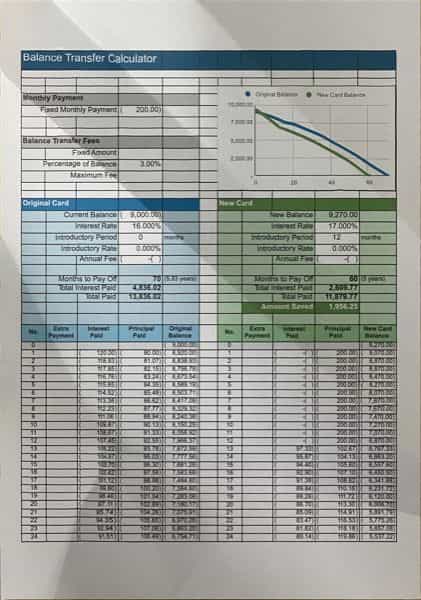}\\
        \vspace{.02\textwidth}%
        \includegraphics[width=1\textwidth, height=1.414\textwidth]{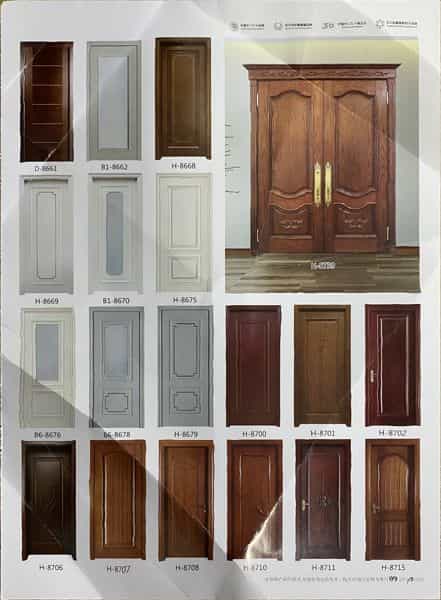}\\
        \vspace{.02\textwidth}%
        \includegraphics[width=1\textwidth, height=1.414\textwidth]{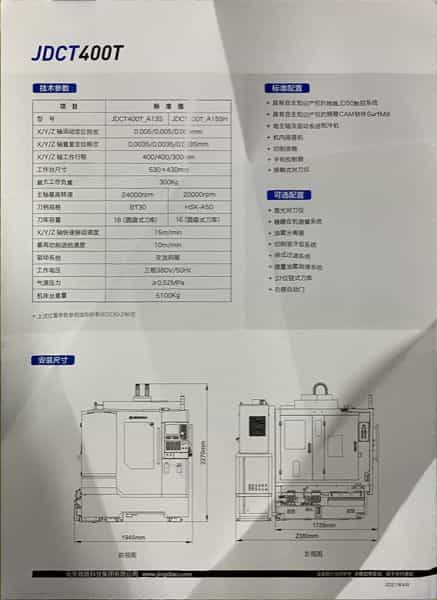}\\
        \vspace{.02\textwidth}%
        \includegraphics[width=1\textwidth, height=1.414\textwidth]{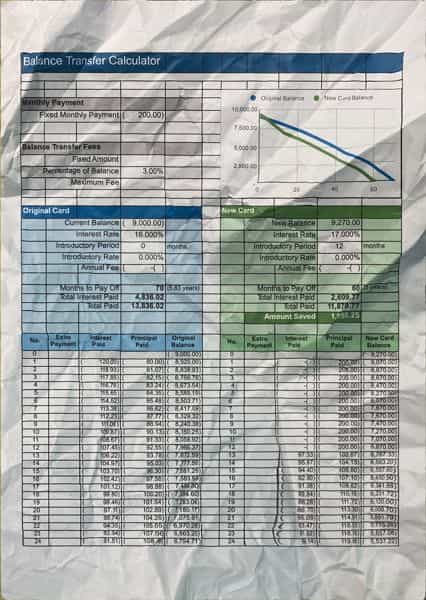}\\
        \vspace{.02\textwidth}%
        \includegraphics[width=1\textwidth, height=1.414\textwidth]{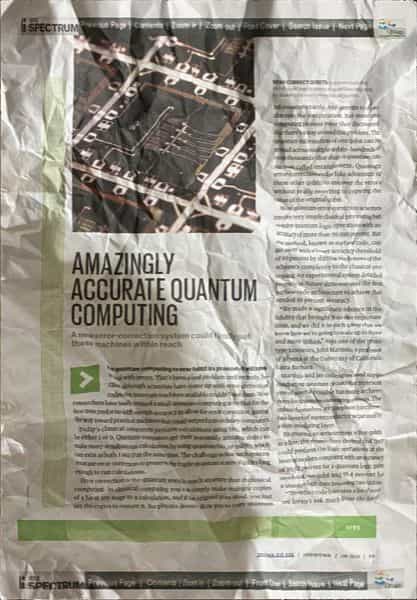} 
    \end{minipage}}\hspace{.01\textwidth}%
    \subfigure[]{\centering
    \begin{minipage}[b]{.09\textwidth}\centering
        \includegraphics[width=1\textwidth, height=1.414\textwidth]{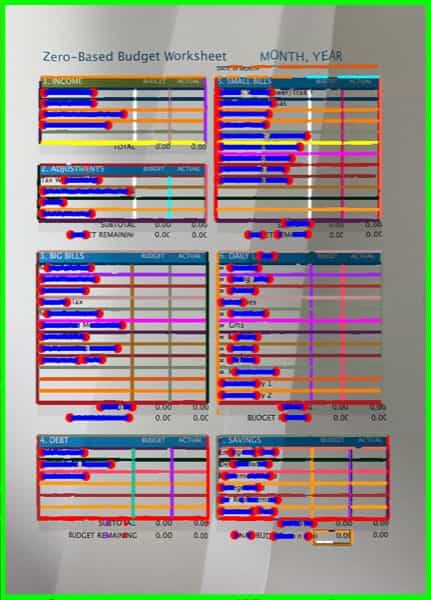}\\
        \vspace{.02\textwidth}%
        \includegraphics[width=1\textwidth, height=1.414\textwidth]{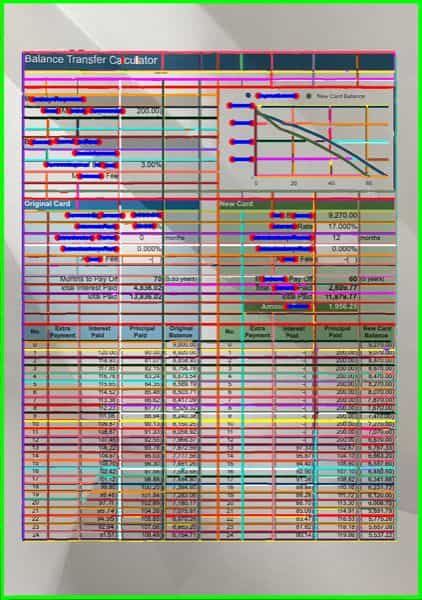}\\
        \vspace{.02\textwidth}%
        \includegraphics[width=1\textwidth, height=1.414\textwidth]{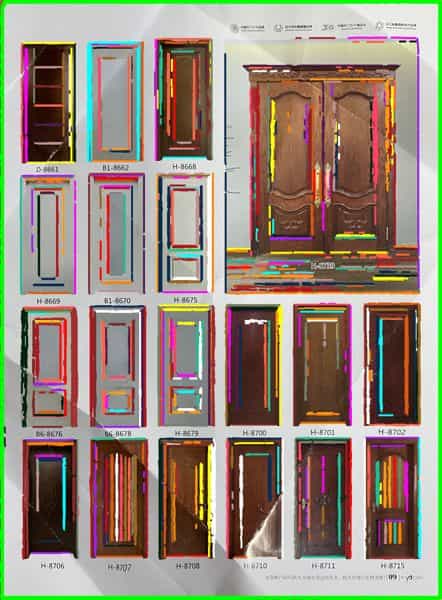}\\
        \vspace{.02\textwidth}%
        \includegraphics[width=1\textwidth, height=1.414\textwidth]{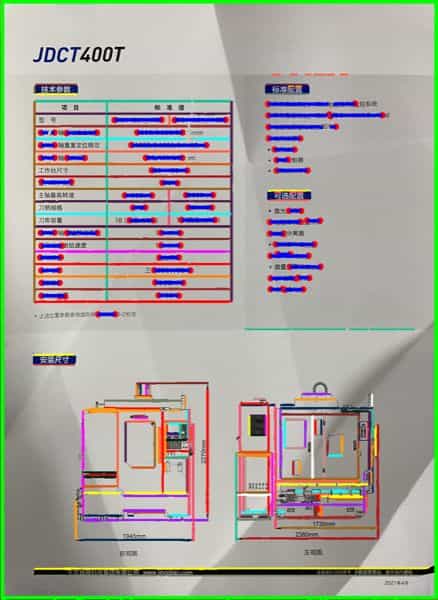}\\
        \vspace{.02\textwidth}%
        \includegraphics[width=1\textwidth, height=1.414\textwidth]{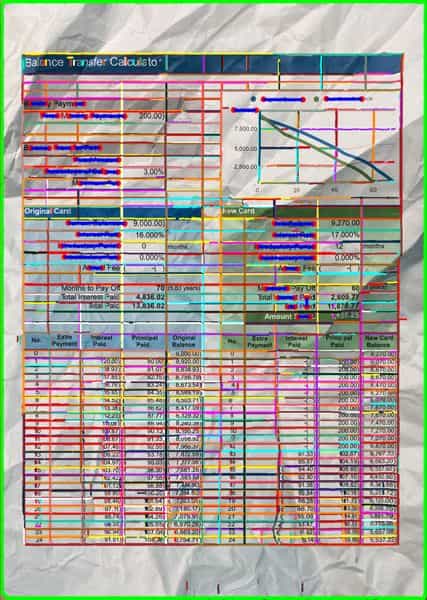}\\
        \vspace{.02\textwidth}%
        \includegraphics[width=1\textwidth, height=1.414\textwidth]{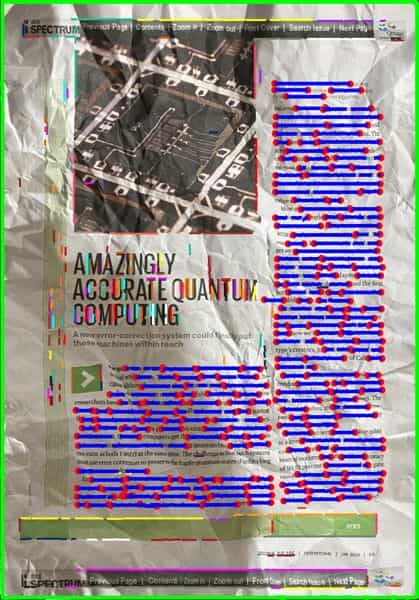}
    \end{minipage}}\hspace{.01\textwidth}%
    \subfigure[]{\centering
    \begin{minipage}[b]{.09\textwidth}\centering
        \includegraphics[width=1\textwidth, height=1.414\textwidth]{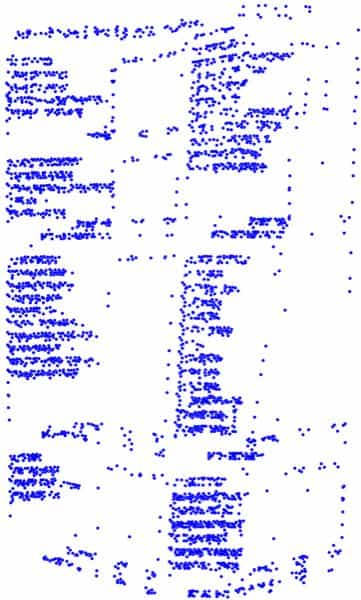}\\
        \vspace{.02\textwidth}%
        \includegraphics[width=1\textwidth, height=1.414\textwidth]{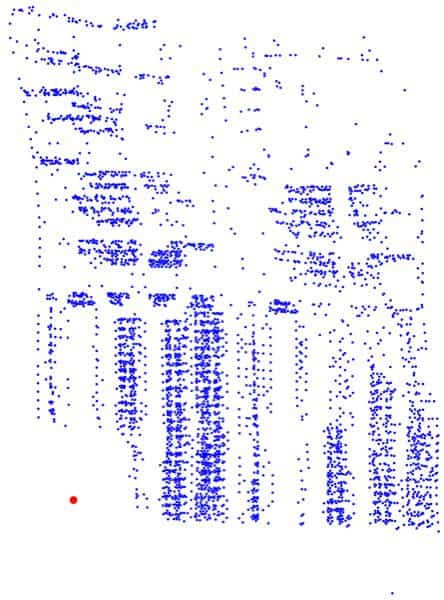}\\
        \vspace{.02\textwidth}%
        \includegraphics[width=1\textwidth, height=1.414\textwidth]{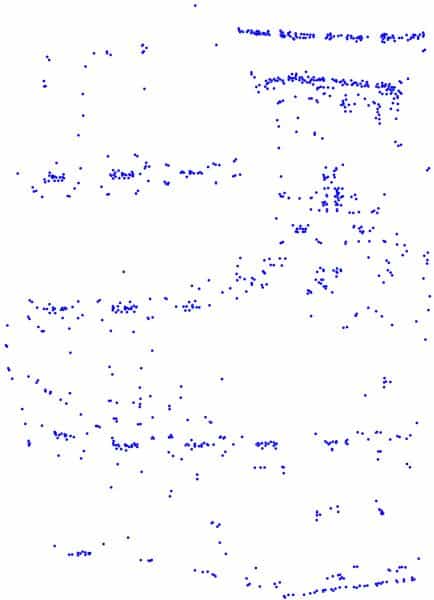}\\
        \vspace{.02\textwidth}%
        \includegraphics[width=1\textwidth, height=1.414\textwidth]{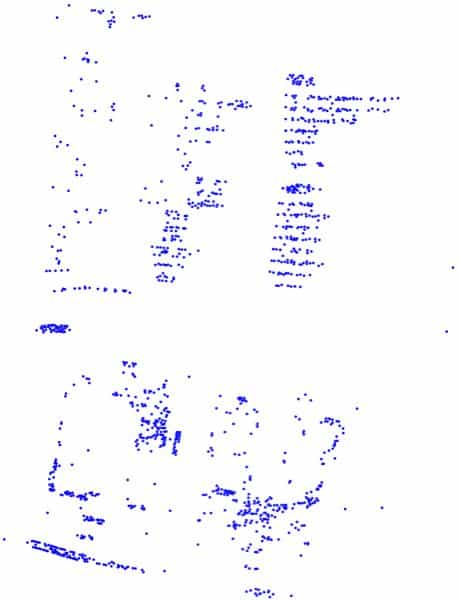}\\
        \vspace{.02\textwidth}%
        \includegraphics[width=1\textwidth, height=1.414\textwidth]{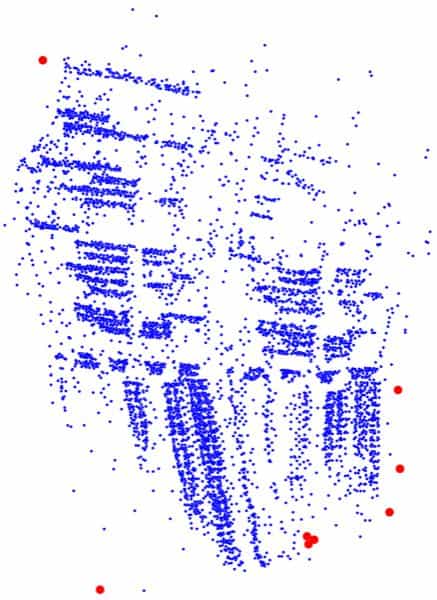}\\
        \vspace{.02\textwidth}%
        \includegraphics[width=1\textwidth, height=1.414\textwidth]{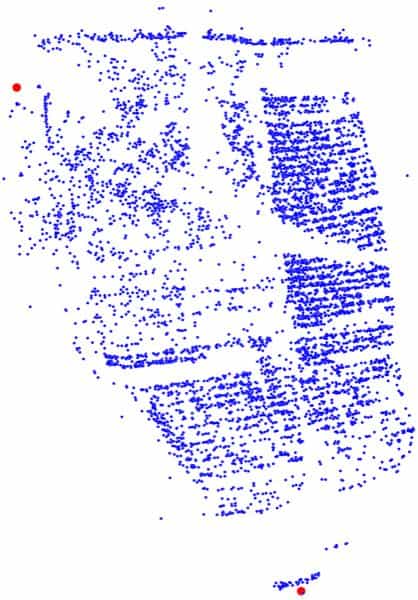}
    \end{minipage}}\hspace{.01\textwidth}%
    \subfigure[]{\centering
    \begin{minipage}[b]{.09\textwidth}\centering
        \includegraphics[width=1\textwidth, height=1.414\textwidth]{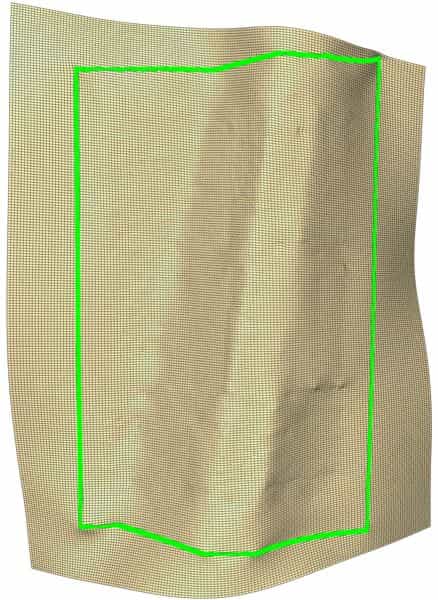}\\
        \vspace{.02\textwidth}%
        \includegraphics[width=1\textwidth, height=1.414\textwidth]{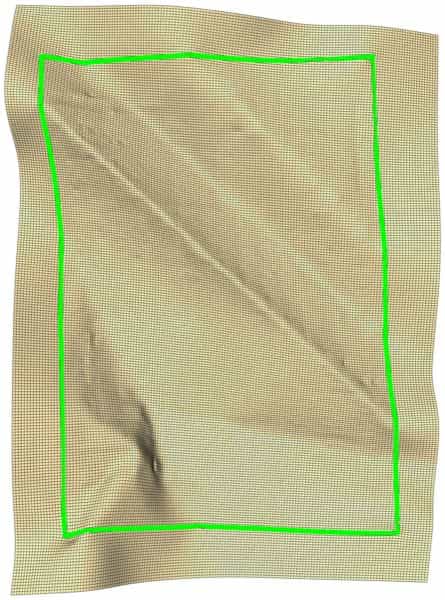}\\
        \vspace{.02\textwidth}%
        \includegraphics[width=1\textwidth, height=1.414\textwidth]{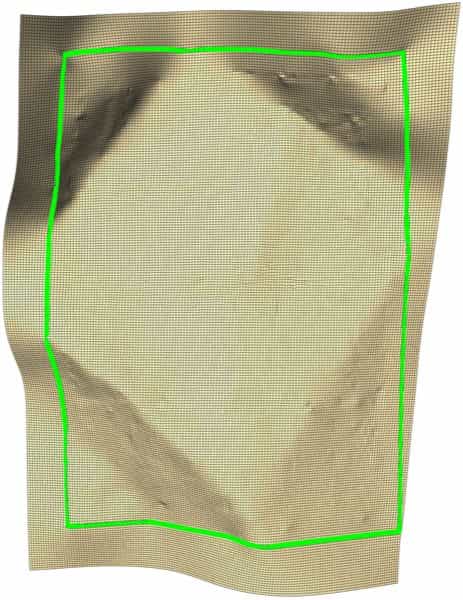}\\
        \vspace{.02\textwidth}%
        \includegraphics[width=1\textwidth, height=1.414\textwidth]{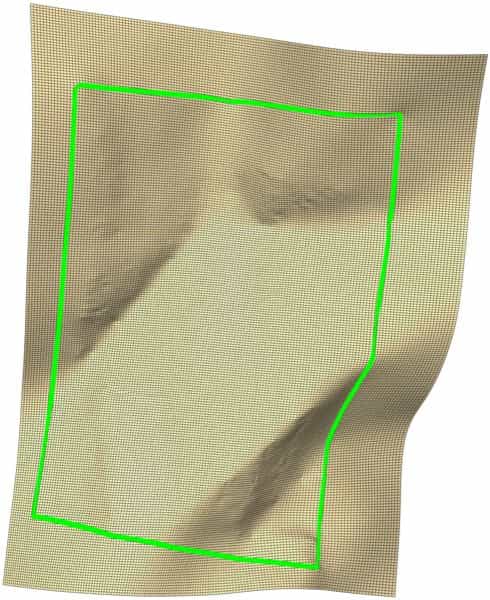}\\
        \vspace{.02\textwidth}%
        \includegraphics[width=1\textwidth, height=1.414\textwidth]{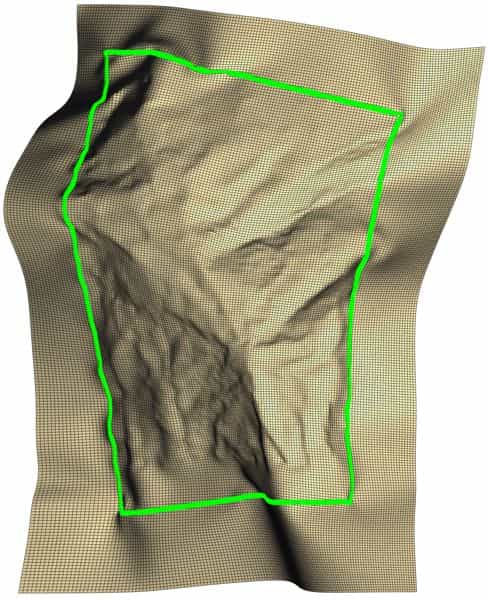}\\
        \vspace{.02\textwidth}%
        \includegraphics[width=1\textwidth, height=1.414\textwidth]{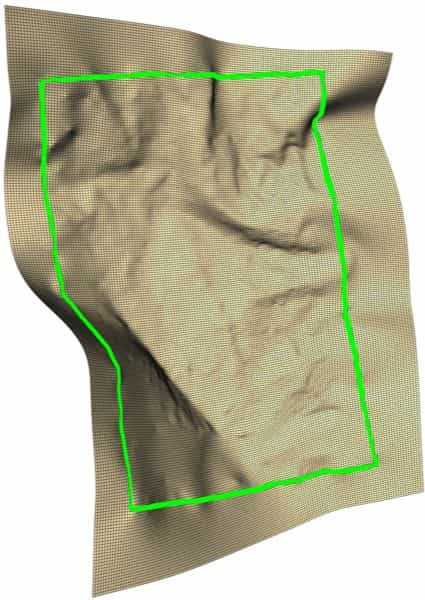}
    \end{minipage}}
\vspace{-0.015\textwidth}
\caption{Dataset \uppercase\expandafter{\romannumeral1}. (a) Original images. (b) Results of  DewarpNet \cite{das2019dewarpnet}. (c) Results of FCN-based \cite{xie2021dewarping}. (d) Results of Points-based~\cite{xie2022document}. (e) Results of DocTr \cite{feng2021doctr}. (f) Results of DocScanner \cite{feng2021docscanner}. (g) Our results. (h) Our results with feature lines. (i) Our final $\mc{P}$. 
 (j) Our final $\mc{M}$ with boundary.}\label{f:results1}
\end{figure}
\begin{figure}[!t]
\centering
    \subfigure[]{\centering 
    \begin{minipage}[b]{.09\textwidth}\centering
        \includegraphics[width=1\textwidth, height=1.75\textwidth]{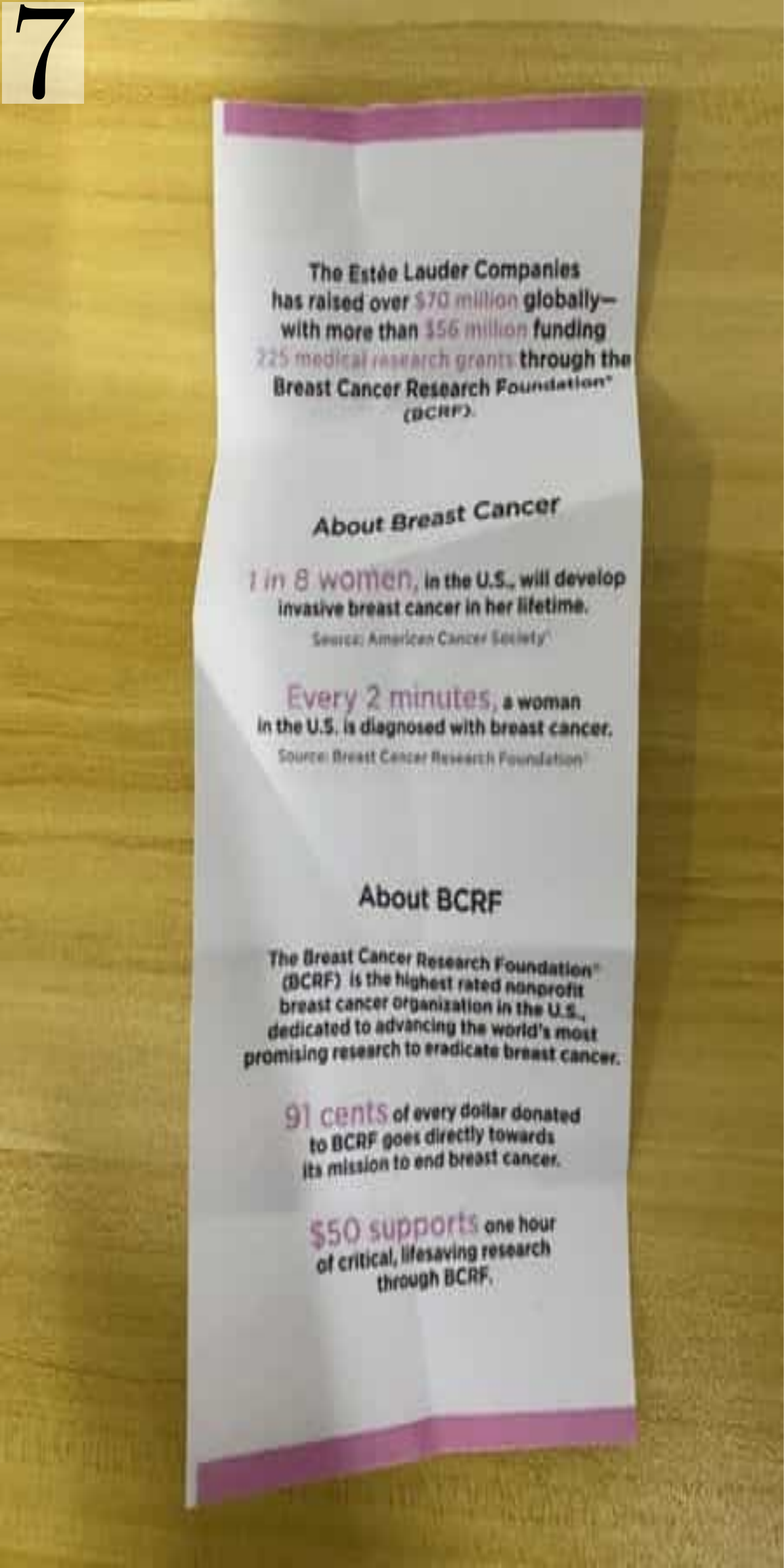}\\
        \vspace{.02\textwidth}%
       \includegraphics[width=1\textwidth, height=1.75\textwidth]{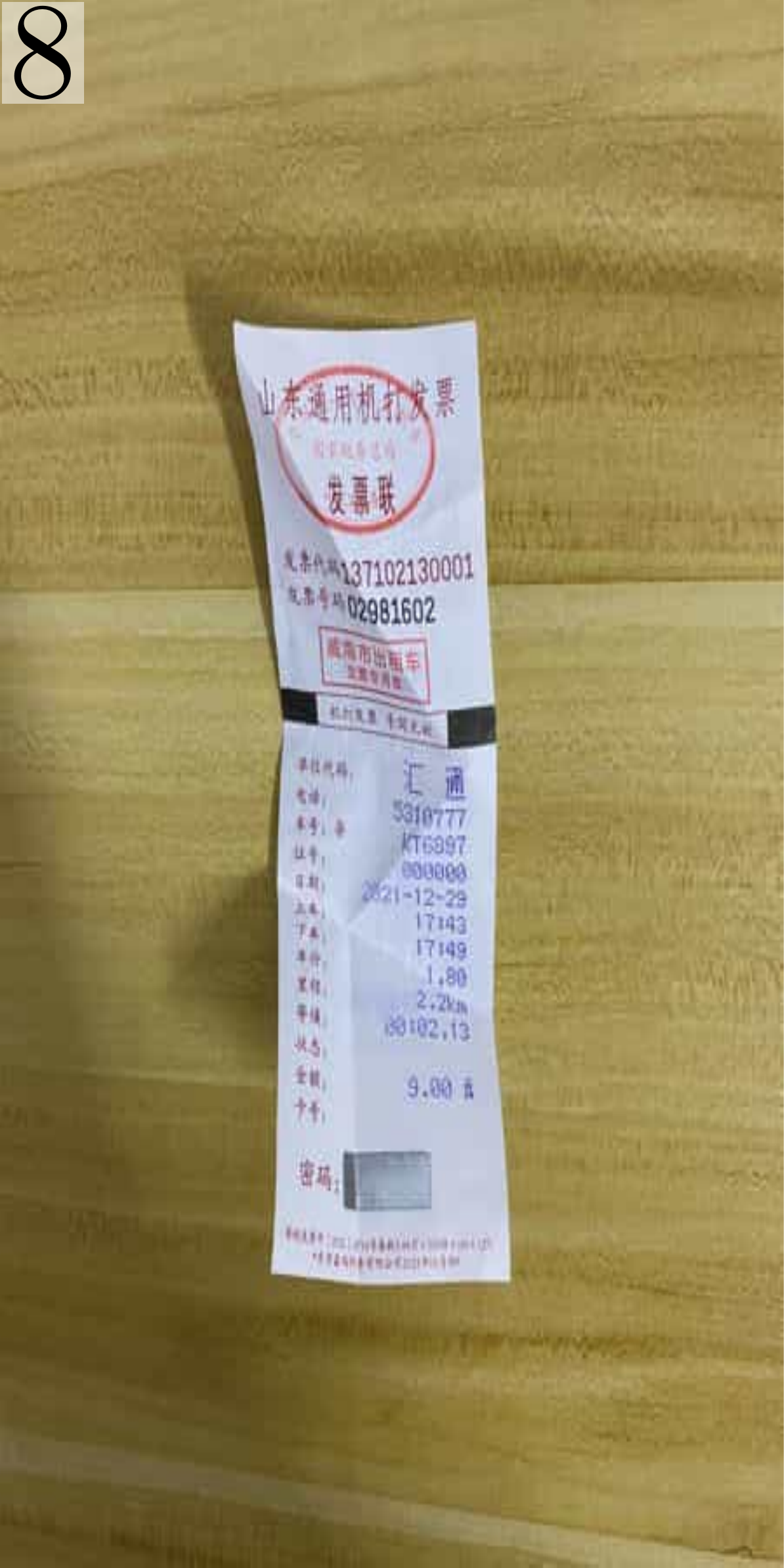}\\
        \vspace{.02\textwidth}%
       \includegraphics[width=1\textwidth, height=1.75\textwidth]{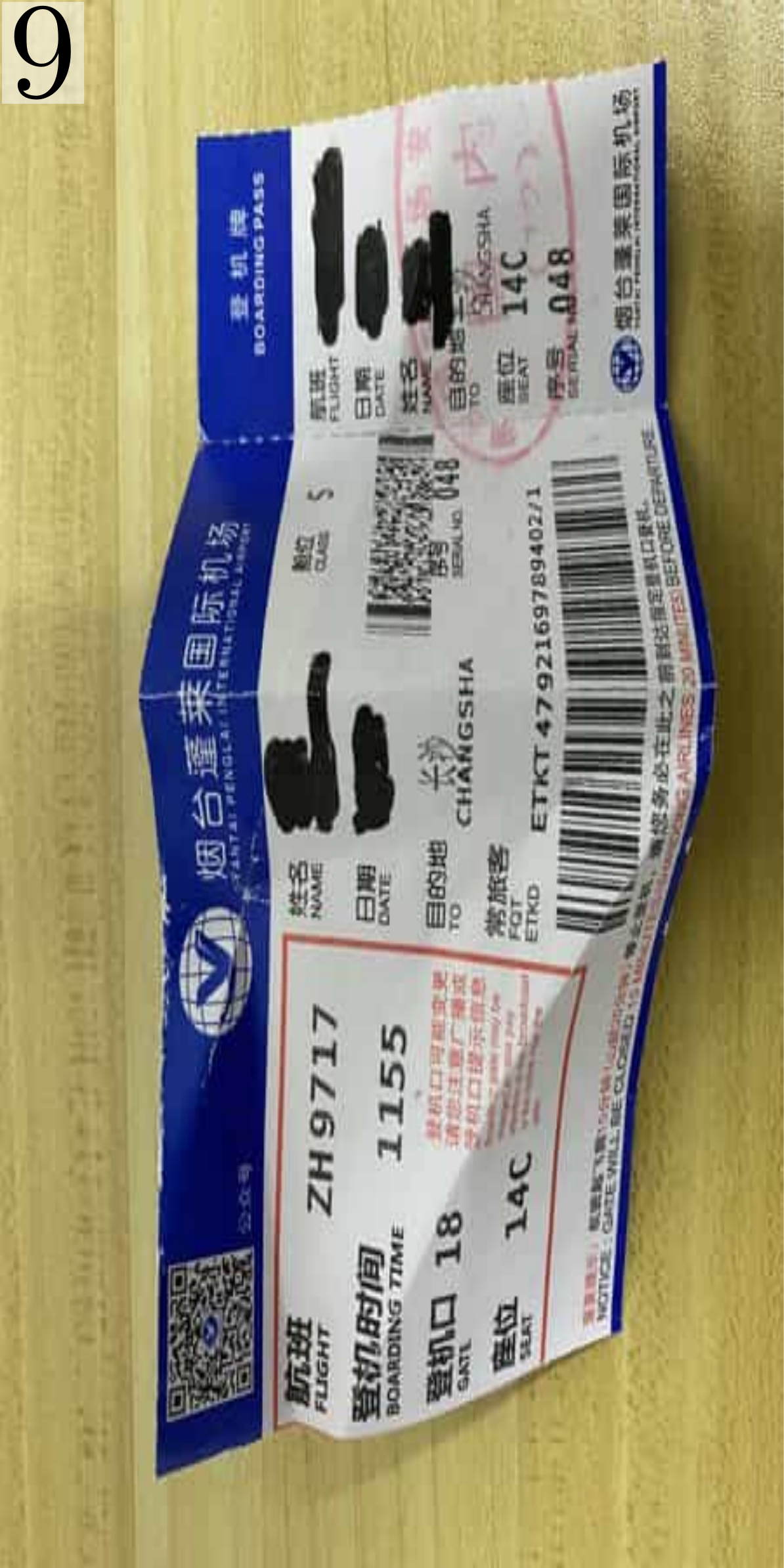}\\
        \vspace{.02\textwidth}%
       \includegraphics[width=1\textwidth, height=1.4524\textwidth]{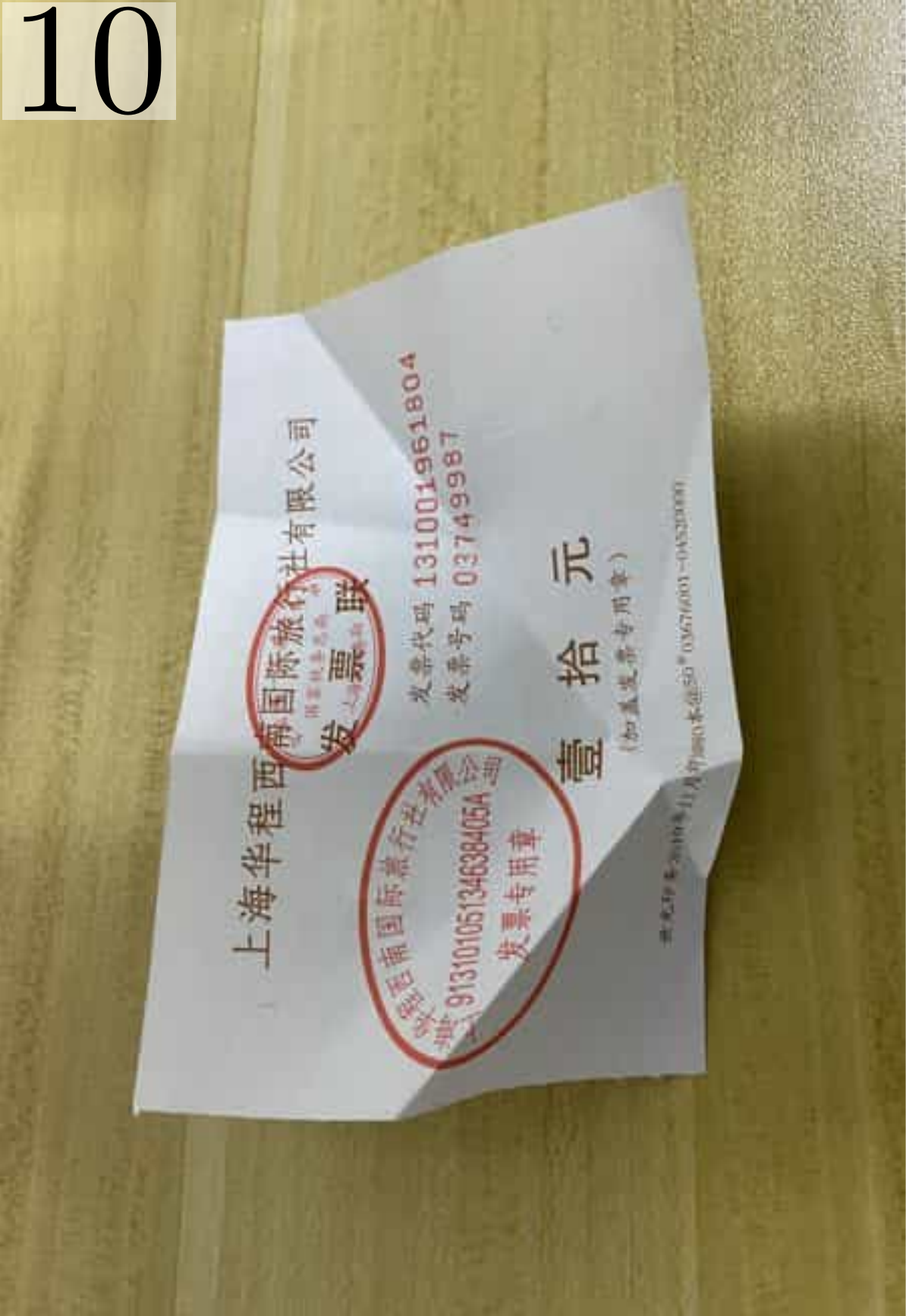}\\
        \vspace{.02\textwidth}%
       \includegraphics[width=1\textwidth, height=1.75\textwidth]{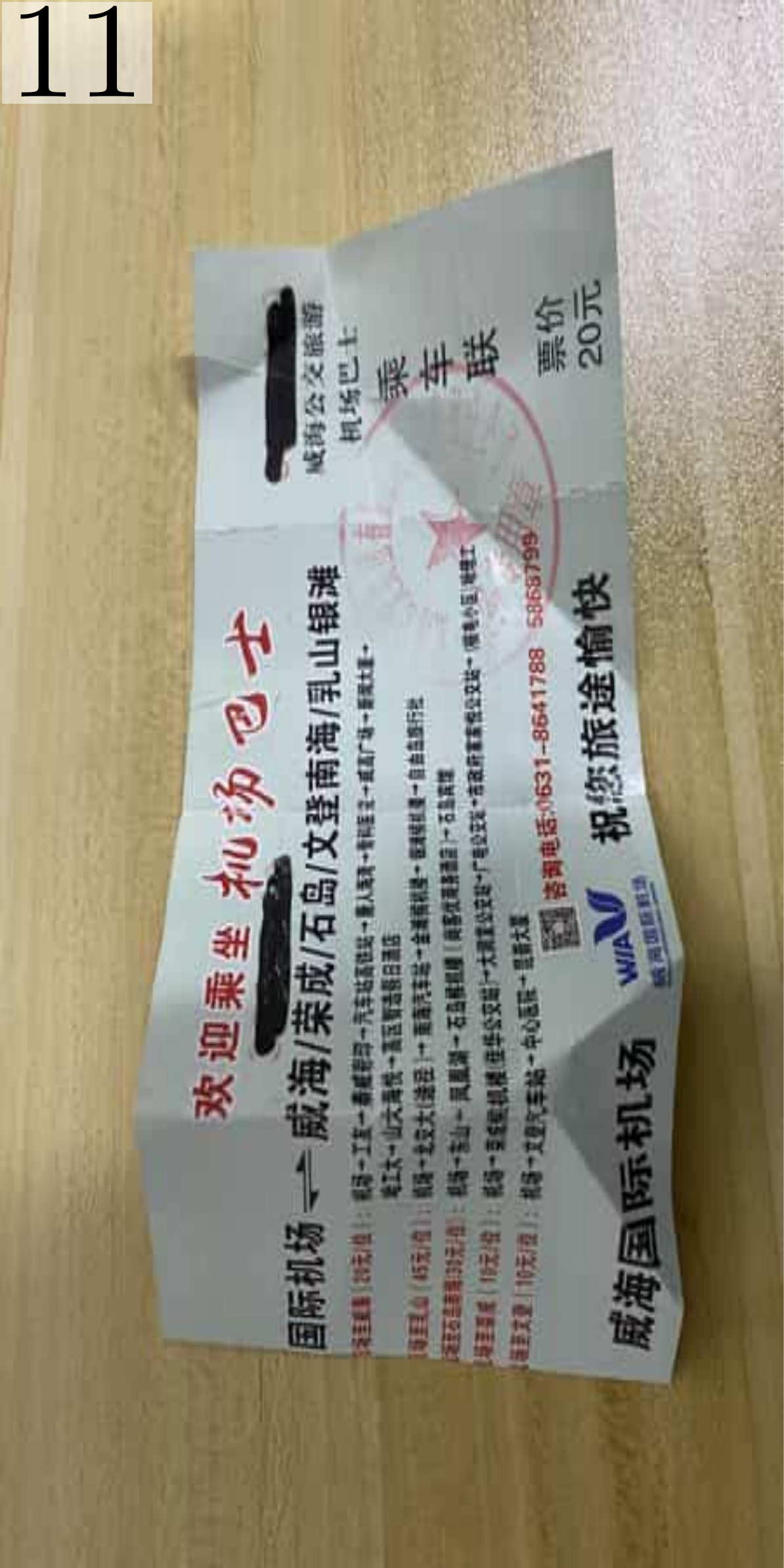}\\
        \vspace{.02\textwidth}%
       \includegraphics[width=1\textwidth, height=1.6\textwidth]{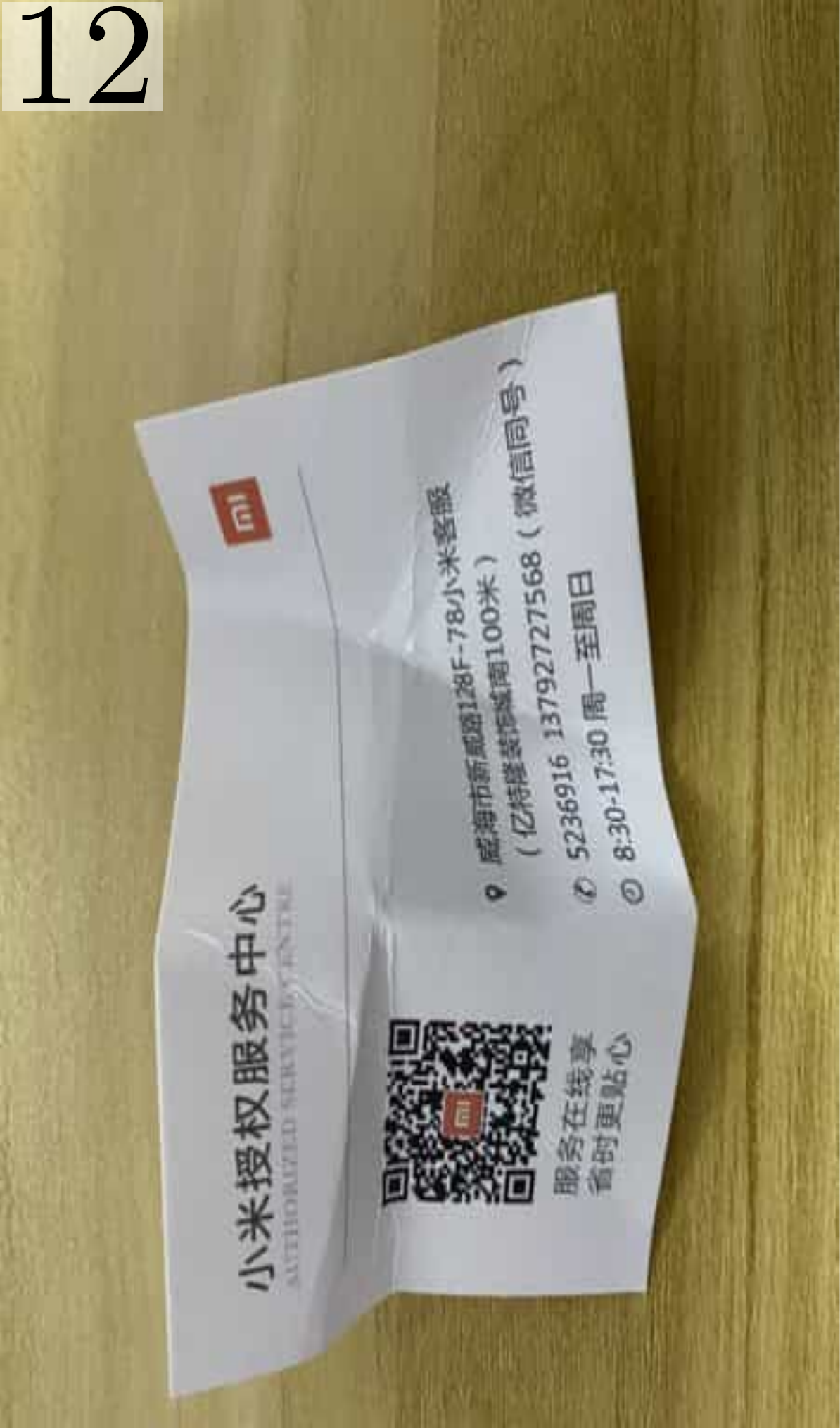}\\
        \vspace{.02\textwidth}%
       \includegraphics[width=1\textwidth, height=1.75\textwidth]{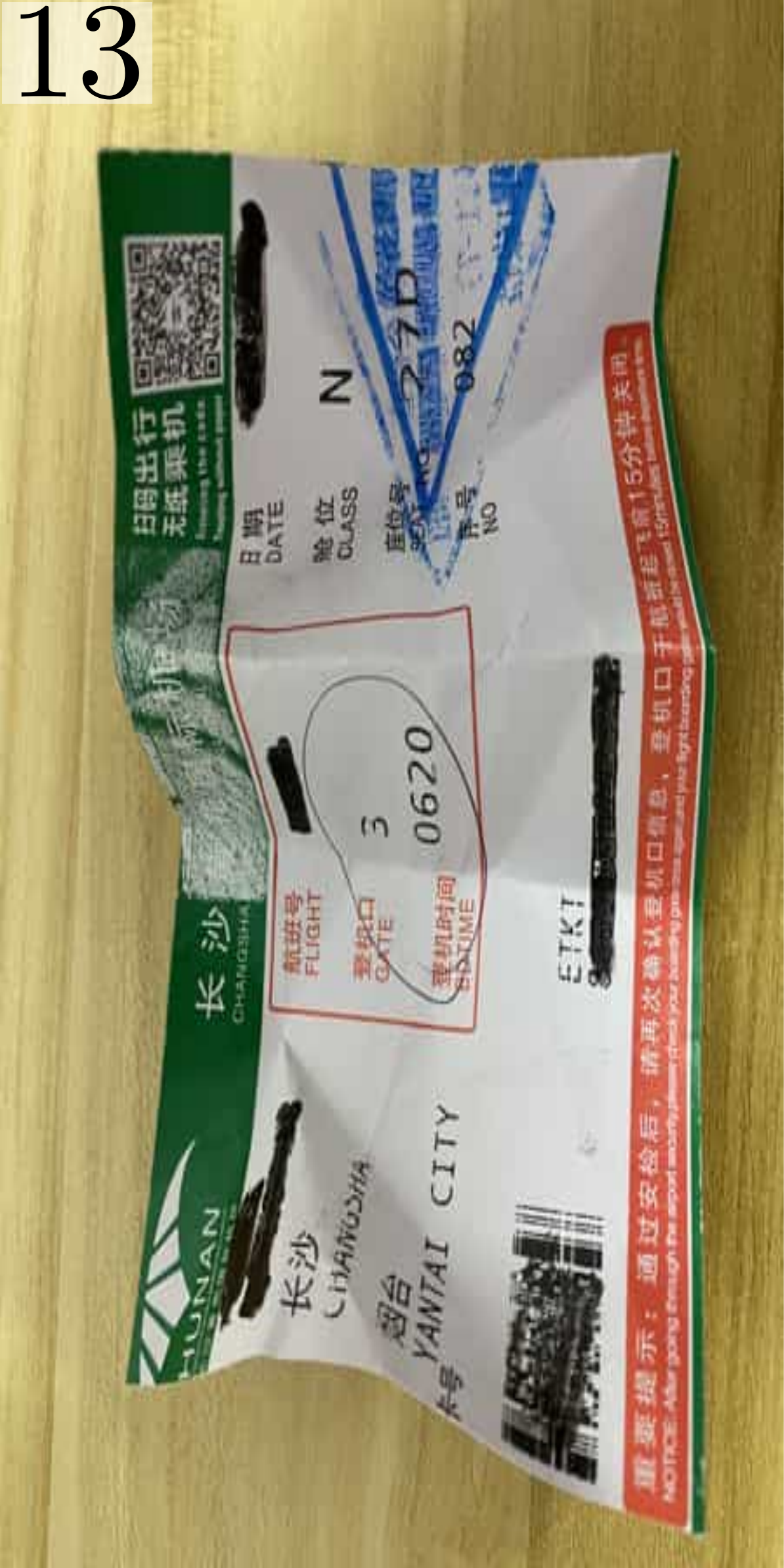}
    \end{minipage}}\hspace{.01\textwidth}%
    \subfigure[]{\centering
    \begin{minipage}[b]{.09\textwidth}\centering
        \includegraphics[width=.938\textwidth, height=1.75\textwidth]{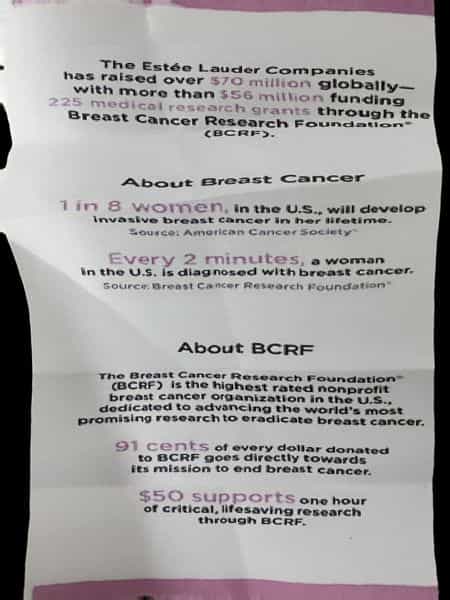}\\
        \vspace{.02\textwidth}%
        \includegraphics[width=.71846847\textwidth, height=1.75\textwidth]{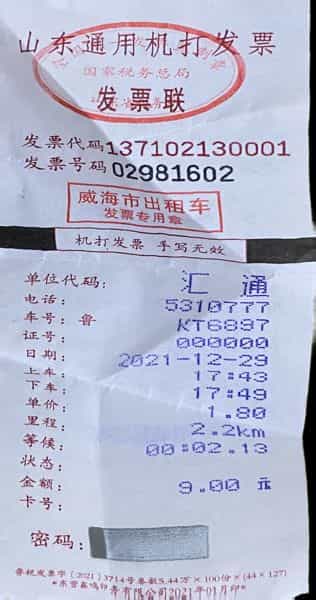}\\
        \vspace{.02\textwidth}%
        \includegraphics[angle=90,width=1\textwidth, height=1.75\textwidth]{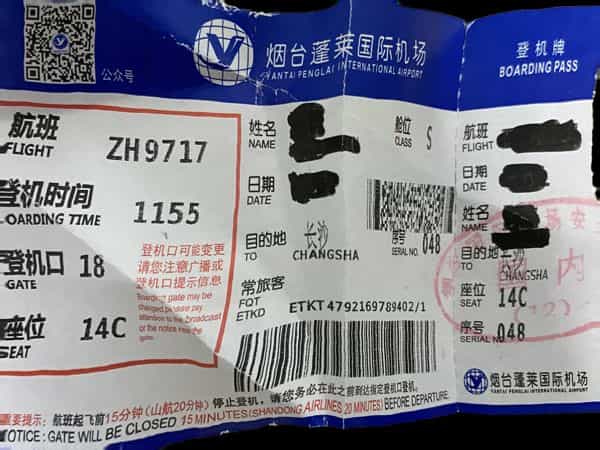}\\
        \vspace{.02\textwidth}%
        \includegraphics[angle=90,width=1\textwidth, height=1.4524\textwidth]{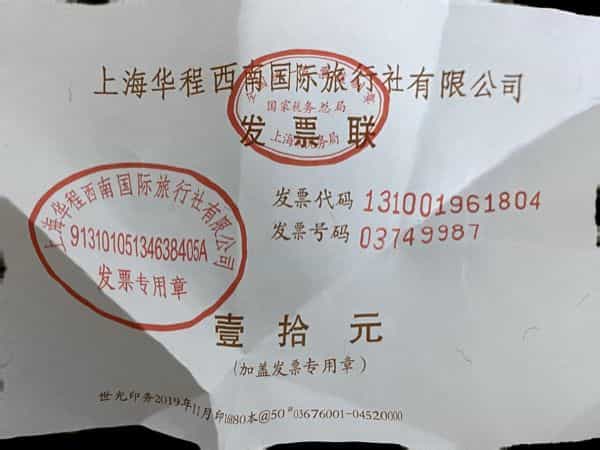}\\
        \vspace{.02\textwidth}%
        \includegraphics[angle=90,width=1\textwidth, height=1.75\textwidth]{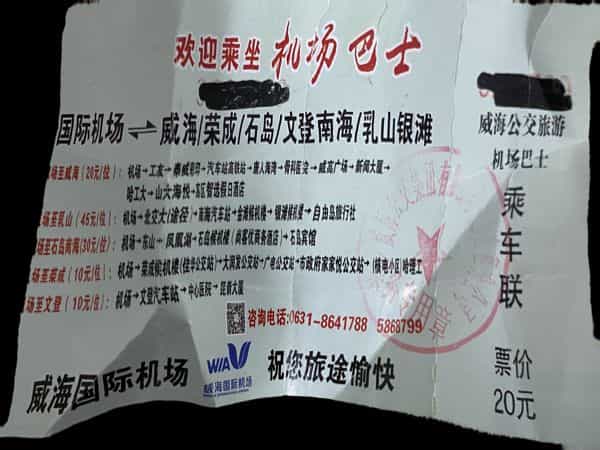}\\
        \vspace{.02\textwidth}%
        \includegraphics[angle=90,width=1\textwidth, height=1.6\textwidth]{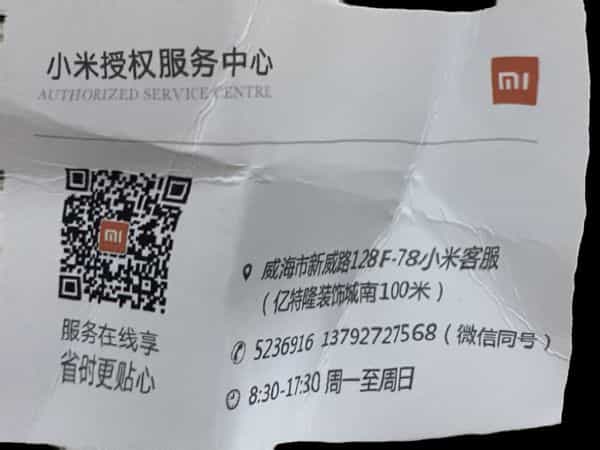}\\
        \vspace{.02\textwidth}%
        \includegraphics[angle=90,width=1\textwidth, height=1.75\textwidth]{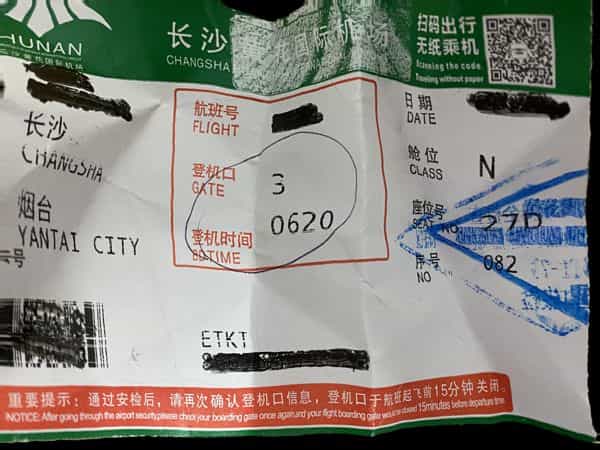}
    \end{minipage}}\hspace{.01\textwidth}%
    \subfigure[]{\centering
    \begin{minipage}[b]{.09\textwidth}\centering
        \includegraphics[width=.938\textwidth, height=1.75\textwidth]{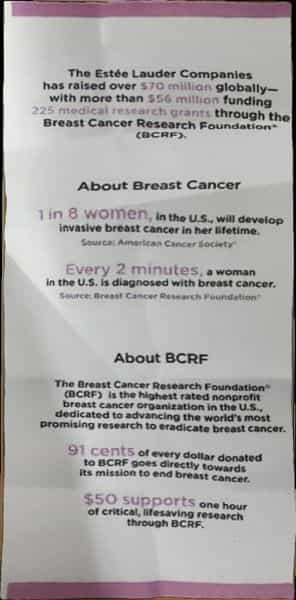}\\
        \vspace{.02\textwidth}%
        \includegraphics[width=.71846847\textwidth, height=1.75\textwidth]{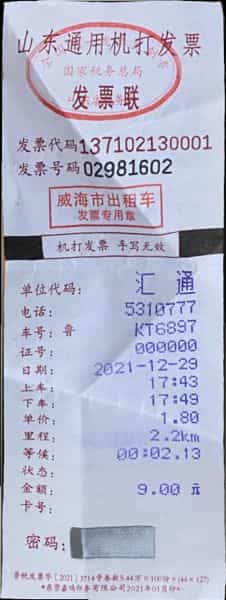}\\
        \vspace{.02\textwidth}%
        \includegraphics[angle=90,width=1\textwidth, height=1.75\textwidth]{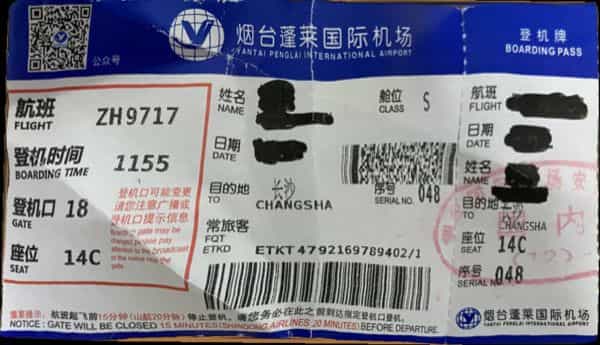}\\
        \vspace{.02\textwidth}%
        \includegraphics[angle=90,width=1\textwidth, height=1.4524\textwidth]{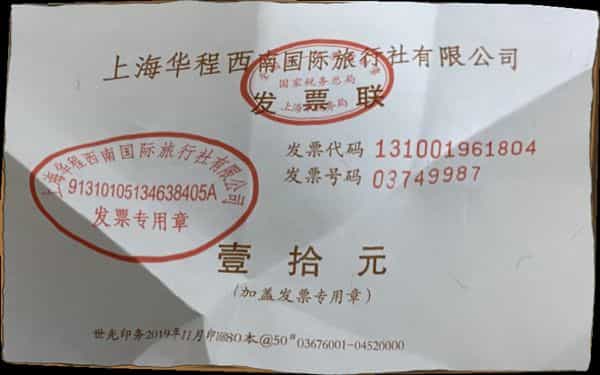}\\
        \vspace{.02\textwidth}%
        \includegraphics[angle=90,width=1\textwidth, height=1.75\textwidth]{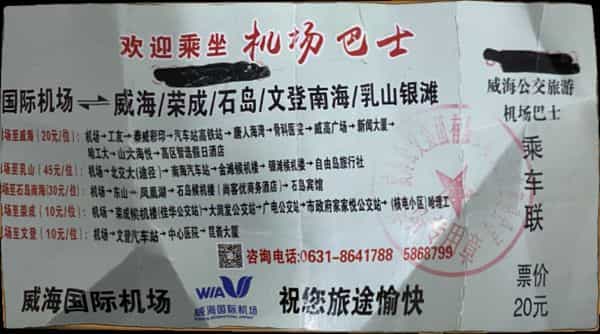}\\
        \vspace{.02\textwidth}%
        \includegraphics[angle=90,width=1\textwidth, height=1.6\textwidth]{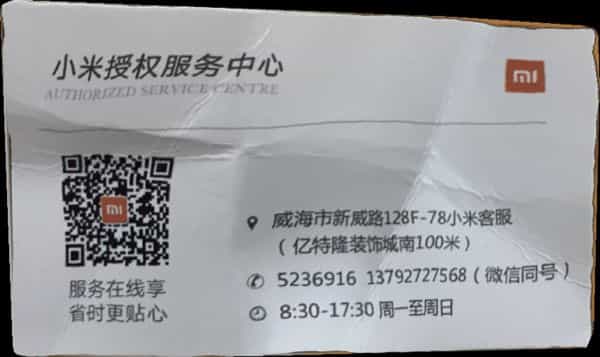}\\
        \vspace{.02\textwidth}%
        \includegraphics[angle=90,width=1\textwidth, height=1.75\textwidth]{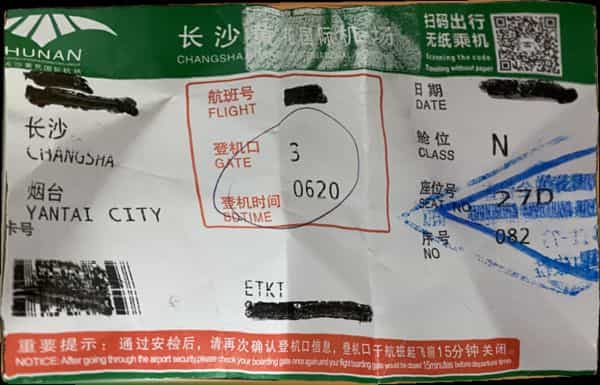}
    \end{minipage}}\hspace{.01\textwidth}%
    \subfigure[]{\centering
    \begin{minipage}[b]{.09\textwidth}\centering
        \includegraphics[width=.938\textwidth, height=1.75\textwidth]{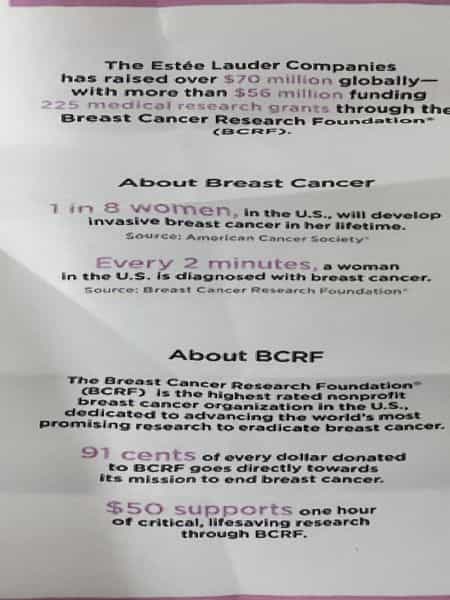}\\
        \vspace{.02\textwidth}%
        \includegraphics[width=.71846847\textwidth, height=1.75\textwidth]{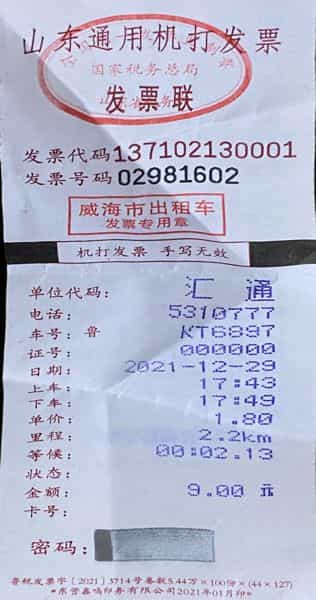}\\
        \vspace{.02\textwidth}%
        \includegraphics[angle=90,width=1\textwidth, height=1.75\textwidth]{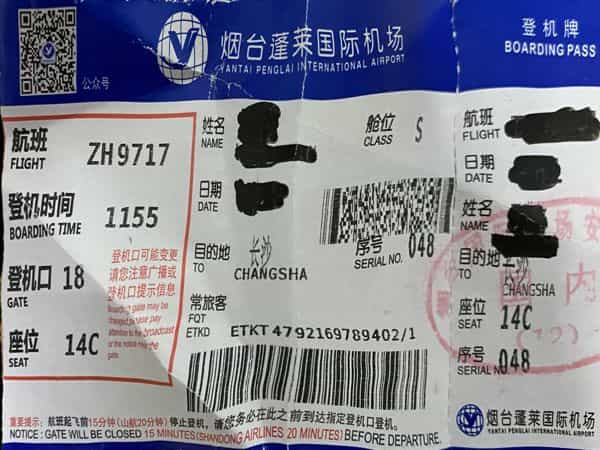}\\
        \vspace{.02\textwidth}%
        \includegraphics[angle=90,width=1\textwidth, height=1.4524\textwidth]{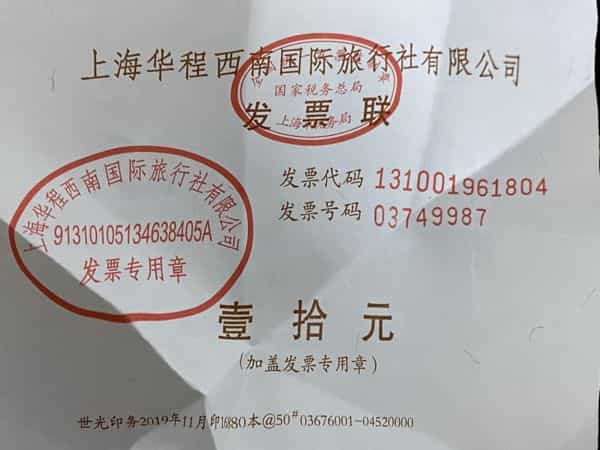}\\
        \vspace{.02\textwidth}%
        \includegraphics[angle=90,width=1\textwidth, height=1.75\textwidth]{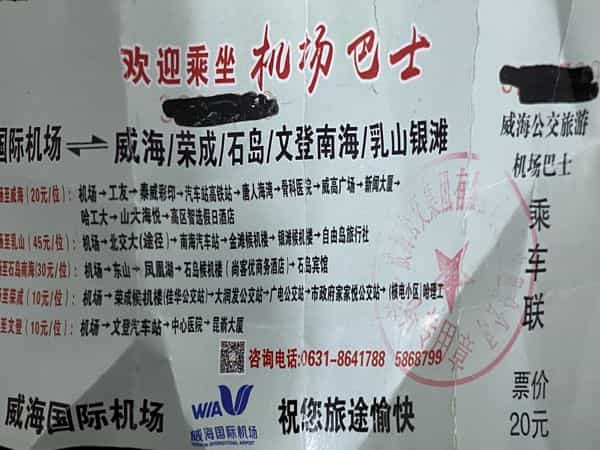}\\
        \vspace{.02\textwidth}%
        \includegraphics[angle=90,width=1\textwidth, height=1.6\textwidth]{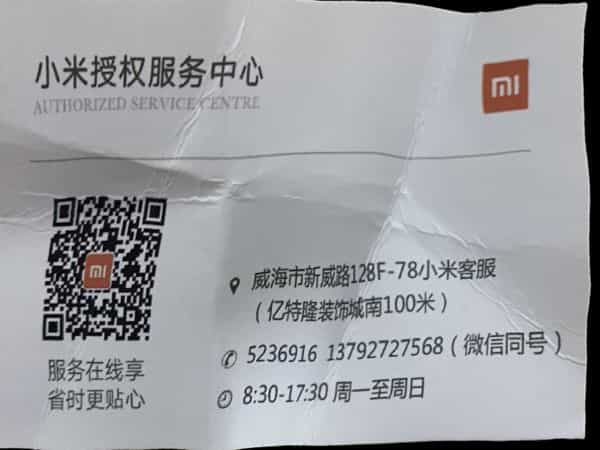}\\
        \vspace{.02\textwidth}%
        \includegraphics[angle=90,width=1\textwidth, height=1.75\textwidth]{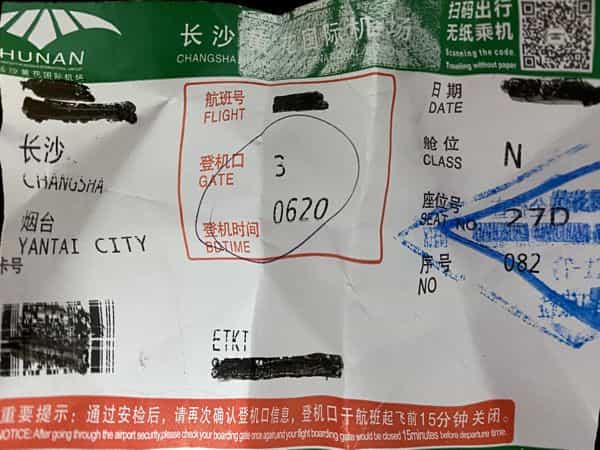}
    \end{minipage}}\hspace{.01\textwidth}%
    \subfigure[]{\centering
    \begin{minipage}[b]{.09\textwidth}\centering
        \includegraphics[width=.938\textwidth, height=1.75\textwidth]{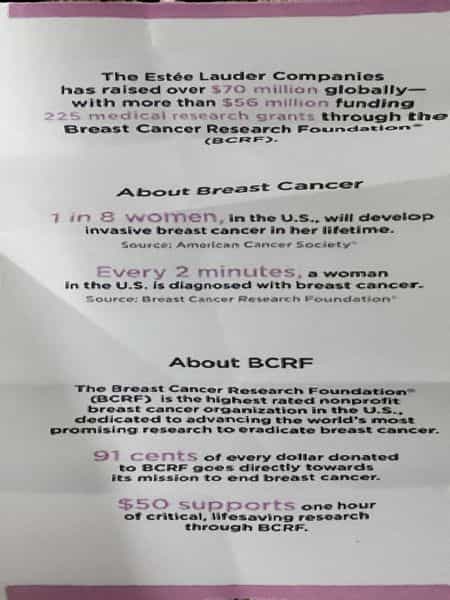}\\
        \vspace{.02\textwidth}%
        \includegraphics[width=.71846847\textwidth, height=1.75\textwidth]{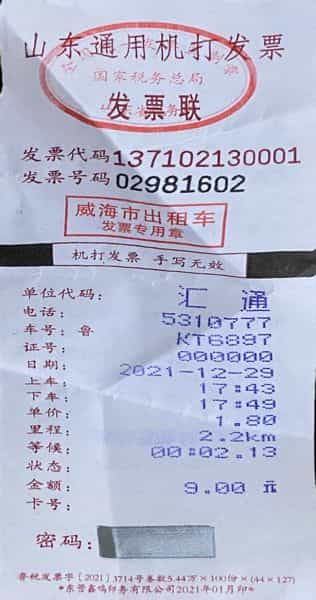}\\
        \vspace{.02\textwidth}%
        \includegraphics[angle=90,width=1\textwidth, height=1.75\textwidth]{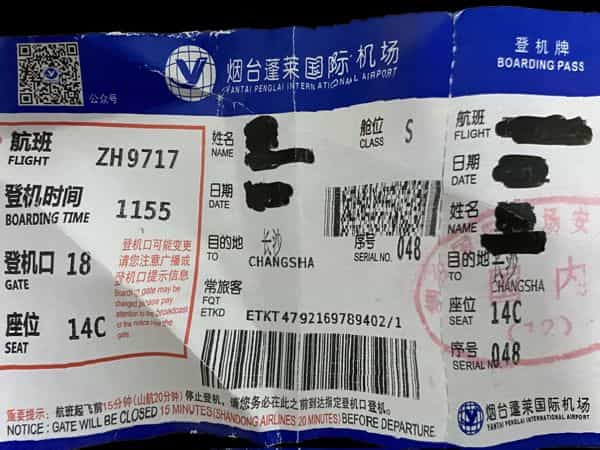}\\
        \vspace{.02\textwidth}%
        \includegraphics[angle=90,width=1\textwidth, height=1.4524\textwidth]{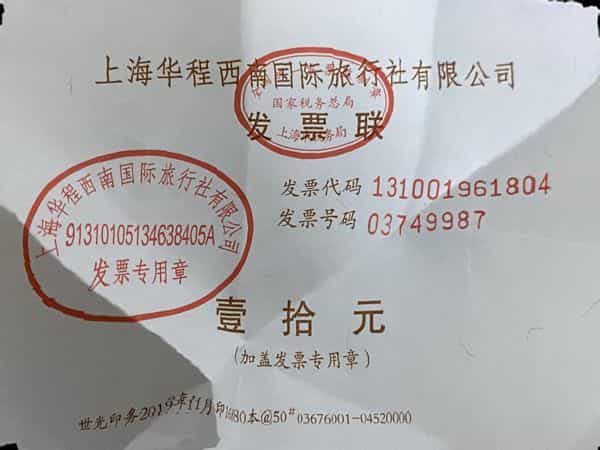}\\
        \vspace{.02\textwidth}%
        \includegraphics[angle=90,width=1\textwidth, height=1.75\textwidth]{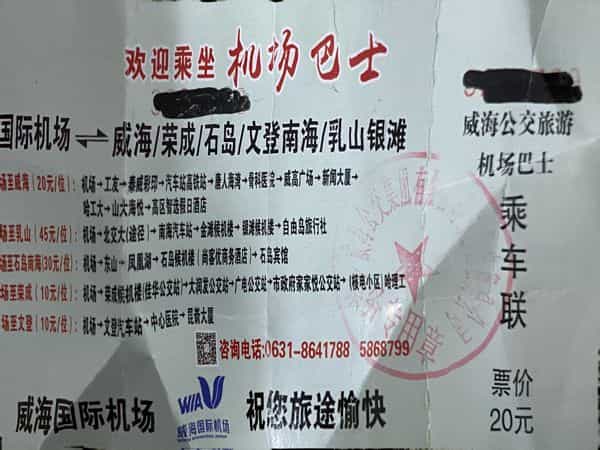}\\
        \vspace{.02\textwidth}%
        \includegraphics[angle=90,width=1\textwidth, height=1.6\textwidth]{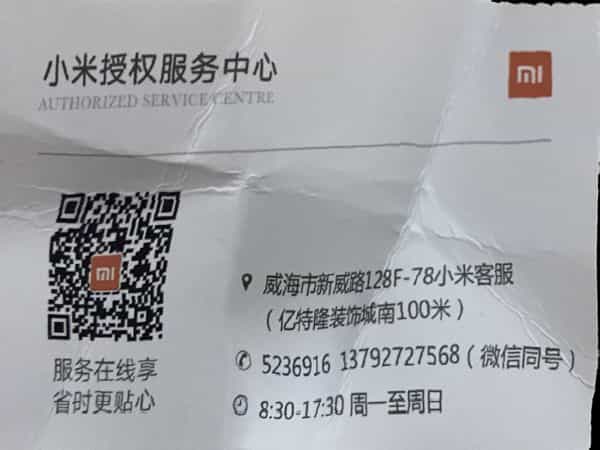}\\
        \vspace{.02\textwidth}%
        \includegraphics[angle=90,width=1\textwidth, height=1.75\textwidth]{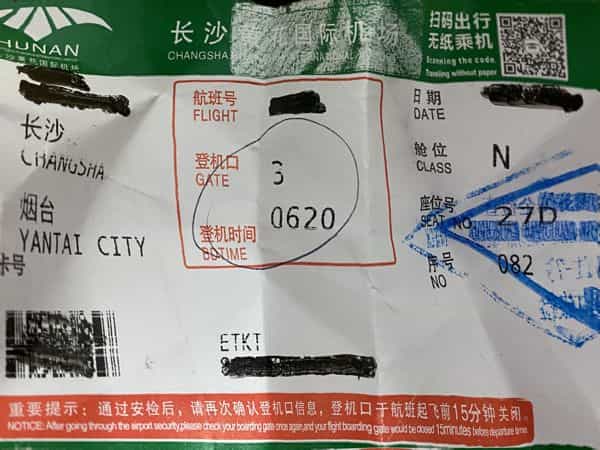}
    \end{minipage}}\hspace{.01\textwidth}%
    \subfigure[]{\centering
    \begin{minipage}[b]{.09\textwidth}\centering
        \includegraphics[width=.938\textwidth, height=1.75\textwidth]{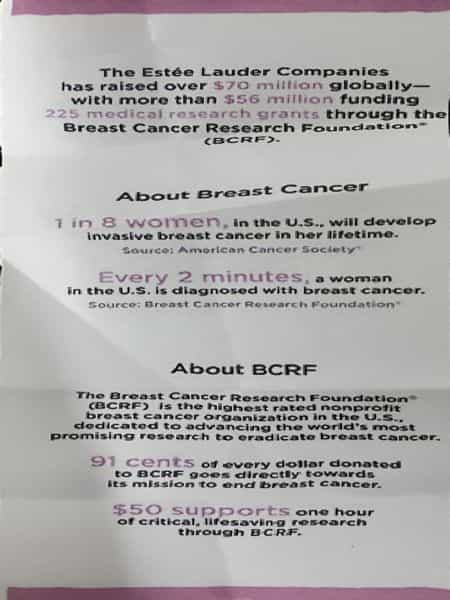}\\
        \vspace{.02\textwidth}%
        \includegraphics[width=.71846847\textwidth, height=1.75\textwidth]{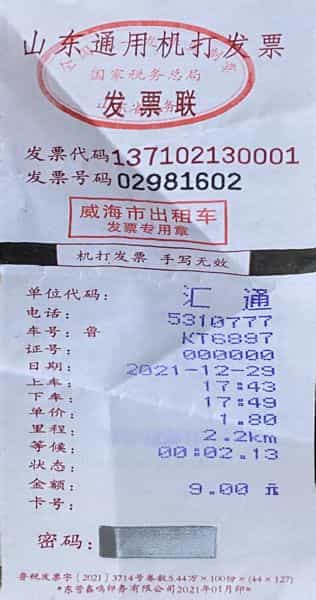}\\
        \vspace{.02\textwidth}%
        \includegraphics[angle=90,width=1\textwidth, height=1.75\textwidth]{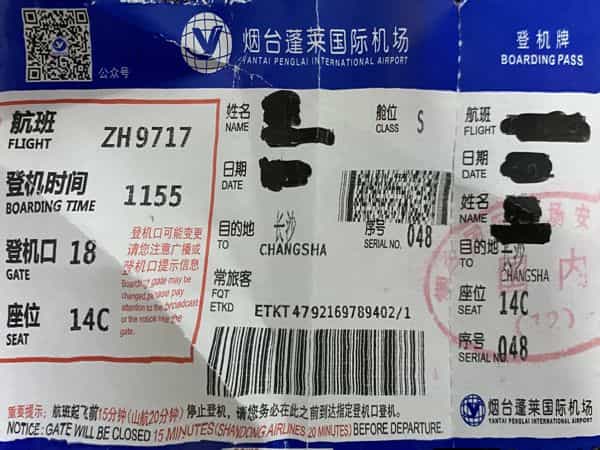}\\
        \vspace{.02\textwidth}%
        \includegraphics[angle=90,width=1\textwidth, height=1.4524\textwidth]{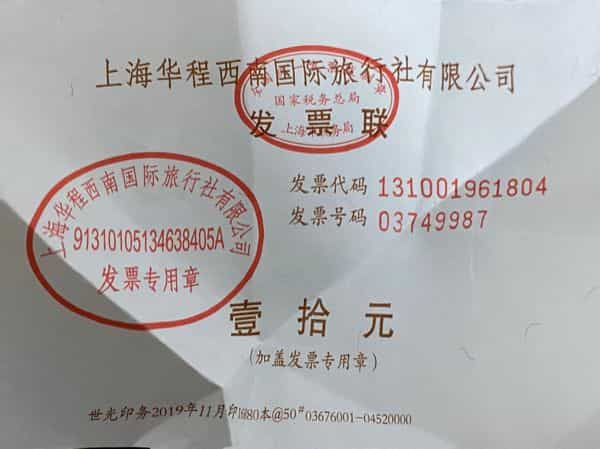}\\
        \vspace{.02\textwidth}%
        \includegraphics[angle=90,width=1\textwidth, height=1.75\textwidth]{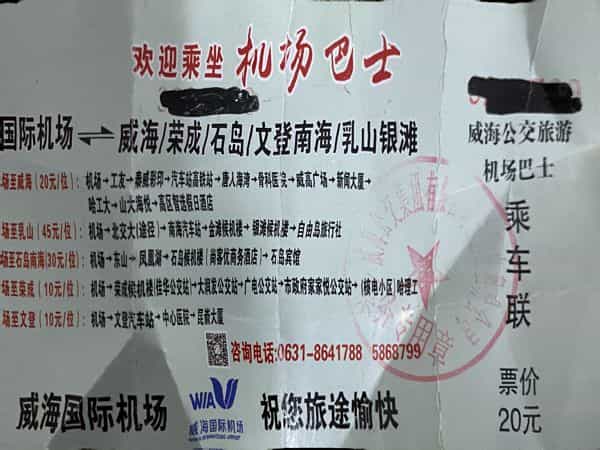}\\
        \vspace{.02\textwidth}%
        \includegraphics[angle=90,width=1\textwidth, height=1.6\textwidth]{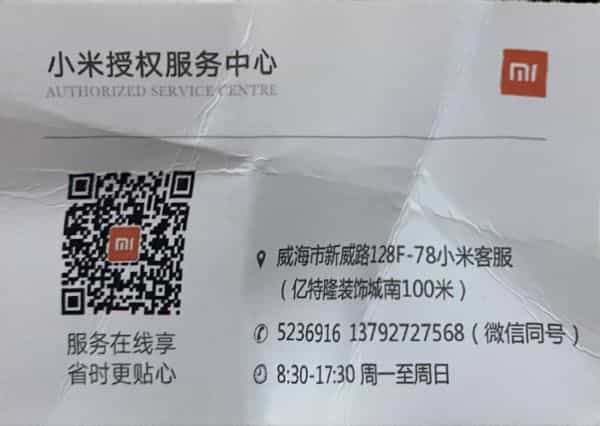}\\
        \vspace{.02\textwidth}%
        \includegraphics[angle=90,width=1\textwidth, height=1.75\textwidth]{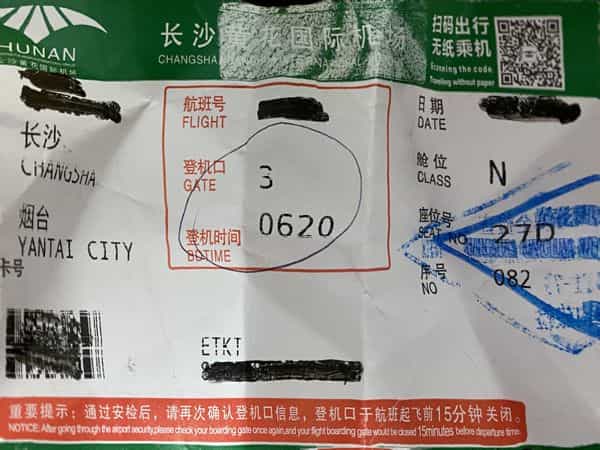}
    \end{minipage}}\hspace{.01\textwidth}%
    \subfigure[]{\centering
    \begin{minipage}[b]{.09\textwidth}\centering
        \includegraphics[width=.938\textwidth, height=1.75\textwidth]{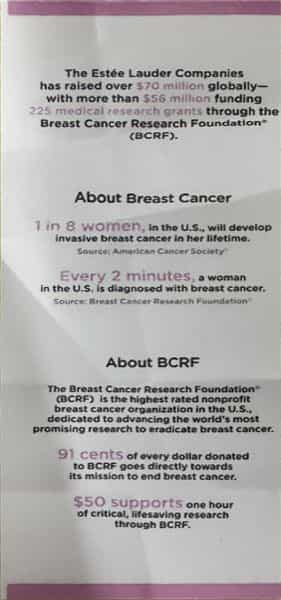}\\
        \vspace{.02\textwidth}%
        \includegraphics[width=.71846847\textwidth, height=1.75\textwidth]{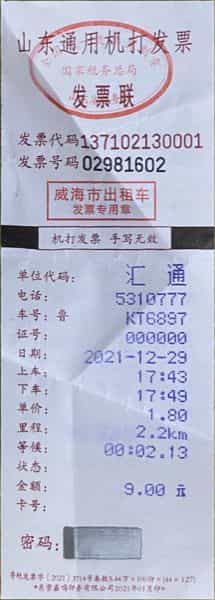}\\
        \vspace{.02\textwidth}%
        \includegraphics[angle=90,width=1\textwidth, height=1.75\textwidth]{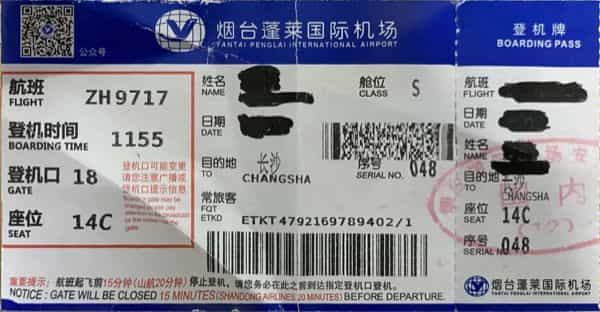}\\
        \vspace{.02\textwidth}%
        \includegraphics[angle=90,width=1\textwidth, height=1.4524\textwidth]{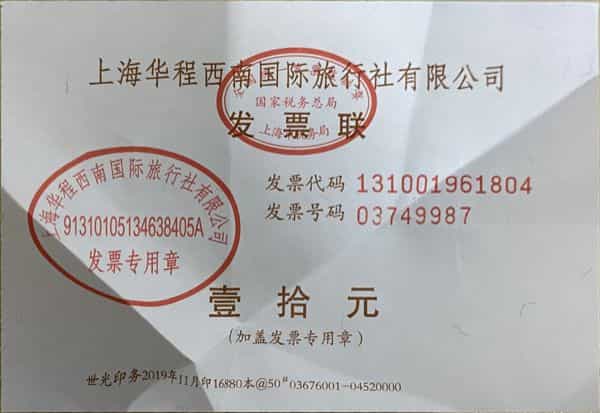}\\
        \vspace{.02\textwidth}%
        \includegraphics[angle=90,width=1\textwidth, height=1.75\textwidth]{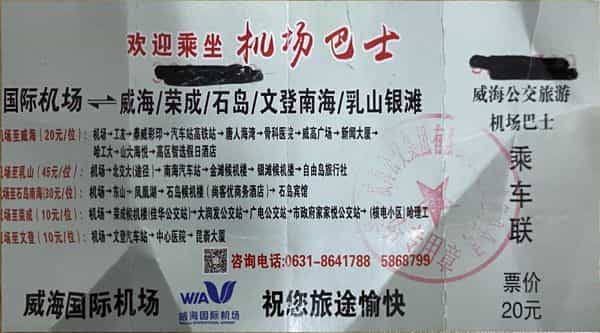}\\
        \vspace{.02\textwidth}%
        \includegraphics[angle=90,width=1\textwidth, height=1.6\textwidth]{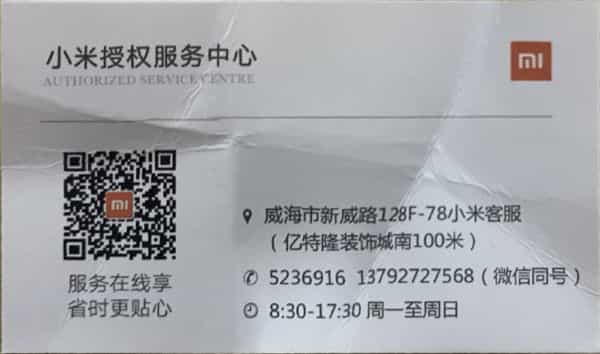}\\
        \vspace{.02\textwidth}%
        \includegraphics[angle=90,width=1\textwidth, height=1.75\textwidth]{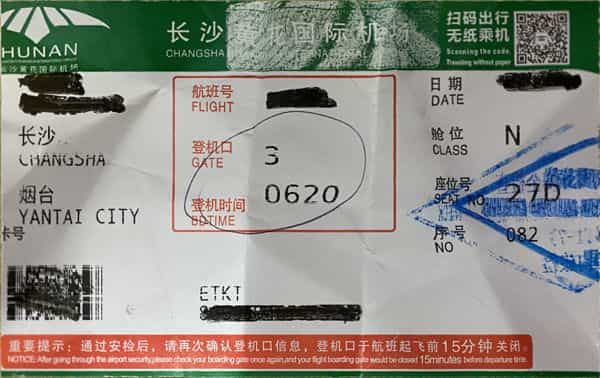}
    \end{minipage}}\hspace{.01\textwidth}%
    \subfigure[]{\centering
    \begin{minipage}[b]{.09\textwidth}\centering
        \includegraphics[width=.938\textwidth, height=1.75\textwidth]{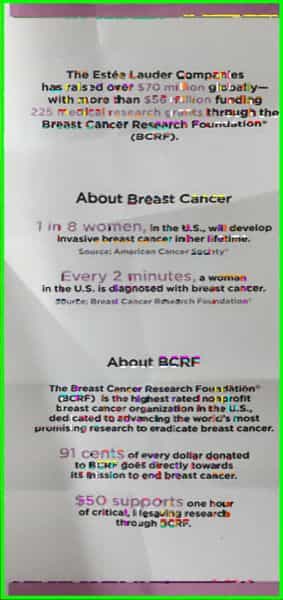}\\
        \vspace{.02\textwidth}%
        \includegraphics[width=.71846847\textwidth, height=1.75\textwidth]{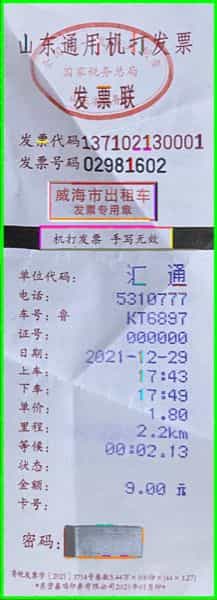}\\
        \vspace{.02\textwidth}%
        \includegraphics[angle=90,width=1\textwidth, height=1.75\textwidth]{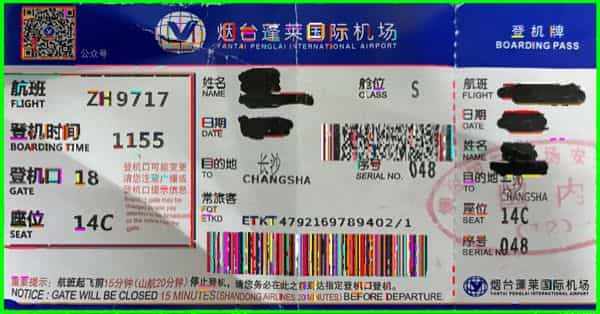}\\
        \vspace{.02\textwidth}%
        \includegraphics[angle=90,width=1\textwidth, height=1.4524\textwidth]{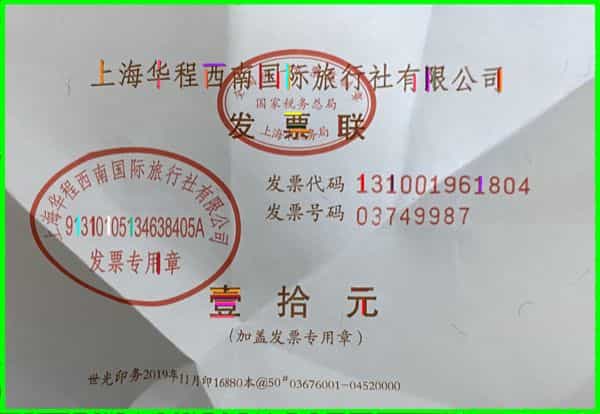}\\
        \vspace{.02\textwidth}%
        \includegraphics[angle=90,width=1\textwidth, height=1.75\textwidth]{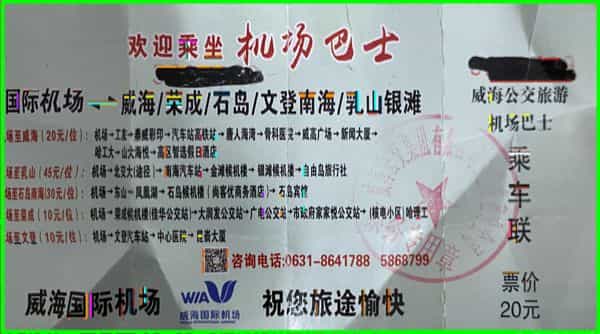}\\
        \vspace{.02\textwidth}%
        \includegraphics[angle=90,width=1\textwidth, height=1.6\textwidth]{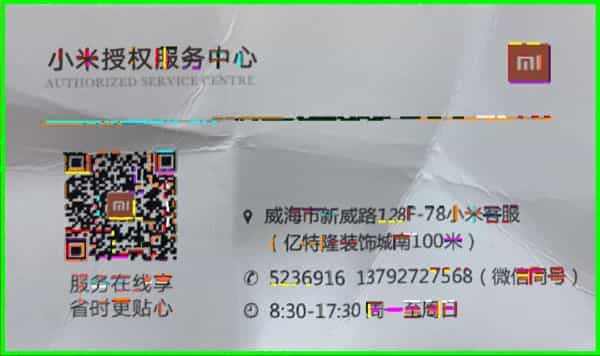}\\
        \vspace{.02\textwidth}%
        \includegraphics[angle=90,width=1\textwidth, height=1.75\textwidth]{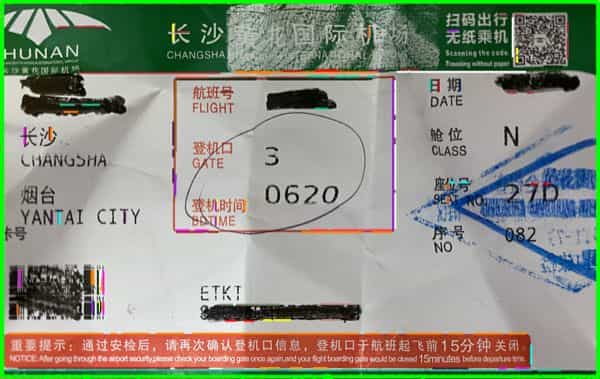}
    \end{minipage}}\hspace{.01\textwidth}%
    \subfigure[]{\centering
    \begin{minipage}[b]{.09\textwidth}\centering
        \includegraphics[width=.938\textwidth, height=1.75\textwidth]{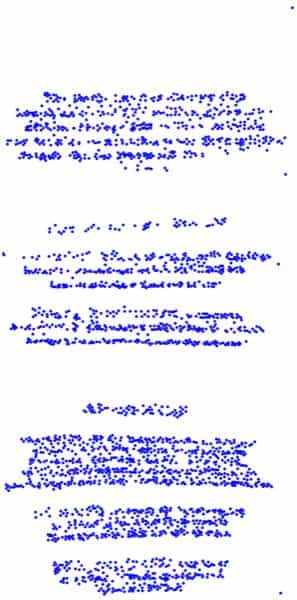}\\
        \vspace{.02\textwidth}%
        \includegraphics[width=.71846847\textwidth, height=1.75\textwidth]{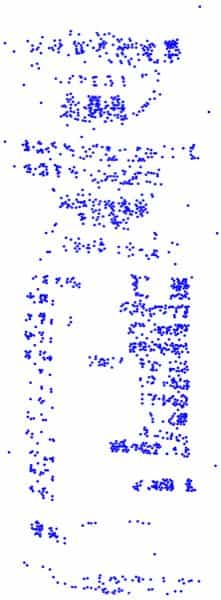}\\
        \vspace{.02\textwidth}%
        \includegraphics[width=1\textwidth, height=1.75\textwidth]{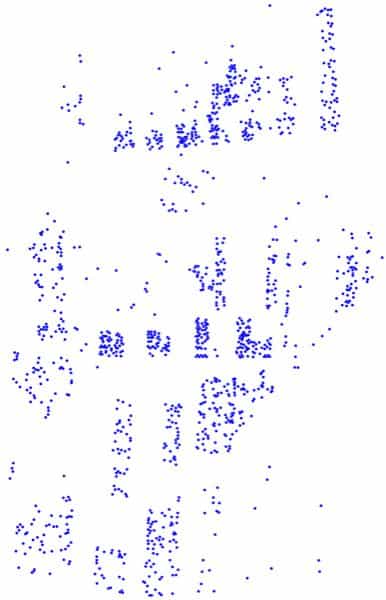}\\
        \vspace{.02\textwidth}%
        \includegraphics[width=1\textwidth, height=1.4524\textwidth]{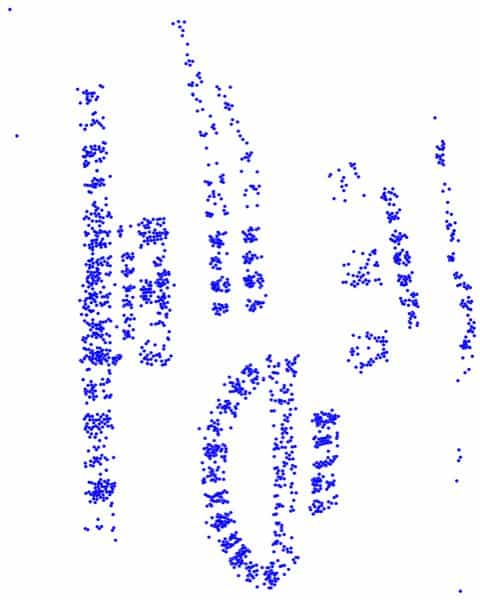}\\
        \vspace{.02\textwidth}%
        \includegraphics[width=1\textwidth, height=1.75\textwidth]{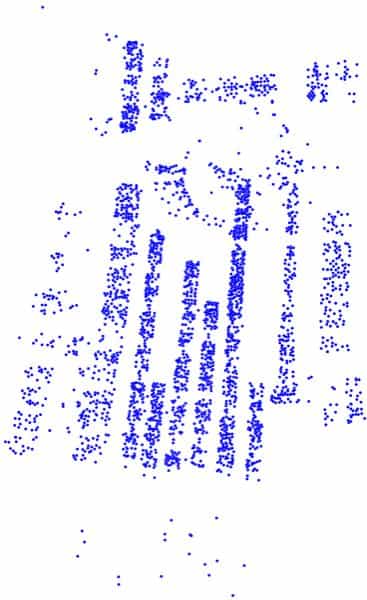}\\
        \vspace{.02\textwidth}%
        \includegraphics[width=1\textwidth, height=1.6\textwidth]{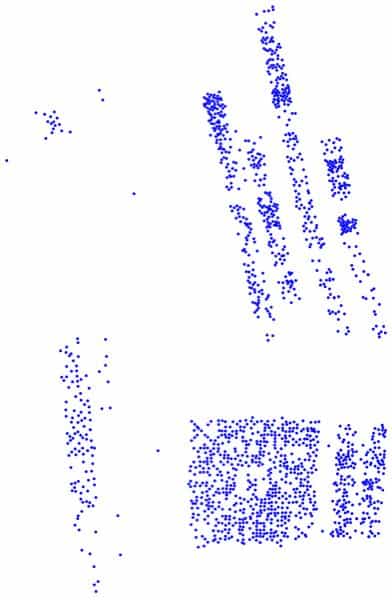}\\
        \vspace{.02\textwidth}%
        \includegraphics[width=1\textwidth, height=1.75\textwidth]{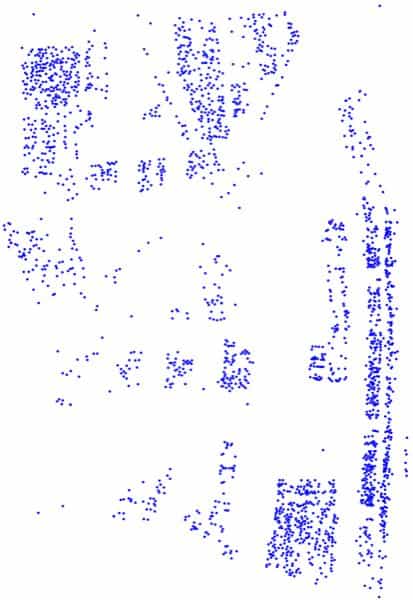}
    \end{minipage}}\hspace{.01\textwidth}%
    \subfigure[]{\centering
    \begin{minipage}[b]{.09\textwidth}\centering
        \includegraphics[width=.938\textwidth, height=1.75\textwidth]{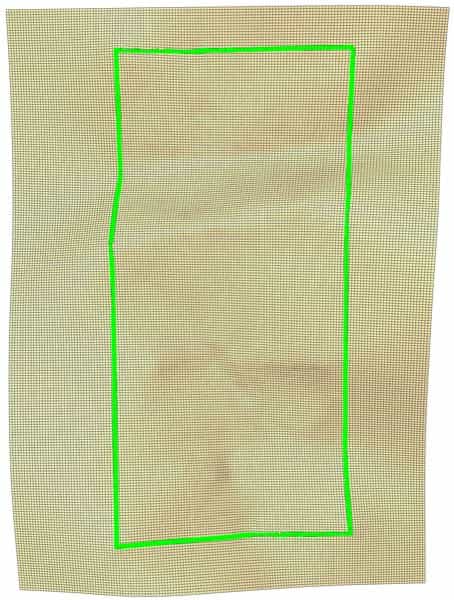}\\
        \vspace{.02\textwidth}%
        \includegraphics[width=.71846847\textwidth, height=1.75\textwidth]{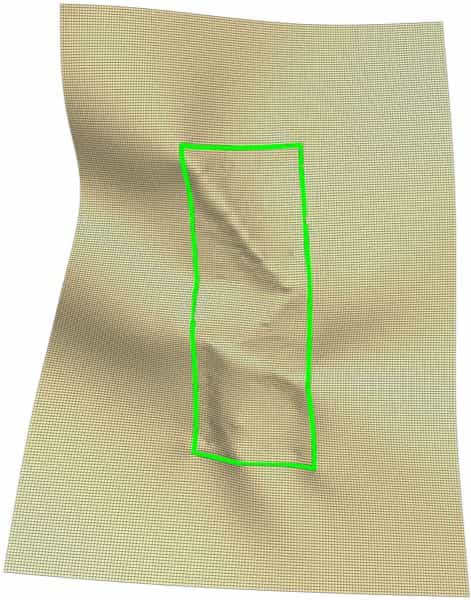}\\
        \vspace{.02\textwidth}%
        \includegraphics[width=1\textwidth, height=1.75\textwidth]{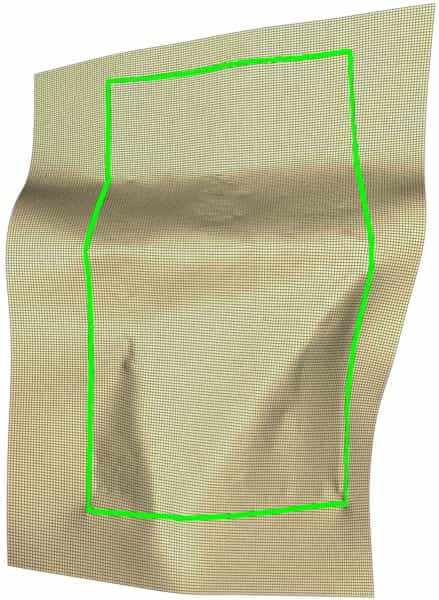}\\
        \vspace{.02\textwidth}%
        \includegraphics[width=1\textwidth, height=1.4524\textwidth]{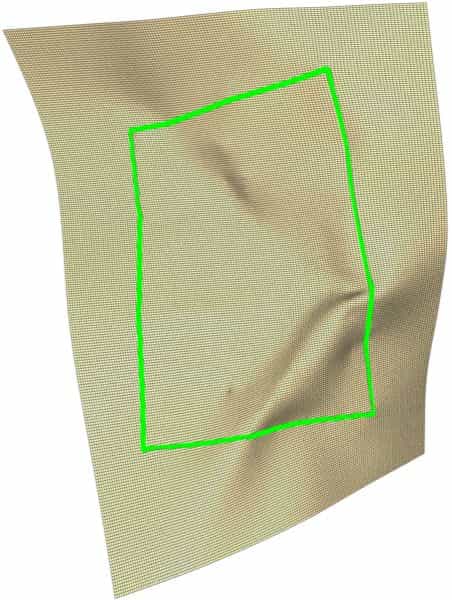}\\
        \vspace{.02\textwidth}%
        \includegraphics[width=1\textwidth, height=1.75\textwidth]{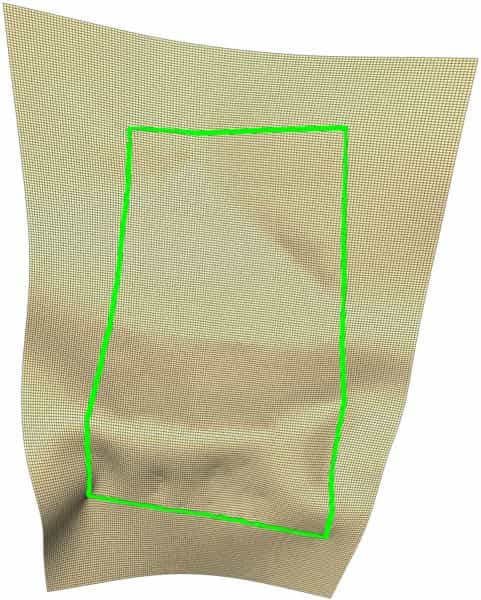}\\
        \vspace{.02\textwidth}%
        \includegraphics[width=1\textwidth, height=1.6\textwidth]{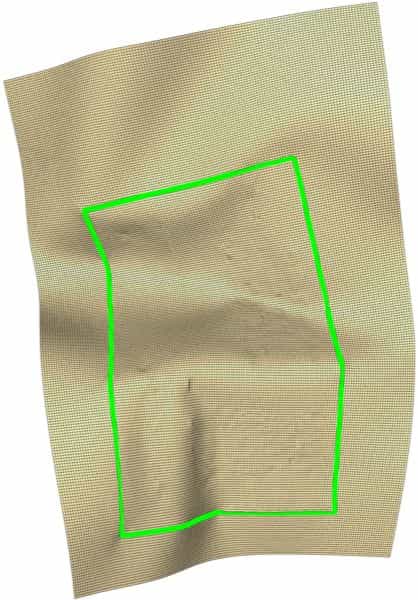}\\
        \vspace{.02\textwidth}%
        \includegraphics[width=1\textwidth, height=1.75\textwidth]{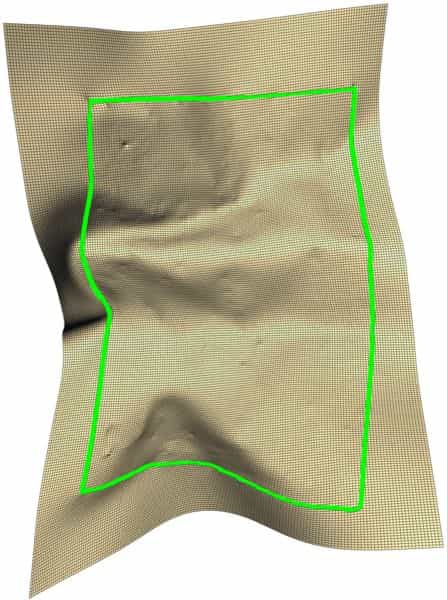}
    \end{minipage}}
\vspace{-0.015\textwidth}
\caption{Dataset \uppercase\expandafter{\romannumeral2}.  (a) Original images. (b) Results of  DewarpNet \cite{das2019dewarpnet}. (c) Results of FCN-based \cite{xie2021dewarping}. (d) Results of Points-based~\cite{xie2022document}. (e) Results of DocTr \cite{feng2021doctr}. (f) Results of DocScanner \cite{feng2021docscanner}. (g) Our results. (h) Our results with feature lines. (i) Our final $\mc{P}$. (j) Our final $\mc{M}$ with boundary.}\label{f:results2}
\end{figure}
\begin{figure}[!t]
\centering
    \subfigure[]{\centering 
    \begin{minipage}[b]{.09\textwidth}\centering
        \includegraphics[width=1\textwidth, height=1.414\textwidth]{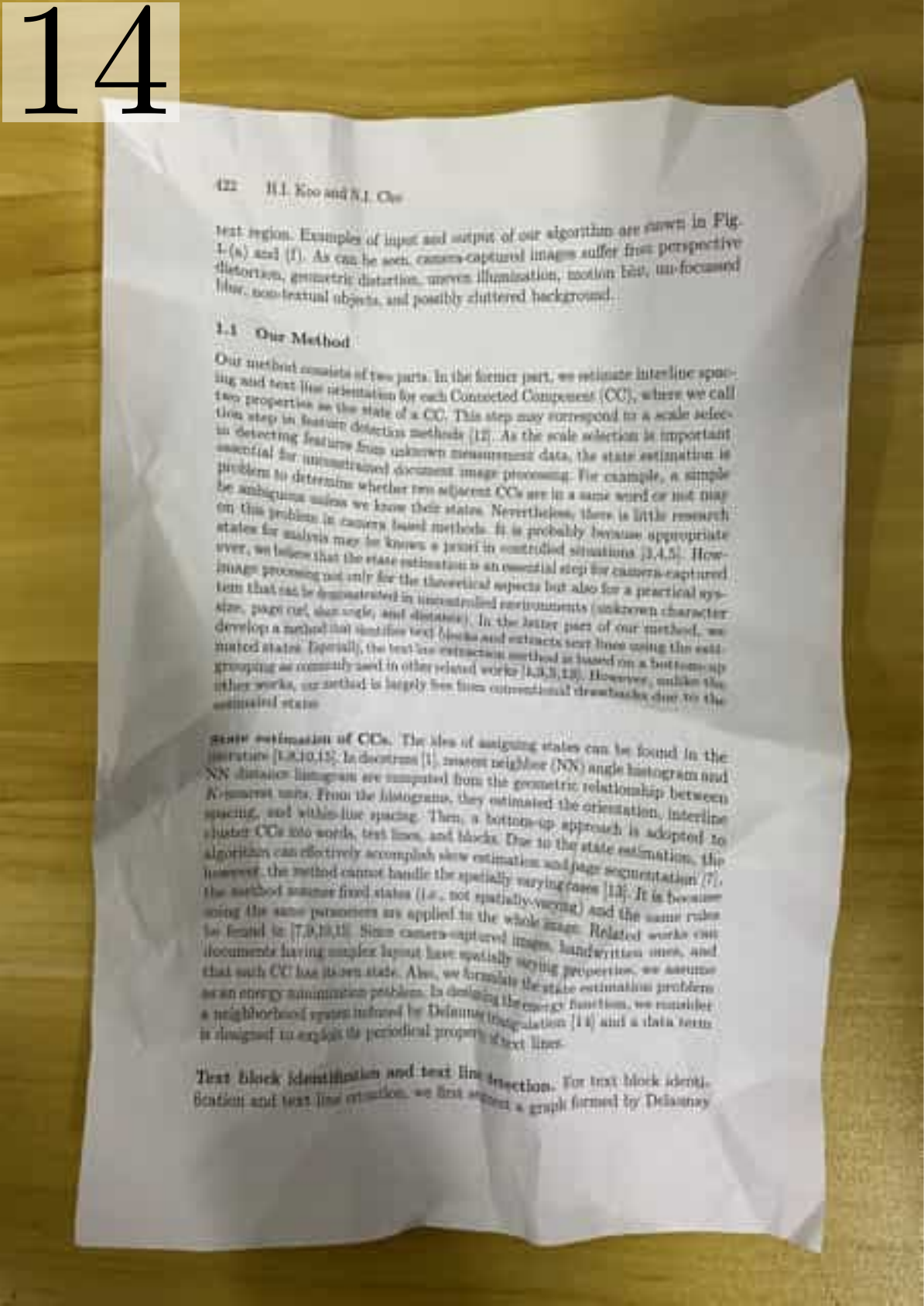}\\
        \vspace{.02\textwidth}%
       \includegraphics[width=1\textwidth, height=1.414\textwidth]{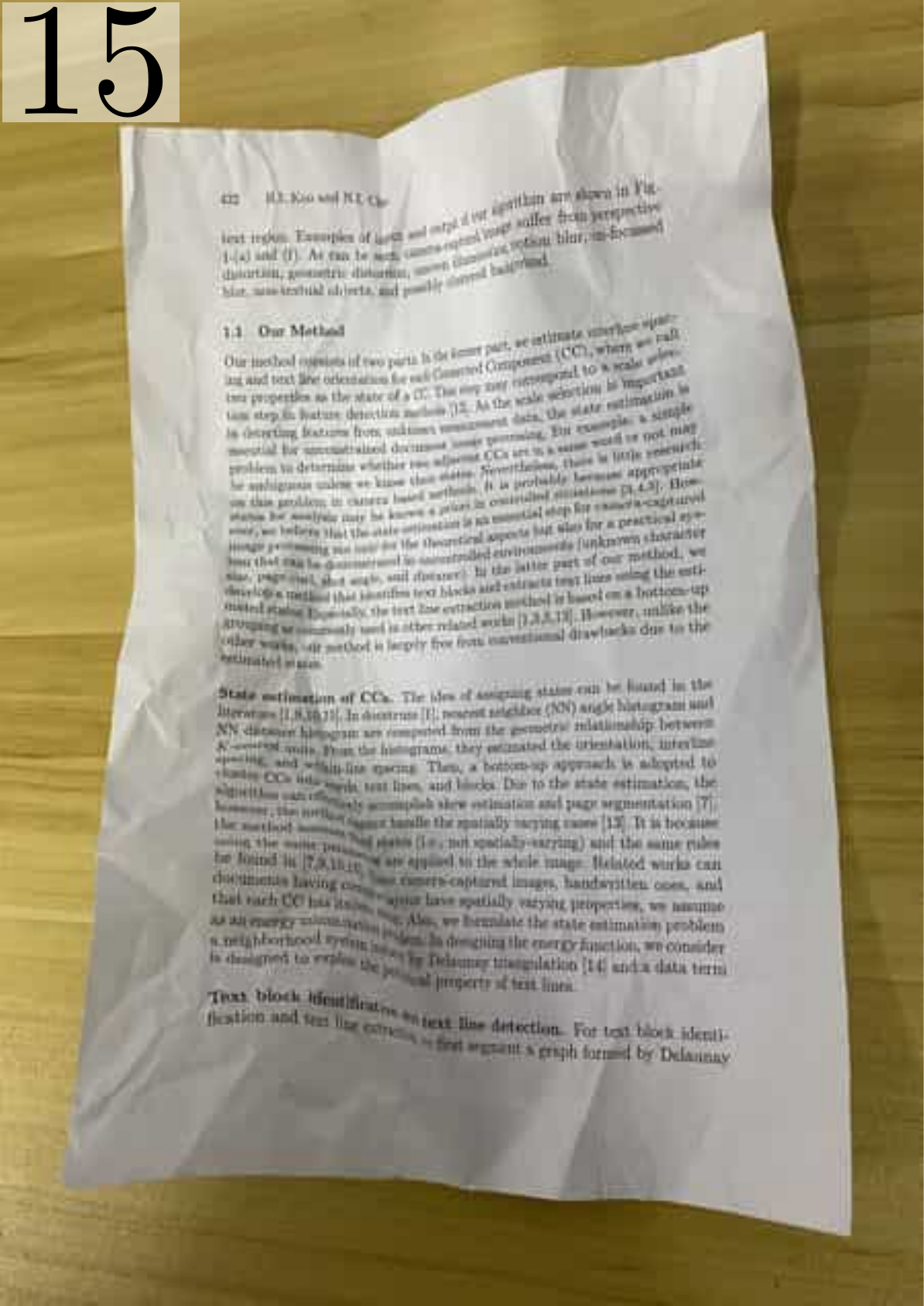}\\
        \vspace{.02\textwidth}%
       \includegraphics[width=1\textwidth, height=1.414\textwidth]{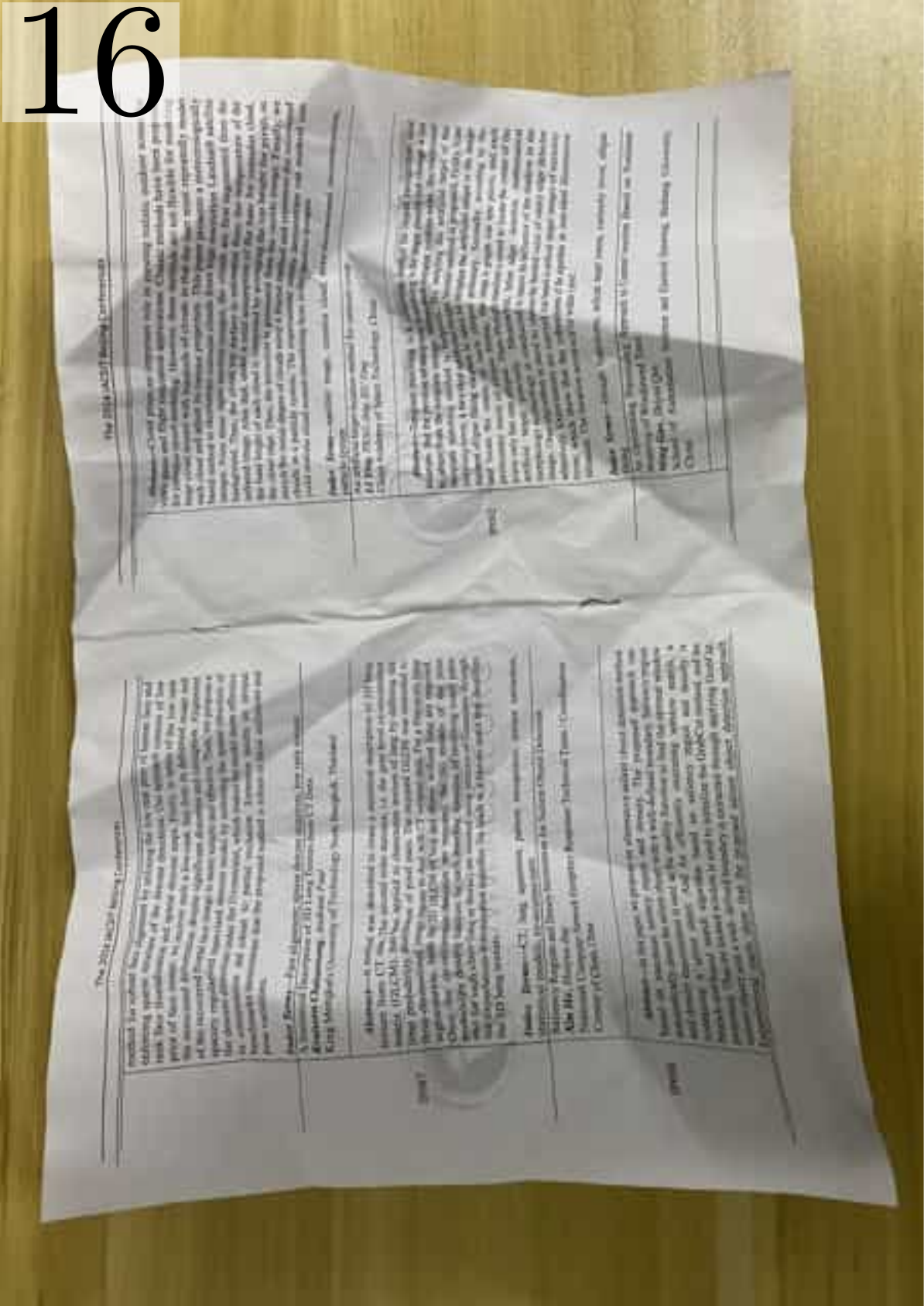}\\
        \vspace{.02\textwidth}%
      \includegraphics[width=1\textwidth, height=1.414\textwidth]{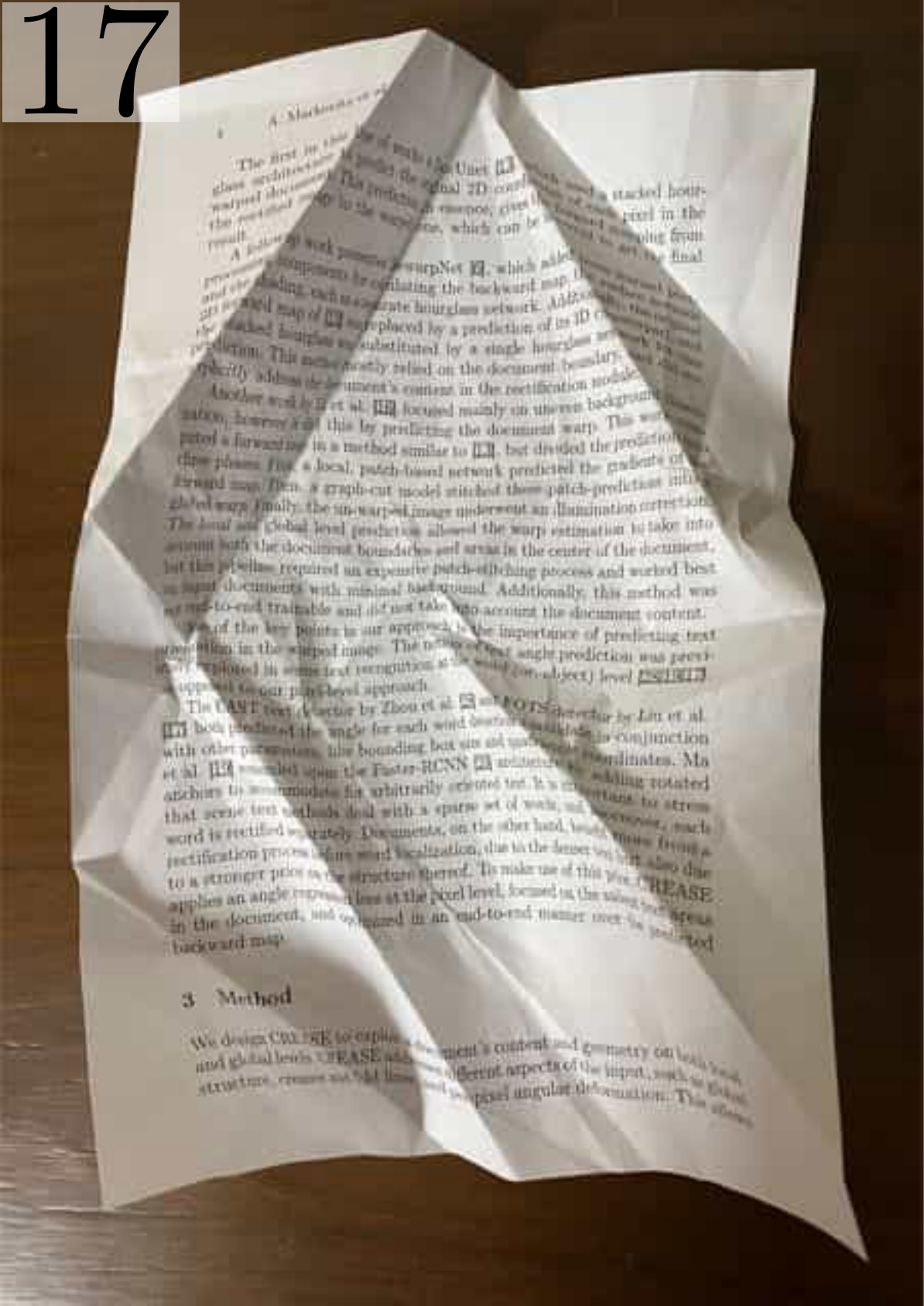}\\
        \vspace{.02\textwidth}%
       \includegraphics[width=1\textwidth, height=1.414\textwidth]{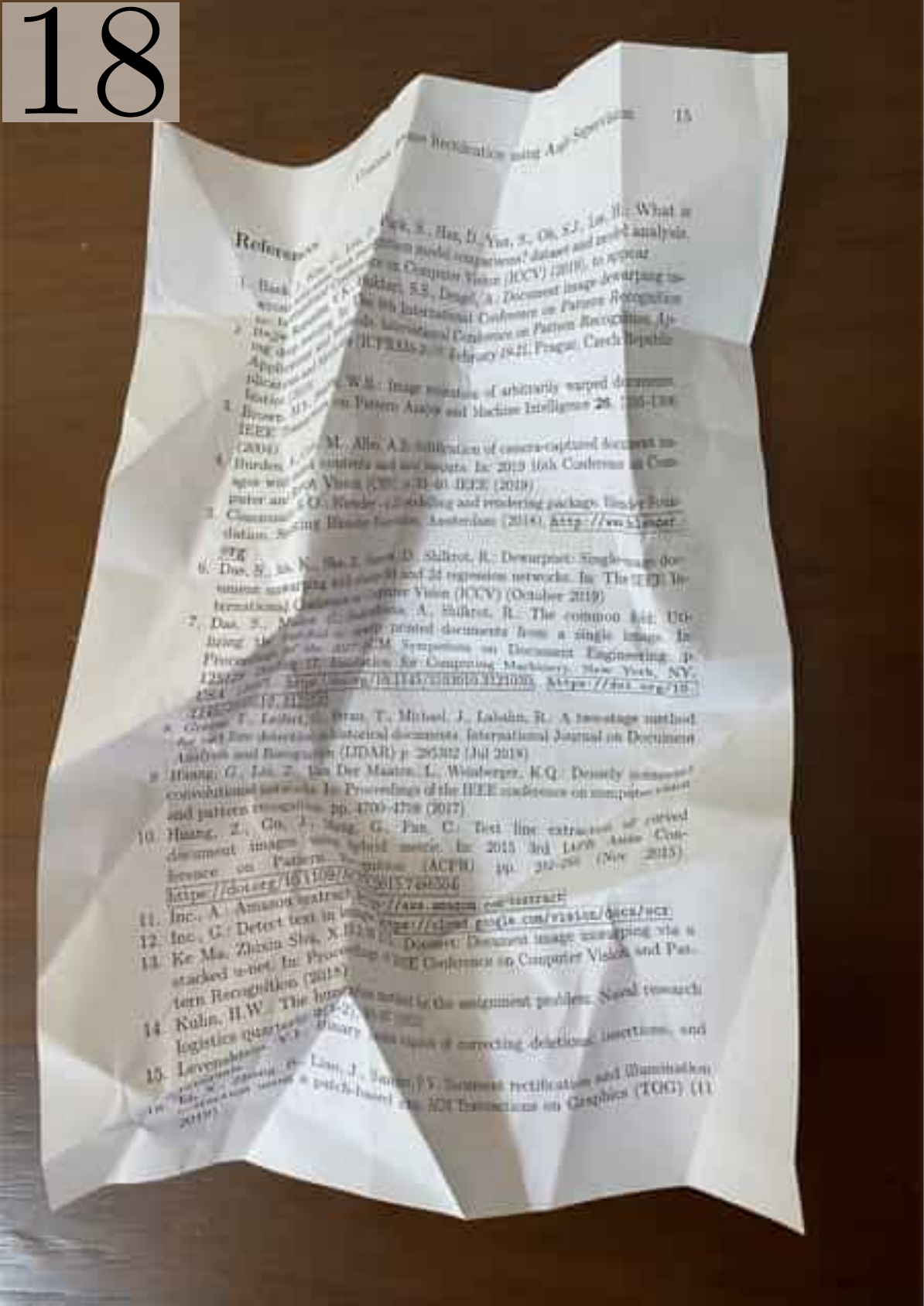}\\
        \vspace{.02\textwidth}%
       \includegraphics[width=1\textwidth, height=1.414\textwidth]{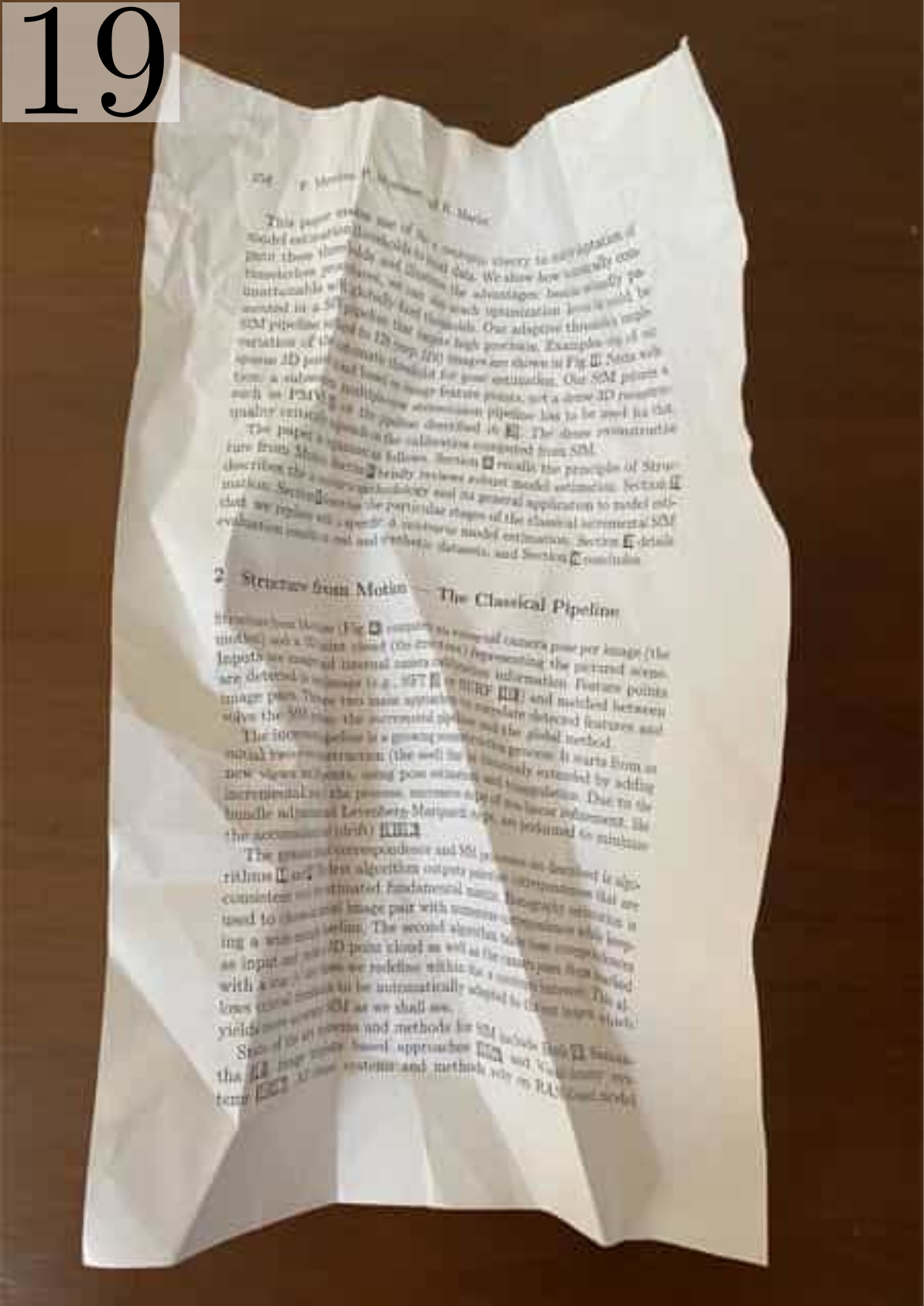}\\
        \vspace{.02\textwidth}%
       \includegraphics[width=1\textwidth, height=1.414\textwidth]{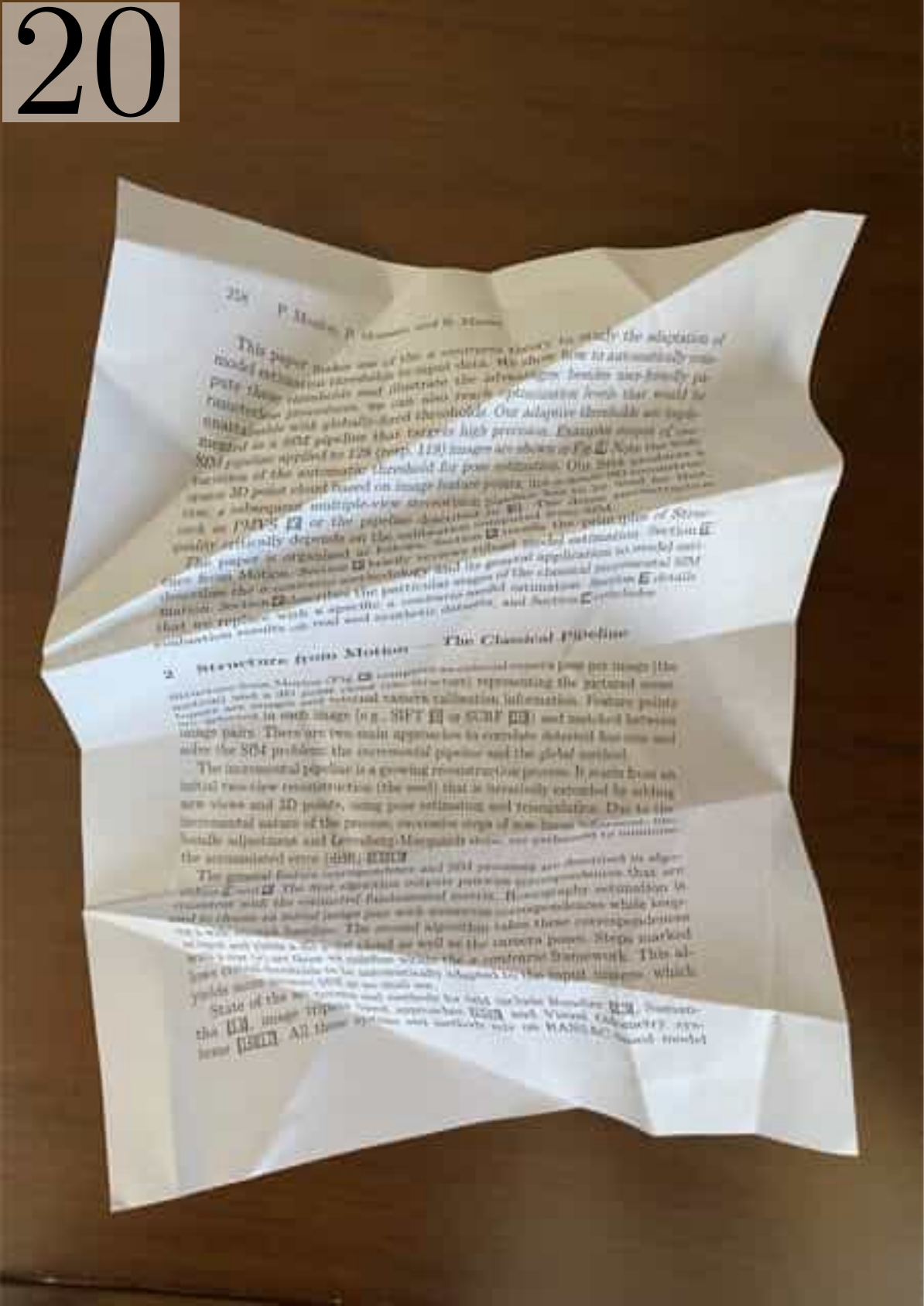}
    \end{minipage}}\hspace{.01\textwidth}%
    \subfigure[]{\centering
    \begin{minipage}[b]{.09\textwidth}\centering
        \includegraphics[width=1\textwidth, height=1.414\textwidth]{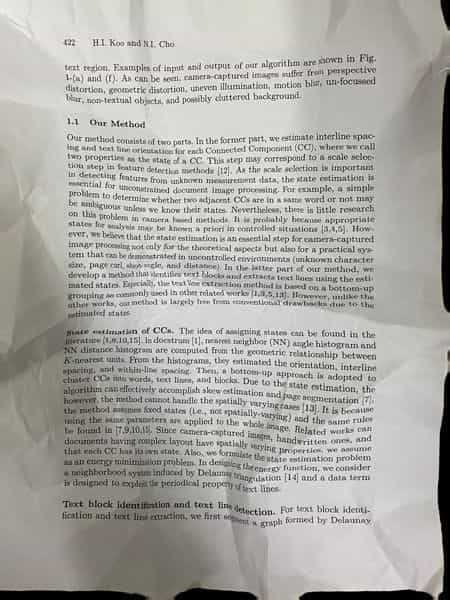}\\
        \vspace{.02\textwidth}%
        \includegraphics[width=1\textwidth, height=1.414\textwidth]{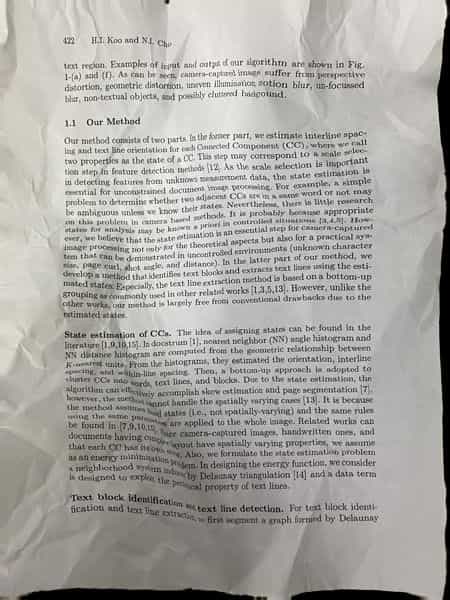}\\
        \vspace{.02\textwidth}%
        \includegraphics[angle=90,width=1\textwidth, height=1.414\textwidth]{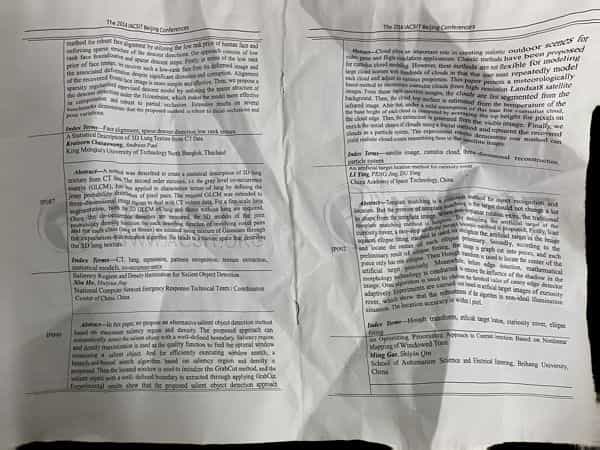}\\
        \vspace{.02\textwidth}%
        \includegraphics[width=1\textwidth, height=1.414\textwidth]{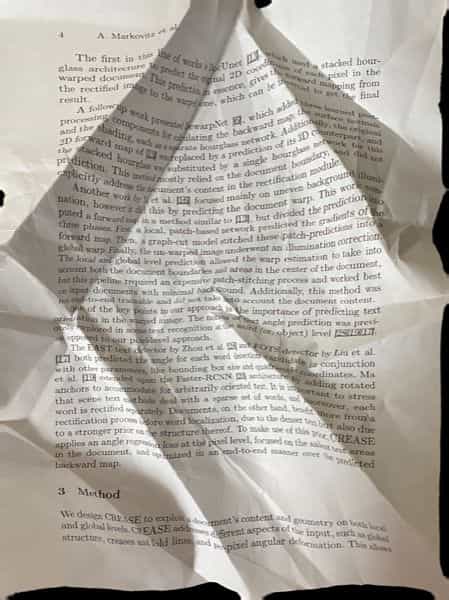}\\
        \vspace{.02\textwidth}%
        \includegraphics[width=1\textwidth, height=1.414\textwidth]{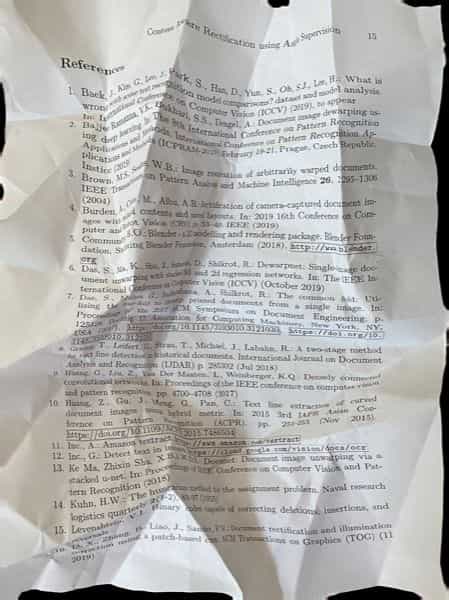}\\
        \vspace{.02\textwidth}%
        \includegraphics[width=1\textwidth, height=1.414\textwidth]{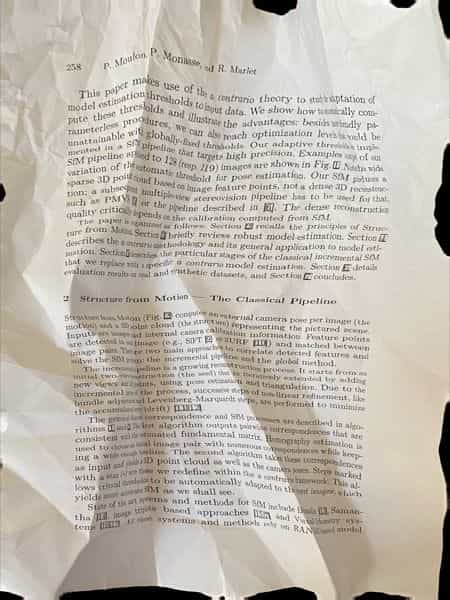}\\
        \vspace{.02\textwidth}%
        \includegraphics[width=1\textwidth, height=1.414\textwidth]{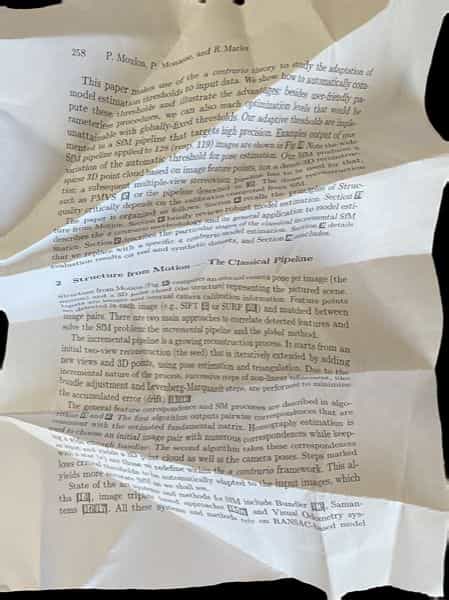}
    \end{minipage}}\hspace{.01\textwidth}%
    \subfigure[]{\centering
    \begin{minipage}[b]{.09\textwidth}\centering
        \includegraphics[width=1\textwidth, height=1.414\textwidth]{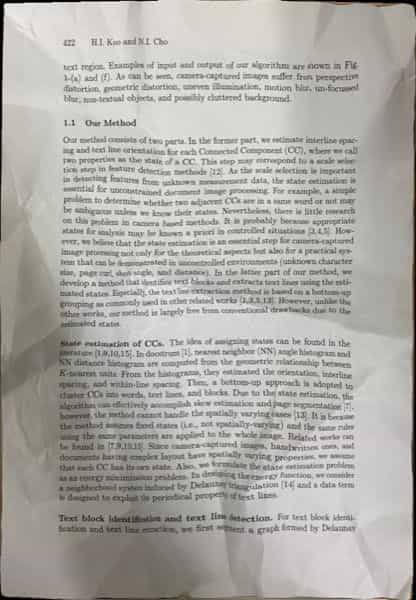}\\
        \vspace{.02\textwidth}%
        \includegraphics[width=1\textwidth, height=1.414\textwidth]{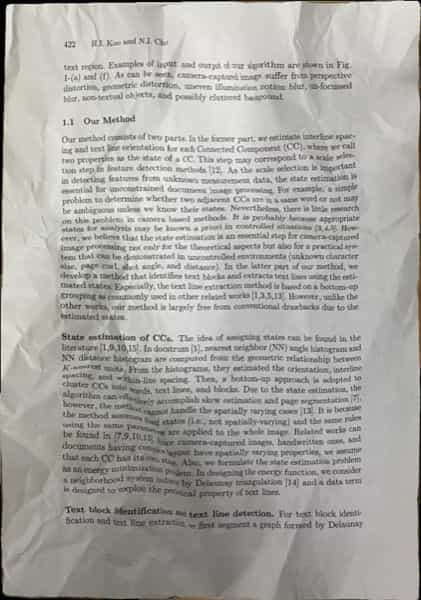}\\
        \vspace{.02\textwidth}%
        \includegraphics[angle=90,width=1\textwidth, height=1.414\textwidth]{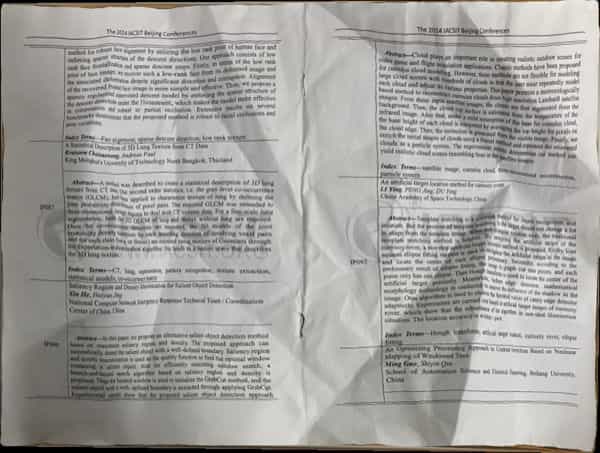}\\
        \vspace{.02\textwidth}%
        \includegraphics[width=1\textwidth, height=1.414\textwidth]{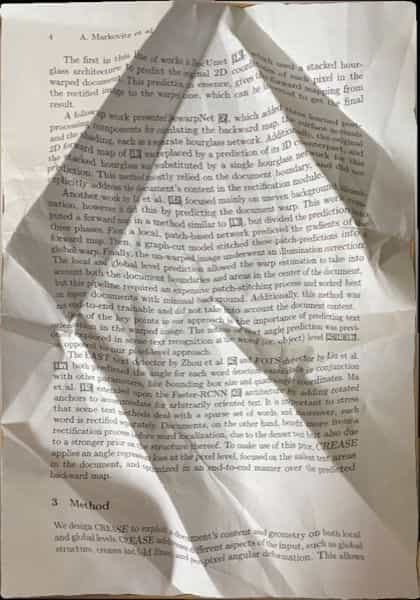}\\
        \vspace{.02\textwidth}%
        \includegraphics[width=1\textwidth, height=1.414\textwidth]{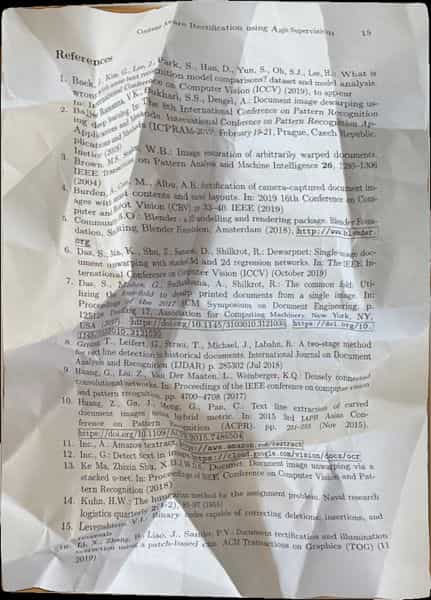}\\
        \vspace{.02\textwidth}%
        \includegraphics[width=1\textwidth, height=1.414\textwidth]{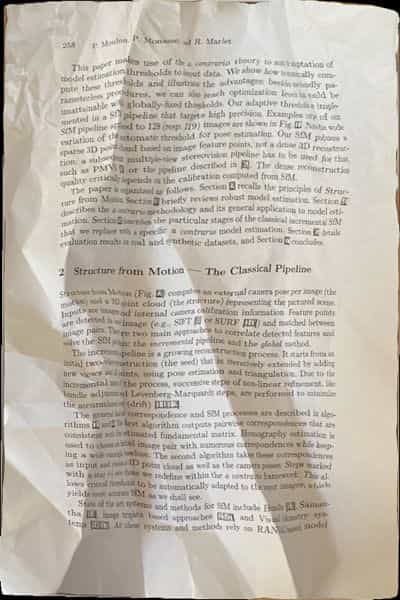}\\
        \vspace{.02\textwidth}%
        \includegraphics[width=1\textwidth, height=1.414\textwidth]{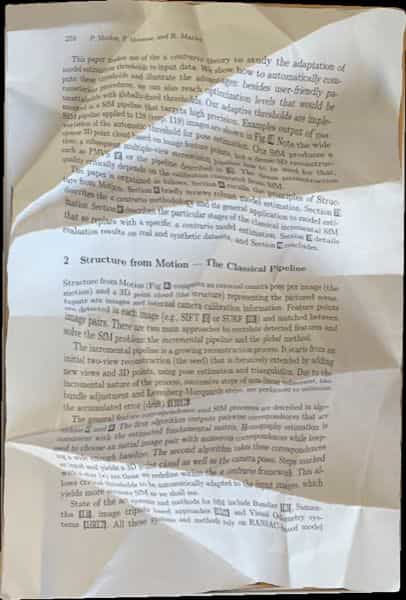}
    \end{minipage}}\hspace{.01\textwidth}%
    \subfigure[]{\centering
    \begin{minipage}[b]{.09\textwidth}\centering
        \includegraphics[width=1\textwidth, height=1.414\textwidth]{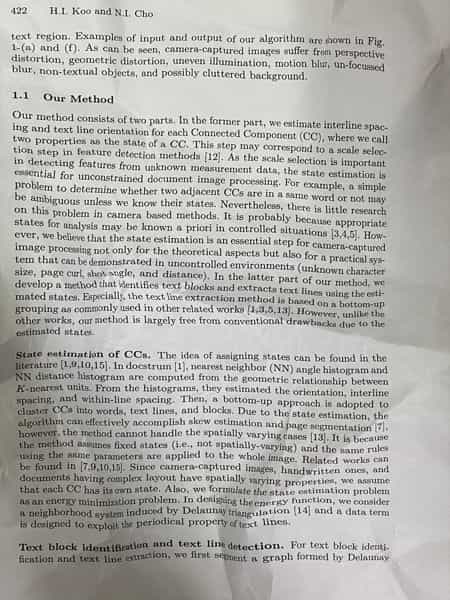}\\
        \vspace{.02\textwidth}%
        \includegraphics[width=1\textwidth, height=1.414\textwidth]{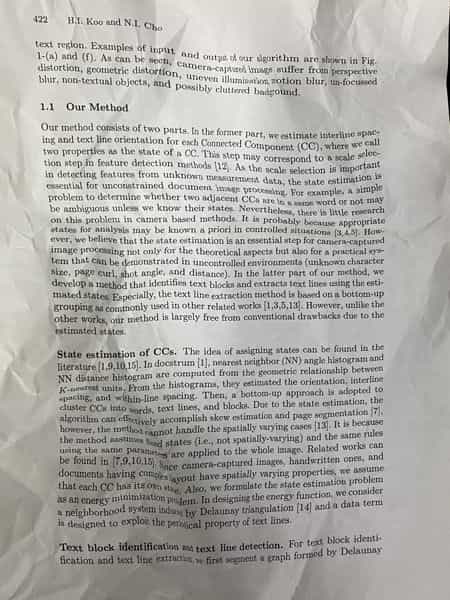}\\
        \vspace{.02\textwidth}%
        \includegraphics[angle=90,width=1\textwidth, height=1.414\textwidth]{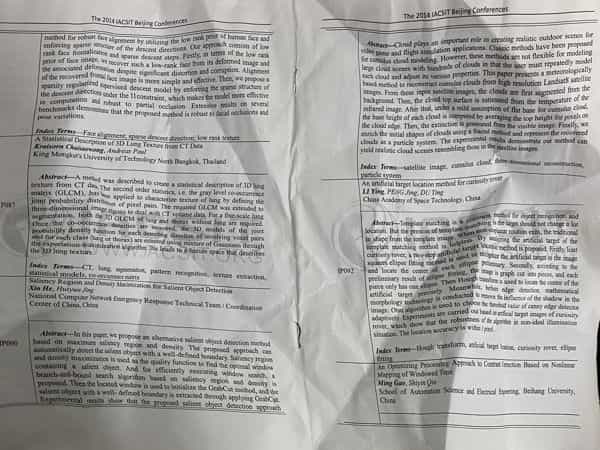}\\
        \vspace{.02\textwidth}%
        \includegraphics[width=1\textwidth, height=1.414\textwidth]{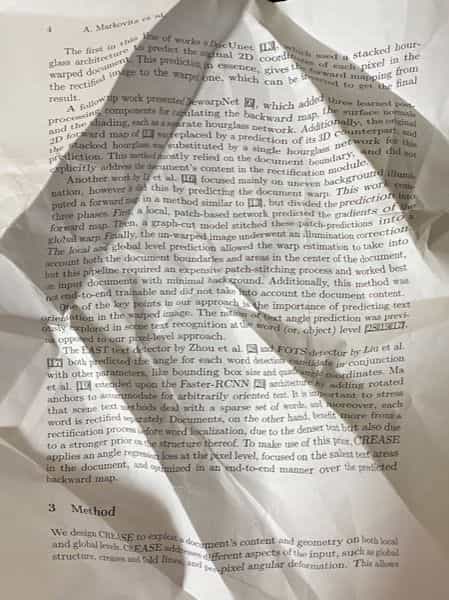}\\
        \vspace{.02\textwidth}%
        \includegraphics[width=1\textwidth, height=1.414\textwidth]{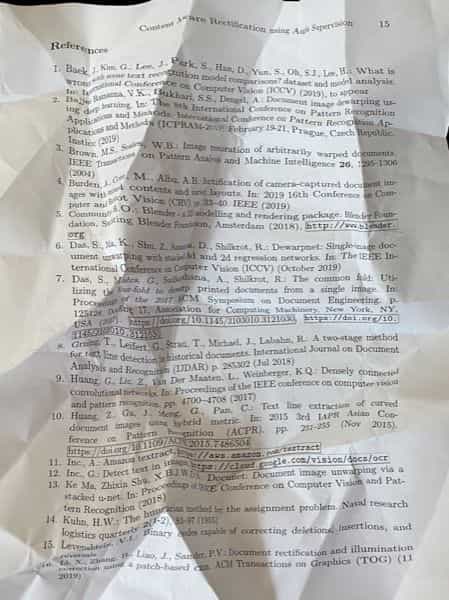}\\
        \vspace{.02\textwidth}%
        \includegraphics[width=1\textwidth, height=1.414\textwidth]{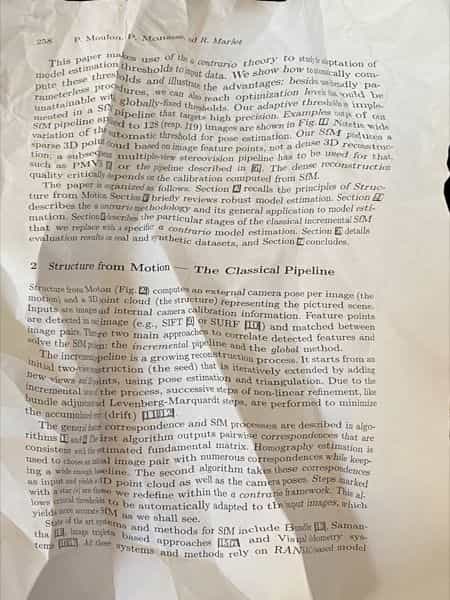}\\
        \vspace{.02\textwidth}%
        \includegraphics[width=1\textwidth, height=1.414\textwidth]{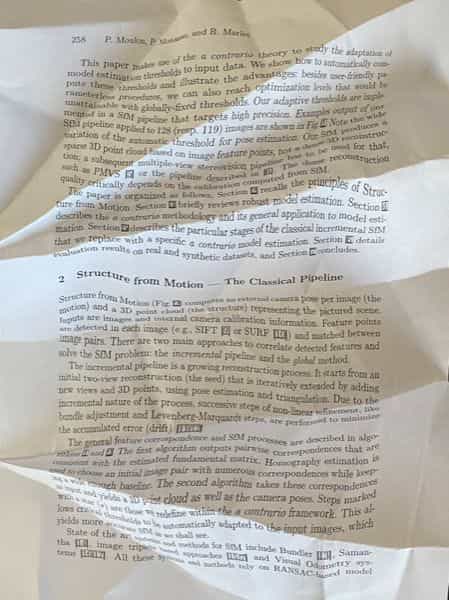}
    \end{minipage}}\hspace{.01\textwidth}%
    \subfigure[]{\centering
    \begin{minipage}[b]{.09\textwidth}\centering
        \includegraphics[width=1\textwidth, height=1.414\textwidth]{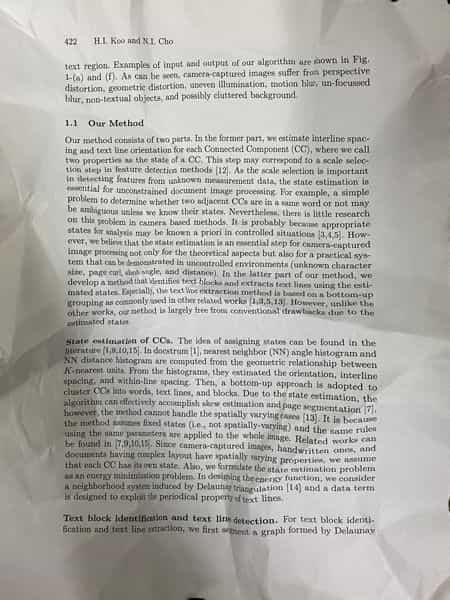}\\
        \vspace{.02\textwidth}%
        \includegraphics[width=1\textwidth, height=1.414\textwidth]{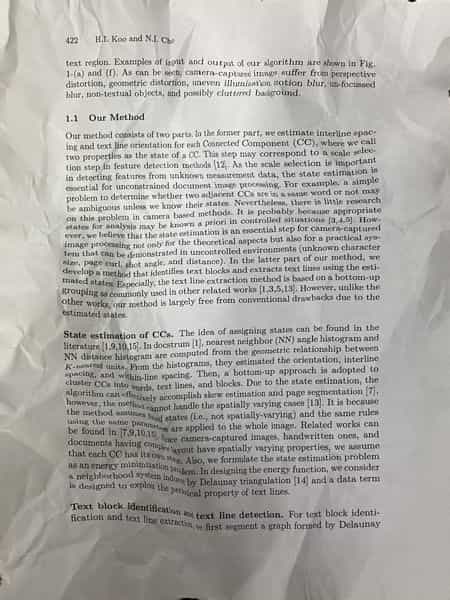}\\
        \vspace{.02\textwidth}%
        \includegraphics[angle=90,width=1\textwidth, height=1.414\textwidth]{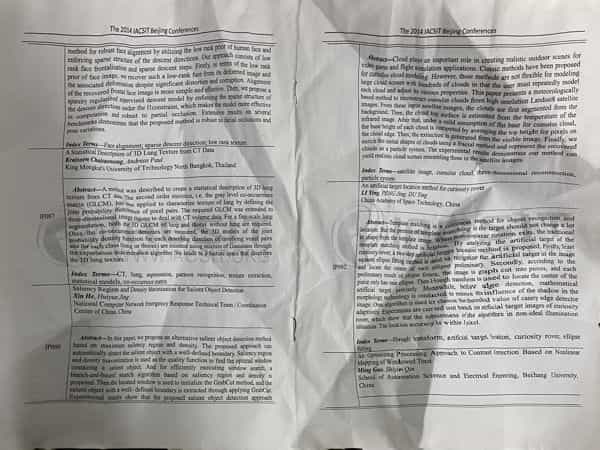}\\
        \vspace{.02\textwidth}%
        \includegraphics[width=1\textwidth, height=1.414\textwidth]{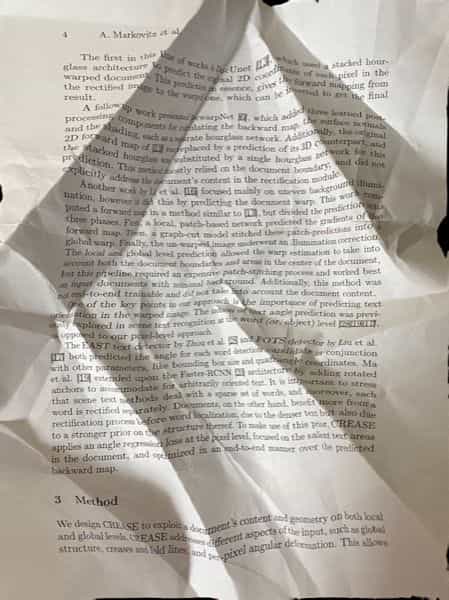}\\
        \vspace{.02\textwidth}%
        \includegraphics[width=1\textwidth, height=1.414\textwidth]{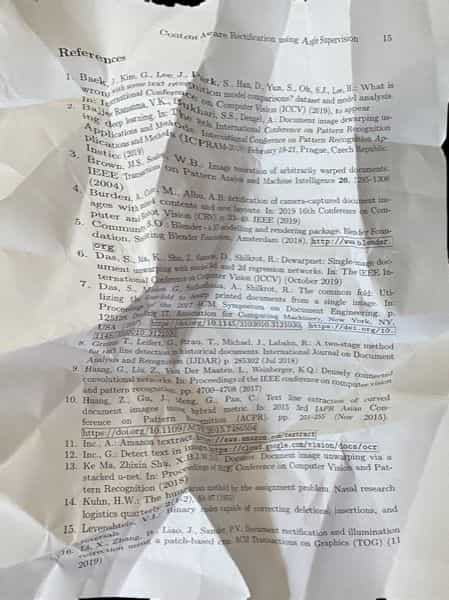}\\
        \vspace{.02\textwidth}%
        \includegraphics[width=1\textwidth, height=1.414\textwidth]{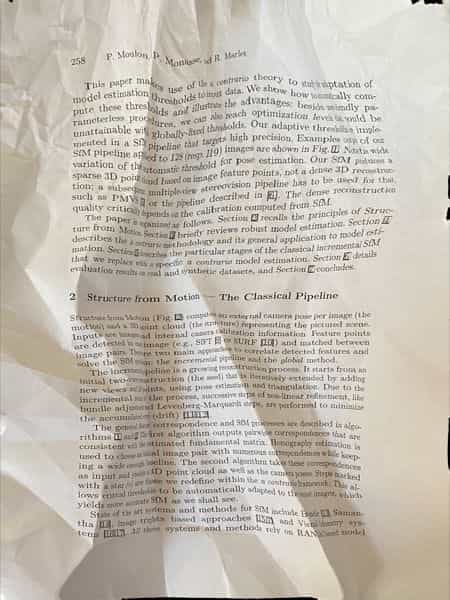}\\
        \vspace{.02\textwidth}%
        \includegraphics[width=1\textwidth, height=1.414\textwidth]{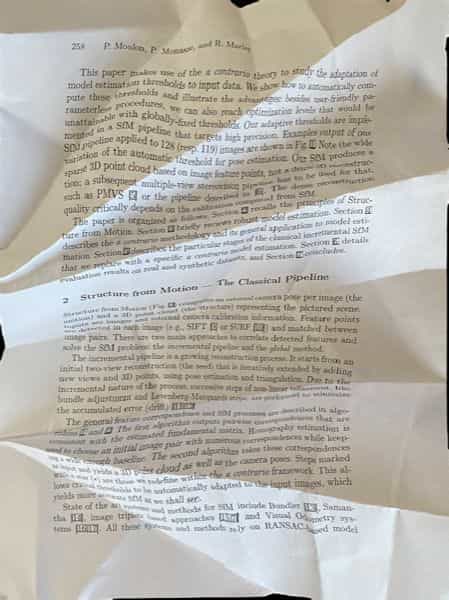}
    \end{minipage}}\hspace{.01\textwidth}%
    \subfigure[]{\centering
    \begin{minipage}[b]{.09\textwidth}\centering
        \includegraphics[width=1\textwidth, height=1.414\textwidth]{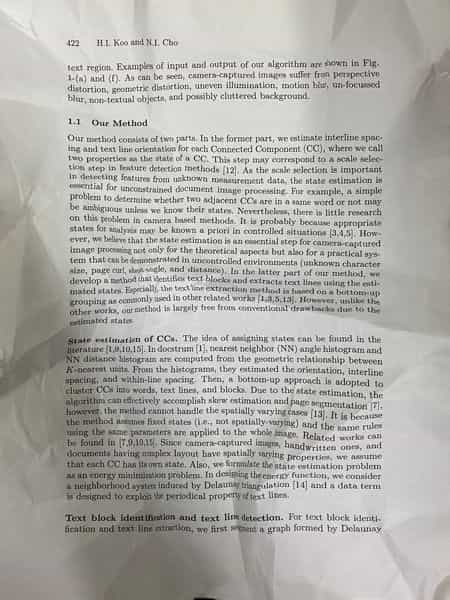}\\
        \vspace{.02\textwidth}%
        \includegraphics[width=1\textwidth, height=1.414\textwidth]{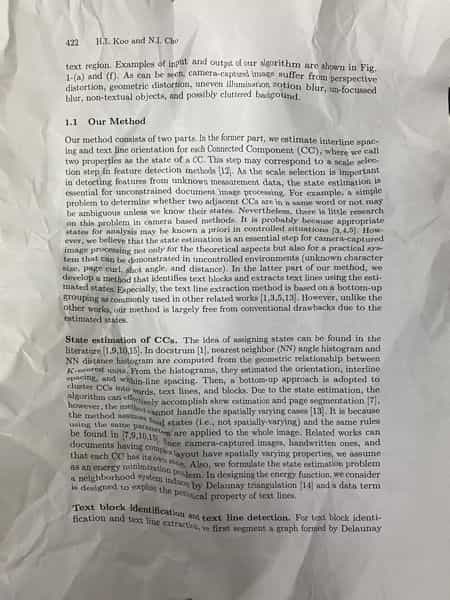}\\
        \vspace{.02\textwidth}%
        \includegraphics[angle=90,width=1\textwidth, height=1.414\textwidth]{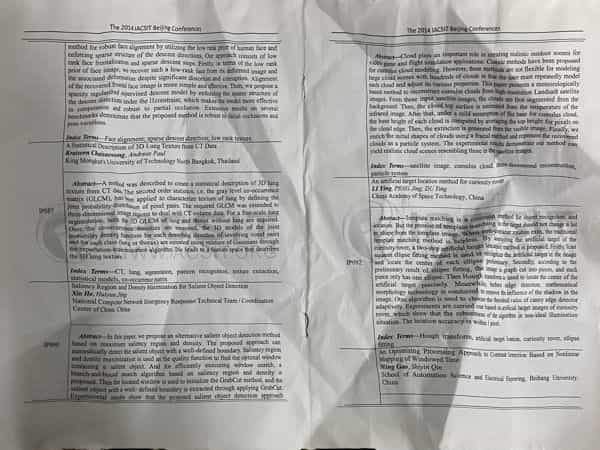}\\
        \vspace{.02\textwidth}%
        \includegraphics[width=1\textwidth, height=1.414\textwidth]{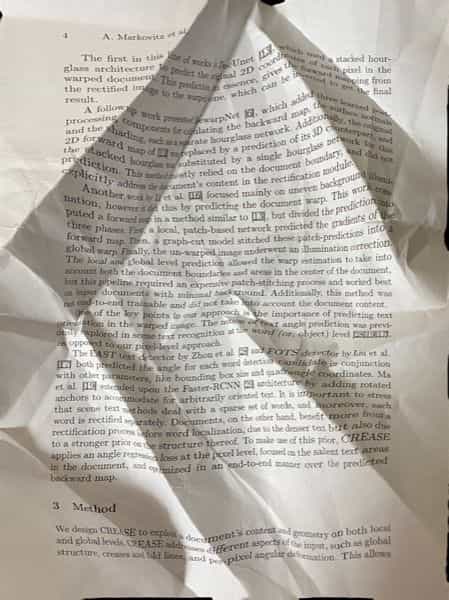}\\
        \vspace{.02\textwidth}%
        \includegraphics[width=1\textwidth, height=1.414\textwidth]{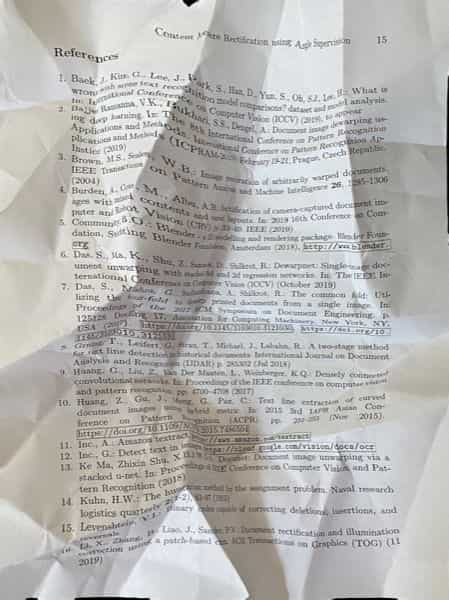}\\
        \vspace{.02\textwidth}%
        \includegraphics[width=1\textwidth, height=1.414\textwidth]{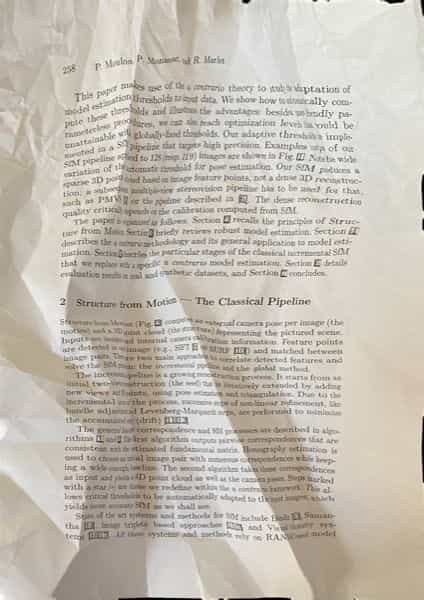}\\
        \vspace{.02\textwidth}%
        \includegraphics[width=1\textwidth, height=1.414\textwidth]{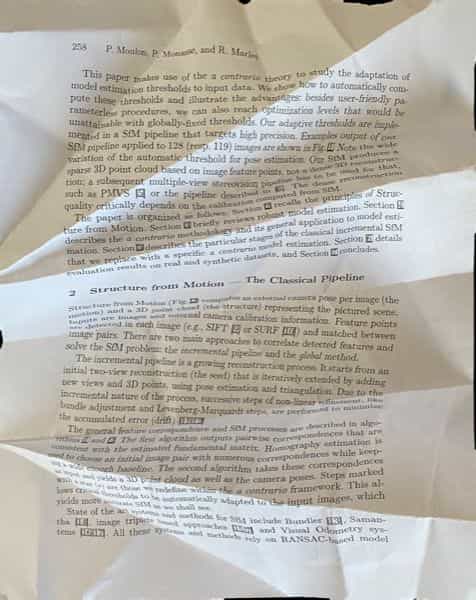}
    \end{minipage}}\hspace{.01\textwidth}%
    \subfigure[]{\centering
    \begin{minipage}[b]{.09\textwidth}\centering
        \includegraphics[width=1\textwidth, height=1.414\textwidth]{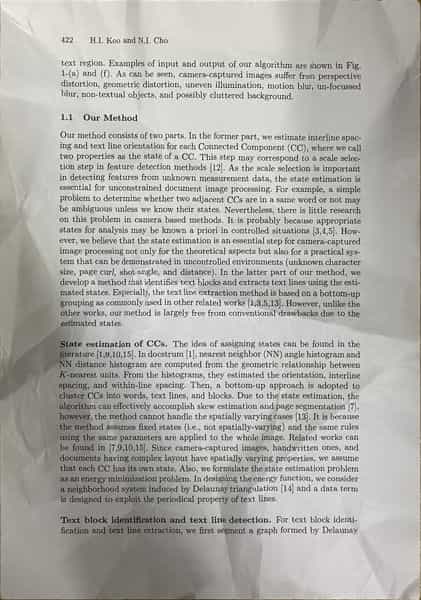}\\
        \vspace{.02\textwidth}%
        \includegraphics[width=1\textwidth, height=1.414\textwidth]{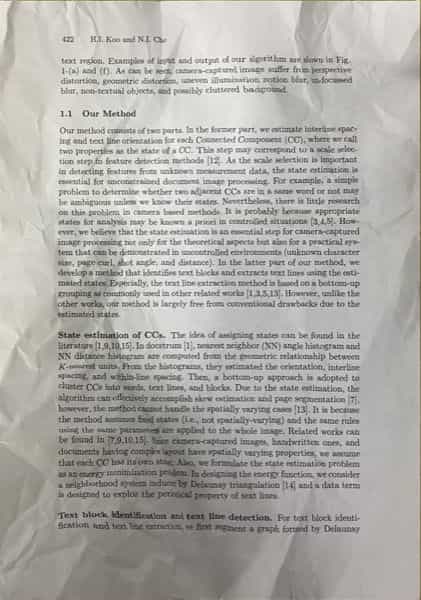}\\
        \vspace{.02\textwidth}%
        \includegraphics[angle=90,width=1\textwidth, height=1.414\textwidth]{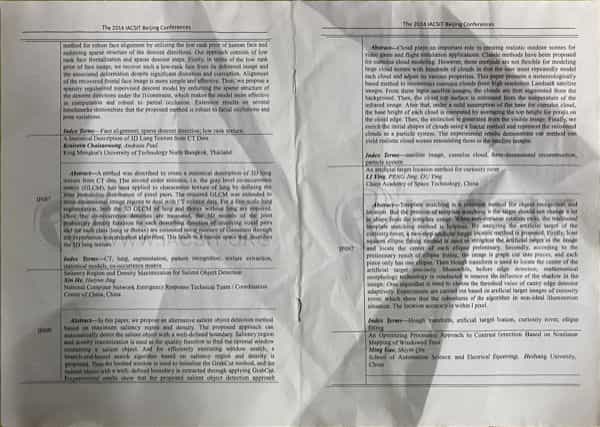}\\
        \vspace{.02\textwidth}%
        \includegraphics[width=1\textwidth, height=1.414\textwidth]{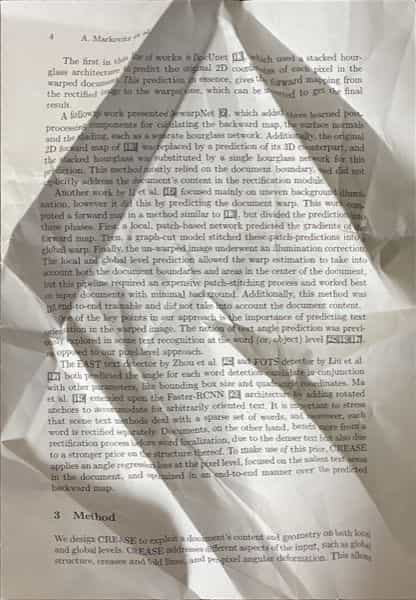}\\
        \vspace{.02\textwidth}%
        \includegraphics[width=1\textwidth, height=1.414\textwidth]{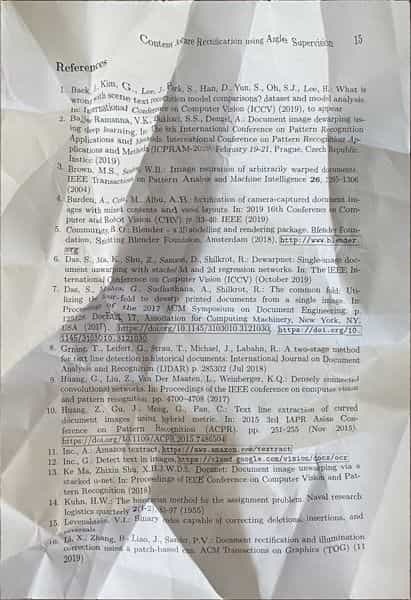}\\
        \vspace{.02\textwidth}%
        \includegraphics[width=1\textwidth, height=1.414\textwidth]{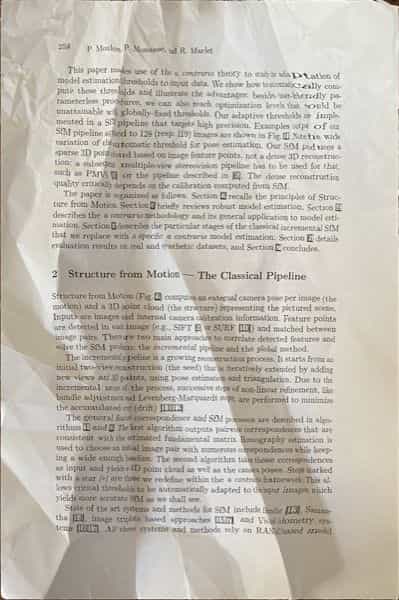}\\
        \vspace{.02\textwidth}%
        \includegraphics[width=1\textwidth, height=1.414\textwidth]{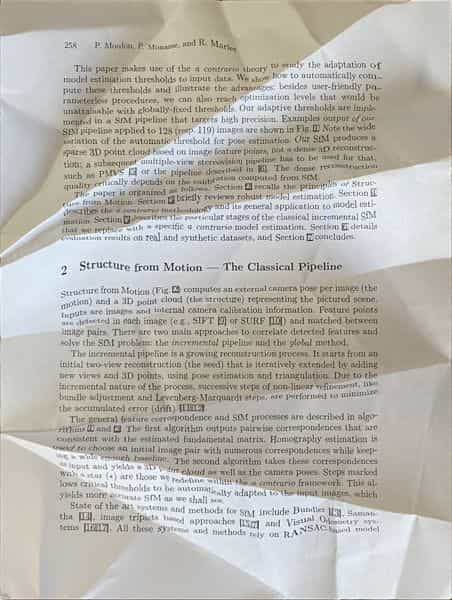}
    \end{minipage}}\hspace{.01\textwidth}%
    \subfigure[]{\centering
    \begin{minipage}[b]{.09\textwidth}\centering
        \includegraphics[width=1\textwidth, height=1.414\textwidth]{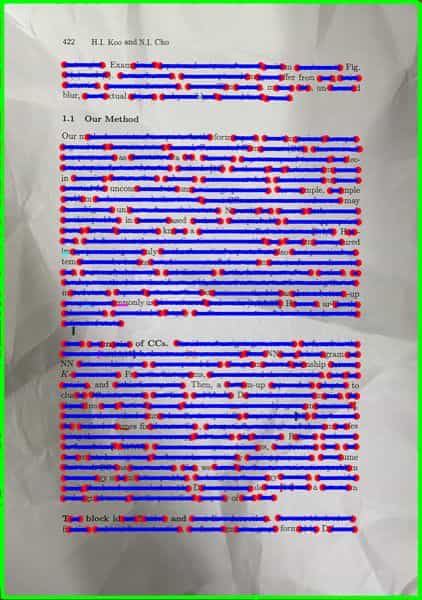}\\
        \vspace{.02\textwidth}%
        \includegraphics[width=1\textwidth, height=1.414\textwidth]{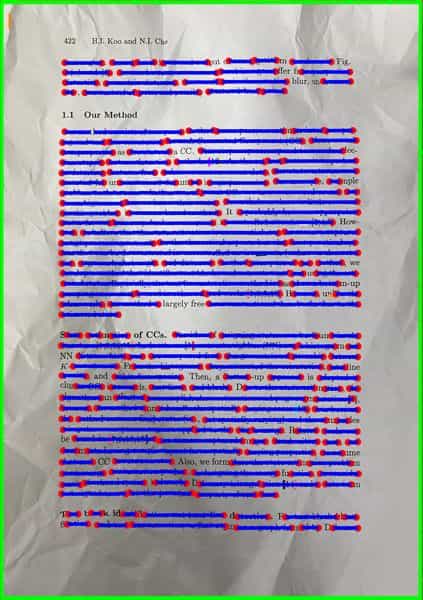}\\
        \vspace{.02\textwidth}%
        \includegraphics[angle=90,width=1\textwidth, height=1.414\textwidth]{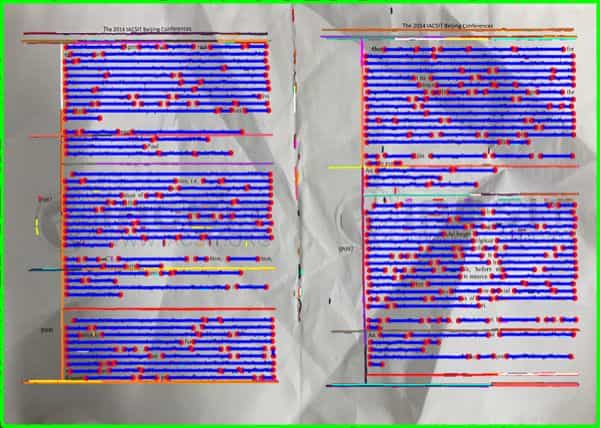}\\
        \vspace{.02\textwidth}%
        \includegraphics[width=1\textwidth, height=1.414\textwidth]{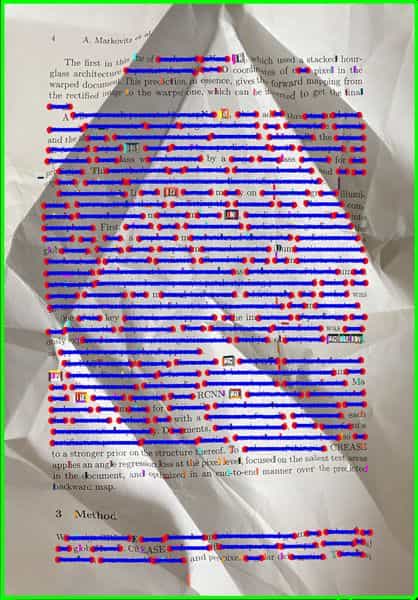}\\
        \vspace{.02\textwidth}%
        \includegraphics[width=1\textwidth, height=1.414\textwidth]{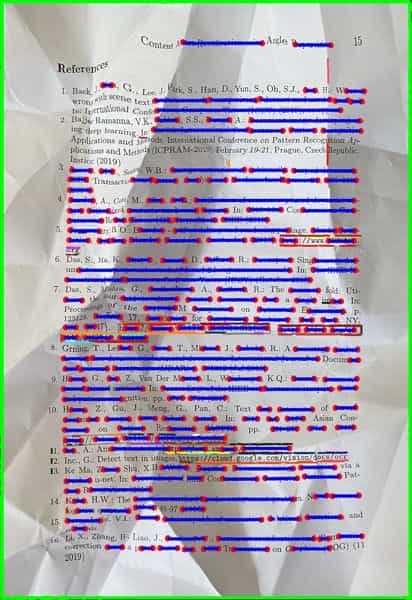}\\
        \vspace{.02\textwidth}%
        \includegraphics[width=1\textwidth, height=1.414\textwidth]{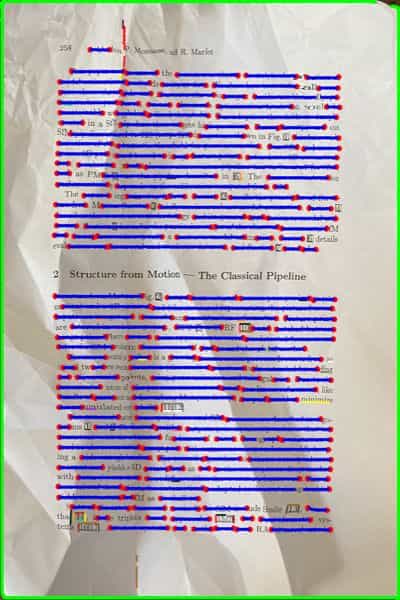}\\
        \vspace{.02\textwidth}%
        \includegraphics[width=1\textwidth, height=1.414\textwidth]{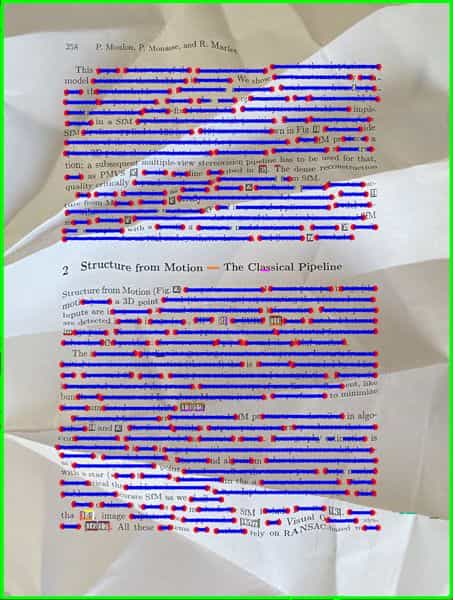}
    \end{minipage}}\hspace{.01\textwidth}%
    \subfigure[]{\centering
    \begin{minipage}[b]{.09\textwidth}\centering
        \includegraphics[width=1\textwidth, height=1.414\textwidth]{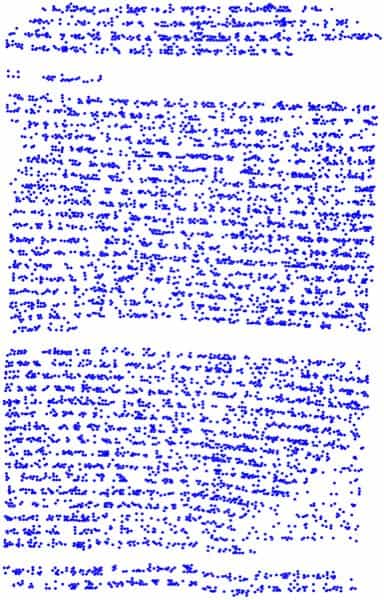}\\
        \vspace{.02\textwidth}%
        \includegraphics[width=1\textwidth, height=1.414\textwidth]{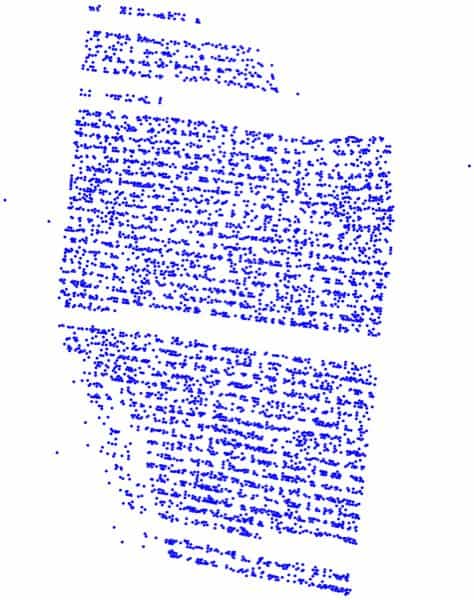}\\
        \vspace{.02\textwidth}%
        \includegraphics[width=1\textwidth, height=1.414\textwidth]{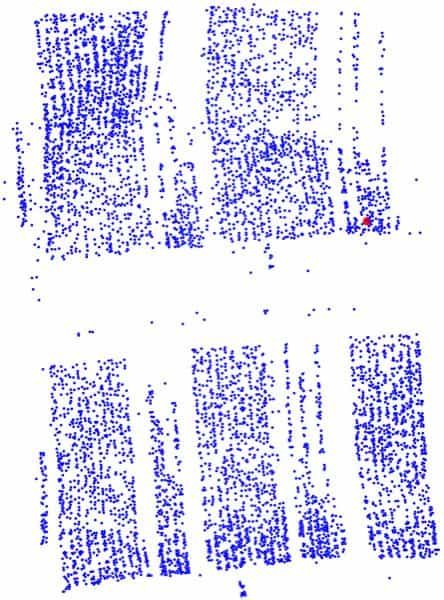}\\
        \vspace{.02\textwidth}%
        \includegraphics[width=1\textwidth, height=1.414\textwidth]{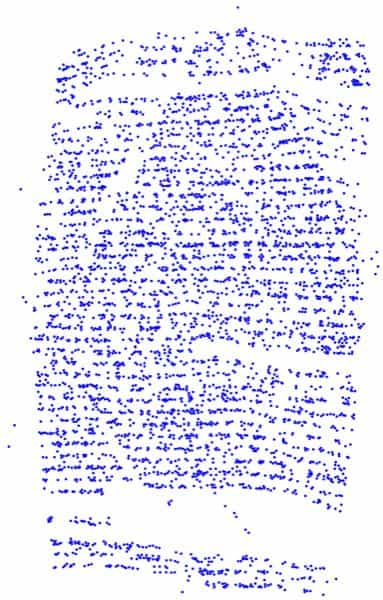}\\
        \vspace{.02\textwidth}%
        \includegraphics[width=1\textwidth, height=1.414\textwidth]{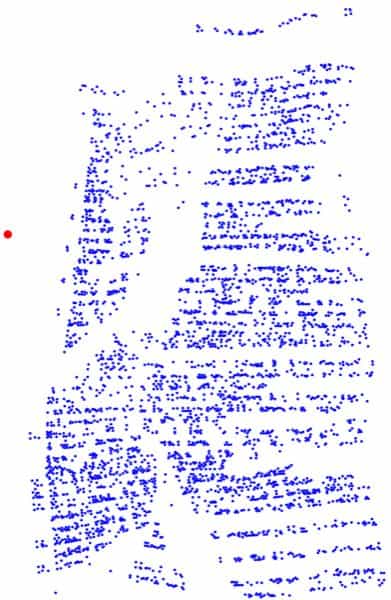}\\
        \vspace{.02\textwidth}%
        \includegraphics[width=1\textwidth, height=1.414\textwidth]{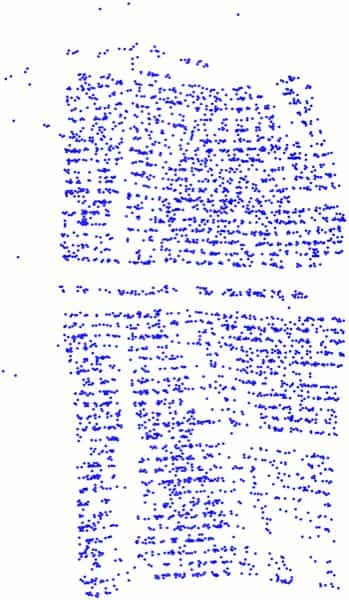}\\
        \vspace{.02\textwidth}%
        \includegraphics[width=1\textwidth, height=1.414\textwidth]{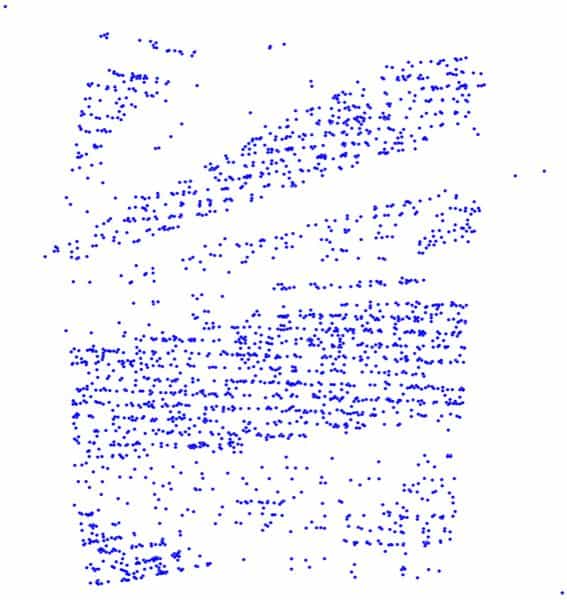}
    \end{minipage}}\hspace{.01\textwidth}%
    \subfigure[]{\centering
    \begin{minipage}[b]{.09\textwidth}\centering
        \includegraphics[width=1\textwidth, height=1.414\textwidth]{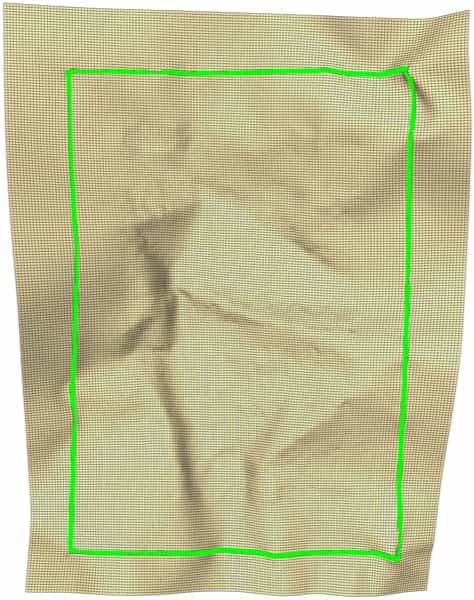}\\
        \vspace{.02\textwidth}%
        \includegraphics[width=1\textwidth, height=1.414\textwidth]{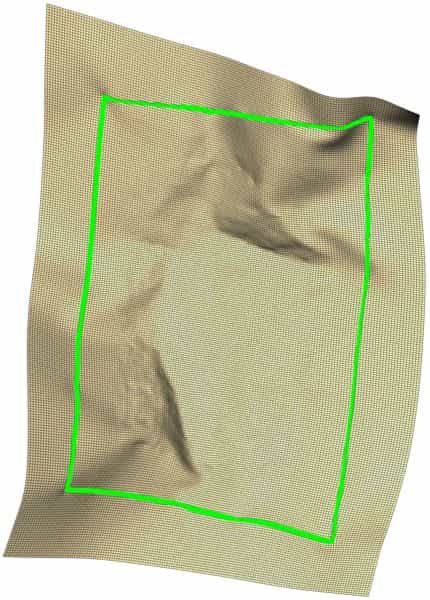}\\
        \vspace{.02\textwidth}%
        \includegraphics[width=1\textwidth, height=1.414\textwidth]{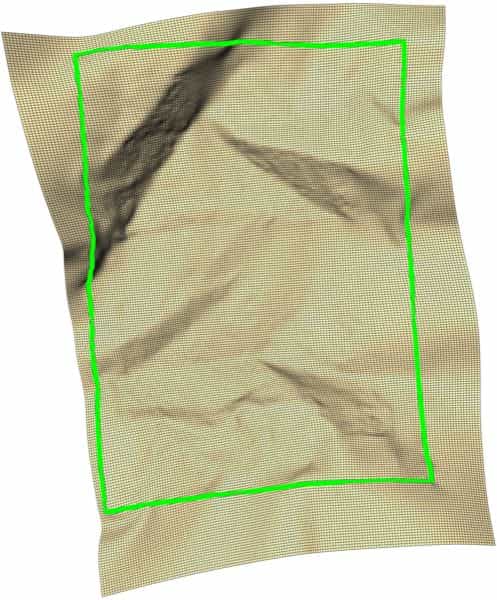}\\
        \vspace{.02\textwidth}%
        \includegraphics[width=1\textwidth, height=1.414\textwidth]{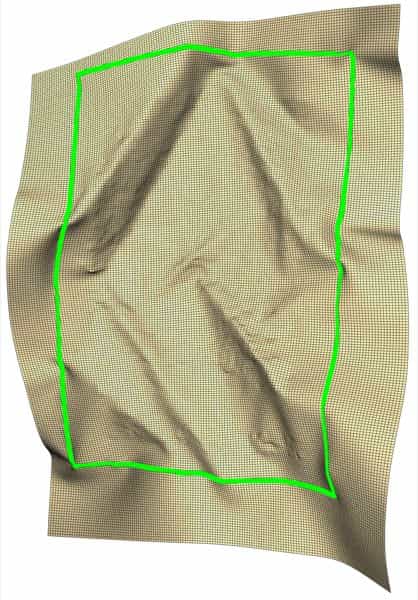}\\
        \vspace{.02\textwidth}%
        \includegraphics[width=1\textwidth, height=1.414\textwidth]{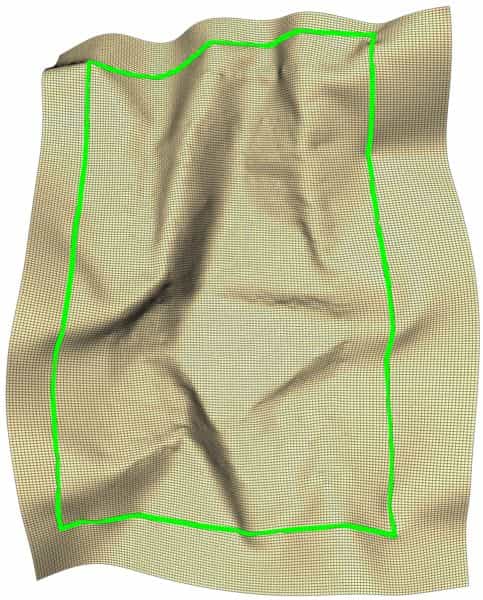}\\
        \vspace{.02\textwidth}%
        \includegraphics[width=1\textwidth, height=1.414\textwidth]{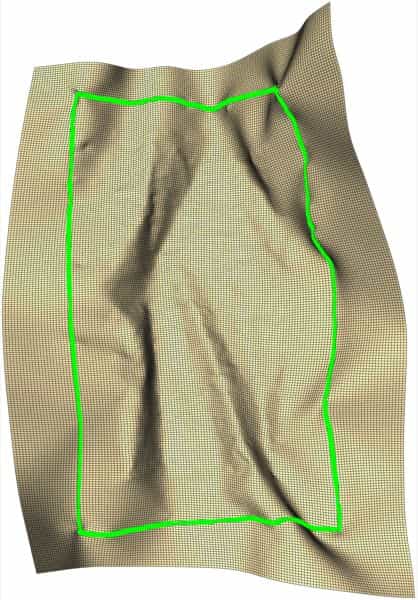}\\
        \vspace{.02\textwidth}%
        \includegraphics[width=1\textwidth, height=1.414\textwidth]{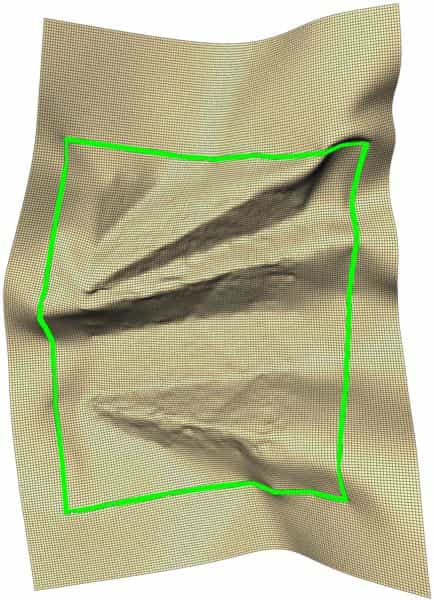}
    \end{minipage}}
\vspace{-0.015\textwidth}
\caption{Dataset \uppercase\expandafter{\romannumeral3}. (a) Original images. (b) Results of  DewarpNet \cite{das2019dewarpnet}. (c) Results of FCN-based \cite{xie2021dewarping}. (d) Results of Points-based~\cite{xie2022document}. (e) Results of DocTr \cite{feng2021doctr}. (f) Results of DocScanner \cite{feng2021docscanner}. (g) Our results. (h) Our results with feature lines. (i) Our final $\mc{P}$. (j) Our final $\mc{M}$ with boundary.}\label{f:results3}
\end{figure}

\begin{enumerate}
\item Dataset \uppercase\expandafter{\romannumeral1}:  statements and posters, including the data 1–6  shown in Fig.~\ref{f:results1}.  This class of documents contains tables or rectangle regions in which edge lines can be utilized as feature line constraints in our method. The texts do not dominate in these documents, and it is difficult to detect text lines. Data 1–4 contain large folding deformations and data 5–6 contain complex creases, all of which are challenging for existing methods.
\item Dataset \uppercase\expandafter{\romannumeral2}: receipts, tickets, and business cards, including the data 7–13  shown in Fig.~\ref{f:results2}. The foldings in these deformed documents are large compared to the size of the document. In these documents, there is  only a small number of text lines or edge lines  can be extracted to serve as  feature line constraints. 
\item  Dataset \uppercase\expandafter{\romannumeral3}: text-based documents, including data 14–20 shown in Fig.~\ref{f:results3}.  These documents contain creases of varying complexity and large folding angles. The text around the creases is distorted significantly, making accurate extraction of the text lines challenging. 
\end{enumerate}

The deep learning methods only require a single image as input, and the original reference image without optical distortion correction is used.  A segmented image with a solid-color background is utilized to remove the effect of background noise. We also crop the images with large background regions. These preprocessing operations on the input images can benefit the deep learning methods. For the FCN-based~\cite{xie2021dewarping}, the input images are scaled to $1024 \times 960$ while retaining the aspect ratio. 

\paragraph{Experimental results and observations} The results of rectified images are shown in Fig.~\ref{f:results1}, Fig.~\ref{f:results2} and Fig.~\ref{f:results3}. We also show the reconstructed 3D models of the documents produced by our method.

The rectified image produced by our method is visually the best among all methods in comparison for every example.  The positions and shapes of the creases are successfully recovered in the reconstructed 3D models, although the creases are not extremely sharp.  The improvements in the results by our methods are even more evident for the document regions with complicated creases. One of the reasons that the deep learning methods cannot give satisfactory results for largely creased documents is that the documents with complicated creases are not considered in particular by these methods, although the datasets used for training contain a large variety of distorted documents. Moreover, the deformation exhibiting complicated creases is definitely more challenging than simple curving or folding, and more advanced neural networks are required to achieve better performance.

The outstanding performance of our method results from the variety of constraints employed in our method, that is, the isometric mapping constraint, the feature line constraints, and the approximation to target data points. The experiments show that our method can produce satisfactory results for the cases where the amount of feature line constraints are quite small, demonstrating the effectiveness of the isometric mapping constraint. This can be observed in the results of data 7, 8, 10,11, and 13 in Fig.~\ref{f:results1}, where the feature lines are quite sparse. There are even more challenging cases with data 17–20 in Fig.~\ref{f:results3} that contain regions with complicated creases without text lines. Data 5 in Fig.~\ref{f:results1}, and data 18–19 in Fig.~\ref{f:results3} contain regions with severe creasing deformations and large folding angles where the target data points are absent due to the lack of textural features in these regions. The pleasing results of our method in the experiments is an evidence of the effectiveness of the proposed approach. We observe that recovering the document regions  with large folding angles is extremely difficult due to the lack of  accurate feature lines extracted in these regions. A more effective feature line detection method can benefit our method by introducing more feature line constraints in optimization. 

\subsection{Quantitative evaluations}
\paragraph{Metrics}
\begin{table*}[t]
\caption{Statistics of our method. The first column shows the index of data. The second column shows the number of data points. The third column shows the number of text feature segments. The fourth column shows the number of edge feature segments. The next four columns show the computation time for each round of optimization in seconds. ${T_n}$ is the total time in the $n$th round of optimization. ${It_n}$ is the iteration number, and ${per_n}$ is the average computation time. The last column shows the total computation time of the algorithm. All data in the table have $\left|\mc{L}^r_B\right|=4$.}\label{tab:my-table}
\centering
\begin{adjustbox}{max width=\textwidth}
\begin{tabular}{cccccrrrrr}
\toprule
Data &
$\left|\hat{\mc{P}}\right|$ &
$\left|\mc{L}^r_T\right|$ &   
$\left|\mc{L}^r_E\right|$ &
${T_0} /{It_0}={per_0}$ &
${T_1} /{It_1}={per_1}$ &
  ${T_2} /{It_2}={per_2}$ &
  ${T_3} /{It_3}={per_3}$ &
  ${T_{all}} /{It_{all}}={per_{all}}$ \\
\midrule
1 &4.7k &98 &1541 &5.45/14=0.38 &3.11/7=0.44 &3.68/5=0.73 &6.77/3=2.25 &19.02/29=0.65 \\
2 &5.3k &42 &2567 &22.25/45=0.49 &10.74/19=0.56 &8.14/9=0.9 &14.73/6=2.45 &55.88/79=0.7 \\
3 &1.1k &0 &2713 &3.23/6=0.53 &4.89/8=0.61 &3.67/4=0.91 &7.41/3=2.47 &19.21/21=0.91 \\
4 &1.6k &73 &1103 &16.92/61=0.27 &2.63/8=0.32 &2.97/5=0.59 &8.57/4=2.14 &31.1/78=0.39 \\
5 &7.6k &35 &4711 &15.79/24=0.65 &20.8/28=0.74 &18.31/16=1.14 &48.92/18=2.71 &103.84/86=1.2 \\
6 &6.2k &406 &1010 &5.01/15=0.33 &5.21/13=0.4 &5.07/7=0.72 &46.97/20=2.34 &62.27/55=1.13 \\
7 &3.5k &0 &1346 &4.33/18=0.24 &2.33/8=0.29 &2.69/5=0.53 &10.75/5=2.15 &20.12/36=0.55 \\
8 &2.0k &0 &266 &0.59/7=0.08 &1.09/8=0.13 &1.77/5=0.35 &13.02/7=1.86 &16.48/27=0.61 \\
9 &1.4k &0 &1479 &5.59/22=0.25 &2.13/7=0.3 &3.39/6=0.56 &12.58/6=2.09 &23.71/41=0.57 \\
10 &2.6k &0 &325 &1.86/17=0.1 &1.1/7=0.15 &1.93/5=0.38 &7.57/4=1.89 &12.47/33=0.37 \\
11 &3.5k &0 &2393 &9.25/21=0.44 &3.55/7=0.5 &3.81/5=0.76 &11.48/5=2.29 &28.11/38=0.73 \\
12 &2.0k &0 &1085 &8.93/42=0.21 &1.82/7=0.26 &2.51/5=0.5 &18.53/9=2.05 &31.8/63=0.5 \\
13 &2.6k &0 &544 &5.74/39=0.14 &1.41/7=0.2 &2.63/6=0.43 &11.91/6=1.98 &21.7/58=0.37 \\
14 &10.3k &242 &42 &4.86/16=0.3 &6.68/18=0.37 &7.46/11=0.67 &13.57/6=2.26 &32.58/51=0.63 \\
15 &8.6k &210 &39 &9.54/33=0.28 &6.28/18=0.34 &7.05/11=0.64 &13.62/6=2.27 &36.51/68=0.53 \\
16 &9.5k &279 &1006 &13.89/30=0.46 &11.16/20=0.55 &17.96/19=0.94 &46.32/18=2.57 &89.34/87=1.02 \\
17 &5.9k &231 &624 &3.77/13=0.29 &5.4/14=0.38 &7.13/10=0.71 &18.43/8=2.3 &34.75/45=0.77 \\
18 &5.7k &272 &538 &4.67/17=0.27 &4.62/14=0.33 &6.74/11=0.61 &15.68/7=2.24 &31.73/49=0.64 \\
19 &5.2k &208 &382 &5.13/18=0.28 &5.33/15=0.35 &5.62/8=0.7 &36.92/16=2.3 &53.01/57=0.93 \\
20 &2.8k &211 &235 &2.28/9=0.25 &5.18/15=0.34 &6.05/9=0.67 &18.6/8=2.32 &32.11/41=0.78 \\
\bottomrule
\end{tabular}
\end{adjustbox}
\end{table*}
\begin{figure}[!htbp]
\centering
\includegraphics[width=.67\textwidth]{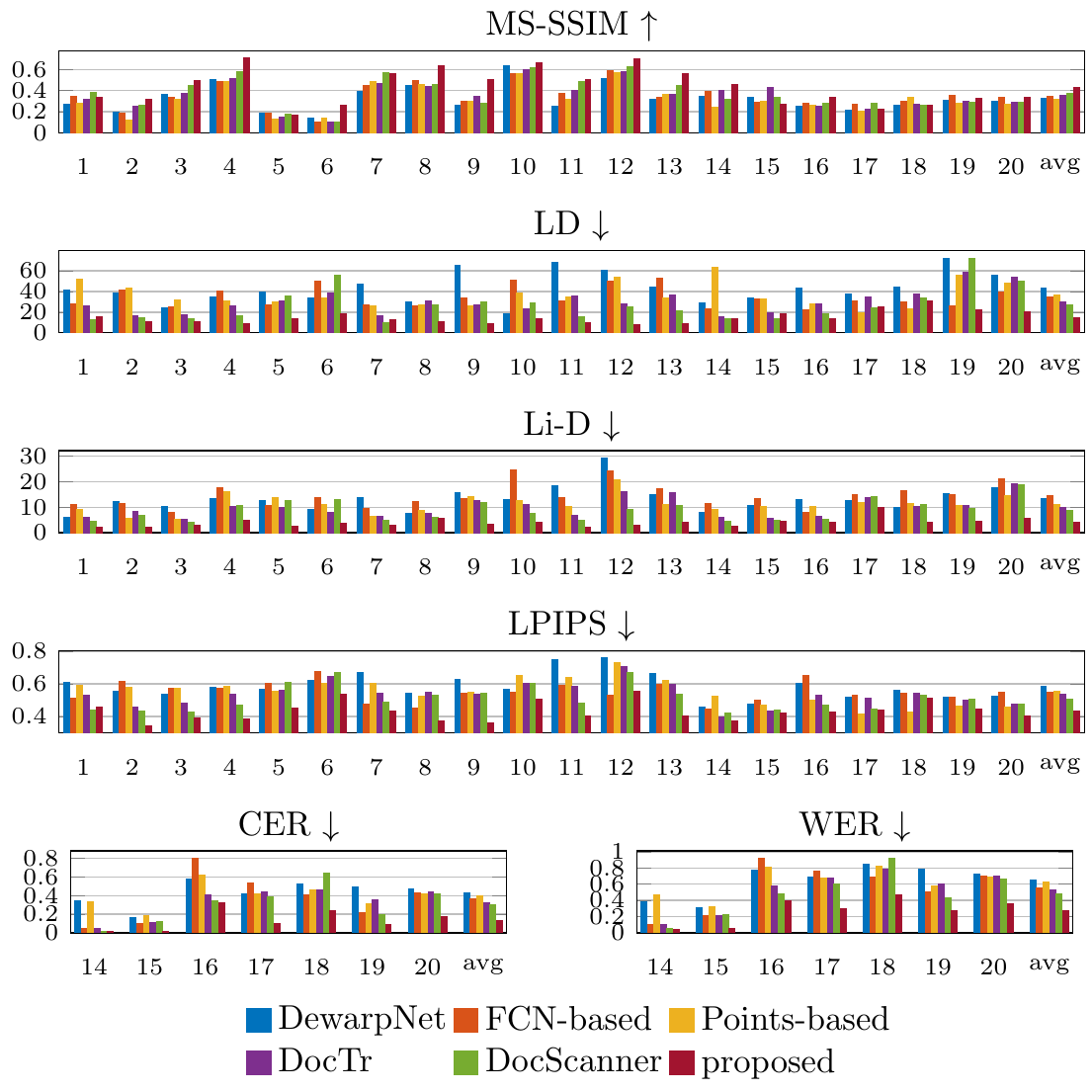}
\vspace{-.015\textwidth}
\caption{Quantitative comparisons by a variety of metrics, evaluating the similarity between the rectified image and the ground-truth image. $\uparrow$ signifies that a higher value means a larger similarity, whereas $\downarrow$ means the reverse. }\label{f:metrics}
\end{figure}

The metrics are defined to evaluate the difference between the rectified image and the ground truth image.  We obtain the ground truth of each example by scanning the flat document.  Various metrics are proposed in the literature for image comparison and no one is optimal. Therefore, we select several widely used metrics listed in the following, including two OCR metrics for text-based documents, and four metrics for image similarity and pixel displacement.


\begin{itemize}
\item{MS-SSIM.} The MS-SSIM \cite{wang2003multiscale} metric has been used in prevailing works \citep{das2019dewarpnet, xie2021dewarping, xie2022document, feng2021doctr, feng2021docscanner} to evaluate image similarity. MS-SSIM is a multiscale extension of SSIM \cite{wang2004image} measuring image similarity by luminance, contrast, and structure. 

\item{LD.} Local Distortion (LD) ~\cite{you2018multiview} measures pixel displacements $(\Delta \boldsymbol{x}, \Delta \boldsymbol{y})$ between two images  by the mean $L_2$ distance between the matched pixels, in which the dense SIFT-flow approach~\cite{celiu2011sift} is employed for image registration. To improve the accuracy of image registration in our work, we apply the more recent  approach  for computing the dense flow, namely the DeepFlow~\cite{weinzaepfel2013deepflow}, which is more robust and accurate than the SIFT-flow.

\item{Li-D.} Line Distortion (Li-D) is introduced in DocScanner~\cite{feng2021docscanner}, concerning the global distortions in the horizontal and vertical directions. The standard deviation of each row and column in the image is calculated based on the results of DeepFlow.

\item{LPIPS.} Learned Perceptual Image Patch Similarity (LPIPS)~\cite{zhang2018unreasonable} calculates a value that represents the subjective perception of the human eye, by comparing the activations of two image patches for some particular network.

\item{CER and WER.}
To evaluate OCR performance for text-based documents, the OCR accuracy is calculated in terms of Character Error Rate (CER) or Word Error Rate (WER) between the reference and recognized text. We use PyTesseract \cite{smith2007overview} as our OCR engine to calculate these metrics.
\end{itemize}  


\paragraph{Evaluation results and discussions}

Before calculating the metrics, all rectified and ground truth images are resized to the same resolution of $1000 \times 1500$. In addition, the shadows caused by creases are removed for OCR-based metrics.  Since high-resolution images of the results in~\cite{you2018multiview}  are unavailable, quantitative comparisons to them are not presented. The quantitative metrics for all examples are shown in Fig.~\ref{f:metrics}.

Our method almost produces the best value with respect to each metric, in terms of every single example and the average statistic. The superiority of our method becomes even more apparent with more complicated creases (e.g., data 5–6 in Fig.~\ref{f:results1} and data 17–20 in Fig.~\ref{f:results3}). However, since every metric has certain limitations, the metric values are not always consistent with the visual performance (e.g. data 1 in Fig.~\ref{f:results1}). Statistically, the average value is more reliable. In the dataset \uppercase\expandafter{\romannumeral3}, the metric values based on image similarity or pixel displacement are not always the best with our method. The performance of the OCR metrics is more consistent with the visual effect, demonstrating that, for text-based documents, OCR-based measurements are more robust than similarity-based metrics.

\paragraph{Time efficiency}

Table \ref{tab:my-table} shows some statistics and computation time for each data. We employ four rounds of optimization for each example, separated by three subdivision operations. For each round of optimization, the total time, the number of iterations, and the average time per iteration are given.  It is not surprising that the average time increases  with more subdivisions introducing more variables in optimization. It is also observed that more feature lines lead to longer computation time. The total time and number of iterations in each round of optimization depend on the complexity of a particular example.

\subsection{Method evaluation}
\label{Sec:added}
In this part, we investigate several factors that affect our method's performance, including the quality of the point cloud, the quality and quantity of feature lines, and incomplete document boundaries. We present some experiments to demonstrate the flexibility of our method in dealing with target data points and feature lines with different qualities.

\paragraph{Point cloud quality} 
The quality of the point cloud is highly dependent on the quality and amount of the input images. The point cloud generated from a small number of photos often contains holes that are challenging for document rectification.  We have demonstrated that our method can produce better results than existing methods with only 3–5 images, in which the data points are sparse, possibly with missing parts. In this section, we conduct more experiments to demonstrate the performance of our method with denser point clouds, generated using 8–14 images taken from various viewpoints. 

\begin{figure}[!htbp]
\centering
    \subfigure[]{\centering
    \begin{minipage}[b]{.10\textwidth}\centering
    \includegraphics[width=1\textwidth, height=1.414\textwidth]{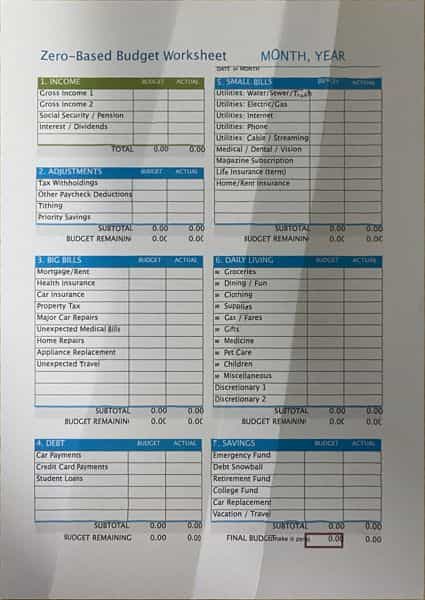}\\
        \vspace{.02\textwidth}%
        \includegraphics[width=1\textwidth, height=1.414\textwidth]{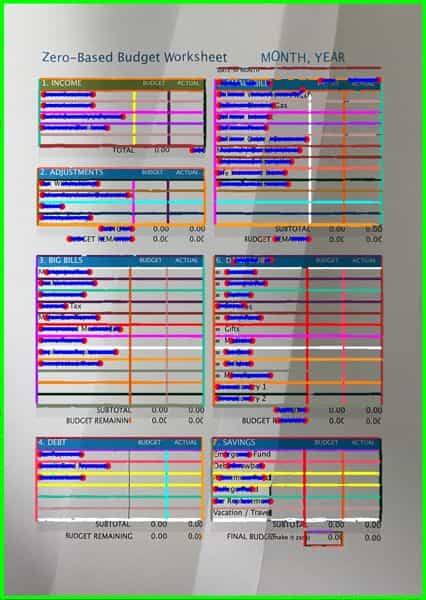}\\
        \vspace{.02\textwidth}%
        \includegraphics[width=1\textwidth, height=1.414\textwidth]{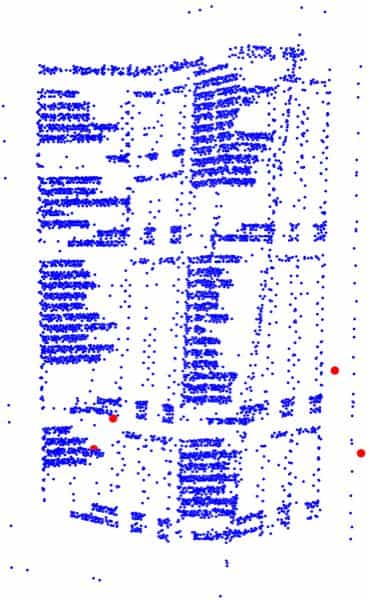}\\
        \vspace{.02\textwidth}%
        \includegraphics[width=1\textwidth, height=1.414\textwidth]{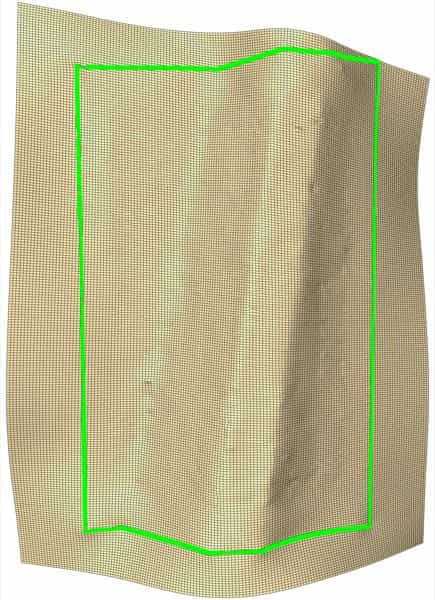}
    \end{minipage}}\hspace{.01\textwidth}%
    \subfigure[]{\centering
    \begin{minipage}[b]{.10\textwidth}\centering
    	\includegraphics[width=1\textwidth, height=1.414\textwidth]{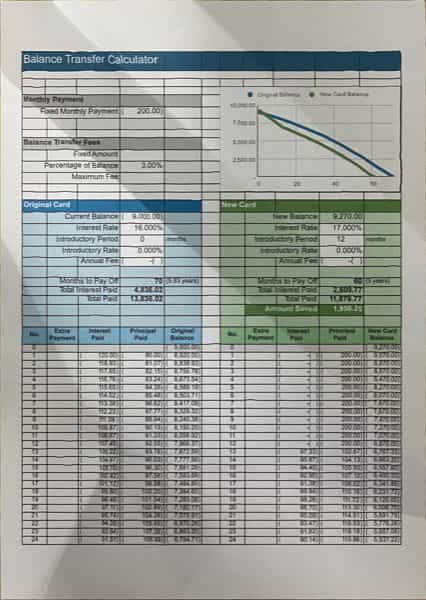}\\
        \vspace{.02\textwidth}%
        \includegraphics[width=1\textwidth, height=1.414\textwidth]{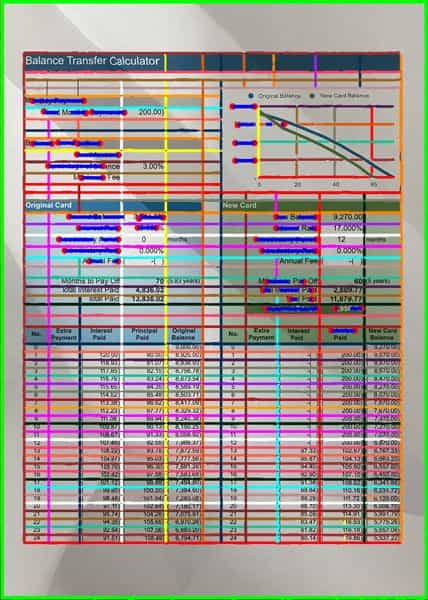}\\
        \vspace{.02\textwidth}%
        \includegraphics[width=1\textwidth, height=1.414\textwidth]{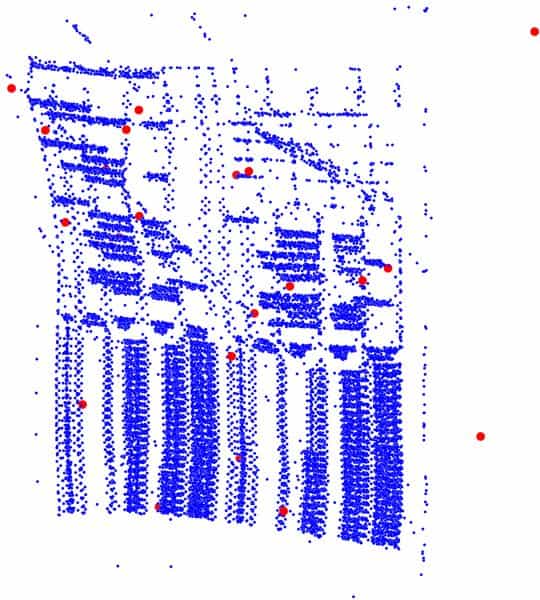}\\
        \vspace{.02\textwidth}%
        \includegraphics[width=1\textwidth, height=1.414\textwidth]{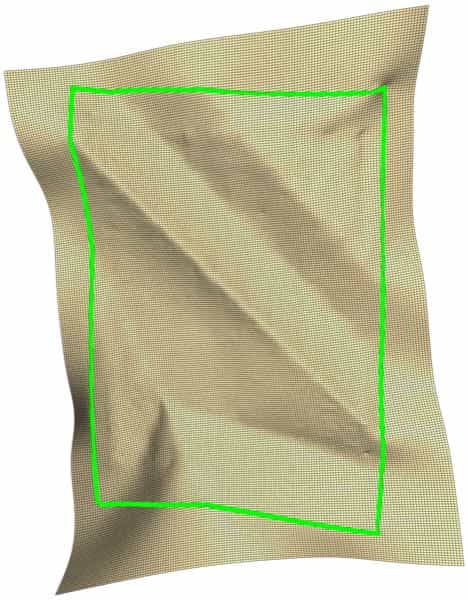}
    \end{minipage}}\hspace{.01\textwidth}%
    \subfigure[]{\centering
    \begin{minipage}[b]{.10\textwidth}\centering
    	\includegraphics[width=1\textwidth, height=1.414\textwidth]{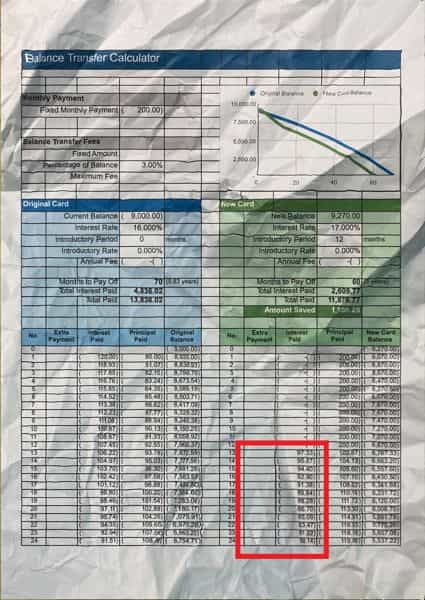}\\
        \vspace{.02\textwidth}%
        \includegraphics[width=1\textwidth, height=1.414\textwidth]{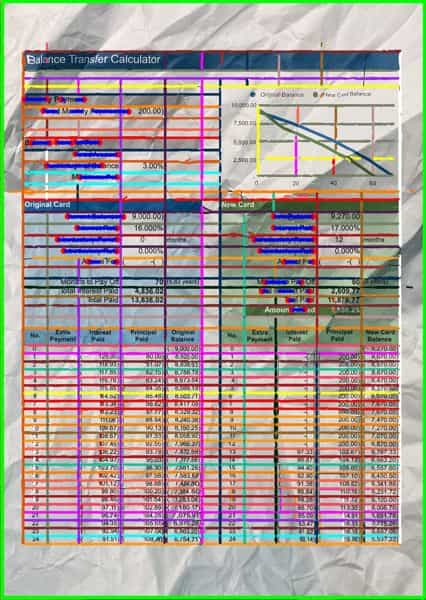}\\
        \vspace{.02\textwidth}%
        \includegraphics[width=1\textwidth, height=1.414\textwidth]{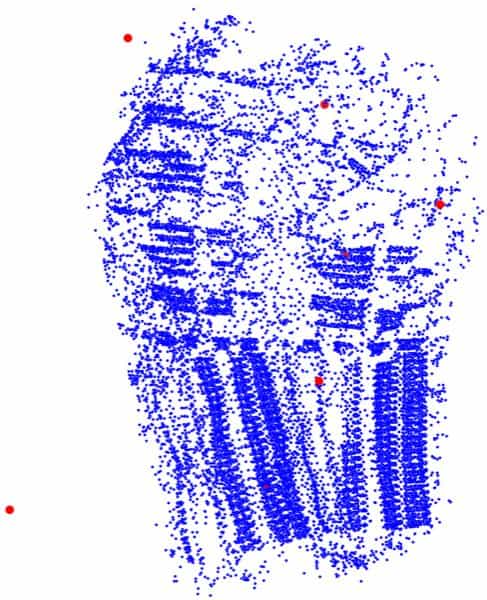}\\
        \vspace{.02\textwidth}%
        \includegraphics[width=1\textwidth, height=1.414\textwidth]{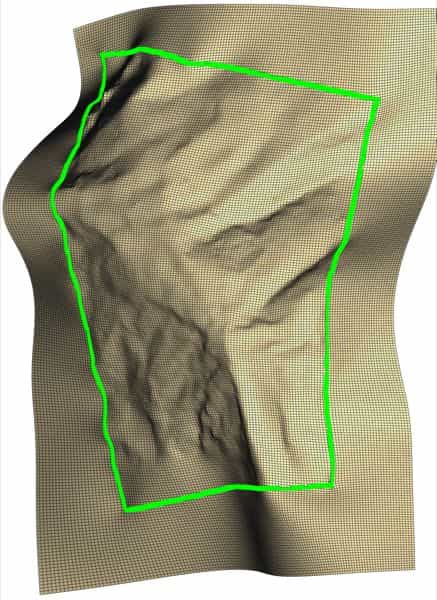}
    \end{minipage}}\hspace{.01\textwidth}%
   \subfigure[]{\centering
    \begin{minipage}[b]{.10\textwidth}\centering
    	\includegraphics[width=1\textwidth, height=1.414\textwidth]{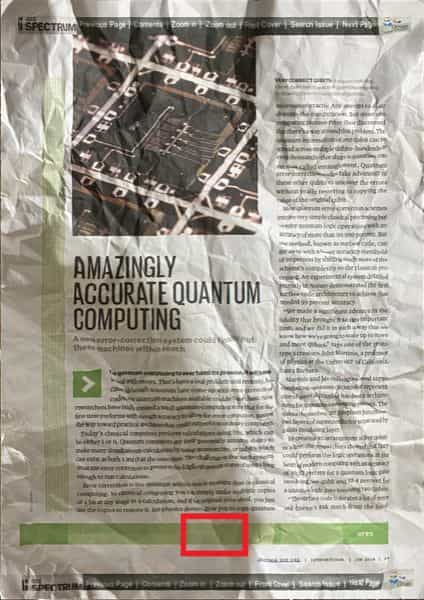}\\
        \vspace{.02\textwidth}%
        \includegraphics[width=1\textwidth, height=1.414\textwidth]{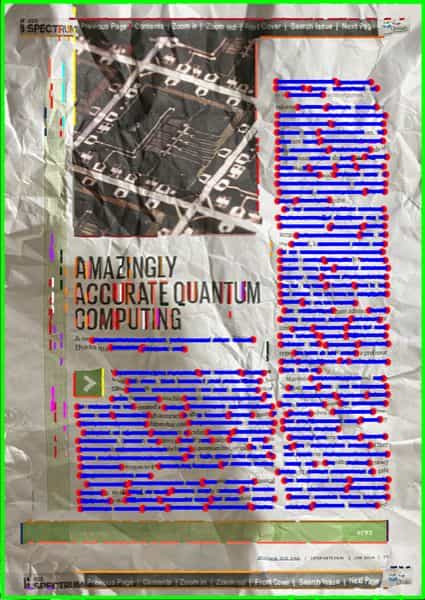}\\
        \vspace{.02\textwidth}%
        \includegraphics[width=1\textwidth, height=1.414\textwidth]{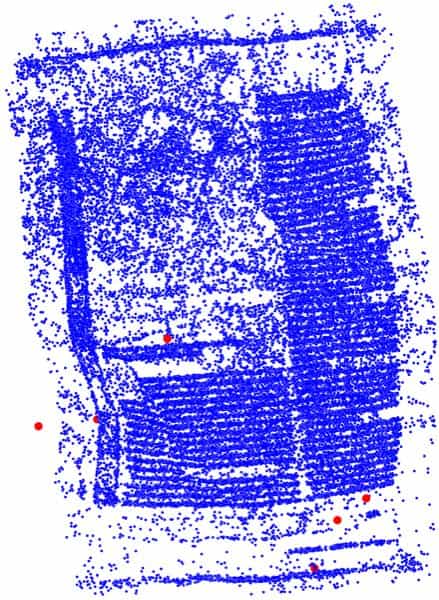}\\
        \vspace{.02\textwidth}%
        \includegraphics[width=1\textwidth, height=1.414\textwidth]{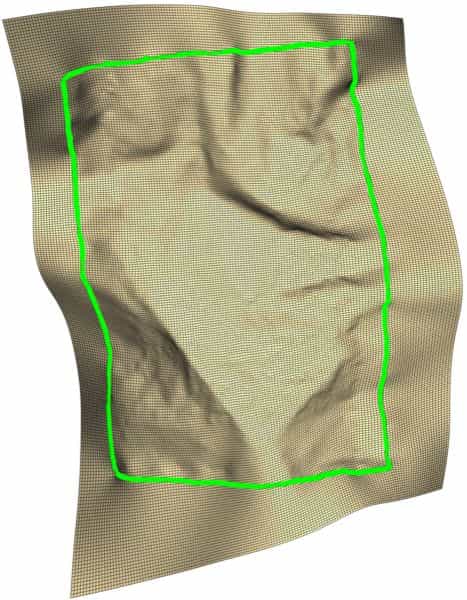}
    \end{minipage}}\hspace{.01\textwidth}%
    \subfigure[]{\centering
    \begin{minipage}[b]{.10\textwidth}\centering
    	\includegraphics[width=1\textwidth, height=1.414\textwidth]{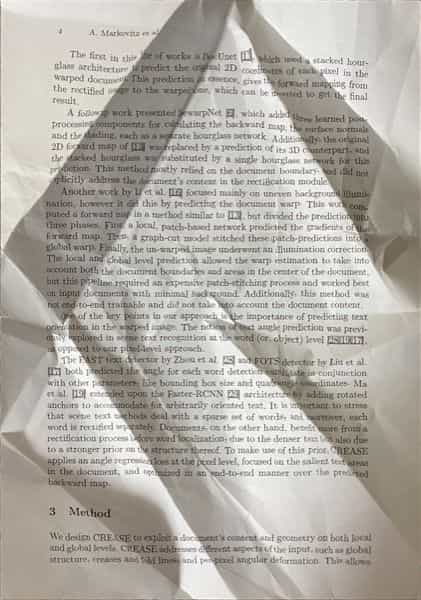}\\
        \vspace{.02\textwidth}%
        \includegraphics[width=1\textwidth, height=1.414\textwidth]{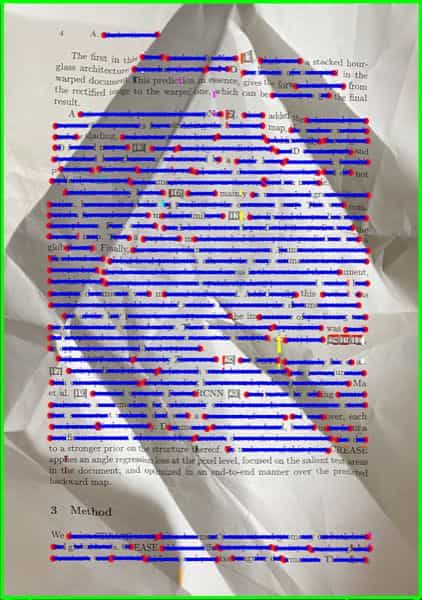}\\
        \vspace{.02\textwidth}%
        \includegraphics[width=1\textwidth, height=1.414\textwidth]{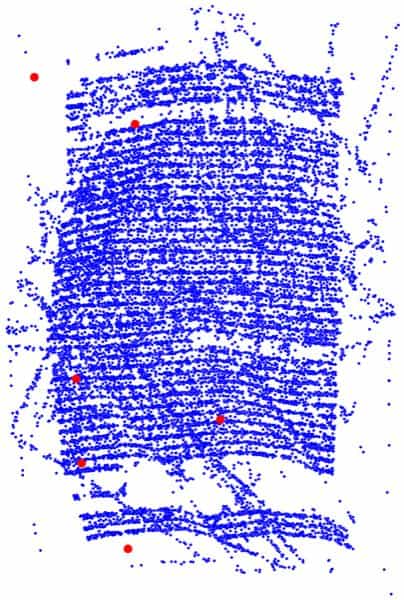}\\
        \vspace{.02\textwidth}%
        \includegraphics[width=1\textwidth, height=1.414\textwidth]{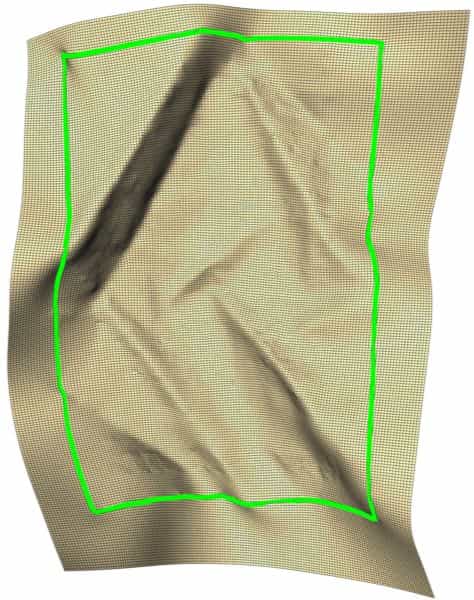}
    \end{minipage}}\hspace{.01\textwidth}%
    \subfigure[]{\centering
    \begin{minipage}[b]{.10\textwidth}\centering
    	\includegraphics[width=1\textwidth, height=1.414\textwidth]{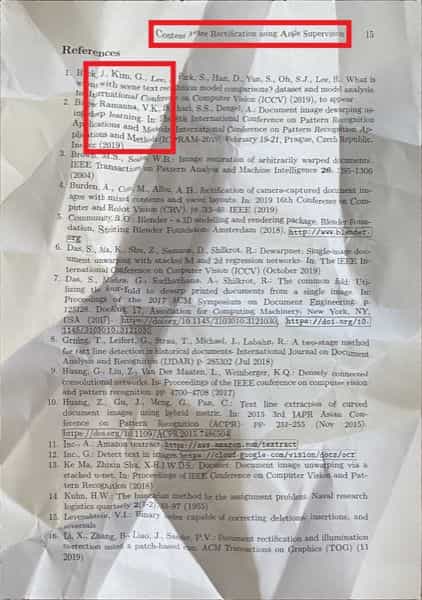}\\
        \vspace{.02\textwidth}%
        \includegraphics[width=1\textwidth, height=1.414\textwidth]{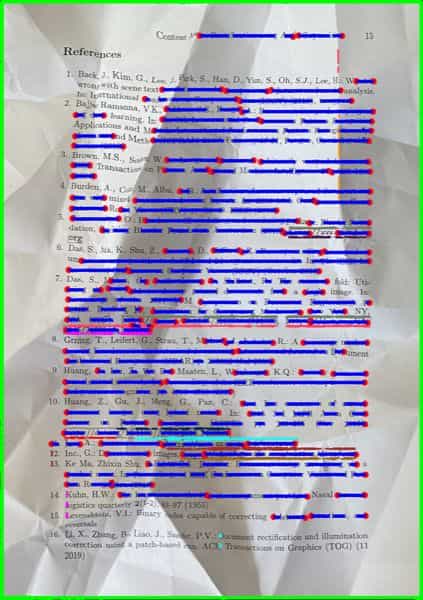}\\
        \vspace{.02\textwidth}%
        \includegraphics[width=1\textwidth, height=1.414\textwidth]{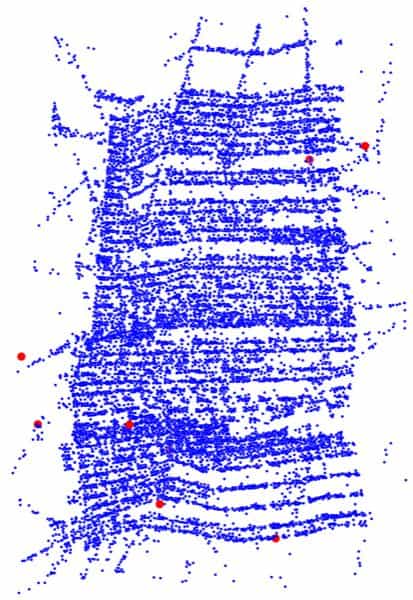}\\
        \vspace{.02\textwidth}%
        \includegraphics[width=1\textwidth, height=1.414\textwidth]{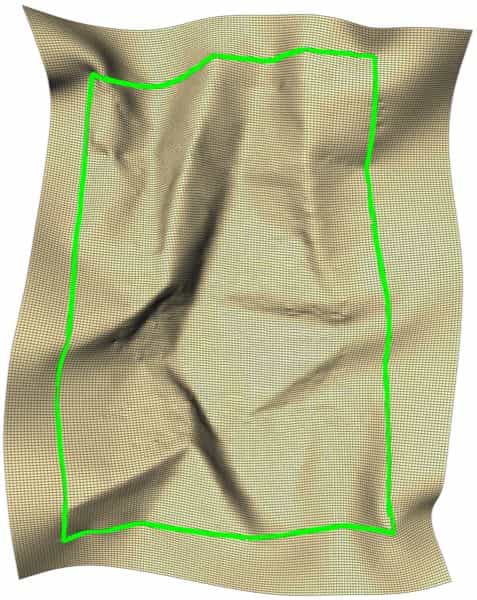}
    \end{minipage}}\hspace{.01\textwidth}%
   \subfigure[]{\centering
    \begin{minipage}[b]{.10\textwidth}\centering
    	\includegraphics[width=1\textwidth, height=1.414\textwidth]{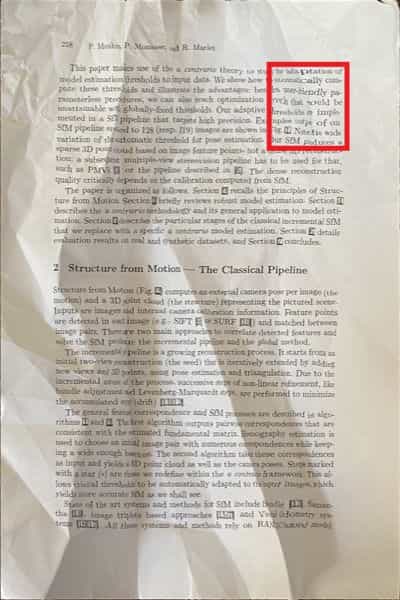}\\
        \vspace{.02\textwidth}%
        \includegraphics[width=1\textwidth, height=1.414\textwidth]{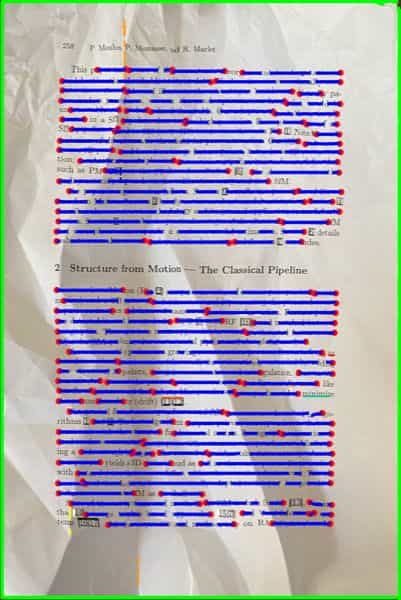}\\
        \vspace{.02\textwidth}%
        \includegraphics[width=1\textwidth, height=1.414\textwidth]{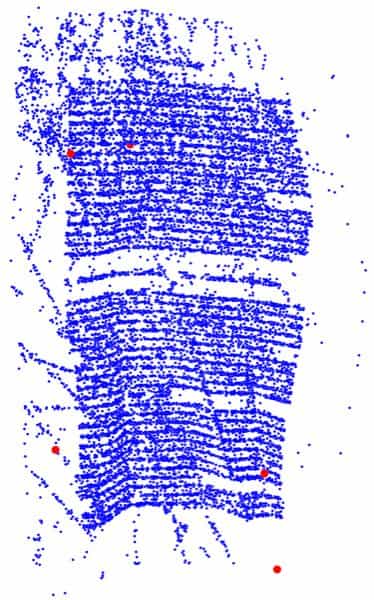}\\
        \vspace{.02\textwidth}%
        \includegraphics[width=1\textwidth, height=1.414\textwidth]{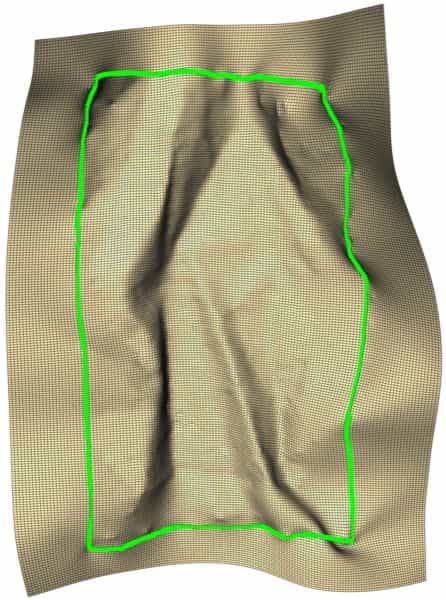}
    \end{minipage}}\hspace{.01\textwidth}%
    \subfigure[]{\centering
    \begin{minipage}[b]{.10\textwidth}\centering
    	\includegraphics[width=1\textwidth, height=1.414\textwidth]{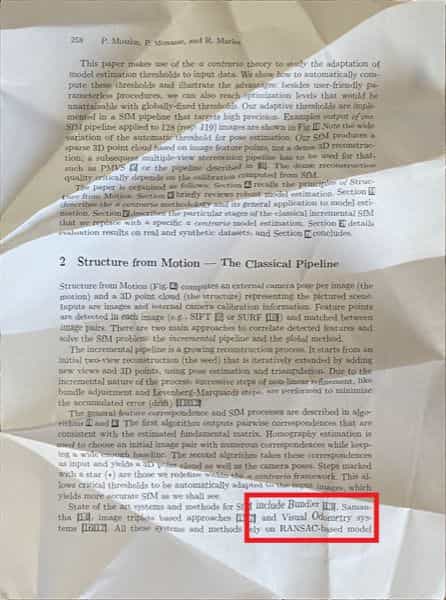}\\
        \vspace{.02\textwidth}%
        \includegraphics[width=1\textwidth, height=1.414\textwidth]{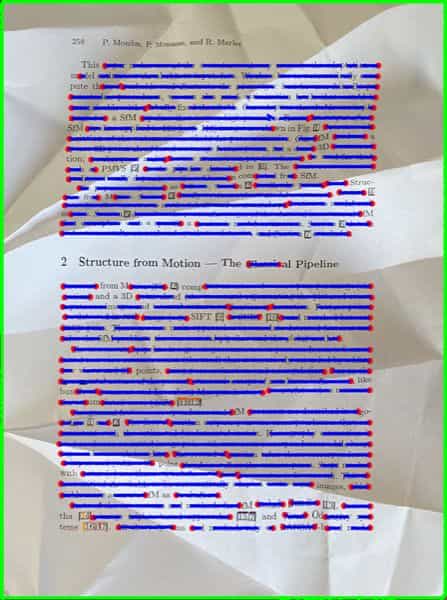}\\
        \vspace{.02\textwidth}%
        \includegraphics[width=1\textwidth, height=1.414\textwidth]{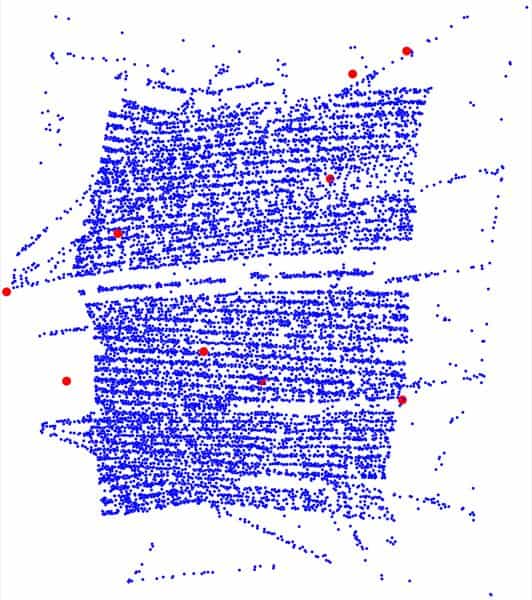}\\
        \vspace{.02\textwidth}%
        \includegraphics[width=1\textwidth, height=1.414\textwidth]{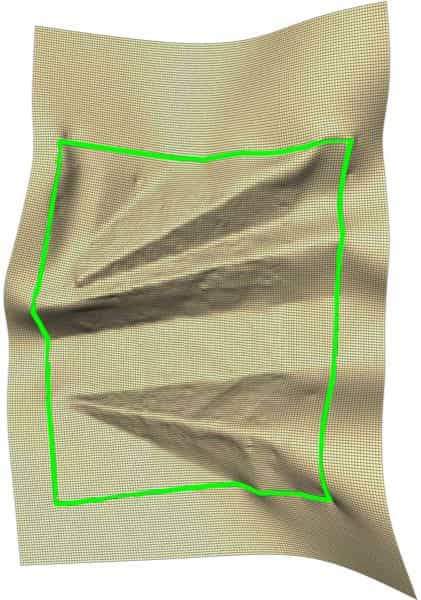}
    \end{minipage}}
\vspace{-.015\textwidth}
\caption{Point cloud quality experiments. Columns (a) through (h) show the new results with more photos, corresponding to data 1, 2, 5, 6, 17, 18, 19, and 20, respectively. Row 1 shows our new rectification results, and the areas with significant differences from the previous results are marked in red. Row 2 shows the generated feature lines. Row 3 shows our final $\mc{P}$. Row 4 shows our final $\mc{M}$ with boundary.}
\label{f:points}
\end{figure}

We select the data 1, 2, 5, 6, and 17–20 used in the previous experiments, most of which are cases with complicated creases. For these cases, a smaller number of images might be insufficient to generate a point cloud of high quality, possibly with holes, due to occlusions at certain viewports, or lack of texture features in the images.  These challenging cases are partially solved when additional images are included.  Referring to Fig.~\ref{f:points},  the marked region of data 6 and the title part of data 18 both have sparse textures, making it difficult to generate dense data points using a small number of images. Improvements are also observed around the marked regions of data 5, the upper left corner of data 18, data 19, and data 20.

In Table~\ref{tab:my-table-points}, we give some statistics, including the number of data points and the computation time.  We observe that a denser point cloud, generated from a larger number of images,  generally increases the computation time for each round of optimization, but speeds up the convergence of the optimization, resulting in a decrease in the total computation time.

\begin{figure}[!t]
\centering
    \subfigure[]{\centering
    \begin{minipage}[b]{.11\textwidth}\centering
    \includegraphics[width=1\textwidth, height=1.414\textwidth]{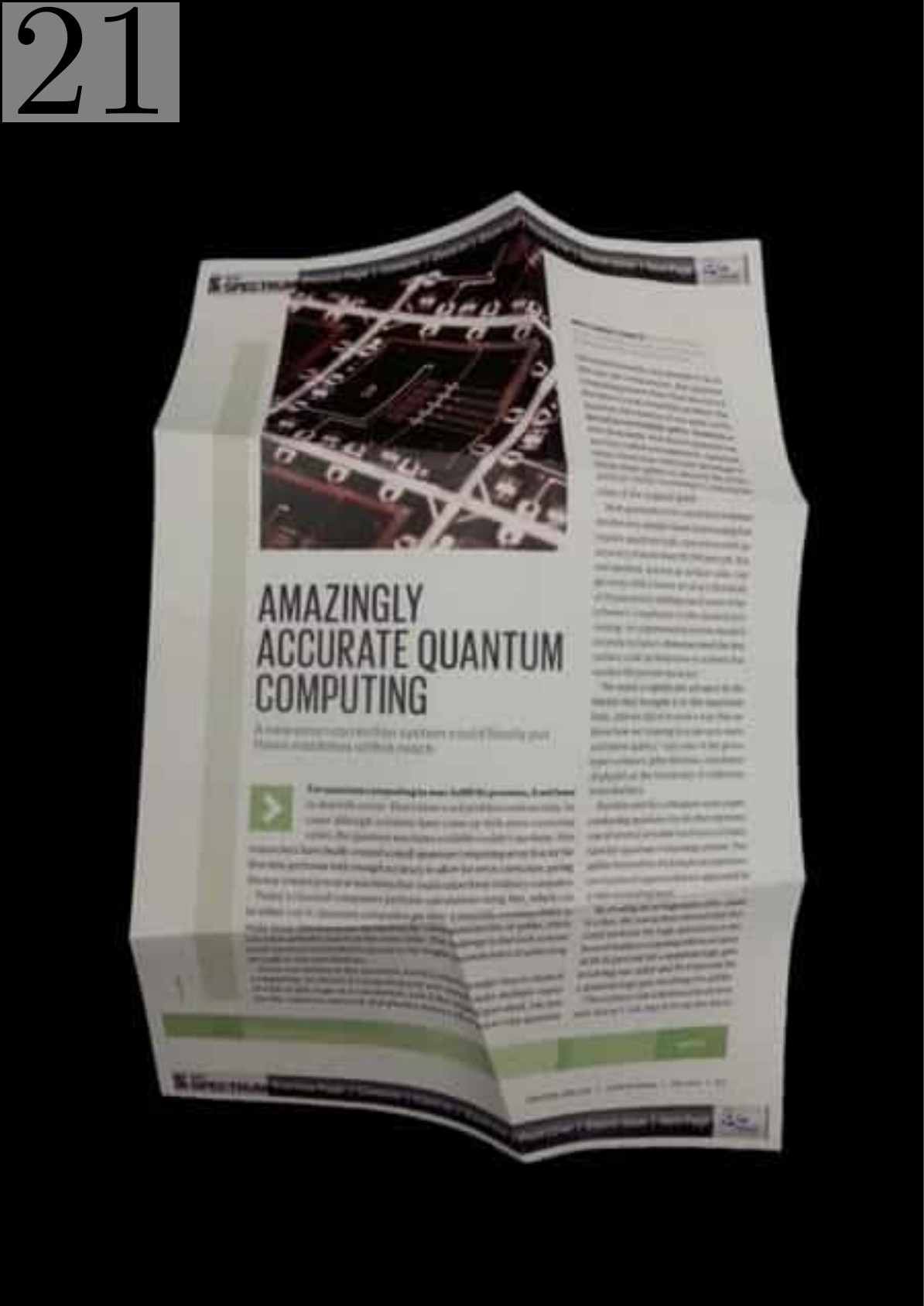}\\
    \vspace{.02\textwidth}%
    \includegraphics[width=1\textwidth, height=1.414\textwidth]{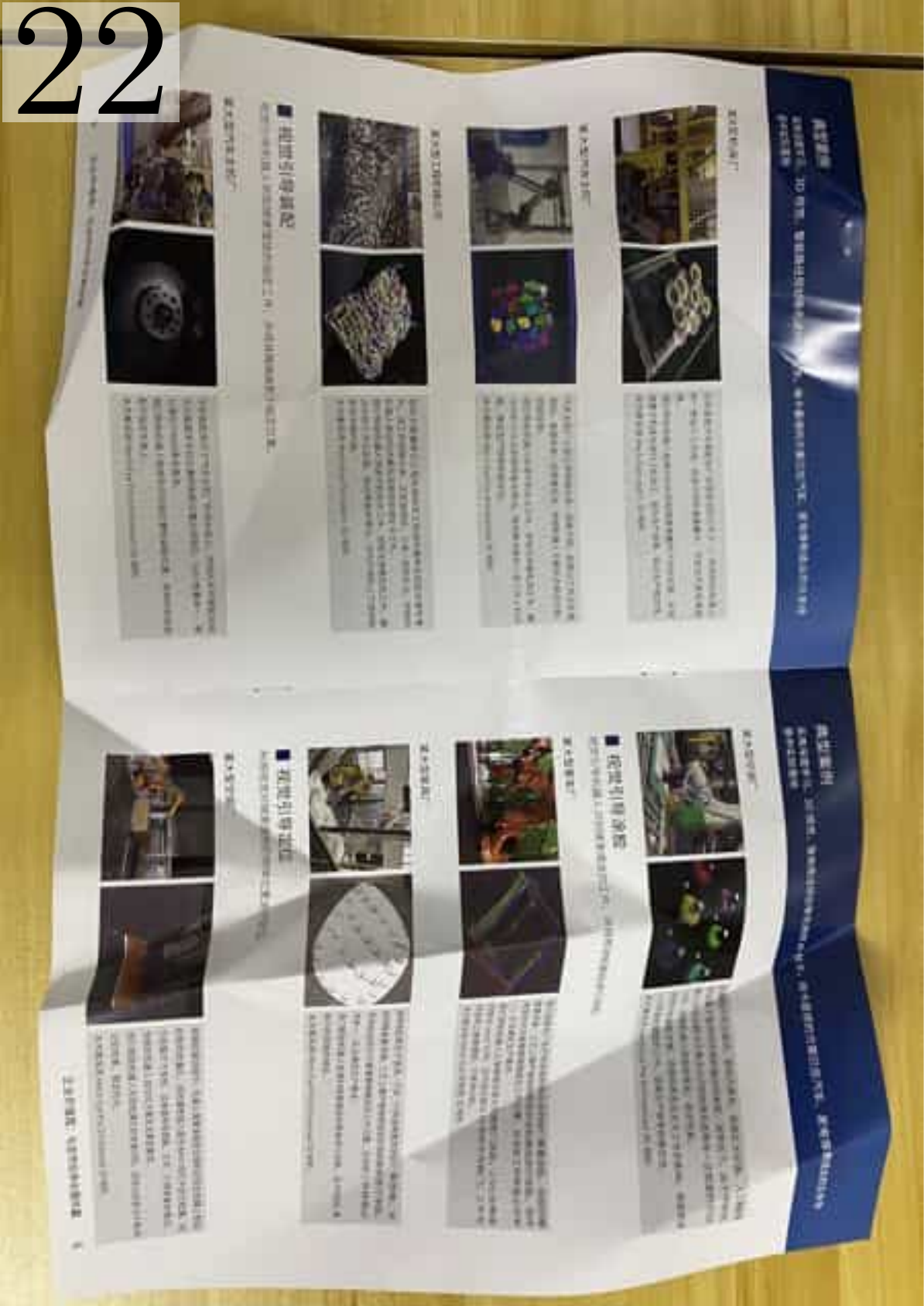}
    \end{minipage}}\hspace{.01\textwidth}%
    \subfigure[]{\centering
    \begin{minipage}[b]{.11\textwidth}\centering
        \includegraphics[width=1\textwidth, height=1.414\textwidth]{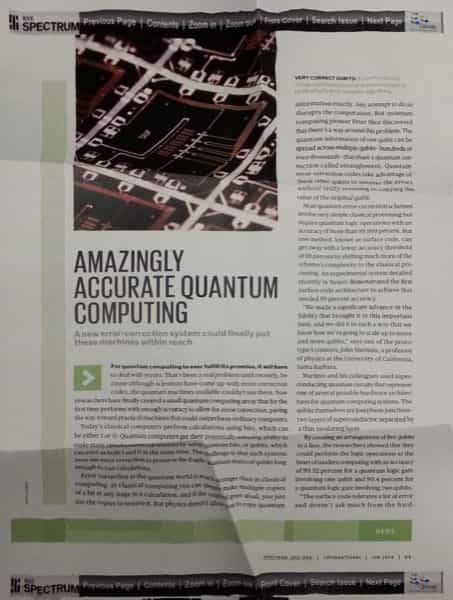}\\
        \vspace{.02\textwidth}%
        \includegraphics[angle=90,width=1\textwidth, height=1.414\textwidth]{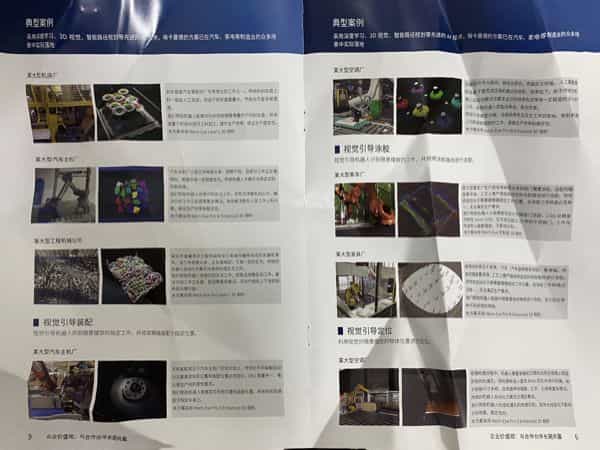}
    \end{minipage}}
    \vspace{-.0\textwidth}%
    \subfigure[]{\centering
    \begin{minipage}[b]{.11\textwidth}\centering
        \includegraphics[width=1\textwidth, height=1.414\textwidth]{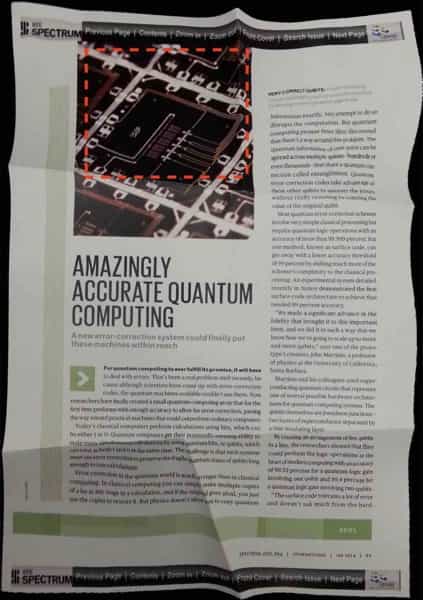}\\
        \vspace{.02\textwidth}%
        \includegraphics[angle=90, width=1\textwidth, height=1.414\textwidth]{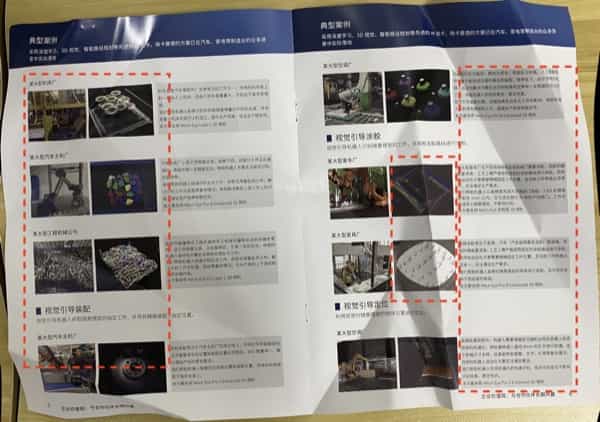}
    \end{minipage}}\hspace{.01\textwidth}%
    \subfigure[]{\centering
    \begin{minipage}[b]{.11\textwidth}\centering
        \includegraphics[width=1\textwidth, height=1.414\textwidth]{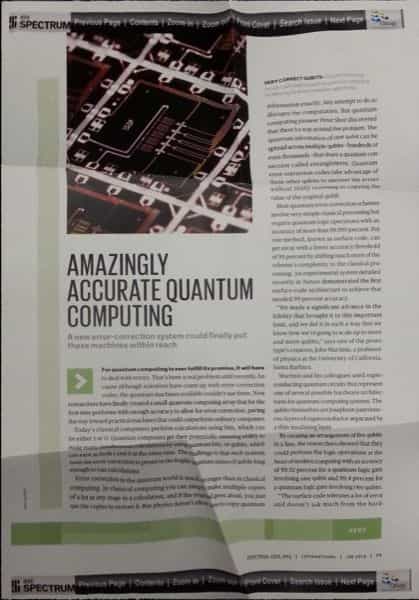}\\
        \vspace{.02\textwidth}%
        \includegraphics[angle=90, width=1\textwidth, height=1.414\textwidth]{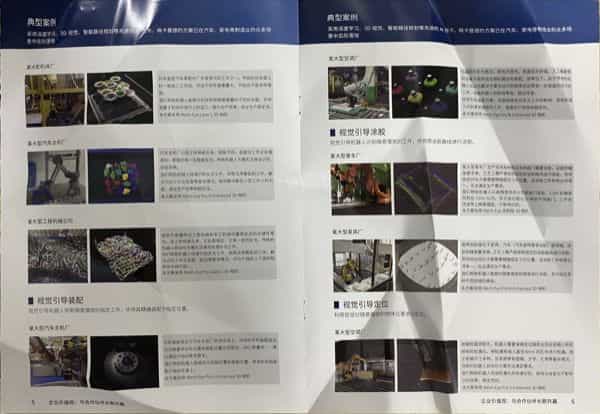}
    \end{minipage}}\hspace{.01\textwidth}%
    \subfigure[]{\centering
    \begin{minipage}[b]{.11\textwidth}\centering
        \includegraphics[width=1\textwidth, height=1.414\textwidth]{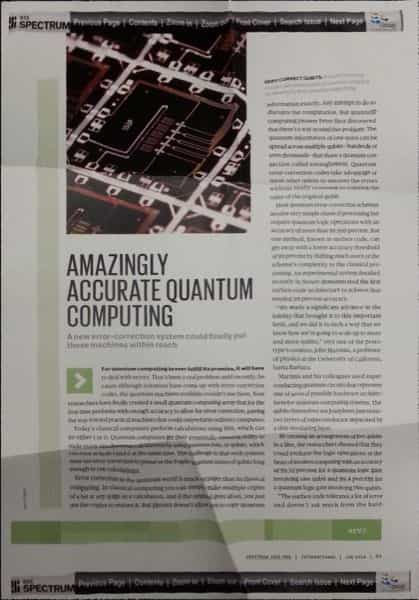}\\
        \vspace{.02\textwidth}%
        \includegraphics[angle=90, width=1\textwidth, height=1.414\textwidth]{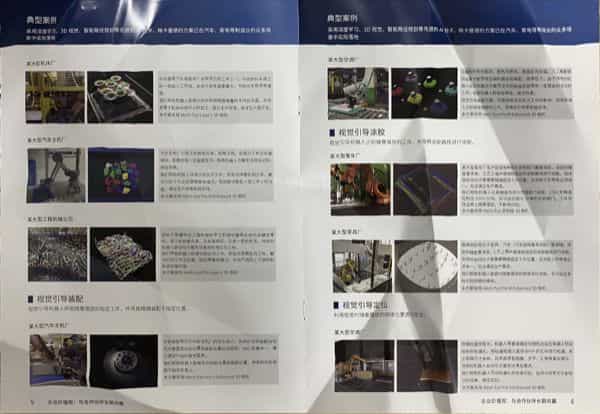}
    \end{minipage}}\hspace{.01\textwidth}%
    \subfigure[]{\centering
    \begin{minipage}[b]{.11\textwidth}\centering
        \includegraphics[width=1\textwidth, height=1.414\textwidth]{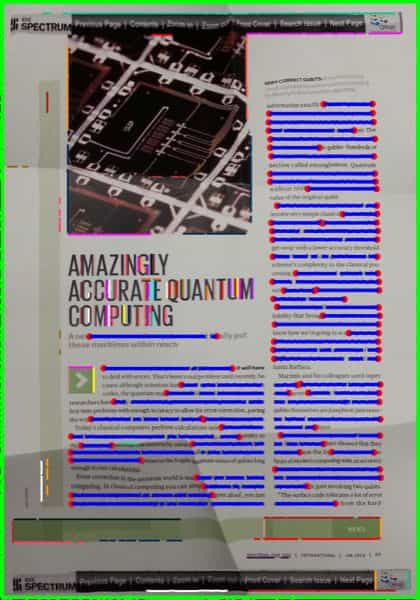}\\
        \vspace{.02\textwidth}%
        \includegraphics[angle=90, width=1\textwidth, height=1.414\textwidth]{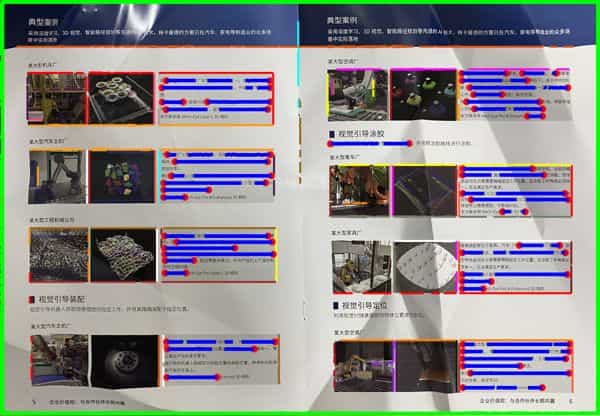}
    \end{minipage}}\hspace{.01\textwidth}%
    \subfigure[]{\centering
    \begin{minipage}[b]{.11\textwidth}\centering
        \includegraphics[width=1\textwidth, height=1.414\textwidth]{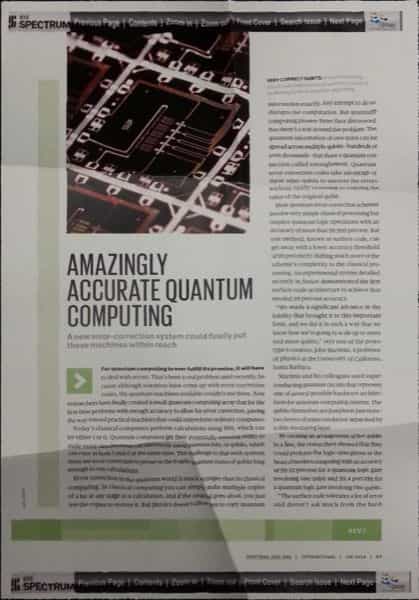}\\
        \vspace{.02\textwidth}%
        \includegraphics[angle=90, width=1\textwidth, height=1.414\textwidth]{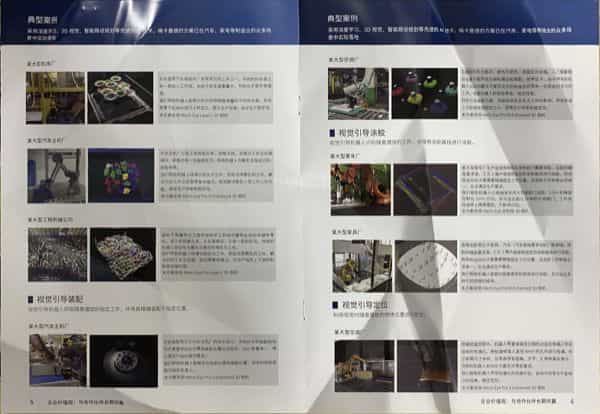}
    \end{minipage}}\hspace{.01\textwidth}%
    \subfigure[]{\centering
    \begin{minipage}[b]{.11\textwidth}\centering
        \includegraphics[width=1\textwidth, height=1.414\textwidth]{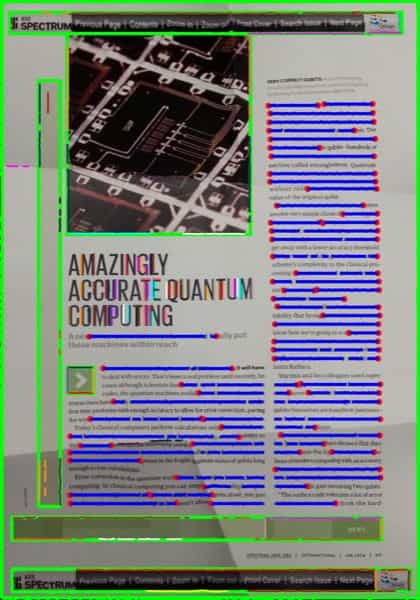}\\
        \vspace{.02\textwidth}%
        \includegraphics[angle=90, width=1\textwidth, height=1.414\textwidth]{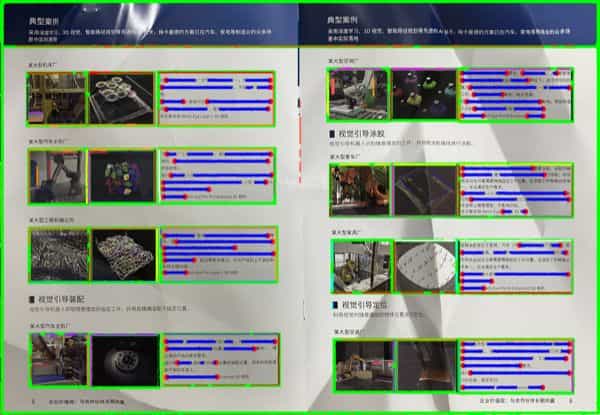}
    \end{minipage}}
\vspace{-.015\textwidth}
\caption{Feature line quality and quantity experiments. (a) Original images. (b) Results of DocScanner \cite{feng2021docscanner}. (c) Our results without using any feature lines. (d) Our results using boundary. (e) Our results using automatically extracted feature lines. (f) The results in (e), feature lines are shown. (g) Our results using additionally manually extracted feature lines based on (e). (h) The results in (g), 
 feature lines are shown.}
\label{f:texture1}
\end{figure}

\paragraph{Feature line quality and quantity.}
In this  part, we show some experiments in Fig.~\ref{f:texture1} to demonstrate the effect of feature lines of different levels of quality and quantity. 
\begin{table*}[!t]
	\caption{Statistics of point cloud quality experiments. The first column shows the index of data. The second column shows the number of data points. The third column shows the number of text feature segments. The fourth column shows the number of edge feature segments. The next four columns show the computation time for each round of optimization in seconds. ${T_n}$ is the total time in the $n$th round of optimization. ${It_n}$ is the iteration number, and ${per_n}$ is the average computation time.  The last column shows the total computation time of the algorithm. All data in the table have $\left|\mc{L}^r_B\right|=4$.}\label{tab:my-table-points}
 \centering
\begin{adjustbox}{max width=\textwidth}
\begin{tabular}{cccccrrrrr}
\toprule
Data &
$\left|\hat{\mc{P}}\right|$ &
$\left|\mc{L}^r_T\right|$ &  
$\left|\mc{L}^r_E\right|$ &
  ${T_0} /{It_0}={per_0}$ &
  ${T_1} /{It_1}={per_1}$ &
  ${T_2} /{It_2}={per_2}$ &
  ${T_3} /{It_3}={per_3}$ &
  ${T_{all}} /{It_{all}}={per_{all}}$ \\
\midrule
1 &11.3k &119 &1599 &8.59/18=0.47 &3.24/6=0.54 &3.48/4=0.87 &9.98/4=2.49 &25.3/32=0.79 \\
2 &20.8k &57 &2556 &5.03/7=0.71 &5.29/7=0.75 &4.7/4=1.17 &5.74/2=2.87 &20.77/20=1.03 \\
5 &22.2k &36 &3629 &10.95/12=0.91 &10.91/11=0.99 &8.78/6=1.46 &18.94/5=3.78 &49.6/34=1.45 \\
6 &41.6k &414 &978 &15.05/29=0.51 &6.67/10=0.66 &13.79/12=1.14 &29.02/10=2.9 &64.53/61=1.05 \\
17 &27.3k &288 &668 &3.25/8=0.4 &4.41/9=0.49 &5.25/6=0.87 &21.84/8=2.73 &34.77/31=1.12 \\
18 &23.0k &310 &514 &2.6/7=0.37 &4.69/10=0.46 &7.44/9=0.82 &17.52/7=2.5 &32.26/33=0.97 \\
19 &24.6k &273 &369 &6.07/16=0.37 &6.15/13=0.47 &5.95/7=0.85 &28.19/11=2.56 &46.36/47=0.98 \\
20 &17.0k &224 &236 &1.72/5=0.34 &3.43/8=0.42 &5.07/6=0.84 &18.44/7=2.63 &28.67/26=1.1 \\
\bottomrule
\end{tabular}
\end{adjustbox}
\end{table*}
\begin{table*}[!t]
	\caption{Statistics of the experiments for feature line quality and quantity. The first column shows the index of data. The second column shows the number of data points. The third column shows the number of boundary feature lines. The fourth column shows the number of text feature segments. The fifth column shows the number of edge feature segments. The next four columns show the computation time in seconds for each round of optimization. ${T_n}$ is the total time in the $n$th round of optimization. ${It_n}$ is the iteration number, and ${per_n}$ is the average computation time. The last column shows the total computation time of the algorithm. We denote the manually extracted feature lines by $\mc{L}^r_B$ for simplicity and use them as the boundary lines.}
\label{tab:my-table-textures}
\centering
\begin{adjustbox}{max width=\textwidth}
\begin{tabular}{ccccccrrrrr}
\toprule
Data &
$\left|\hat{\mc{P}}\right|$ &
$\left|\mc{L}^r_B\right|$ &  
$\left|\mc{L}^r_T\right|$ &  
$\left|\mc{L}^r_E\right|$ &
  ${T_0} /{It_0}={per_0}$ &
  ${T_1} /{It_1}={per_1}$ &
  ${T_2} /{It_2}={per_2}$ &
  ${T_3} /{It_3}={per_3}$ &
  ${T_{all}} /{It_{all}}={per_{all}}$ \\
\midrule
21 &1.6k &0 &0 &0 &3.68/21=0.17 &0.42/4=0.1 &1.58/5=0.31 &13.8/8=1.72 &19.48/38=0.51 \\
21 &1.6k &4 &0 &0 &0.88/12=0.07 &0.87/7=0.12 &2.05/6=0.34 &10.92/6=1.82 &14.74/31=0.47 \\
21 &1.6k &4 &264 &963 &2.93/10=0.29 &5.59/16=0.34 &7.84/12=0.65 &13.22/6=2.2 &29.59/44=0.67 \\
21 &1.6k &27 &264 &963 &4.08/10=0.4 &5.68/12=0.47 &6.57/8=0.82 &14.09/6=2.34 &30.44/36=0.84 \\
22 &8.8k &0 &0 &0 &6.79/21=0.32 &0.81/5=0.16 &2.41/6=0.4 &11.52/6=1.92 &21.54/38=0.56 \\
22 &8.8k &4 &0 &0 &1.47/11=0.13 &1.19/6=0.19 &3.2/7=0.45 &12.35/6=2.05 &18.22/30=0.6 \\
22 &8.8k &4 &147 &771 &3.24/12=0.27 &2.71/8=0.33 &16.4/26=0.63 &13.48/6=2.24 &35.83/52=0.68 \\
22 &8.8k &104 &147 &771 &9.19/15=0.61 &6.22/9=0.69 &7.93/7=1.13 &33.38/12=2.78 &56.74/43=1.31 \\
\bottomrule
\end{tabular}
\end{adjustbox}
\end{table*}

Firstly, we observe that even without any feature line, the results produced by the method are  much better than existing methods, see the creased regions denoted by red dashed lines in Fig.~\ref{f:texture1}. This means that the isometric mapping constraint is effective for rectifying the documents even without using any structural information in the image. The overall quality of the results gets improved significantly when document boundary lines are included as constraints. The rectification results become more accurate in several local areas when some feature lines around these regions are added. Finally, by including some manually generated feature lines in the places where automatic feature line extraction fails, the quality of the rectified image around the regions with newly added feature lines is significantly enhanced. In summary, the quality of rectification by our method gets improved when the quality and quantity of the feature lines are improved. This demonstrates the effectiveness of including feature lines as constraints and also renders the flexibility of our method for dealing with a large number of feature lines.

Table \ref{tab:my-table-textures} gives some statistics on the experiments. The optimization  converges  quickly without any feature line constraints. Interestingly, the first round of optimization in our approach converges even faster when the document boundary constraints are included. When more feature lines are included,  the optimization time in each round increases, but fewer rounds are required to reach convergence. The average time of each iteration and the total computation time increases slightly when manually generated feature lines are included.


\paragraph{Incomplete document boundaries.} 

In the previous experiments, the boundaries of the documents are successfully extracted. The data points and the detected edge segments outside the document region can be removed. But there are cases the whole document boundary is unavailable  and special treatments are needed to deal with such cases. 

We use the same examples (data 16 and data 19) as in previous experiments, but with different reference images from which complete boundary extraction failed. We denote the two cases as data $16^\star$ and $19^\star$, respectively. For the deep learning methods in comparison~\cite{das2019dewarpnet, xie2021dewarping, xie2022document, feng2021doctr}, we continue to deliver images with segmented backgrounds as in previous experiments.

\begin{figure}[!t]
\centering
    \subfigure[]{\centering
    \begin{minipage}[b]{.09\textwidth}\centering
    \includegraphics[width=1\textwidth, height=1.414\textwidth]{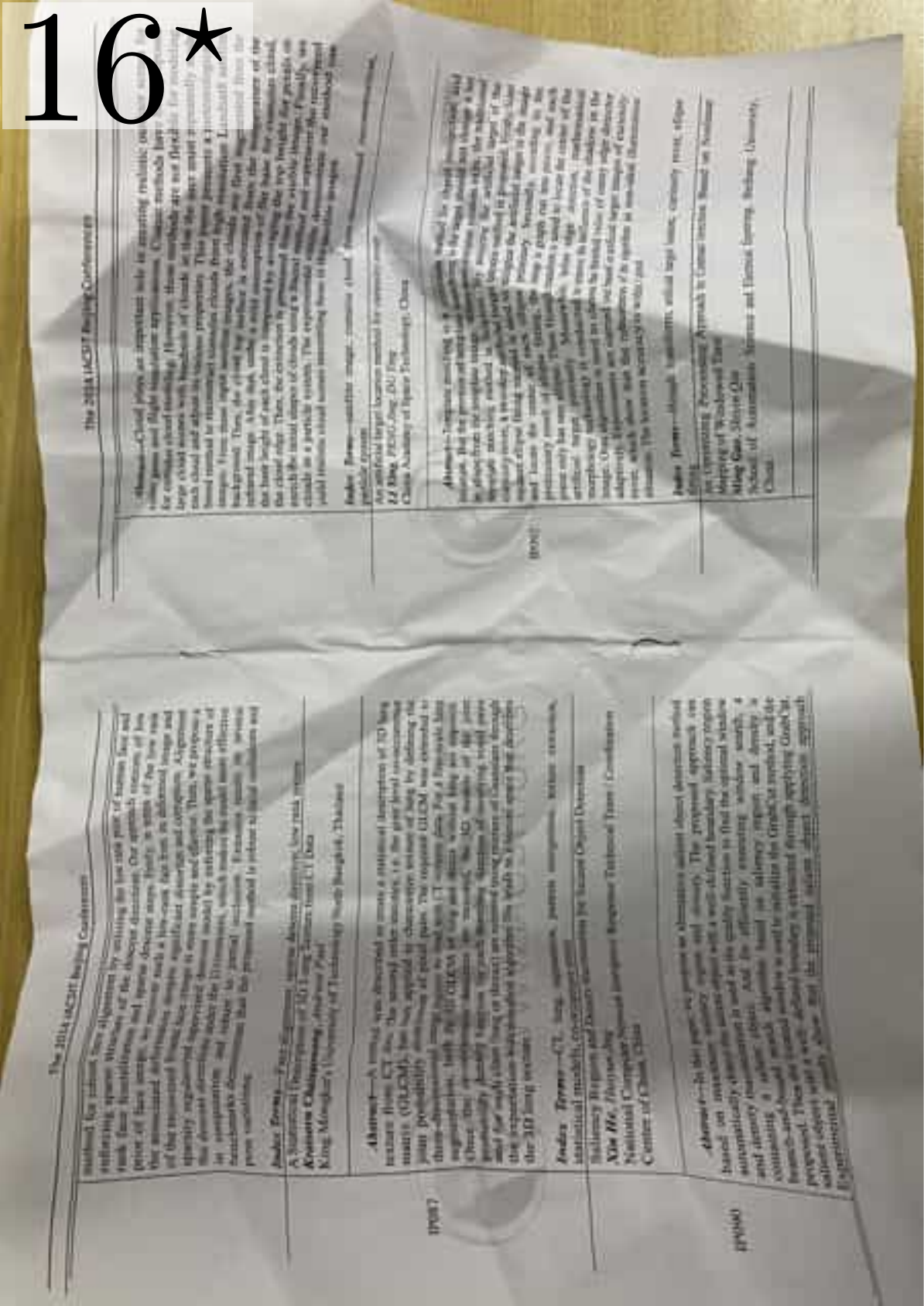}\\	
    \vspace{.02\textwidth}%
    \includegraphics[width=1\textwidth, height=1.414\textwidth]{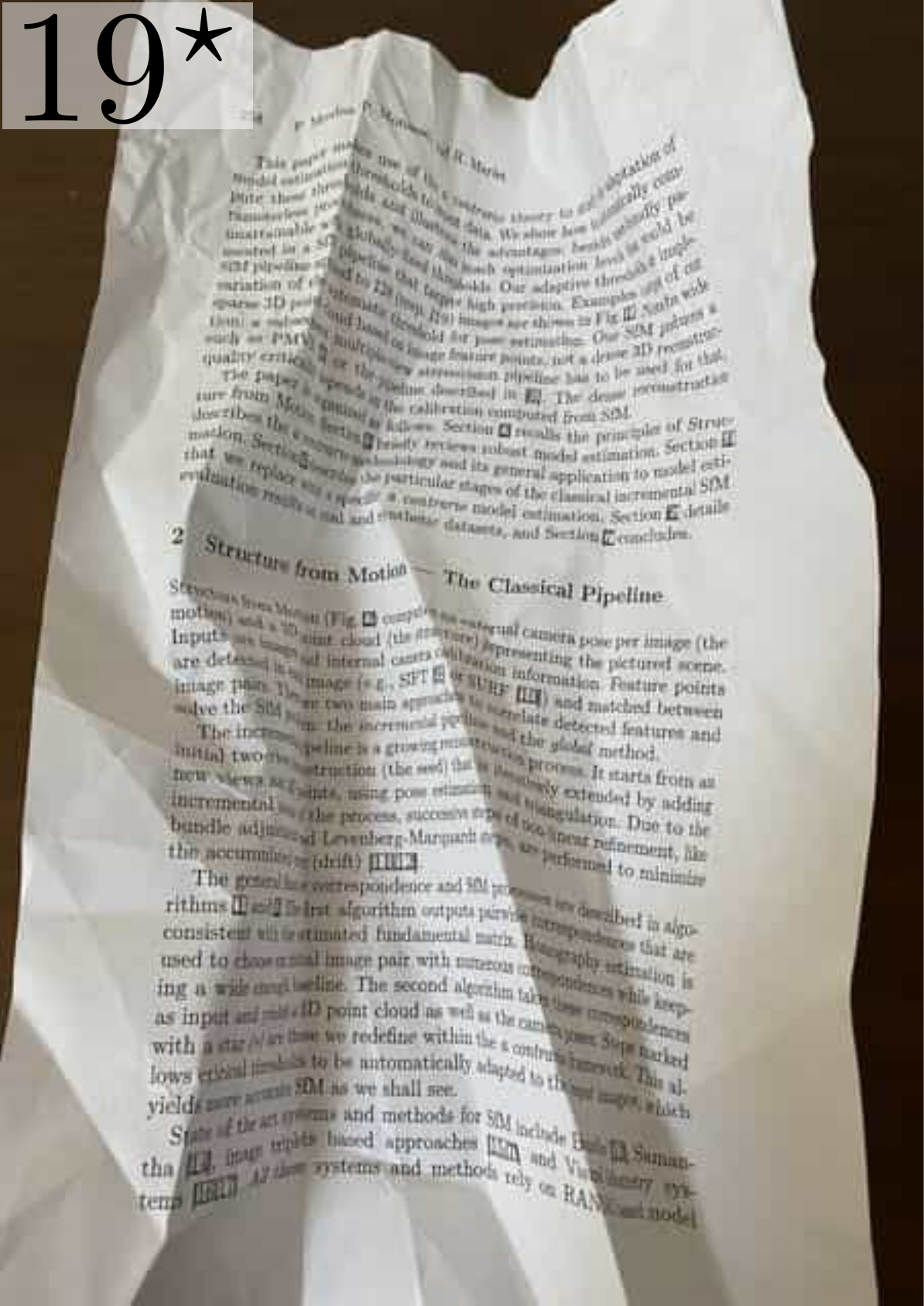}
    \end{minipage}}\hspace{.01\textwidth}%
     \subfigure[]{\centering
    \begin{minipage}[b]{.09\textwidth}\centering
        \includegraphics[angle=90, width=1\textwidth, height=1.414\textwidth]{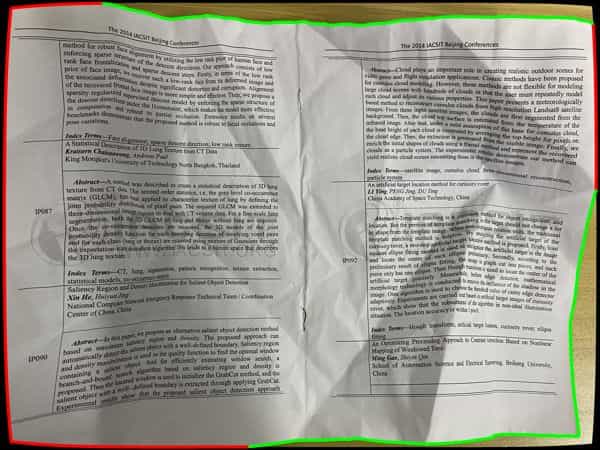}\\
        \vspace{.02\textwidth}%
        \includegraphics[width=1\textwidth, height=1.414\textwidth]{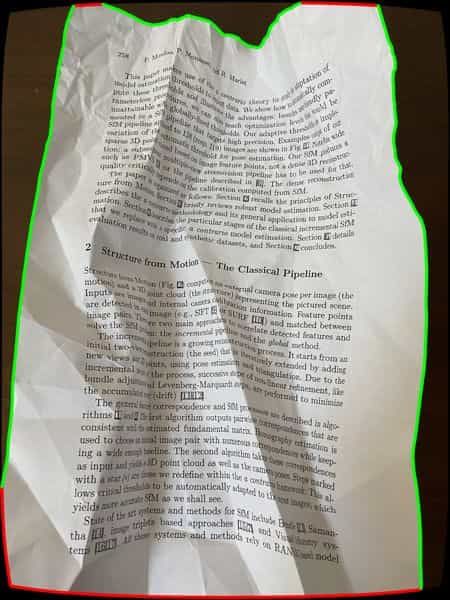}
    \end{minipage}}\hspace{.01\textwidth}%
   \subfigure[]{\centering
    \begin{minipage}[b]{.09\textwidth}\centering
        \includegraphics[angle=90, width=1\textwidth, height=1.414\textwidth]{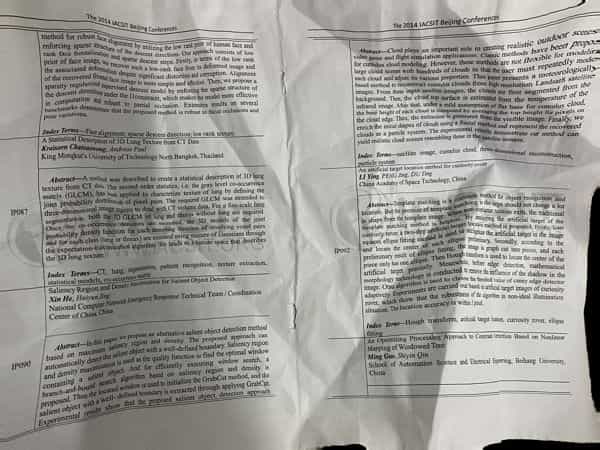}\\
        \vspace{.02\textwidth}%
        \includegraphics[width=1\textwidth, height=1.414\textwidth]{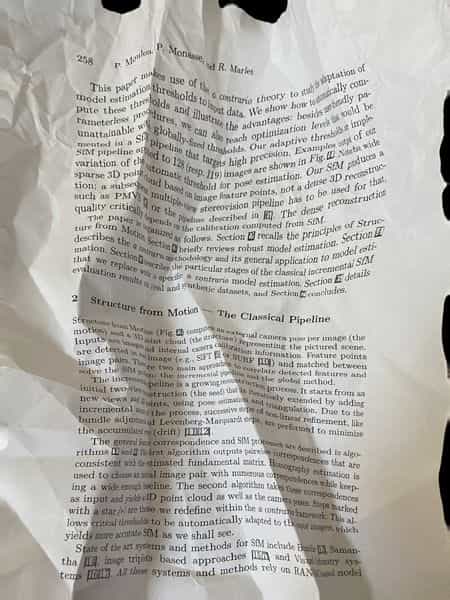}
    \end{minipage}}\hspace{.01\textwidth}%
    \subfigure[]{\centering
    \begin{minipage}[b]{.09\textwidth}\centering
        \includegraphics[angle=90, width=1\textwidth, height=1.414\textwidth]{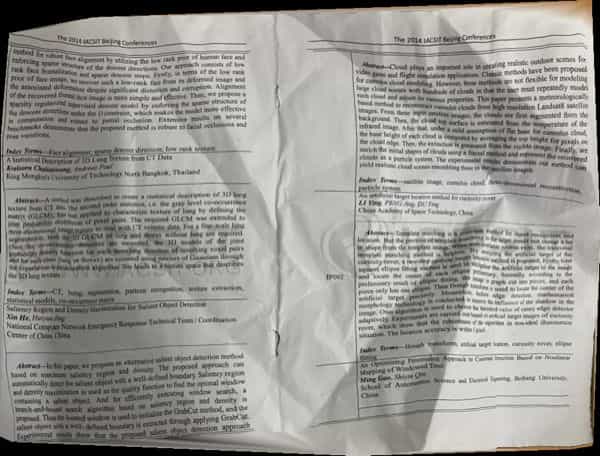}\\
        \vspace{.02\textwidth}%
        \includegraphics[width=1\textwidth, height=1.414\textwidth]{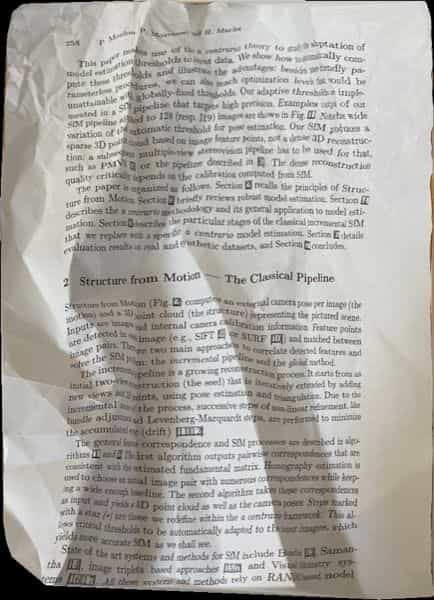}
    \end{minipage}}\hspace{.01\textwidth}%
   \subfigure[]{\centering
    \begin{minipage}[b]{.09\textwidth}\centering
        \includegraphics[angle=90, width=1\textwidth, height=1.414\textwidth]{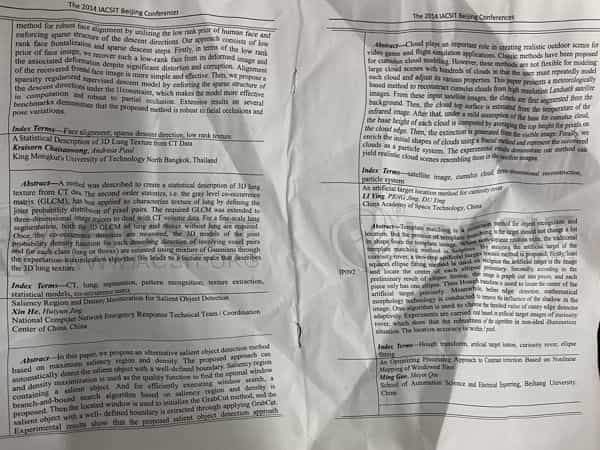}\\
        \vspace{.02\textwidth}%
        \includegraphics[width=1\textwidth, height=1.414\textwidth]{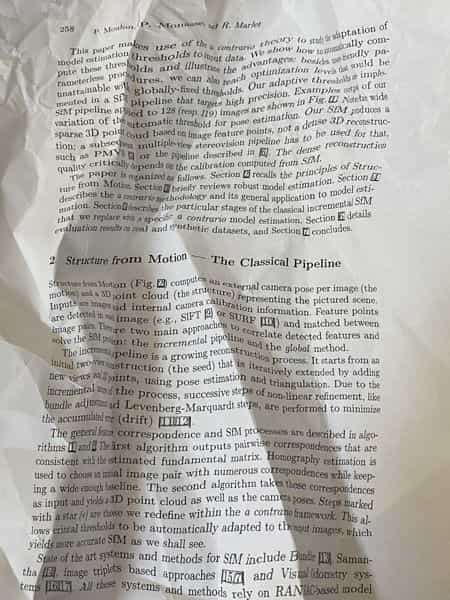}
    \end{minipage}}\hspace{.01\textwidth}%
    \subfigure[]{\centering
    \begin{minipage}[b]{.09\textwidth}\centering
        \includegraphics[angle=90, width=1\textwidth, height=1.414\textwidth]{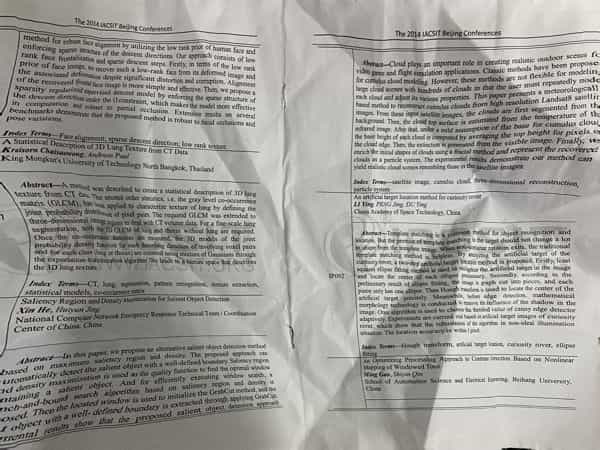}\\
        \vspace{.02\textwidth}%
        \includegraphics[width=1\textwidth, height=1.414\textwidth]{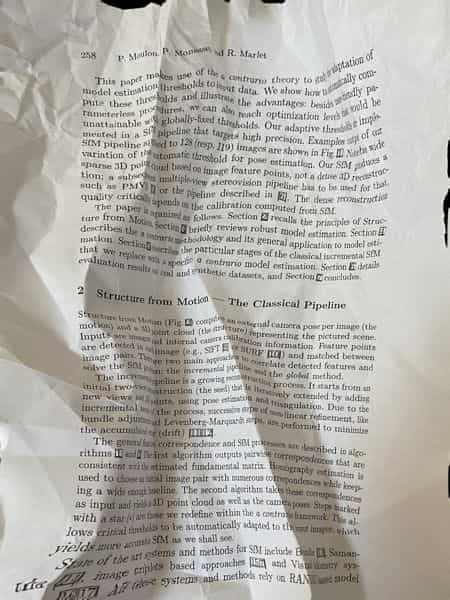}
    \end{minipage}}\hspace{.01\textwidth}%
    \subfigure[]{\centering
    \begin{minipage}[b]{.09\textwidth}\centering
        \includegraphics[angle=90, width=1\textwidth, height=1.414\textwidth]{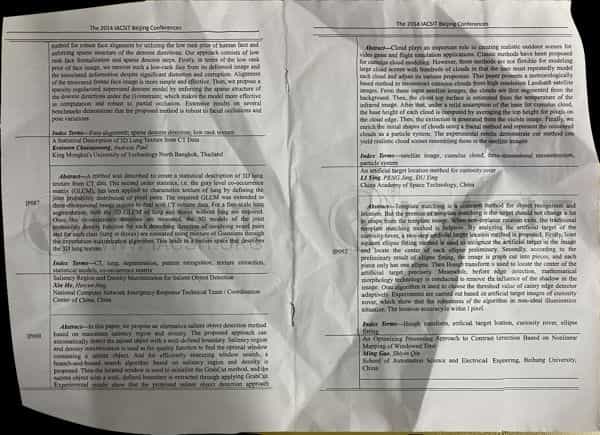}\\
        \vspace{.02\textwidth}%
        \includegraphics[width=1\textwidth, height=1.414\textwidth]{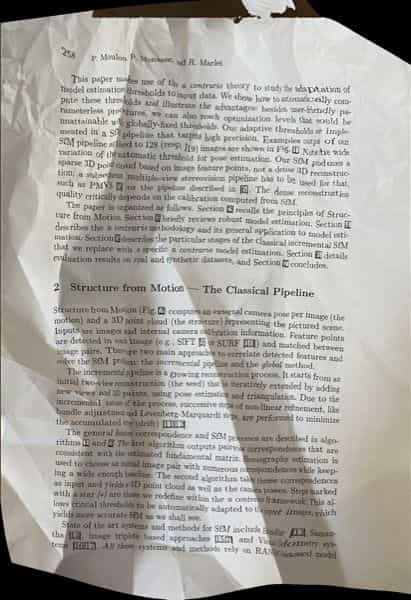}
    \end{minipage}}\hspace{.01\textwidth}%
    \subfigure[]{\centering
    \begin{minipage}[b]{.09\textwidth}\centering
        \includegraphics[angle=90, width=1\textwidth, height=1.414\textwidth]{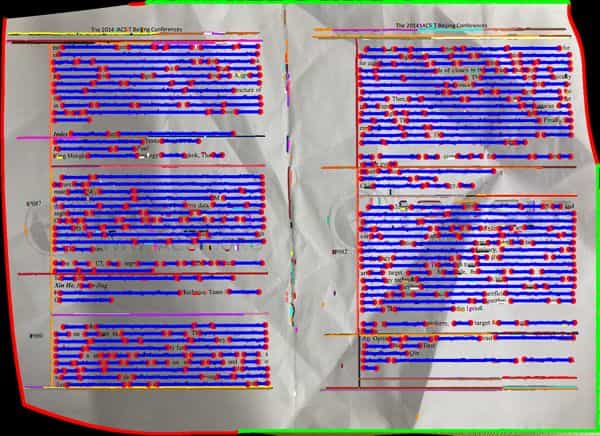}\\
        \vspace{.02\textwidth}%
        \includegraphics[width=1\textwidth, height=1.414\textwidth]{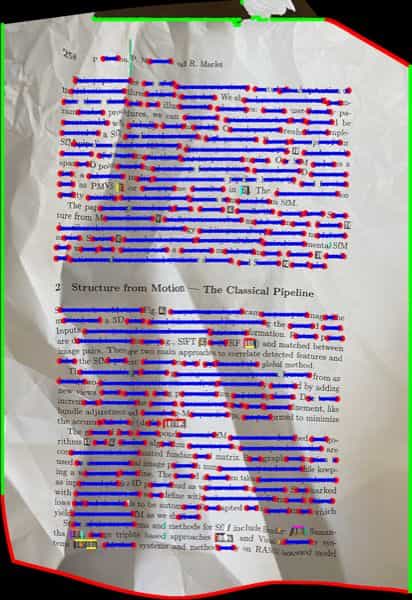}
    \end{minipage}}\hspace{.01\textwidth}%
    \subfigure[]{\centering
    \begin{minipage}[b]{.09\textwidth}\centering
        \includegraphics[width=1\textwidth, height=1.414\textwidth]{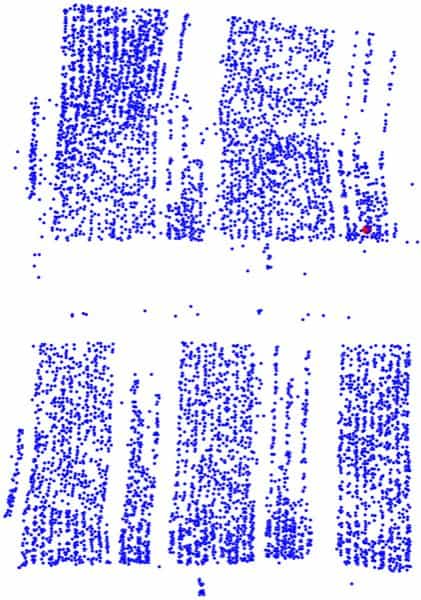}\\
        \vspace{.02\textwidth}%
        \includegraphics[width=1\textwidth, height=1.414\textwidth]{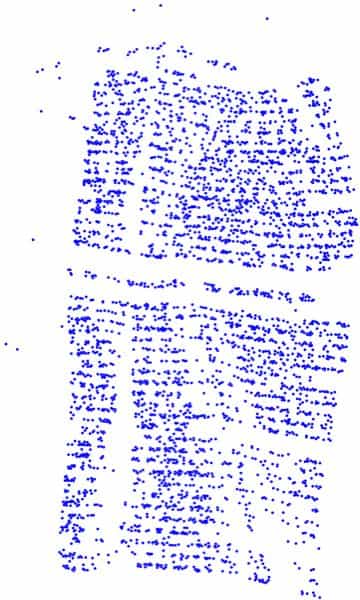}
    \end{minipage}}\hspace{.01\textwidth}%
    \subfigure[]{\centering
    \begin{minipage}[b]{.09\textwidth}\centering
        \includegraphics[width=1\textwidth, height=1.414\textwidth]{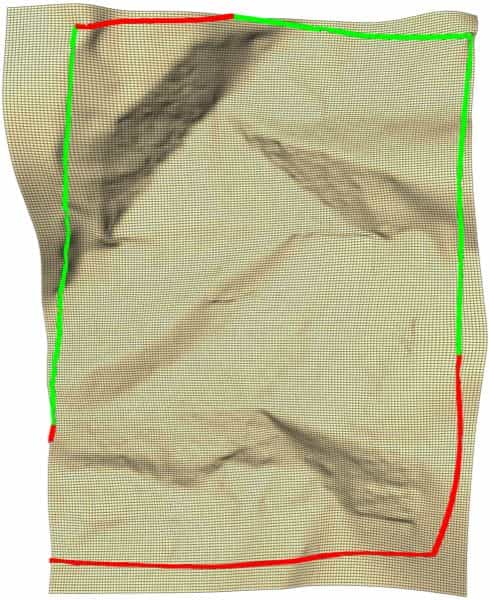}\\
        \vspace{.02\textwidth}%
        \includegraphics[width=1\textwidth, height=1.414\textwidth]{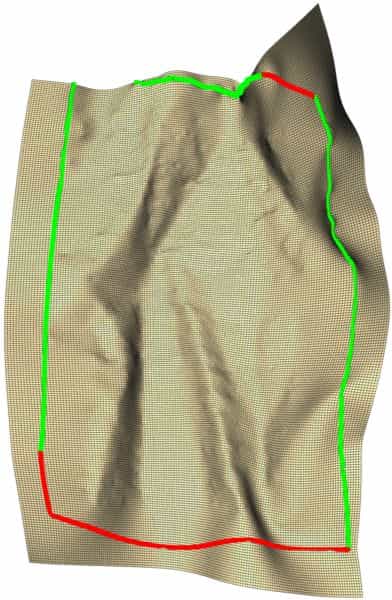}
    \end{minipage}}
\vspace{-.015\textwidth}
\caption{Incomplete document boundaries experiments. (a) The original images. (b) our reference images with boundary. Green and red are used to indicate the document boundary and image boundary, respectively. (c) Results of  DewarpNet \cite{das2019dewarpnet}. (d) Results of  FCN-based \cite{xie2021dewarping}. (e) Results of Points-based~\cite{xie2022document}. (f) Results of DocTr \cite{feng2021doctr}. 
 (g): Our results. (h) Our results with feature lines. (i) Our final $\mc{P}$. (j) Our final $\mc{M}$ with boundary.}
\label{f:none}
\end{figure}

In Fig.~\ref{f:none}, the document region is incomplete and part of the document boundary does not exist in the image. In the places where the document boundary is missing, we connect the document boundary with the image boundary to make a closed region. Fig. \ref{f:none} shows the results where the valid document boundaries are used in the optimization. Since our method only straightens the real document boundaries,  effective rectification can be obtained even with limited boundary information. 

As comparisons, existing methods need complete boundary lines of the document and boundary missing can affect the rectification results. DewarpNet~\cite{das2019dewarpnet} on data $16^\star$ and DocTr~\cite{feng2021doctr} on both data $16^\star$ and $19^\star$ show different degrees of distortion in the boundary portions due to the influence of incomplete boundaries. FCN-based~\cite{xie2021dewarping} and  Points-based~\cite{xie2022document}  do not exhibit significant distortion, but remove a piece of the left region on data $16^\star$, resulting in the loss of document content. 

%% file: Conclusion.tex
\section{Conclusions}

We have proposed a novel approach to image rectification of distorted documents which is effective and capable of tackling challenging cases with complex creases and large foldings.  Our method is based on a framework of general document image rectification which is realized by integrating the computational isometric mapping model with the pinhole camera model. The quad meshes in the computation of the isometric mapping serve as flexible representations for creased or folded document models.  Moreover, based on this general computational framework, both textural features in the input images and  the developability property of the 3D document model are considered in the computation, leading to resulting rectified images with considerably better quality than existing approaches.

Extensive experiments demonstrated that the proposed method is effective even without any feature line constraints, due to the isometric mapping constraints leading to a developable document model and also its accurate development in the plane.  When  the document contains feature lines that can be detected reliably, the feature line constraints can greatly improve the quality of image rectification. The proposed method is flexible and effective in dealing with a large number of feature lines. The detection of feature lines and other feature elements in the image is a challenging task and is worthy of further investigation. Any feature detection method can be used in our framework and the results can be enhanced by better methods for detecting feature lines in the document image.

%% file: Main.bbl
\begin{thebibliography}{55}
\providecommand{\natexlab}[1]{#1}
\providecommand{\url}[1]{\texttt{#1}}
\expandafter\ifx\csname urlstyle\endcsname\relax
  \providecommand{\doi}[1]{doi: #1}\else
  \providecommand{\doi}{doi: \begingroup \urlstyle{rm}\Url}\fi

\bibitem[Lavialle et~al.(2001)Lavialle, Molines, Angella, and
  Baylou]{lavialle2001active}
O.~Lavialle, X.~Molines, F.~Angella, and P.~Baylou.
\newblock Active contours network to straighten distorted text lines.
\newblock In \emph{Proceedings 2001 International Conference on Image
  Processing (Cat. No.01CH37205)}, volume~3, pages 748--751 vol.3, 2001.
\newblock \doi{10.1109/ICIP.2001.958227}.

\bibitem[Sakoe et~al.(2005)Sakoe, Asano, Ezaki, and Uchida]{ezaki2005dewarping}
H.~Sakoe, A.~Asano, H.~Ezaki, and S.~Uchida.
\newblock Dewarping of document image by global optimization.
\newblock In \emph{Proceedings. Eighth International Conference on Document
  Analysis and Recognition}, pages 302--306, Los Alamitos, CA, USA, sep 2005.
  IEEE Computer Society.
\newblock \doi{10.1109/ICDAR.2005.87}.
\newblock URL \url{https://doi.ieeecomputersociety.org/10.1109/ICDAR.2005.87}.

\bibitem[Zhang et~al.(2008)Zhang, Liu, Ding, and Zou]{zhang2008arbitrary}
Yu~Zhang, Changsong Liu, Xiaoqing Ding, and Yanming Zou.
\newblock Arbitrary warped document image restoration based on segmentation and
  thin-plate splines.
\newblock In \emph{2008 19th International Conference on Pattern Recognition},
  pages 1--4, 2008.
\newblock \doi{10.1109/ICPR.2008.4761528}.

\bibitem[Liu et~al.(2015)Liu, Zhang, Wang, and Ding]{liu2015restoring}
Changsong Liu, Yu~Zhang, Baokang Wang, and Xiaoqing Ding.
\newblock Restoring camera-captured distorted document images.
\newblock \emph{International Journal on Document Analysis and Recognition},
  18\penalty0 (2):\penalty0 111--124, 2015.
\newblock \doi{10.1007/s10032-014-0233-8}.

\bibitem[Ulges et~al.(2005)Ulges, Lampert, and Breuel]{ulges2005document}
A.~Ulges, C.H. Lampert, and T.M. Breuel.
\newblock Document image dewarping using robust estimation of curled text
  lines.
\newblock In \emph{Eighth International Conference on Document Analysis and
  Recognition (ICDAR'05)}, pages 1001--1005 Vol. 2, 2005.
\newblock \doi{10.1109/ICDAR.2005.90}.

\bibitem[Lu and Tan(2006)]{shijianlu2006document}
Shijian Lu and Chew~Lim Tan.
\newblock Document flattening through grid modeling and regularization.
\newblock In \emph{18th International Conference on Pattern Recognition
  (ICPR'06)}, volume~1, pages 971--974, 2006.
\newblock \doi{10.1109/ICPR.2006.458}.

\bibitem[Lu et~al.(2005)Lu, Chen, and Ko]{lu2005perspective}
Shijian Lu, Ben~M. Chen, and C.C. Ko.
\newblock Perspective rectification of document images using fuzzy set and
  morphological operations.
\newblock \emph{Image and Vision Computing}, 23\penalty0 (5):\penalty0
  541--553, 2005.
\newblock ISSN 0262-8856.
\newblock \doi{https://doi.org/10.1016/j.imavis.2005.01.003}.
\newblock URL
  \url{https://www.sciencedirect.com/science/article/pii/S0262885605000119}.

\bibitem[Schneider et~al.(2007)Schneider, Block, and
  Rojas]{schneider2007robust}
D.~Schneider, M.~Block, and R.~Rojas.
\newblock Robust document warping with interpolated vector fields.
\newblock In \emph{Ninth International Conference on Document Analysis and
  Recognition (ICDAR 2007)}, volume~1, pages 113--117, 2007.
\newblock \doi{10.1109/ICDAR.2007.4378686}.

\bibitem[Stamatopoulos et~al.(2011)Stamatopoulos, Gatos, Pratikakis, and
  Perantonis]{stamatopoulos2011goaloriented}
N.~Stamatopoulos, B.~Gatos, I.~Pratikakis, and S.~J. Perantonis.
\newblock Goal-oriented rectification of camera-based document images.
\newblock \emph{Trans. Img. Proc.}, 20\penalty0 (4):\penalty0 910–920, apr
  2011.
\newblock ISSN 1057-7149.
\newblock \doi{10.1109/TIP.2010.2080280}.
\newblock URL \url{https://doi.org/10.1109/TIP.2010.2080280}.

\bibitem[Wada et~al.(1997)Wada, Ukida, and Matsuyama]{wada1997shape}
Toshikazu Wada, Hiroyuki Ukida, and Takashi Matsuyama.
\newblock Shape from shading with interreflections under a proximal light
  source: Distortion-free copying of an unfolded book.
\newblock \emph{International Journal of Computer Vision}, 24\penalty0
  (2):\penalty0 125--135, 1997.
\newblock \doi{10.1023/A:1007906904009}.

\bibitem[Tan et~al.(2006)Tan, Zhang, Zhang, and Xia]{chewlimtan2006restoring}
Chew~Lim Tan, Li~Zhang, Zheng Zhang, and Tao Xia.
\newblock Restoring warped document images through 3d shape modeling.
\newblock \emph{IEEE Transactions on Pattern Analysis and Machine
  Intelligence}, 28\penalty0 (2):\penalty0 195--208, 2006.
\newblock \doi{10.1109/TPAMI.2006.40}.

\bibitem[Courteille et~al.(2004)Courteille, Crouzil, Durou, and
  Gurdjos]{courteille2004shape}
F.~Courteille, A.~Crouzil, J.-D. Durou, and P.~Gurdjos.
\newblock Towards shape from shading under realistic photographic conditions.
\newblock In \emph{Proceedings of the 17th International Conference on Pattern
  Recognition, 2004. ICPR 2004.}, volume~2, pages 277--280 Vol.2, 2004.
\newblock \doi{10.1109/ICPR.2004.1334160}.

\bibitem[Courteille et~al.(2007)Courteille, Crouzil, Durou, and
  Gurdjos]{courteille2007shape}
Frédéric Courteille, Alain Crouzil, Jean-Denis Durou, and Pierre Gurdjos.
\newblock Shape from shading for the digitization of curved documents.
\newblock \emph{Machine Vision and Applications}, 18\penalty0 (5):\penalty0
  301--316, 2007.
\newblock \doi{10.1007/s00138-006-0062-y}.

\bibitem[Zhang et~al.(2004)Zhang, Lim, and Fan]{zhang2004estimation}
Z.~Zhang, C.L. Lim, and L.~Fan.
\newblock Estimation of 3d shape of warped document surface for image
  restoration.
\newblock In \emph{Proceedings of the 17th International Conference on Pattern
  Recognition, 2004. ICPR 2004.}, volume~1, pages 486--489 Vol.1, 2004.
\newblock \doi{10.1109/ICPR.2004.1334172}.

\bibitem[Zhang et~al.(2009)Zhang, Yip, Brown, and Tan]{zhang2009unified}
Li~Zhang, A.M. Yip, M.S. Brown, and Chew~Lim Tan.
\newblock A unified framework for document restoration using inpainting and
  shape-from-shading.
\newblock \emph{Pattern Recognition}, 42\penalty0 (11):\penalty0 2961--2978,
  2009.
\newblock ISSN 0031-3203.
\newblock \doi{https://doi.org/10.1016/j.patcog.2009.03.025}.
\newblock URL
  \url{https://www.sciencedirect.com/science/article/pii/S0031320309001277}.

\bibitem[Cao et~al.(2003)Cao, Ding, and Liu]{cao2003cylindrical}
Huaigu Cao, Xiaoqing Ding, and Changsong Liu.
\newblock A cylindrical surface model to rectify the bound document image.
\newblock In \emph{Proceedings Ninth IEEE International Conference on Computer
  Vision}, pages 228--233 vol.1, 2003.
\newblock \doi{10.1109/ICCV.2003.1238346}.

\bibitem[Koo(2013)]{koo2013segmentation}
Hyung~Il Koo.
\newblock Segmentation and rectification of pictures in the camera-captured
  images of printed documents.
\newblock \emph{IEEE Transactions on Multimedia}, 15\penalty0 (3):\penalty0
  647--660, 2013.
\newblock \doi{10.1109/TMM.2012.2236305}.

\bibitem[Koo et~al.(2009)Koo, Kim, and Cho]{koo2009composition}
Hyung~Il Koo, Jinho Kim, and Nam~Ik Cho.
\newblock Composition of a dewarped and enhanced document image from two view
  images.
\newblock \emph{IEEE Transactions on Image Processing}, 18\penalty0
  (7):\penalty0 1551--1562, 2009.
\newblock \doi{10.1109/TIP.2009.2019301}.

\bibitem[Fu et~al.(2007)Fu, Wu, Li, Li, Xu, and Yang]{fu2007modelbased}
Bin Fu, Minghui Wu, Rongfeng Li, Wenxin Li, Zhuoqun Xu, and Chunxu Yang.
\newblock A model-{{Based}} book dewarping method using text line detection.
\newblock In \emph{Proceedings of the 2nd International Workshop on
  Camera-Based Document Analysis and Recognition, CBDAR 2007}, pages 63--70,
  Curitiba, Barazil, 2007.

\bibitem[Meng et~al.(2012)Meng, Pan, Xiang, Duan, and Zheng]{meng2012metric}
Gaofeng Meng, Chunhong Pan, Shiming Xiang, Jiangyong Duan, and Nanning Zheng.
\newblock Metric rectification of curved document images.
\newblock \emph{IEEE Transactions on Pattern Analysis and Machine
  Intelligence}, 34\penalty0 (4):\penalty0 707--722, 2012.
\newblock \doi{10.1109/TPAMI.2011.151}.

\bibitem[Kim et~al.(2015)Kim, Koo, and Cho]{kim2015document}
Beom~Su Kim, Hyung~Il Koo, and Nam~Ik Cho.
\newblock Document dewarping via text-line based optimization.
\newblock \emph{Pattern Recognition}, 48\penalty0 (11):\penalty0 3600--3614,
  2015.
\newblock ISSN 0031-3203.
\newblock \doi{https://doi.org/10.1016/j.patcog.2015.04.026}.
\newblock URL
  \url{https://www.sciencedirect.com/science/article/pii/S003132031500165X}.

\bibitem[Kil et~al.(2017)Kil, Seo, Koo, and Cho]{kil2017robust}
Taeho Kil, Wonkyo Seo, Hyung~Il Koo, and Nam~Ik Cho.
\newblock Robust document image dewarping method using text-lines and line
  segments.
\newblock In \emph{2017 14th IAPR International Conference on Document Analysis
  and Recognition (ICDAR)}, volume~01, pages 865--870, 2017.
\newblock \doi{10.1109/ICDAR.2017.146}.

\bibitem[Meng et~al.(2018)Meng, Su, Wu, Xiang, and Pan]{meng2018exploiting}
Gaofeng Meng, Yuanqi Su, Ying Wu, Shiming Xiang, and Chunhong Pan.
\newblock Exploiting vector fields for geometric rectification of distorted
  document images.
\newblock In \emph{Computer Vision – ECCV 2018: 15th European Conference,
  Munich, Germany, September 8-14, 2018, Proceedings, Part XVI}, page
  180–195, Berlin, Heidelberg, 2018. Springer-Verlag.
\newblock ISBN 978-3-030-01269-4.
\newblock \doi{10.1007/978-3-030-01270-0_11}.
\newblock URL \url{https://doi.org/10.1007/978-3-030-01270-0_11}.

\bibitem[He et~al.(2013)He, Pan, Xie, Sun, and Naoi]{he2013book}
Yuan He, Pan Pan, Shufu Xie, Jun Sun, and Satoshi Naoi.
\newblock A book dewarping system by boundary-based 3d surface reconstruction.
\newblock In \emph{2013 12th International Conference on Document Analysis and
  Recognition}, pages 403--407, 2013.
\newblock \doi{10.1109/ICDAR.2013.88}.

\bibitem[Brown and Tsoi(2006)]{brown2006geometric}
M.S. Brown and Y.-C. Tsoi.
\newblock Geometric and shading correction for images of printed materials
  using boundary.
\newblock \emph{IEEE Transactions on Image Processing}, 15\penalty0
  (6):\penalty0 1544--1554, 2006.
\newblock \doi{10.1109/TIP.2006.871082}.

\bibitem[Tsoi and Brown(2007)]{tsoi2007multiview}
Yau-Chat Tsoi and Michael~S. Brown.
\newblock Multi-view document rectification using boundary.
\newblock In \emph{2007 IEEE Conference on Computer Vision and Pattern
  Recognition}, pages 1--8, 2007.
\newblock \doi{10.1109/CVPR.2007.383251}.

\bibitem[Meng et~al.(2020)Meng, Pan, Xiang, and Wu]{meng2020baselines}
Gaofeng Meng, Chunhong Pan, Shiming Xiang, and Ying Wu.
\newblock Baselines extraction from curved document images via slope fields
  recovery.
\newblock \emph{IEEE Transactions on Pattern Analysis and Machine
  Intelligence}, 42\penalty0 (4):\penalty0 793--808, 2020.
\newblock \doi{10.1109/TPAMI.2018.2886900}.

\bibitem[Liang et~al.(2005)Liang, DeMenthon, and
  Doermann]{jianliang2005flattening}
Jian Liang, D.~DeMenthon, and D.~Doermann.
\newblock Flattening curved documents in images.
\newblock In \emph{2005 IEEE Computer Society Conference on Computer Vision and
  Pattern Recognition (CVPR'05)}, volume~2, pages 338--345 vol. 2, 2005.
\newblock \doi{10.1109/CVPR.2005.163}.

\bibitem[Liang et~al.(2008)Liang, DeMenthon, and Doermann]{liang2008geometric}
Jian Liang, Daniel DeMenthon, and David Doermann.
\newblock Geometric rectification of camera-captured document images.
\newblock \emph{IEEE Transactions on Pattern Analysis and Machine
  Intelligence}, 30\penalty0 (4):\penalty0 591--605, 2008.
\newblock \doi{10.1109/TPAMI.2007.70724}.

\bibitem[Tian and Narasimhan(2011)]{tian2011rectification}
Yuandong Tian and Srinivasa~G. Narasimhan.
\newblock Rectification and 3d reconstruction of curved document images.
\newblock In \emph{CVPR 2011}, pages 377--384, 2011.
\newblock \doi{10.1109/CVPR.2011.5995540}.

\bibitem[Perriollat and Bartoli(2013)]{perriollat2013computational}
Mathieu Perriollat and Adrien Bartoli.
\newblock A computational model of bounded developable surfaces with
  application to image-based three-dimensional reconstruction.
\newblock \emph{Computer Animation and Virtual Worlds}, 24\penalty0
  (5):\penalty0 459--476, 2013.
\newblock \doi{https://doi.org/10.1002/cav.1478}.

\bibitem[You et~al.(2018)You, Matsushita, Sinha, Bou, and
  Ikeuchi]{you2018multiview}
Shaodi You, Yasuyuki Matsushita, Sudipta Sinha, Yusuke Bou, and Katsushi
  Ikeuchi.
\newblock Multiview rectification of folded documents.
\newblock \emph{IEEE Transactions on Pattern Analysis and Machine
  Intelligence}, 40\penalty0 (2):\penalty0 505--511, 2018.
\newblock \doi{10.1109/TPAMI.2017.2675980}.

\bibitem[Li et~al.(2019)Li, Zhang, Liao, and Sander]{li2019document}
Xiaoyu Li, Bo~Zhang, Jing Liao, and Pedro~V. Sander.
\newblock Document rectification and illumination correction using a
  patch-based cnn.
\newblock \emph{ACM Trans. Graph.}, 38\penalty0 (6), nov 2019.
\newblock ISSN 0730-0301.
\newblock \doi{10.1145/3355089.3356563}.
\newblock URL \url{https://doi.org/10.1145/3355089.3356563}.

\bibitem[Liu et~al.(2020)Liu, Meng, Fan, Xiang, and Pan]{liu2020geometric}
Xiyan Liu, Gaofeng Meng, Bin Fan, Shiming Xiang, and Chunhong Pan.
\newblock Geometric rectification of document images using adversarial gated
  unwarping network.
\newblock \emph{Pattern Recognition}, 108:\penalty0 107576, 2020.
\newblock ISSN 0031-3203.
\newblock \doi{https://doi.org/10.1016/j.patcog.2020.107576}.
\newblock URL
  \url{https://www.sciencedirect.com/science/article/pii/S0031320320303794}.

\bibitem[Ma et~al.(2018)Ma, Shu, Bai, Wang, and Samaras]{ma2018docunet}
Ke~Ma, Zhixin Shu, Xue Bai, Jue Wang, and Dimitris Samaras.
\newblock Docunet: Document image unwarping via a stacked u-net.
\newblock In \emph{2018 IEEE/CVF Conference on Computer Vision and Pattern
  Recognition}, pages 4700--4709, 2018.
\newblock \doi{10.1109/CVPR.2018.00494}.

\bibitem[Xie et~al.(2020)Xie, Yin, Zhang, and Liu]{xie2021dewarping}
Guo-Wang Xie, Fei Yin, Xu-Yao Zhang, and Cheng-Lin Liu.
\newblock Dewarping document image by displacement flow estimation with fully
  convolutional network.
\newblock In Xiang Bai, Dimosthenis Karatzas, and Daniel Lopresti, editors,
  \emph{Document Analysis Systems}, pages 131--144, Cham, 2020. Springer
  International Publishing.
\newblock ISBN 978-3-030-57058-3.
\newblock \doi{10.1007/978-3-030-57058-3_10}.

\bibitem[Xie et~al.(2021)Xie, Yin, Zhang, and Liu]{xie2022document}
Guo-Wang Xie, Fei Yin, Xu-Yao Zhang, and Cheng-Lin Liu.
\newblock Document dewarping with control points.
\newblock In \emph{Document Analysis and Recognition – ICDAR 2021: 16th
  International Conference, Lausanne, Switzerland, September 5–10, 2021,
  Proceedings, Part I}, page 466–480, Berlin, Heidelberg, 2021.
  Springer-Verlag.
\newblock ISBN 978-3-030-86548-1.
\newblock \doi{10.1007/978-3-030-86549-8_30}.
\newblock URL \url{https://doi.org/10.1007/978-3-030-86549-8_30}.

\bibitem[Das et~al.(2017)Das, Mishra, Sudharshana, and Shilkrot]{das2017common}
Sagnik Das, Gaurav Mishra, Akshay Sudharshana, and Roy Shilkrot.
\newblock The common fold: Utilizing the four-fold to dewarp printed documents
  from a single image.
\newblock In \emph{Proceedings of the 2017 ACM Symposium on Document
  Engineering}, DocEng '17, page 125–128, New York, NY, USA, 2017.
  Association for Computing Machinery.
\newblock ISBN 9781450346894.
\newblock \doi{10.1145/3103010.3121030}.
\newblock URL \url{https://doi.org/10.1145/3103010.3121030}.

\bibitem[Markovitz et~al.(2020)Markovitz, Lavi, Perel, Mazor, and
  Litman]{markovitz2020can}
Amir Markovitz, Inbal Lavi, Or~Perel, Shai Mazor, and Roee Litman.
\newblock Can you read me now? content aware rectification using angle
  supervision.
\newblock In \emph{Computer Vision – ECCV 2020: 16th European Conference,
  Glasgow, UK, August 23–28, 2020, Proceedings, Part XII}, page 208–223,
  Berlin, Heidelberg, 2020. Springer-Verlag.
\newblock ISBN 978-3-030-58609-6.
\newblock \doi{10.1007/978-3-030-58610-2_13}.
\newblock URL \url{https://doi.org/10.1007/978-3-030-58610-2_13}.

\bibitem[Feng et~al.(2021{\natexlab{a}})Feng, Wang, Zhou, Deng, and
  Li]{feng2021doctr}
Hao Feng, Yuechen Wang, Wengang Zhou, Jiajun Deng, and Houqiang Li.
\newblock Doctr: Document image transformer for geometric unwarping and
  illumination correction.
\newblock In \emph{Proceedings of the 29th ACM International Conference on
  Multimedia}, MM '21, page 273–281, New York, NY, USA, 2021{\natexlab{a}}.
  Association for Computing Machinery.
\newblock ISBN 9781450386517.
\newblock \doi{10.1145/3474085.3475388}.
\newblock URL \url{https://doi.org/10.1145/3474085.3475388}.

\bibitem[Feng et~al.(2021{\natexlab{b}})Feng, Zhou, Deng, Tian, and
  Li]{feng2021docscanner}
Hao Feng, Wengang Zhou, Jiajun Deng, Qi~Tian, and Houqiang Li.
\newblock Docscanner: Robust document image rectification with progressive
  learning, 2021{\natexlab{b}}.
\newblock URL \url{https://arxiv.org/abs/2110.14968}.

\bibitem[Das et~al.(2019)Das, Ma, Shu, Samaras, and Shilkrot]{das2019dewarpnet}
Sagnik Das, Ke~Ma, Zhixin Shu, Dimitris Samaras, and Roy Shilkrot.
\newblock Dewarpnet: Single-image document unwarping with stacked 3d and 2d
  regression networks.
\newblock In \emph{2019 IEEE/CVF International Conference on Computer Vision
  (ICCV)}, pages 131--140, 2019.
\newblock \doi{10.1109/ICCV.2019.00022}.

\bibitem[Das et~al.(2021)Das, Singh, Wu, Bas, Mahadevan, Bhotika, and
  Samaras]{das2021endtoend}
Sagnik Das, Kunwar~Yashraj Singh, Jon Wu, Erhan Bas, Vijay Mahadevan, Rahul
  Bhotika, and Dimitris Samaras.
\newblock End-to-end piece-wise unwarping of document images.
\newblock In \emph{2021 IEEE/CVF International Conference on Computer Vision
  (ICCV)}, pages 4248--4257, 2021.
\newblock \doi{10.1109/ICCV48922.2021.00423}.

\bibitem[Jiang et~al.(2020)Jiang, Wang, Rist, Wallner, and
  Pottmann]{jiang2020quadmesh}
Caigui Jiang, Cheng Wang, Florian Rist, Johannes Wallner, and Helmut Pottmann.
\newblock Quad-mesh based isometric mappings and developable surfaces.
\newblock \emph{ACM Trans. Graph.}, 39\penalty0 (4), aug 2020.
\newblock ISSN 0730-0301.
\newblock \doi{10.1145/3386569.3392430}.
\newblock URL \url{https://doi.org/10.1145/3386569.3392430}.

\bibitem[Bo et~al.(2012)Bo, Ling, and Wang]{bo2012Revisit}
Pengbo Bo, Ruotian Ling, and Wenping Wang.
\newblock A revisit to fitting parametric surfaces to point clouds.
\newblock \emph{Computers \& Graphics}, 36\penalty0 (5):\penalty0 534--540,
  2012.
\newblock ISSN 00978493.
\newblock \doi{10.1016/j.cag.2012.03.036}.

\bibitem[Sawhney and Crane(2017)]{sawhney2018boundary}
Rohan Sawhney and Keenan Crane.
\newblock Boundary first flattening.
\newblock \emph{ACM Trans. Graph.}, 37\penalty0 (1), dec 2017.
\newblock ISSN 0730-0301.
\newblock \doi{10.1145/3132705}.
\newblock URL \url{https://doi.org/10.1145/3132705}.

\bibitem[Koo and Cho(2010)]{koo2010state}
Hyung~Il Koo and Nam~Ik Cho.
\newblock State estimation in a document image and its application in text
  block identification and text line extraction.
\newblock \emph{Lecture Notes in Computer Science (including subseries Lecture
  Notes in Artificial Intelligence and Lecture Notes in Bioinformatics)}, 6312
  LNCS\penalty0 (PART 2):\penalty0 421--434, 2010.
\newblock ISSN 3642155510.
\newblock \doi{10.1007/978-3-642-15552-9_31}.

\bibitem[Grompone~von Gioi et~al.(2012)Grompone~von Gioi, Jakubowicz, Morel,
  and Randall]{gromponevongioi2012lsd}
Rafael Grompone~von Gioi, Jérémie Jakubowicz, Jean-Michel Morel, and Gregory
  Randall.
\newblock {LSD: a Line Segment Detector}.
\newblock \emph{{Image Processing On Line}}, 2:\penalty0 35--55, 2012.
\newblock \url{https://doi.org/10.5201/ipol.2012.gjmr-lsd}.

\bibitem[Moulon et~al.(2017)Moulon, Monasse, Perrot, and
  Marlet]{moulon2017openmvg}
Pierre Moulon, Pascal Monasse, Romuald Perrot, and Renaud Marlet.
\newblock Openmvg: Open multiple view geometry.
\newblock In Bertrand Kerautret, Miguel Colom, and Pascal Monasse, editors,
  \emph{Reproducible Research in Pattern Recognition}, pages 60--74, Cham,
  2017. Springer International Publishing.
\newblock ISBN 978-3-319-56414-2.

\bibitem[Wang et~al.(2003)Wang, Simoncelli, and Bovik]{wang2003multiscale}
Z.~Wang, E.P. Simoncelli, and A.C. Bovik.
\newblock Multiscale structural similarity for image quality assessment.
\newblock In \emph{The Thrity-Seventh Asilomar Conference on Signals, Systems
  \& Computers, 2003}, volume~2, pages 1398--1402 Vol.2, 2003.
\newblock \doi{10.1109/ACSSC.2003.1292216}.

\bibitem[Wang et~al.(2004)Wang, Bovik, Sheikh, and Simoncelli]{wang2004image}
Zhou Wang, A.C. Bovik, H.R. Sheikh, and E.P. Simoncelli.
\newblock Image quality assessment: from error visibility to structural
  similarity.
\newblock \emph{IEEE Transactions on Image Processing}, 13\penalty0
  (4):\penalty0 600--612, 2004.
\newblock \doi{10.1109/TIP.2003.819861}.

\bibitem[Liu et~al.(2011)Liu, Yuen, and Torralba]{celiu2011sift}
Ce~Liu, Jenny Yuen, and Antonio Torralba.
\newblock Sift flow: Dense correspondence across scenes and its applications.
\newblock \emph{IEEE Transactions on Pattern Analysis and Machine
  Intelligence}, 33\penalty0 (5):\penalty0 978--994, 2011.
\newblock \doi{10.1109/TPAMI.2010.147}.

\bibitem[Weinzaepfel et~al.(2013)Weinzaepfel, Revaud, Harchaoui, and
  Schmid]{weinzaepfel2013deepflow}
Philippe Weinzaepfel, Jerome Revaud, Zaid Harchaoui, and Cordelia Schmid.
\newblock Deepflow: Large displacement optical flow with deep matching.
\newblock In \emph{2013 IEEE International Conference on Computer Vision},
  pages 1385--1392, 2013.
\newblock \doi{10.1109/ICCV.2013.175}.

\bibitem[Zhang et~al.(2018)Zhang, Isola, Efros, Shechtman, and
  Wang]{zhang2018unreasonable}
Richard Zhang, Phillip Isola, Alexei~A. Efros, Eli Shechtman, and Oliver Wang.
\newblock The unreasonable effectiveness of deep features as a perceptual
  metric.
\newblock In \emph{2018 IEEE/CVF Conference on Computer Vision and Pattern
  Recognition}, pages 586--595, 2018.
\newblock \doi{10.1109/CVPR.2018.00068}.

\bibitem[Smith(2007)]{smith2007overview}
R.~Smith.
\newblock An overview of the tesseract ocr engine.
\newblock In \emph{Ninth International Conference on Document Analysis and
  Recognition (ICDAR 2007)}, volume~2, pages 629--633, 2007.
\newblock \doi{10.1109/ICDAR.2007.4376991}.

\end{thebibliography}
